\documentclass[11pt,table]{article}
\pdfoutput=1

\usepackage[utf8]{inputenc} % allow utf-8 input
\usepackage[T1]{fontenc}    % use 8-bit T1 fonts
\usepackage[in]{fullpage}
\usepackage{xurl}
\usepackage{booktabs}       % professional-quality tables
\usepackage{amsfonts}       % blackboard math symbols
\usepackage{nicefrac}       % compact symbols for 1/2, etc.
\usepackage{microtype}      % microtypography
\usepackage{amsmath,amssymb,amsthm}
\usepackage{mathtools}
\usepackage{mathrsfs}
\usepackage{thmtools}
\usepackage{thm-restate}
\usepackage{comment}
\usepackage{parskip}
\usepackage{siunitx}
\usepackage{etoolbox}
\usepackage{enumitem}
\usepackage{graphicx}
\usepackage{subcaption}
\usepackage[textsize=scriptsize,textwidth=2.2cm]{todonotes}
\usepackage{tabularx}
\usepackage{xcolor} 
\usepackage[numbers,sort]{natbib}
\usepackage{hyperref}
\usepackage{longtable}
\usepackage{afterpage}
\usepackage{etoc}

\extrafloats{100}

\newtoggle{showtodos}
\togglefalse{showtodos}

\newtoggle{showappendix}
\toggletrue{showappendix}

\newtoggle{isicml}
\togglefalse{isicml}

\newcommand{\kwfontsize}{\scriptsize}

\newcommand{\dist}{\ensuremath{\mathcal{D}}}
\newcommand{\fhat}{\ensuremath{\hat{f}}}
\newcommand{\ermloss}{\ensuremath{L}}
\newcommand{\Eop}{\ensuremath{\mathop{\mathbb{E}}}}
\newcommand{\E}{\Eop}
\newcommand{\R}{\ensuremath{\mathbb{R}}}
\newcommand{\N}{\ensuremath{\mathcal{N}}}

\newcommand{\Dtest}{\ensuremath{S}}
\newcommand{\D}[1]{\ensuremath{D_{\text{#1}}}}
\newcommand{\ind}{\ensuremath{\mathbb{I}}}
\newcommand{\accorig}{\ensuremath{\text{acc}_\text{orig}}}
\newcommand{\accnew}{\ensuremath{\text{acc}_\text{new}}}
\newcommand{\acc}{\ensuremath{\text{acc}}}

\newcommand{\keyword}[1]{\texttt{#1}}
\newcommand{\model}[1]{\texttt{#1}}
\newcommand{\class}[1]{\texttt{#1}}
\newcommand{\airplane}{\class{airplane}}

\newcommand{\dataseta}{\textsf{Threshold0.7}}
\newcommand{\datasetb}{\textsf{MatchedFrequency}}
\newcommand{\datasetc}{\textsf{TopImages}}

\newcolumntype{L}[1]{>{\raggedright\arraybackslash}p{#1}}
\newcolumntype{C}[1]{>{\centering\arraybackslash}p{#1}}
\newcolumntype{R}[1]{>{\raggedleft\arraybackslash}p{#1}}

\DeclarePairedDelimiter{\abs}{\lvert}{\rvert}

\DeclarePairedDelimiter{\brackets}{[}{]}

\iftoggle{showtodos}{
\newcommand{\becca}[1]{\todo[color=green!40]{Becca: #1}}
\newcommand{\ludwig}[1]{\todo[color=blue!40]{Ludwig: #1}}
}{
\newcommand{\becca}[1]{}
\newcommand{\ludwig}[1]{}
}

\title{Do ImageNet Classifiers Generalize to ImageNet?}
\author{Benjamin Recht\thanks{Authors ordered alphabetically. Ben did none of the work.} \\ UC Berkeley\and Rebecca Roelofs\\ UC Berkeley\and Ludwig Schmidt\\ UC Berkeley\and Vaishaal Shankar\\ UC Berkeley}

\begin{document}

\date{}

\maketitle

\begin{abstract}
We build new test sets for the CIFAR-10 and ImageNet datasets.
Both benchmarks have been the focus of intense research for almost a decade, raising the danger of overfitting to excessively re-used test sets.
By closely following the original dataset creation processes, we test to what extent current classification models generalize to new data.
We evaluate a broad range of models and find accuracy drops of 3\% -- 15\% on CIFAR-10 and 11\% -- 14\% on ImageNet.
However, accuracy gains on the original test sets translate to larger gains on the new test sets.
Our results suggest that the accuracy drops are not caused by adaptivity, but by the models' inability to generalize to slightly ``harder'' images than those found in the original test sets.

\end{abstract}

\etocdepthtag.toc{mtsection}

\section{Introduction}
\label{sec:intro}
\newlength{\negspaceint}
\iftoggle{isicml}{
\setlength{\negspaceint}{-0.4cm}
}{
\setlength{\negspaceint}{-0.0cm}
}

The overarching goal of machine learning is to produce models that \emph{generalize}.
We usually quantify generalization by measuring the performance of a model on a held-out test set.
What does good performance on the test set then imply?
At the very least, one would hope that the model also performs well on a new test set assembled from the same data source by following the same data cleaning protocol.

In this paper, we realize this thought experiment by replicating the dataset creation process for two prominent benchmarks, CIFAR-10 and ImageNet \cite{krizhevsky2009learning,imagenet}.
In contrast to the ideal outcome, we find that a wide range of classification models fail to reach their original accuracy scores.
The accuracy drops range from 3\% to 15\% on CIFAR-10 and 11\% to 14\% on ImageNet.
On ImageNet, the accuracy loss amounts to approximately five years of progress in a highly active period of machine learning research.

Conventional wisdom suggests that such drops arise because the models have been adapted to the specific images in the original test sets, e.g., via extensive hyperparameter tuning.
However, our experiments show that the relative order of models is almost exactly preserved on our new test sets: the models with highest accuracy on the original test sets are still the models with highest accuracy on the new test sets.
Moreover, there are no diminishing returns in accuracy.
In fact, every percentage point of accuracy improvement on the original test set translates to a \emph{larger} improvement on our new test sets.
So although later models could have been adapted more to the test set, they see smaller drops in accuracy.
These results provide evidence that exhaustive test set evaluations are an effective way to improve image classification models.
Adaptivity is therefore an unlikely explanation for the accuracy drops.

Instead, we propose an alternative explanation based on the relative difficulty of the original and new test sets.
We demonstrate that it is possible to recover the original ImageNet accuracies almost exactly if we only include the easiest images from our candidate pool.
This suggests that the accuracy scores of even the best image classifiers are still highly sensitive to minutiae of the data cleaning process.
This brittleness puts claims about human-level performance into context \cite{karpathycifarblog,superhuman,RDSKSMHKKBBL15}.
It also shows that current classifiers still do not generalize reliably even in the benign environment of a carefully controlled reproducibility experiment.

\begin{figure*}[ht!]
  \centering
  \begin{subfigure}{0.48\textwidth}
    \includegraphics[width=\linewidth]{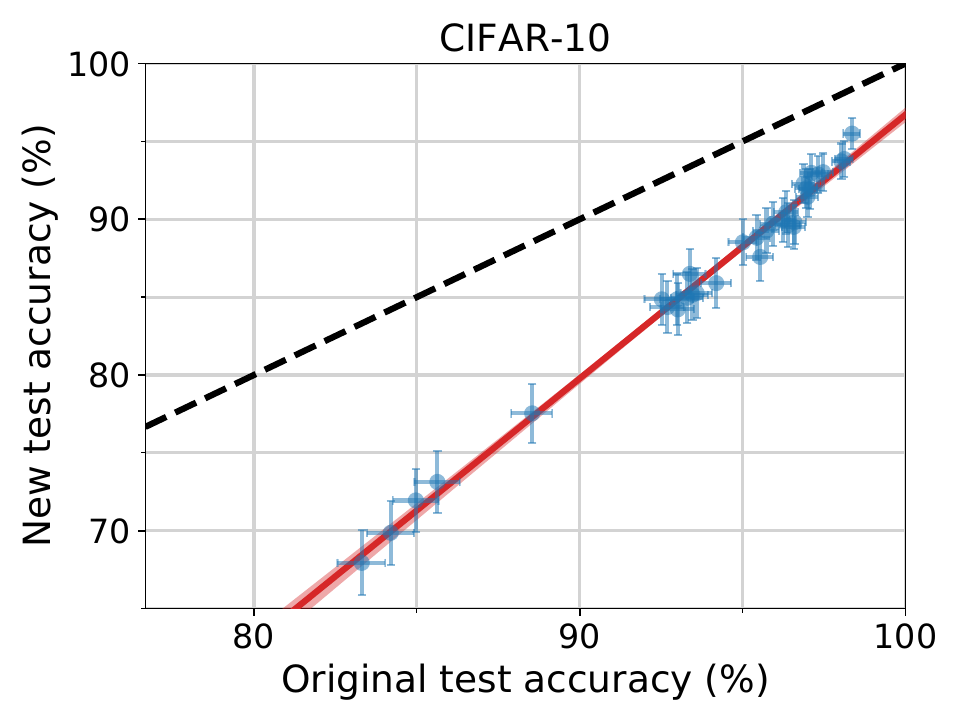}
  \end{subfigure}
  \hfill
  \begin{subfigure}{0.48\textwidth}
    \includegraphics[width=\linewidth]{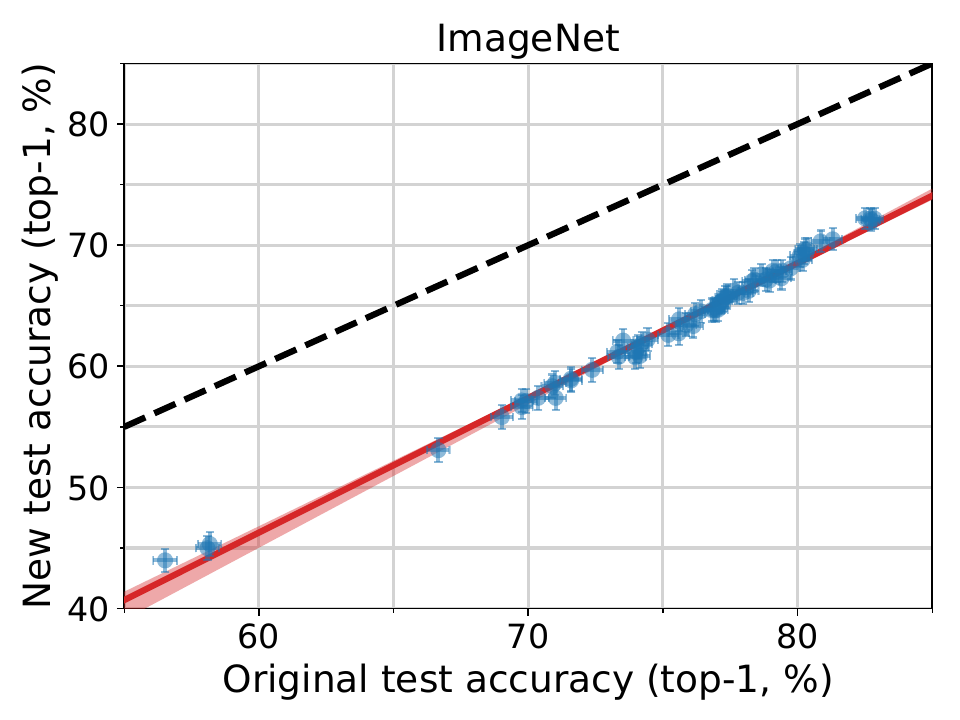}
  \end{subfigure}
  \begin{subfigure}{\textwidth}
    \vspace{-.15cm}
    \centering
    \includegraphics[width=.75\linewidth]{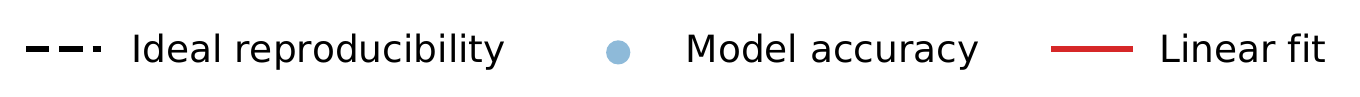}
  \end{subfigure}
  \vspace{-.6cm}
  \caption{
    Model accuracy on the original test sets vs.\ our new test sets.
    Each data point corresponds to one model in our testbed (shown with 95\% Clopper-Pearson confidence intervals).
    The plots reveal two main phenomena:
    (i) There is a significant drop in accuracy from the original to the new test sets.
    (ii) The model accuracies closely follow a linear function with slope \emph{greater} than 1 ($1.7$ for CIFAR-10 and $1.1$ for ImageNet).
    This means that every percentage point of progress on the original test set translates into more than one percentage point on the new test set.
    The two plots are drawn so that their aspect ratio is the same, i.e., the slopes of the lines are visually comparable.
    The red shaded region is a 95\% confidence region for the linear fit from 100,000 bootstrap samples.
    \vspace{-.3cm}
  }
  \label{fig:intro_plot}
\end{figure*}

Figure \ref{fig:intro_plot} shows the main result of our experiment.
Before we describe our methodology in Section \ref{sec:overview}, the next section provides relevant background.
To enable future research, we release both our new test sets and the corresponding code.\footnote{\url{https://github.com/modestyachts/CIFAR-10.1} and \url{https://github.com/modestyachts/ImageNetV2}}

\section{Potential Causes of Accuracy Drops}
\label{sec:formal}
We adopt the standard classification setup and posit the existence of a ``true'' underlying data distribution $\dist$ over labeled examples $(x, y)$.
The overall goal in classification is to find a model $\fhat$ that minimizes the population loss
\begin{equation}
  \label{eq:pop_loss}
  \ermloss_\dist(\fhat) \; = \; \Eop_{(x, y) \sim \dist} \brackets*{\ind\brackets{\fhat(x) \neq y}} \; .
\end{equation}

Since we usually do not know the distribution $\dist$, we instead measure the performance of a trained classifier via a \emph{test set} $\Dtest$ drawn from the distribution $\dist$:
\begin{equation}
  \label{eq:emp_loss}
  \ermloss_{\Dtest}(\fhat) \; = \; \frac{1}{\abs{\Dtest}} \sum_{(x, y) \in \Dtest} \ind\brackets{\fhat(x) \neq y} \; .
\end{equation}

We then use this test error $\ermloss_{\Dtest}(\fhat)$ as a proxy for the population loss $\ermloss_\dist(\fhat)$.
If a model $\fhat$ achieves a low test error, we assume that it will perform similarly well on future examples from the distribution $\dist$.
This assumption underlies essentially all empirical evaluations in machine learning since it allows us to argue that the model $\fhat$ generalizes.

In our experiments, we test this assumption by collecting a new test set $\Dtest'$ from a data distribution $\dist'$ that we carefully control to resemble the original distribution $\dist$.
Ideally, the original test accuracy $\ermloss_{\Dtest}(\fhat)$ and new test accuracy $\ermloss_{\Dtest'}(\fhat)$ would then match up to the random sampling error.
In contrast to this idealized view, our results in Figure \ref{fig:intro_plot} show a large drop in accuracy from the original test set $\Dtest$ set to our new test set $\Dtest'$.
To understand this accuracy drop in more detail, we decompose the difference between $\ermloss_{\Dtest}(\fhat)$ and $\ermloss_{\Dtest'}(\fhat)$ into three parts (dropping the dependence on $\fhat$ to simplify notation):
\begin{align*}
  \ermloss_{\Dtest} - \ermloss_{\Dtest'} \,  = \, \underbrace{(\ermloss_{\Dtest}  - \ermloss_{\dist})}_{\text{Adaptivity gap}} \iftoggle{isicml}{}{\;} + \iftoggle{isicml}{}{\;} \underbrace{( \ermloss_{\dist} -  \ermloss_{\dist'})}_{\text{Distribution Gap}} \iftoggle{isicml}{}{\;} + \iftoggle{isicml}{}{\;} \underbrace{(\ermloss_{\dist'} - \ermloss_{\Dtest'})}_{\text{Generalization gap}} \iftoggle{isicml}{}{\,} .
\end{align*}
We now discuss to what extent each of the three terms can lead to accuracy drops.

\parskip=2.5pt
\vspace{\negspaceint}
\paragraph{Generalization Gap.} By construction, our new test set $\Dtest'$ is independent of the existing classifier $\fhat$.
Hence the third term $\ermloss_{\dist'} - \ermloss_{\Dtest'}$ is the standard \emph{generalization gap} commonly studied in machine learning.
It is determined solely by the random sampling error.
%\iftoggle{isicml}{The size of our test sets makes the generalization gap small enough so we can ignore it here (see Appendix \ref{app:randomness} for details).}{

A first guess is that this inherent sampling error suffices to explain the accuracy drops in Figure \ref{fig:intro_plot} (e.g., the new test set $\Dtest'$ could have sampled certain ``harder'' modes of the distribution $\dist$ more often).
However, random fluctuations of this magnitude are unlikely for the size of our test sets.
With 10,000 data points (as in our new ImageNet test set), a Clopper-Pearson 95\% confidence interval for the test accuracy has size of at most $\pm \;\! 1\%$.
Increasing the confidence level to 99.99\% yields a confidence interval of size at most $\pm \;\! 2\%$.
Moreover, these confidence intervals become smaller for higher accuracies, which is the relevant regime for the best-performing models.
Hence random chance alone cannot explain the accuracy drops observed in our experiments.\footnote{We remark that the sampling process for the new test set $\Dtest'$ could indeed \emph{systematically} sample harder modes more often than under the original data distribution $\dist$.
Such a systematic change in the sampling process would not be an effect of random chance but captured by the distribution gap described below.}
%}

\vspace{\negspaceint}
\paragraph{Adaptivity Gap.} 
We call the term $\ermloss_{\Dtest}  - \ermloss_{\dist}$ the \emph{adaptivity gap}.
It measures how much adapting the model $\fhat$ to the test set $\Dtest$ causes the test error $\ermloss_{\Dtest}$ to underestimate the population loss $\ermloss_{\dist}$.
If we assumed that our model $\fhat$ is independent of the test set $\Dtest$, this terms would follow the same concentration laws as the generalization gap $\ermloss_{\dist'}  - \ermloss_{\Dtest'}$ above. %\iftoggle{isicml}{ (see Appendix \ref{app:randomness})}{}.
But this assumption is undermined by the common practice of tuning model hyperparameters directly on the test set, which introduces dependencies between the model $\fhat$ and the test set $\Dtest$. In the extreme case, this can be seen as training directly on the test set.
But milder forms of adaptivity may also artificially inflate accuracy scores by increasing the gap between $\ermloss_{\Dtest}$ and $\ermloss_{\dist}$ beyond the purely random error.
\paragraph{Distribution Gap.} We call the term $\ermloss_{\dist} -  \ermloss_{\dist'}$ the \emph{distribution gap}.
  It quantifies how much the change from the original distribution $\dist$ to our new distribution $\dist'$ affects the model $\fhat$.
Note that this term is not influenced by random effects but quantifies the systematic difference between sampling the original and new test sets.
While we went to great lengths to minimize such systematic differences, in practice it is hard to argue whether two high-dimensional distributions are exactly the same.
We typically lack a precise definition of either distribution, and collecting a real dataset involves a plethora of design choices.
\subsection{Distinguishing Between the Two Mechanisms}
\label{sec:formal_multiple}
For a single model $\fhat$, it is unclear how to disentangle the adaptivity and distribution gaps.
To gain a more nuanced understanding, we measure accuracies for \emph{multiple} models $\fhat_1, \ldots, \fhat_k$.
This provides additional insights because it allows us to determine how the two gaps have evolved over time.

For both CIFAR-10 and ImageNet, the classification models come from a long line of papers that incrementally improved accuracy scores over the past decade.
A natural assumption is that later models have experienced more adaptive overfitting since they are the result of more successive hyperparameter tuning on the same test set.
Their higher accuracy scores would then come from an increasing adaptivity gap and reflect progress only on the specific examples in the test set $\Dtest$ but not on the actual distribution $\dist$.
In an extreme case, the population accuracies $\ermloss_{\dist}(\fhat_i)$ would plateau (or even decrease) while the test accuracies $\ermloss_{\Dtest}(\fhat_i)$ would continue to grow for successive models $\fhat_i$.

However, this idealized scenario is in stark contrast to our results in Figure \ref{fig:intro_plot}.
Later models do not see diminishing returns but an \emph{increased} advantage over earlier models.
Hence we view our results as evidence that the accuracy drops mainly stem from a large distribution gap.
After presenting our results in more detail in the next section, we will further discuss this point in Section \ref{sec:discussion}.
\parskip=4.5pt

\section{Summary of Our Experiments}
\label{sec:overview}
\newlength{\negspaceow}
\iftoggle{isicml}{
\setlength{\negspaceow}{-0.4cm}
}{
\setlength{\negspaceow}{-0.0cm}
}

We now give an overview of the main steps in our reproducibility experiment.
Appendices \ref{app:cifar} and \ref{app:imagenet} describe our methodology in more detail.
We begin with the first decision, which was to choose informative datasets.

\subsection{Choice of Datasets}
We focus on image classification since it has become the most prominent task in machine learning and underlies a broad range of applications.
The cumulative progress on ImageNet is often cited as one of the main breakthroughs in computer vision and machine learning \cite{MalikCACM}.
State-of-the-art models now surpass human-level accuracy by some measure \cite{superhuman,RDSKSMHKKBBL15}.
This makes it particularly important to check if common image classification models can reliably generalize to new data from the same source.

We decided on CIFAR-10 and ImageNet, two of the most widely-used image classification benchmarks \cite{hamnerpopular}.
Both datasets have been the focus of intense research for almost ten years now.
Due to the competitive nature of these benchmarks, they are an excellent example for testing whether adaptivity has led to overfitting.
In addition to their popularity, their carefully documented dataset creation process makes them well suited for a reproducibility experiment \cite{krizhevsky2009learning,imagenet,RDSKSMHKKBBL15}.

Each of the two datasets has specific features that make it especially interesting for our replication study.
CIFAR-10 is small enough so that many researchers developed and tested new models for this dataset.
In contrast, ImageNet requires significantly more computational resources, and experimenting with new architectures has long been out of reach for many research groups.
As a result, CIFAR-10 has likely experienced more hyperparameter tuning, which may also have led to more adaptive overfitting.

On the other hand, the limited size of CIFAR-10 could also make the models more susceptible to small changes in the distribution.
Since the CIFAR-10 models are only exposed to a constrained visual environment, they may be unable to learn a robust representation.
In contrast, ImageNet captures a much broader variety of images: it contains about $24\times$ more training images than CIFAR-10 and roughly $100 \times$ more pixels per image.
So conventional wisdom (such as the claims of human-level performance) would suggest that ImageNet models also generalize more reliably .

As we will see, neither of these conjectures is supported by our data: CIFAR-10 models do not suffer from more adaptive overfitting, and ImageNet models do not appear to be significantly more robust.

\subsection{Dataset Creation Methodology}
One way to test generalization would be to evaluate existing models on new i.i.d.\ data from the original test distribution.
For example, this would be possible if the original dataset authors had collected a larger initial dataset and randomly split it into two test sets, keeping one of the test sets hidden for several years.
Unfortunately, we are not aware of such a setup for CIFAR-10 or ImageNet.

In this paper, we instead mimic the original distribution as closely as possible by repeating the dataset curation process that selected the original test set\footnote{For ImageNet, we repeat the creation process of the \emph{validation set} because most papers developed and tested models on the validation set. We discuss this point in more detail in Appendix \ref{sec:imagenet_building_new_test_set}. In the context to this paper, we use the terms ``validation set'' and ``test set'' interchangeably for ImageNet.} from a larger data source.
While this introduces the difficulty of disentangling the adaptivity gap from the distribution gap, it also enables us to check whether independent replication affects current accuracy scores.
In spite of our efforts, we found that it is astonishingly hard to replicate the test set distributions of CIFAR-10 and ImageNet. At a high level, creating a new test set consists of two parts:
\paragraph{Gathering Data.} To obtain images for a new test set, a simple approach would be to use a different dataset, e.g., Open Images \cite{openimages}.
However, each dataset comes with specific biases \cite{TE11}.
For instance, CIFAR-10 and ImageNet were assembled in the late 2000s, and some classes such as \class{car} or \class{cell\_phone} have changed significantly over the past decade.
We avoided such biases by drawing new images from the same source as CIFAR-10 and ImageNet.
For CIFAR-10, this was the larger Tiny Image dataset \cite{tinyimages}.
For ImageNet, we followed the original process of utilizing the Flickr image hosting service and only considered images uploaded in a similar time frame as for ImageNet.
In addition to the data source and the class distribution, both datasets also have rich structure \emph{within} each class.
For instance, each class in CIFAR-10 consists of images from multiple specific keywords in Tiny Images.
Similarly, each class in ImageNet was assembled from the results of multiple queries to the Flickr API.
We relied on the documentation of the two datasets to closely match the sub-class distribution as well.
\vspace{\negspaceow}
\paragraph{Cleaning Data.} Many images in Tiny Images and the Flickr results are only weakly related to the query (or not at all).
To obtain a high-quality dataset with correct labels, it is therefore necessary to manually select valid images from the candidate pool.
While this step may seem trivial, our results in Section \ref{sec:imagenet_details} will show that it has major impact on the model accuracies.

The authors of CIFAR-10 relied on paid student labelers to annotate their dataset.
The researchers in the ImageNet project utilized Amazon Mechanical Turk (MTurk) to handle the large size of their dataset.
We again replicated both annotation processes.
Two graduate students authors of this paper impersonated the CIFAR-10 labelers, and we employed MTurk workers for our new ImageNet test set.
For both datasets, we also followed the original labeling instructions, MTurk task format, etc.\

After collecting a set of correctly labeled images, we sampled our final test sets from the filtered candidate pool.
We decided on a test set size of 2,000 for CIFAR-10 and 10,000 for ImageNet.
While these are smaller than the original test sets, the sample sizes are still large enough to obtain 95\% confidence intervals of about $\pm 1\%$.
Moreover, our aim was to avoid bias due to CIFAR-10 and ImageNet possibly leaving only ``harder'' images in the respective data sources.
This effect is minimized by building test sets that are small compared to the original datasets (about 3\% of the overall CIFAR-10 dataset and less than 1\% of the overall ImageNet dataset).

\subsection{Results on the New Test Sets}

\begin{table*}[ht!]
    \centering
    
    \begin{subtable}{\linewidth}
      \centering
      \rowcolors{4}{white}{gray!15}
      \begin{tabular}{rp{4.75cm}rrrrr}
\toprule 
\multicolumn{7}{c}{\textbf{CIFAR-10}} \\ 
\midrule
\multicolumn{1}{l}{Orig.} &                                 &                                        &                                          &    & \multicolumn{1}{l}{New} &  \\ 
 \multicolumn{1}{l}{Rank} & Model & Orig. Accuracy & New Accuracy & Gap & \multicolumn{1}{l}{Rank} & $\Delta$ Rank \\
\midrule
 1 &  \model{autoaug\_pyramid\_net\_tf} &  98.4 {\footnotesize \textcolor{gray}{[98.1, 98.6]}} &  95.5 {\footnotesize \textcolor{gray}{[94.5, 96.4]}} &  2.9 &  1 &  0 \\
 6 &  \model{shake\_shake\_64d\_cutout} &  97.1 {\footnotesize \textcolor{gray}{[96.8, 97.4]}} &  93.0 {\footnotesize \textcolor{gray}{[91.8, 94.1]}} &  4.1 &  5 &  1 \\
 16 &  \model{wide\_resnet\_28\_10} &  95.9 {\footnotesize \textcolor{gray}{[95.5, 96.3]}} &  89.7 {\footnotesize \textcolor{gray}{[88.3, 91.0]}} &  6.2 &  14 &  2 \\
 23 &  \model{resnet\_basic\_110} &  93.5 {\footnotesize \textcolor{gray}{[93.0, 93.9]}} &  85.2 {\footnotesize \textcolor{gray}{[83.5, 86.7]}} &  8.3 &  24 &  -1 \\
 27 &  \model{vgg\_15\_BN\_64} &  93.0 {\footnotesize \textcolor{gray}{[92.5, 93.5]}} &  84.9 {\footnotesize \textcolor{gray}{[83.2, 86.4]}} &  8.1 &  27 &  0 \\
 30 &  \model{cudaconvnet} &  88.5 {\footnotesize \textcolor{gray}{[87.9, 89.2]}} &  77.5 {\footnotesize \textcolor{gray}{[75.7, 79.3]}} &  11.0 &  30 &  0 \\
 31 &  \model{random\_features\_256k\_aug} &  85.6 {\footnotesize \textcolor{gray}{[84.9, 86.3]}} &  73.1 {\footnotesize \textcolor{gray}{[71.1, 75.1]}} &  12.5 &  31 &  0 \\
\bottomrule
\end{tabular}

      \label{tab:subsampled_cifar_model_results}
    \end{subtable}
    \begin{subtable}{\linewidth}
      \centering
        \rowcolors{4}{white}{gray!15}
        \begin{tabular}{rp{4.75cm}rrrrr}
\toprule 
\multicolumn{7}{c}{\textbf{ImageNet Top-1 }} \\ 
\midrule
\multicolumn{1}{l}{Orig.} &                        &                                        &                                          &    & \multicolumn{1}{l}{New} &  \\ 
 \multicolumn{1}{l}{Rank} & Model & Orig. Accuracy & New Accuracy & Gap & \multicolumn{1}{l}{Rank} & $\Delta$ Rank \\
\midrule
 1 &  \model{pnasnet\_large\_tf} &  82.9 {\footnotesize \textcolor{gray}{[82.5, 83.2]}} &  72.2 {\footnotesize \textcolor{gray}{[71.3, 73.1]}} &  10.7 &  3 &  -2 \\
 4 &  \model{nasnetalarge} &  82.5 {\footnotesize \textcolor{gray}{[82.2, 82.8]}} &  72.2 {\footnotesize \textcolor{gray}{[71.3, 73.1]}} &  10.3 &  1 &  3 \\
 21 &  \model{resnet152} &  78.3 {\footnotesize \textcolor{gray}{[77.9, 78.7]}} &  67.0 {\footnotesize \textcolor{gray}{[66.1, 67.9]}} &  11.3 &  21 &  0 \\
 23 &  \model{inception\_v3\_tf} &  78.0 {\footnotesize \textcolor{gray}{[77.6, 78.3]}} &  66.1 {\footnotesize \textcolor{gray}{[65.1, 67.0]}} &  11.9 &  24 &  -1 \\
 30 &  \model{densenet161} &  77.1 {\footnotesize \textcolor{gray}{[76.8, 77.5]}} &  65.3 {\footnotesize \textcolor{gray}{[64.4, 66.2]}} &  11.8 &  30 &  0 \\
 43 &  \model{vgg19\_bn} &  74.2 {\footnotesize \textcolor{gray}{[73.8, 74.6]}} &  61.9 {\footnotesize \textcolor{gray}{[60.9, 62.8]}} &  12.3 &  44 &  -1 \\
 64 &  \model{alexnet} &  56.5 {\footnotesize \textcolor{gray}{[56.1, 57.0]}} &  44.0 {\footnotesize \textcolor{gray}{[43.0, 45.0]}} &  12.5 &  64 &  0 \\
 65 &  \model{fv\_64k} &  35.1 {\footnotesize \textcolor{gray}{[34.7, 35.5]}} &  24.1 {\footnotesize \textcolor{gray}{[23.2, 24.9]}} &  11.0 &  65 &  0 \\
\bottomrule
\end{tabular}

        \label{tab:subsampled_imagenet_model_results}
    \end{subtable}
    \vspace{-3mm}
    \caption{Model accuracies on the original CIFAR-10 test set, the original ImageNet validation set, and our new test sets.
      $\Delta$ Rank is the relative difference in the ranking from the original test set to the new test set in the full ordering of all models (see Appendices \ref{apx:cifar10_model_accuracies} and \ref{sec:imagenettable}).
      For example, $\Delta \text{Rank} = -2$ means that a model dropped by two places on the new test set compared to the original test set.
      The confidence intervals are 95\% Clopper-Pearson intervals.
      Due to space constraints, references for the models can be found in Appendices \ref{apx:cifar10_model_descriptions} and \ref{apx:imagenet_model_descriptions}.}
    \label{tab:subsampled_model_results}
    \vspace{-3.0mm}
\end{table*}

After assembling our new test sets, we evaluated a broad range of image classification models spanning a decade of machine learning research.
The models include the seminal AlexNet \cite{alexnet}, widely used convolutional networks \cite{vgg,resnet,densenet,inceptionv3}, and the state-of-the-art \cite{autoaugment, pnasnet}.
For all deep architectures, we used code previously published online. 
We relied on pre-trained models whenever possible and otherwise ran the training commands from the respective repositories.
In addition, we also evaluated the best-performing approaches preceding convolutional networks on each dataset.
These are random features for CIFAR-10 \cite{rahimi2009weighted,rf} and Fisher vectors for ImageNet \cite{fishervectors}.\footnote{We remark that our implementation of Fisher vectors yields top-5 accuracy numbers that are 17\% lower than the published numbers in ILSVRC 2012 \cite{RDSKSMHKKBBL15}.
  Unfortunately, there is no publicly available reference implementation of Fisher vector models achieving this accuracy score.
Hence our implementation should not be seen as an exact reproduction of the state-of-the-art Fisher vector model, but as a baseline inspired by this approach.
The main goal of including Fisher vector models in our experiment is to investigate if they follow the same overall trends as convolutional neural networks.}
We wrote our own implementations for these models, which we also release publicly.\footnote{\url{https://github.com/modestyachts/nondeep}}

Overall, the top-1 accuracies range from 83\% to 98\% on the original CIFAR-10 test set and 21\% to 83\% on the original ImageNet validation set.
We refer the reader to Appendices \ref{apx:imagenet_model_descriptions} and \ref{apx:cifar10_model_descriptions} for a full list of models and source repositories.

Figure \ref{fig:intro_plot} in the introduction plots original vs.\ new accuracies, and Table \ref{tab:subsampled_model_results} in this section summarizes the numbers of key models.
The remaining accuracy scores can be found in Appendices \ref{apx:cifar10_model_accuracies} and \ref{sec:imagenettable}.
We now briefly describe the two main trends and discuss the results further in Section \ref{sec:discussion}.

\paragraph{A Significant Drop in Accuracy.} All models see a large drop in accuracy from the original test sets to our new test sets.
For widely used architectures such as VGG \cite{vgg} and ResNet \cite{resnet}, the drop is 8\% on CIFAR-10 and 11\% on ImageNet.
On CIFAR-10, the state of the art \cite{autoaugment} is more robust and only drops by 3\% from 98.4\% to 95.5\%. 
In contrast, the best model on ImageNet \cite{pnasnet} sees an 11\% drop from 83\% to 72\% in top-1 accuracy and a 6\% drop from 96\% to 90\% in top-5 accuracy.
So the top-1 drop on ImageNet is larger than what we observed on CIFAR-10.

To put these accuracy numbers into perspective, we note that the best model in the ILSVRC\footnote{ILSVRC is the ImageNet Large Scale Visual Recognition Challenge \cite{RDSKSMHKKBBL15}.} 2013 competition achieved 89\% top-5 accuracy, and the best model from ILSVRC 2014 achieved 93\% top-5 accuracy.
So the 6\% drop in \mbox{top-5} accuracy from the 2018 state-of-the-art corresponds to approximately five years of progress in a very active period of machine learning research.

\paragraph{Few Changes in the Relative Order.} 
When sorting the models in order of their original and new accuracy, there are few changes in the respective rankings.
Models with comparable original accuracy tend to see a similar decrease in performance.
In fact, Figure \ref{fig:intro_plot} shows that the original accuracy is highly predictive of the new accuracy and that the relationship can be summarized well with a linear function.
On CIFAR-10, the new accuracy of a model is approximately given by the following formula:
\[
    \accnew \; = \; 1.69 \cdot \accorig - 72.7\% \; .
\]
On ImageNet, the top-1 accuracy of a model is given by 
\[
    \accnew \; = \; 1.11 \cdot \accorig - 20.2\% \; .
\]
Computing a 95\% confidence interval from 100,000 bootstrap samples gives $[1.63, 1.76]$ for the slope and $[-78.6, -67.5]$ for the offset on CIFAR-10, and $[1.07, 1.19]$ and $[-26.0, -17.8]$ respectively for ImageNet.

On both datasets, the slope of the linear fit is \emph{greater} \mbox{than 1.}
So models with higher original accuracy see a smaller drop on the new test sets.
In other words, model robustness \emph{improves} with increasing accuracy. 
This effect is less pronounced on ImageNet (slope 1.1) than on CIFAR-10 (slope 1.7).
In contrast to a scenario with strong adaptive overfitting, neither dataset sees diminishing returns in accuracy scores when going from the original to the new test sets.

\subsection{Experiments to Test Follow-Up Hypotheses}
\label{sec:imagenet_explaining_the_gap}
Since the drop from original to new accuracies is concerningly large, we investigated multiple hypotheses for explaining this drop.
Appendices \ref{apx:explain_gap_cifar} and \ref{apx:imagenetfollowup} list a range of follow-up experiments we conducted, e.g., re-tuning hyperparameters, training on part of our new test set, or performing cross-validation.
However, none of these effects can explain the size of the drop.
We conjecture that the accuracy drops stem from small variations in the human annotation process.
As we will see in the next section, the resulting changes in the test sets can significantly affect model accuracies.

\section{Understanding the Impact of Data Cleaning on ImageNet}
\label{sec:imagenet_details}
A crucial aspect of ImageNet is the use of MTurk.
There is a broad range of design choices for the MTurk tasks and how the resulting annotations determine the final dataset.
To better understand the impact of these design choices, we assembled three different test sets for ImageNet.
All of these test sets consist of images from the same Flickr candidate pool, are correctly labeled, and selected by more than 70\% of the MTurk workers on average.
Nevertheless, the resulting model accuracies vary by 14\%.
To put these numbers in context, we first describe our MTurk annotation pipeline.
\vspace{\negspaceow}
\paragraph{MTurk Tasks.}
We designed our MTurk tasks and user interface to closely resemble those originally used for ImageNet. 
As in ImageNet, each MTurk task contained a grid of 48 candidate images for a given target class.
The task description was derived from the original ImageNet instructions and included the definition of the target class with a link to a corresponding Wikipedia page.  
We asked the MTurk workers to select images belonging to the target class regardless of ``occlusions, other objects, and clutter or text in the scene'' and to avoid drawings or paintings (both as in ImageNet).
Appendix \ref{apx:mturk_ui} shows a screenshot of our UI and a screenshot of the original UI for comparison.

For quality control, we embedded at least six randomly selected images from the original validation set in each MTurk task (three from the same class, three from a class that is nearby in the WordNet hierarchy).
These images appeared in random locations of the image grid for each task.
In total, we collected sufficient MTurk annotations so that we have at least 20 annotated validation images for each class.

The main outcome of the MTurk tasks is a \emph{selection frequency} for each image, i.e., what fraction of MTurk workers selected the image in a task for its target class.
We recruited at least ten MTurk workers for each task (and hence for each image), which is similar to ImageNet.
Since each task contained original validation images, we could also estimate how often images from the original dataset were selected by our MTurk workers.

\vspace{\negspaceow}
\paragraph{Sampling Strategies. } In order to understand how the MTurk selection frequency affects the model accuracies, we explored three sampling strategies.
\begin{itemize}
  \item \textbf{\datasetb:} First, we estimated the selection frequency distribution for each class from the annotated original validation images.
    We then sampled ten images from our candidate pool for each class according to these class-specific distributions (see Appendix \ref{sec:imagenetsampling} for details).
  \item \textbf{\dataseta:} For each class, we sampled ten images with selection frequency at least 0.7.
\item \textbf{\datasetc:} For each class, we chose the ten images with highest selection frequency.
\end{itemize}
In order to minimize labeling errors, we manually reviewed each dataset and removed incorrect images.
The average selection frequencies of the three final datasets range from 0.93 for \datasetc{} over 0.85 for \dataseta{} to 0.73 for \datasetb.
For comparison, the original validation set has an average selection frequency of 0.71 in our experiments.
Hence all three of our new test sets have higher selection frequencies than the original ImageNet validation set.
In the preceding sections, we presented results on \datasetb{} for ImageNet since it is closest to the validation set in terms of selection frequencies.

\vspace{\negspaceow}
\paragraph{Results.} Table \ref{tab:sampling_results_summary} shows that the MTurk selection frequency has significant impact on both top-1 and top-5 accuracy.
In particular, \datasetc{} has the highest average MTurk selection frequency and sees a small \emph{increase} of about 2\% in both average top-1 and top-5 accuracy compared to the original validation set.
This is in stark contrast to \datasetb{}, which has the lowest average selection frequency and exhibits a significant drop of 12\% and 8\%, respectively.
The \dataseta{} dataset is in the middle and sees a small decrease of 3\% in top-1 and 1\% in top-5 accuracy.

In total, going from \datasetc{} to \datasetb{} decreases the accuracies by about 14\% (top-1) and 10\% (top-5).
For comparison, note that after excluding AlexNet (and the SqueezeNet models tuned to match AlexNet \cite{squeezenet}), the range of accuracies spanned by all remaining convolutional networks is roughly 14\% (top-1) and 8\% (top-5).
So the variation in accuracy caused by the three sampling strategies is larger than the variation in accuracy among all post-AlexNet models we tested.

\begin{table*}[tb!]
  \centering
  \rowcolors{2}{white}{gray!15}
  \begin{tabular}{C{3cm} C{3.75cm} C{3.75cm} C{3.75cm} }
   \toprule
   \textbf{Sampling Strategy} & \textbf{Average MTurk Selection Freq.} & \textbf{Average Top-1 Accuracy Change} & \textbf{Average Top-5 Accuracy Change} \\
   \midrule
    \datasetb{} & 0.73	& -11.8\% &	-8.2\% \\
    \dataseta{} & 0.85 & -3.2\% &	-1.2\% \\
    \datasetc{} & 0.93	& +2.1\% &	+1.8\% \\
 \bottomrule
  \end{tabular}
  \caption{Impact of the three sampling strategies for our ImageNet test sets.
    The table shows the average MTurk selection frequency in the resulting datasets and the average changes in model accuracy compared to the original validation set.
    We refer the reader to Section \ref{sec:imagenet_details} for a description of the three sampling strategies.
    All three test sets have an average selection frequency of more than 0.7, yet the model accuracies still vary widely.
  For comparison, the original ImageNet validation set has an average selection frequency of 0.71 in our MTurk experiments.
    The changes in average accuracy span 14\% and 10\% in top-1 and top-5, respectively.
    This shows that details of the sampling strategy have large influence on the resulting accuracies.
  }
  \label{tab:sampling_results_summary}
\end{table*}

\begin{figure*}[ht!]
  \centering
\iftoggle{isicml}{
  \begin{subfigure}{0.39\textwidth}
    \includegraphics[width=\linewidth]{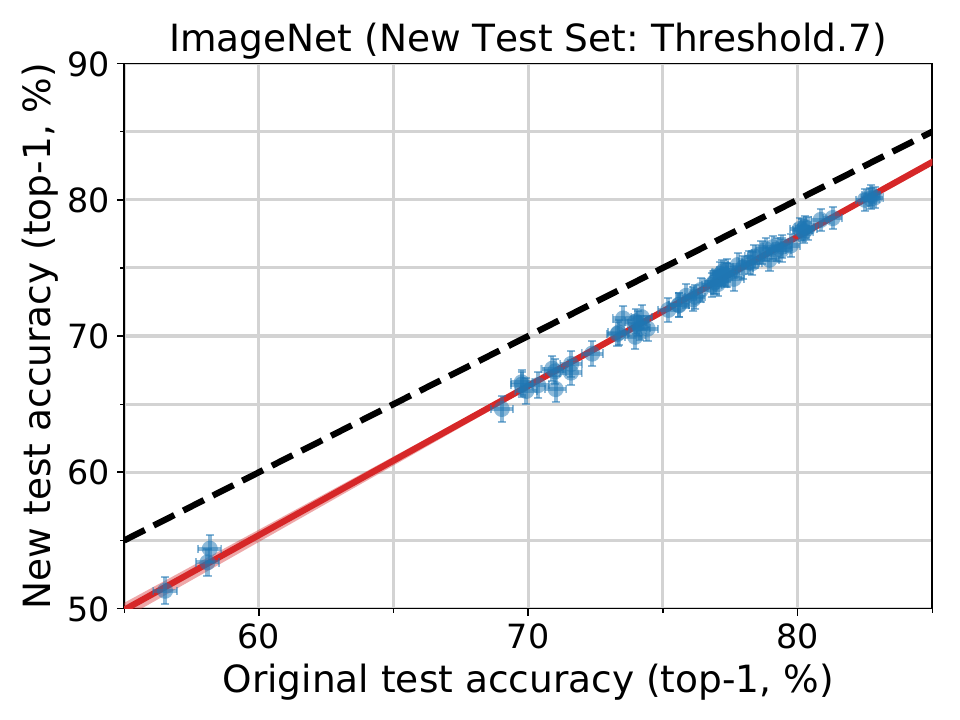}
  \end{subfigure}
  \begin{subfigure}{0.39\textwidth}
    \includegraphics[width=\linewidth]{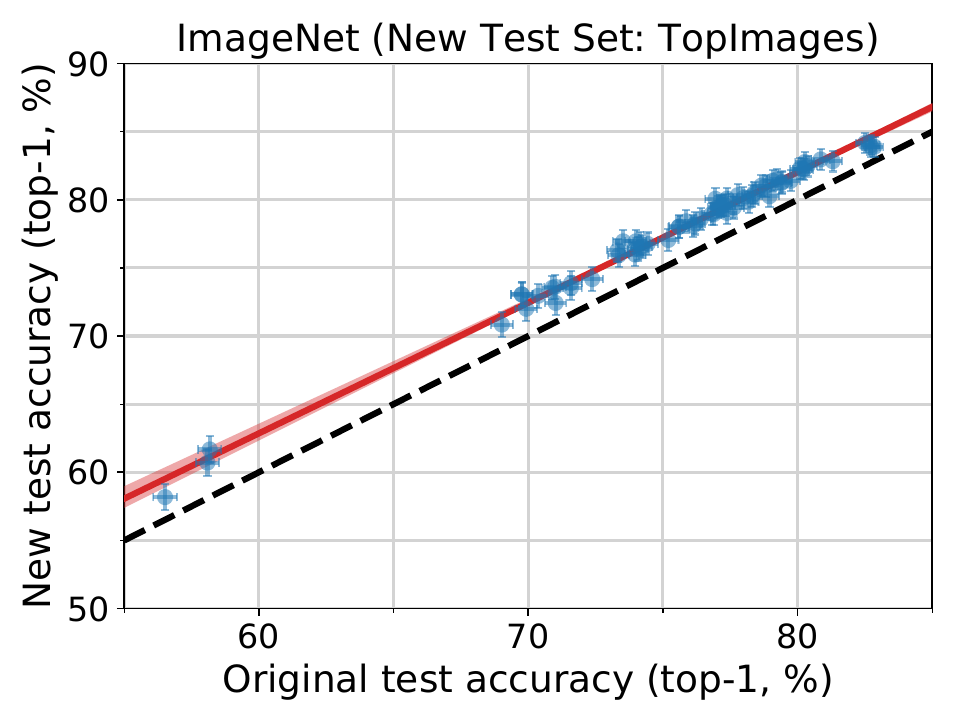}
  \end{subfigure}
  \begin{subfigure}{0.21\textwidth}
    \includegraphics[width=\linewidth]{figures/imagenet_ac_plot_separate_legend_vertical.pdf}
  \end{subfigure}
  \vspace{-.3cm}
}{
  \begin{subfigure}{0.48\textwidth}
    \includegraphics[width=\linewidth]{figures/imagenet_ac_plot_a_without_legend.pdf}
  \end{subfigure}
  \hfill
  \begin{subfigure}{0.48\textwidth}
    \includegraphics[width=\linewidth]{figures/imagenet_ac_plot_c_without_legend.pdf}
  \end{subfigure}
  \begin{subfigure}{\textwidth}
    \vspace{-.15cm}
    \centering
    \includegraphics[width=.75\linewidth]{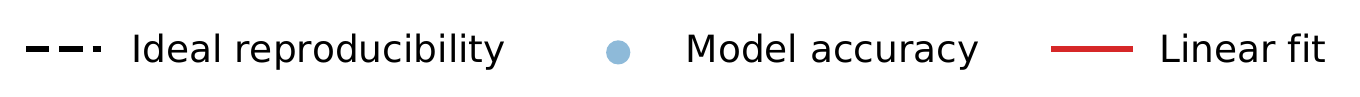}
  \end{subfigure}
  \vspace{-.6cm}
}
  \caption{Model accuracy on the original ImageNet validation set vs.\ accuracy on two variants of our new test set.
    We refer the reader to Section \ref{sec:imagenet_details} for a description of these test sets.
    Each data point corresponds to one model in our testbed (shown with 95\% Clopper-Pearson confidence intervals).
    On \dataseta{}, the model accuracies are 3\% lower than on the original test set.
    On \datasetc{}, which contains the images most frequently selected by MTurk workers, the models perform 2\% \emph{better} than on the original test set.
    The accuracies on both datasets closely follow a linear function, similar to \datasetb{} in Figure \ref{fig:intro_plot}.
    The red shaded region is a 95\% confidence region for the linear fit from 100,000 bootstrap samples.\\[-.7cm]}
  \label{fig:imagenet_a_and_c_top1}
\end{figure*}

Figure \ref{fig:imagenet_a_and_c_top1} plots the new vs.\ original top-1 accuracies on \dataseta{} and \datasetc{}, similar to Figure \ref{fig:intro_plot} for \datasetb{} before.
For easy comparison of top-1 and top-5 accuracy plots on all three datasets, we refer the reader to Figure \ref{fig:intro_plot} in Appendix \ref{sec:imagenettable}.
All three plots show a good linear fit.

\section{Discussion}
\label{sec:discussion}
\newlength{\negspacerw}
\iftoggle{isicml}{
\setlength{\negspacerw}{-0.2cm}
}{
\setlength{\negspacerw}{-0.0cm}
}

\iftoggle{isicml}{
Due to space constraints, we defer a discussion of related work to Appendix \ref{app:related}.
Furthermore, Appendix \ref{app:linfit} contains a theoretical model for the accurate linear fit observed in Figures \ref{fig:intro_plot} and \ref{fig:imagenet_a_and_c_top1}.
Here, we return to the main question from Section \ref{sec:formal}: \emph{What causes the accuracy drops?}}
{We now return to the main question from Section \ref{sec:formal}: \emph{What causes the accuracy drops?}}
As before, we distinguish between two possible mechanisms. \vspace{\negspacerw}
\subsection{Adaptivity Gap}
In its prototypical form, \emph{adaptive} overfitting would manifest itself in diminishing returns observed on the new test set (see Section \ref{sec:formal_multiple}).
However, we do not observe this pattern on either CIFAR-10 or ImageNet.
On both datasets, the slope of the linear fit is \emph{greater} than 1, i.e., each point of accuracy improvement on the original test set translates to more than 1\% on the new test set.
This is the opposite of the standard overfitting scenario.
So at least on CIFAR-10 and ImageNet, multiple years of competitive test set adaptivity did not lead to diminishing accuracy numbers.

While our experiments rule out the most dangerous form of adaptive overfitting, we remark that they do not exclude all variants.
For instance, it could be that any test set adaptivity leads to a roughly constant drop in accuracy.
Then all models are affected equally and we would see no diminishing returns since later models could still be better.
Testing for this form of adaptive overfitting likely requires a new test set that is truly i.i.d.\ and not the result of a separate data collection effort.
Finding a suitable dataset for such an experiment is an interesting direction for future research.

The lack of adaptive overfitting contradicts conventional wisdom in machine learning.
We now describe two mechanisms that could have prevented adaptive overfitting:
\paragraph{The Ladder Mechanism.} Blum and Hardt introduced the Ladder algorithm to protect machine learning competitions against adaptive overfitting \cite{BH15}.
The core idea is that constrained interaction with the test set can allow a large number of model evaluations to succeed, even if the models are chosen adaptively.
Due to the natural form of their algorithm, the authors point out that it can also be seen as a mechanism that the machine learning community \emph{implicitly} follows.
\paragraph{Limited Model Class.} Adaptivity is only a problem if we can choose among models for which the test set accuracy differs significantly from the population accuracy.
Importantly, this argument does not rely on the number of \emph{all} possible models (e.g., all parameter settings of a neural network), but only on those models that could actually be evaluated on the test set.
For instance, the standard deep learning workflow only produces models trained with SGD-style algorithms on a fixed training set, and requires that the models achieve high training accuracy (otherwise we would not consider the corresponding hyperparameters).
Hence the number of different models arising from the current methodology may be small enough so that uniform convergence holds.

Our experiments offer little evidence for favoring one explanation over the other.
One observation is that the convolutional networks shared many errors on CIFAR-10, which could be an indicator that the models are rather similar.
But to gain a deeper understanding into adaptive overfitting, it is likely necessary to gather further data from more machine learning benchmarks, especially in scenarios where adaptive overfitting \emph{does} occur naturally. 

\vspace{\negspacerw}
\subsection{Distribution Gap}
The lack of diminishing returns in our experiments points towards the distribution gap as the primary reason for the accuracy drops.
Moreover, our results on ImageNet show that changes in the sampling strategy can indeed affect model accuracies by a large amount, even if the data source and other parts of the dataset creation process stay the same. 

So in spite of our efforts to match the original dataset creation process, the distribution gap is still our leading hypothesis for the accuracy drops.
This demonstrates that it is surprisingly hard to accurately replicate the distribution of current image classification datasets.
The main difficulty likely is the subjective nature of the human annotation step.
There are many parameters that can affect the quality of human labels such as the annotator population (MTurk vs.\ students, qualifications, location \& time, etc.), the exact task format, and compensation.
Moreover, there are no exact definitions for many classes in ImageNet (e.g., see Appendix \ref{app:ambiguous_imagenet}).
Understanding these aspects in more detail is an important direction for designing future datasets that contain challenging images while still being labeled correctly.

The difficulty of clearly defining the data distribution, combined with the brittle behavior of the tested models, calls into question whether the black-box and i.i.d.\ framework of learning can produce reliable classifiers.
Our analysis of selection frequencies in Figure \ref{fig:rainbow_plot} (Appendix \ref{sec:rainbow}) shows that we could create a new test set with even lower model accuracies.
The images in this hypothetical dataset would still be correct, from Flickr, and selected by more than half of the MTurk labelers on average.
So in spite of the impressive accuracy scores on the original validation set, current ImageNet models still have difficulty generalizing from ``easy'' to ``hard'' images.

\iftoggle{isicml}{}{
  \subsection{A Model for the Linear Fit}
  \label{sec:probitmodel}
Finally, we briefly comment on the striking linear relationship between original and new test accuracies that we observe in all our experiments (for instance, see Figure \ref{fig:intro_plot} in the introduction or Figures \ref{fig:imagenet_plotpage} and \ref{fig:imagenet_probit_plotpage} in the appendix).
To illustrate how this phenomenon could arise, we present a simple data model where  a small modification of the data distribution can lead to significant changes in accuracy, yet the relative order of models is preserved as a linear relationship.
We emphasize that this model should not be seen as the true explanation.
Instead, we hope it can inform future experiments that explore natural variations in test distributions.

First, as we describe in Appendix~\ref{app:imagenetresults}, we find that we achieve better fits to our data under a \emph{probit scaling} of the accuracies.
Over a wide range from 21\% to 83\% (all models in our ImageNet testbed), the accuracies on the new test set, $\alpha_{\mathrm{new}}$, are related to the accuracies on the original test set, $\alpha_{\mathrm{orig}}$, by the relationship
\[
	\Phi^{-1}(\alpha_{\mathrm{new}}) \; = \;  u \cdot \Phi^{-1}(\alpha_{\mathrm{orig}})+v
\]
where $\Phi$ is the Gaussian CDF, and $u$ and $v$ are scalars. The probit scale is in a sense more natural than a linear scale as the accuracy numbers are probabilities. When we plot accuracies on a probit scale in Figures \ref{fig:linear_vs_probit} and \ref{fig:imagenet_probit_plotpage}, we effectively visualize $\Phi^{-1}(\alpha)$ instead of $\alpha$.

We now provide a simple plausible model where the original and new accuracies are related linearly on a probit scale. Assume that every example $i$ has a scalar ``difficulty'' $\tau_i \in \R$ that quantifies how easy it is to classify.
Further assume the probability of a model $j$ correctly classifying an image with difficulty $\tau$ is given by an increasing function $\zeta_j(\tau)$.
We show that for restricted classes of difficulty functions $\zeta_j$, we find a linear relationship between average accuracies after distribution shifts.

To be specific, we focus on the following parameterization. Assume the difficulty distribution of images in a test set follows a normal distribution with mean $\mu$ and variance $\sigma^2$. Further assume that
\[
  \zeta_j(\tau) \; = \; \Phi(s_j - \tau) \; ,
\]
where $\Phi: \R \rightarrow (0, 1)$ is the CDF of a standard normal distribution, and $s_j$ is the ``skill'' of model $j$.
Models with higher skill have higher classification accuracy, and images with higher difficulty lead to smaller classification accuracy.
Again, the choice of $\Phi$ here is somewhat arbitrary: any sigmoidal function that maps $(-\infty, +\infty)$ to $(0, 1)$ is plausible.
But using the Gaussian CDF yields a simple calculation illustrating the linear phenomenon.

Using the above notation, the accuracy $\alpha_{j, \mu,\sigma}$ of a model $j$ on a test set with difficulty mean $\mu$ and variance $\sigma$ is then given by
\[
  \alpha_{j, \mu, \sigma} \; = \; \E_{\tau \sim \N(\mu, \sigma)} \left[ \Phi(s_j - \tau) \right] \; .
\]
We can expand the CDF into an expectation and combine the two expectations by utilizing the fact that a linear combination of two Gaussians is again Gaussian.
This yields:
\[
  \alpha_{j, \mu, \sigma} \; = \; \Phi\left( \frac{s_j - \mu}{\sqrt{\sigma^2 + 1}} \right) \; .
\]
On a probit scale, the quantities we plot are given by
\[
  \tilde{\alpha}_{j, \mu, \sigma}  \; = \; \Phi^{-1}(\alpha_{j, \mu, \sigma}) \; = \; \frac{s_j - \mu}{\sqrt{\sigma^2 + 1}}  \; .
\]

Next, we consider the case where we have multiple models and two test sets with difficulty parameters $\mu_k$ and $\sigma_k$ respectively for $k \in \{1, 2\}$.
Then $\tilde{\alpha}_{j, 2}$, the probit-scaled accuracy on the second test set, is a linear function of the accuracy on the first test set, $\tilde{\alpha}_{j, 1}$:
\[
  \tilde{\alpha}_{j, 2} \; = \; u \cdot \tilde{\alpha}_{j, 1} + v \; ,
\]
with
\begin{align*}
  u \; = \; \frac{\sqrt{\sigma^2_1 + 1}}{\sqrt{\sigma^2_2 + 1}} ~~~\mbox{and} ~~~v \; = \; \frac{\mu_1 - \mu_2}{\sqrt{\sigma^2_2 + 1}} \; .
\end{align*}
Hence, we see that the Gaussian difficulty model above yields a linear relationship between original and new test accuracy in the probit domain. While the Gaussian assumptions here made the calculations simple, a variety of different simple classes of $\zeta_j$ will give rise to the same linear relationship between the accuracies on two different test sets.

  \section{Related Work}
  We now briefly discuss related threads in machine learning.
To the best of our knowledge, there are no reproducibility experiments directly comparable to ours in the literature.

\paragraph{Dataset Biases.}
The computer vision community has a rich history of creating new datasets and discussing their relative merits, e.g., \cite{caltech101,lotushill,PBEFHLMSRTWZZ06,TE11,pascalvoc,imagenet,RDSKSMHKKBBL15,mscoco}.
The paper closest to ours is \cite{TE11}, which studies dataset biases by measuring how models trained on one dataset generalize to other datasets.
The main difference to our work is that the authors test generalization across \emph{different} datasets, where larger changes in the distribution (and hence larger drops in accuracy) are expected.
In contrast, our experiments explicitly attempt to reproduce the original data distribution and demonstrate that even small variations arising in this process can lead to significant accuracy drops.
Moreover, \cite{TE11} do not test on previously unseen data, so their experiments cannot rule out adaptive overfitting.

\paragraph{Transfer Learning From ImageNet.}
\citet{KSL18} study how well accuracy on ImageNet transfers to other image classification datasets.
An important difference from both our work and \cite{TE11} is that the the ImageNet models are re-trained on the target datasets.
The authors find that better ImageNet models usually perform better on the target dataset as well.
Similar to \cite{TE11}, these experiments cannot rule out adaptive overfitting since the authors do not use new data.
Moreover, the experiments do not measure accuracy drops due to small variations in the data generating process since the models are evaluated on a different task with an explicit adaptation step.
Interestingly, the authors also find an approximately linear relationship between ImageNet and transfer accuracy.

\paragraph{Adversarial Examples.}
While adversarial examples \cite{intriguing,biggio2017wild} also show that existing models are brittle, the perturbations have to be finely tuned since models are much more robust to random perturbations.
In contrast, our results demonstrate that even small, benign variations in the data sampling process can already lead to a significant accuracy drop without an adversary.

A natural question is whether adversarially robust models are also more robust to the distribution shifts observed in our work.
As a first data point, we tested the common $\ell_\infty$-robustness baseline from \cite{madry2017towards} for CIFAR-10.
Interestingly, the accuracy numbers of this model fall almost exactly on the linear fit given by the other models in our testbed.
Hence $\ell_\infty$-robustness does not seem to offer benefits for the distribution shift arising from our reproducibility experiment.
However, we note that more forms of adversarial robustness such as spatial transformations or color space changes have been studied \cite{rotations,semanticadversarial,xiao2018spatially,Fawzi2015ManitestAC,Kanbak2018GeometricRO}.
Testing these variants is an interesting direction for future work.

\paragraph{Non-Adversarial Image Perturbations.}
Recent work also explores less adversarial changes to the input, e.g., \cite{Geirhos2018,hendrycks2018benchmarking}.
In these papers, the authors modify the ImageNet validation set via well-specified perturbations such as Gaussian noise, a fixed rotation, or adding a synthetic snow-like pattern.
Standard ImageNet models then achieve significantly lower accuracy on the perturbed examples than on the unmodified validation set.
While this is an interesting test of robustness, the mechanism underlying the accuracy drops is significantly different from our work.
The aforementioned papers rely on an intentional, clearly-visible, and well-defined perturbation of existing validation images.
Moreover, some of the interventions are quite different from the ImageNet validation set (e.g., ImageNet contains few images of falling snow).
In contrast, our experiments use new images and match the distribution of the existing validation set as closely as possible.
Hence it is unclear what properties of our new images cause the accuracy drops.

}

\section{Conclusion \& Future Work}
\label{sec:conclusion}
The expansive growth of machine learning rests on the aspiration to deploy trained systems in a variety of challenging environments.
Common examples include autonomous vehicles, content moderation, and medicine.
In order to use machine learning in these areas responsibly, it is important that we can both train models with sufficient generalization abilities, and also reliably measure their performance.
As our results show, these goals still pose significant hurdles even in a benign environment.

\iftoggle{isicml}{Our experiments are only a first step in addressing this reliability challenge.
One important question is whether other machine learning tasks are also resilient to adaptive overfitting, but similarly brittle under natural variations in the data.
Another direction is developing methods for more comprehensive yet still realistic evaluations of machine learning systems.
Of course, the overarching goal is to develop learning algorithms that generalize reliably.
While this is often a vague goal, our new test sets offer a well-defined instantiation of this challenge that is beyond the reach of current methods.
Generalizing from our ``easy'' to slightly ``harder'' images will hopefully serve as a starting point towards a future generation of more reliable models.}{

Our experiments are only one step in addressing this reliability challenge.
There are multiple promising avenues for future work:

\paragraph{Understanding Adaptive Overfitting.}
In contrast to conventional wisdom, our experiments show that there are no diminishing returns associated with test set re-use on CIFAR-10 and ImageNet.
A more nuanced understanding of this phenomenon will require studying whether other machine learning problems are also resilient to adaptive overfitting.
For instance, one direction would be to conduct similar reproducibility experiments on tasks in natural language processing, or to analyze data from competition platforms such as Kaggle and CodaLab.\footnote{\url{https://www.kaggle.com/competitions} and \url{https://competitions.codalab.org/}.}

\paragraph{Characterizing the Distribution Gap.}
Why do the classification models in our testbed perform worse on the new test sets?
The selection frequency experiments in Section \ref{sec:imagenet_details} suggest that images selected less frequently by the MTurk workers are harder for the models.
However, the selection frequency analysis does not explain what aspects of the images make them harder.
Candidate hypotheses are object size, special filters applied to the images (black \& white, sepia, etc.), or unusual object poses.
Exploring whether there is a succinct description of the difference between the original and new test distributions is an interesting direction for future work.

\paragraph{Learning More Robust Models.} An overarching goal is to make classification models more robust to small variations in the data.
If the change from the original to our new test sets can be characterized accurately, techniques such as data augmentation or robust optimization may be able to close some of the accuracy gap.
Otherwise, one possible approach could be to gather particularly hard examples from Flickr or other data sources and expand the training set this way.
However, it may also be necessary to develop entirely novel approaches to image classification.

\paragraph{Measuring Human Accuracy.} One interesting question is whether our new test sets are also harder for humans.
As a first step in this direction, our human accuracy experiment on CIFAR-10 (see Appendix \ref{app:cifar_human}) shows that average human performance is not affected significantly by the distribution shift between the original and new images that are most difficult for the models.
This suggests that the images are only harder for the trained models and not for humans.
But a more comprehensive understanding of the human baseline will require additional human accuracy experiments on both CIFAR-10 and ImageNet.

\paragraph{Building Further Test Sets.} The dominant paradigm in machine learning is to evaluate the performance of a classification model on a single test set per benchmark.
Our results suggest that this is not comprehensive enough to characterize the reliability of current models.
To understand their generalization abilities more accurately, new test data from various sources may be needed.
One intriguing question here is whether accuracy on other test sets will also follow a linear function of the original test accuracy.

\paragraph{Suggestions For Future Datasets.} We found that it is surprisingly difficult to create a new test set that matches the distribution of an existing dataset.
Based on our experience with this process, we provide some suggestions for improving machine learning datasets in the future:

\begin{itemize}
\item \textbf{Code release.} It is hard to fully document the dataset creation process in a paper because it involves a long list of design choices.
Hence it would be beneficial for reproducibility efforts if future dataset papers released not only the data but also all code used to create the datasets.

\item \textbf{Annotator quality.} Our results show that changes in the human annotation process can have significant impact on the difficulty of the resulting datasets.
To better understand the quality of human annotations, it would be valuable if authors conducted a standardized test with their annotators (e.g., classifying a common set of images) and included the results in the description of the dataset.
Moreover, building variants of the test set with different annotation processes could also shed light on the variability arising from this data cleaning step.

\item \textbf{``Super hold-out''.} Having access to data from the original CIFAR-10 and ImageNet data collection could have clarified the cause of the accuracy drops in our experiments.
  By keeping an additional test set hidden for multiple years, future benchmarks could explicitly test for adaptive overfitting after a certain time period.

\item \textbf{Simpler tasks for humans.} The large number of classes and fine distinctions between them make ImageNet a particularly hard problem for humans without special training.
While classifying a large variety of objects with fine-grained distinctions is an important research goal, there are also trade-offs.
Often it becomes necessary to rely on images with high annotator agreement to ensure correct labels, which in turn leads to bias by excluding harder images.
Moreover, the large number of classes causes difficulties when characterizing human performance.
So an alternative approach for a dataset could be to choose a task that is simpler for humans in terms of class structure (fewer classes, clear class boundaries), but contains a larger variety of object poses, lighting conditions, occlusions, image corruptions, etc.

\item \textbf{Test sets with expert annotations.} Compared to building a full training set, a test set requires a smaller number of human annotations.
This makes it possible to employ a separate labeling process for the test set that relies on more costly expert annotations.
While this violates the assumption that train and test splits are i.i.d.\ from the same distribution, the expert labels can also increase quality both in terms of correctness and example diversity.
\end{itemize}

Finally, we emphasize that our recommendations here should \emph{not} be seen as flaws in CIFAR-10 or ImageNet.
Both datasets were assembled in the late 2000s for an accuracy regime that is very different from the state-of-the-art now.
Over the past decade, especially ImageNet has successfully guided the field to increasingly better models, thereby clearly demonstrating the immense value of this dataset.
But as models have increased in accuracy and our reliability expectations have grown accordingly, it is now time to revisit how we create and utilize datasets in machine learning.}

\section*{Acknowledgements}
We would like to thank Tudor Achim, Alex Berg, Orianna DeMasi, Jia Deng, Alexei Efros, David Fouhey, Moritz Hardt, Piotr Indyk, Esther Rolf, and Olga Russakovsky for helpful discussions while working on this paper.
Moritz Hardt has been particularly helpful in all stages of this project and -- among other invaluable advice -- suggested the title of this paper and a precursor to the data model in Section \ref{sec:probitmodel}.
We also thank the participants of our human accuracy experiment in Appendix \ref{app:cifar_human} (whose names we keep anonymous following our IRB protocol).

This research was generously supported in part by ONR awards N00014-17-1-2191, N00014-17-1-2401, and N00014-18-1-2833, the DARPA Assured Autonomy (FA8750-18-C-0101) and Lagrange (W911NF-16-1-0552) programs, an Amazon AWS AI Research Award, and a gift from Microsoft Research.
In addition, LS was supported by a Google PhD fellowship and a Microsoft Research Fellowship at the Simons Institute for the Theory of Computing.

\bibliographystyle{plainnat}
\bibliography{main}

\iftoggle{showappendix}{

\appendix

\newpage
\clearpage

\etocdepthtag.toc{mtappendix}

\section{Overview of the Appendix}
The following appendix contains a detailed description of our experiments on CIFAR-10 and ImageNet.
To ease navigation, we first provide a table of contents for the appendix.

\vspace{-.5cm}

\etocsettagdepth{mtsection}{none}
\etocsettagdepth{mtappendix}{subsubsection}
\tableofcontents

\section{Details of the CIFAR-10 Experiments}
\label{app:cifar}
We first present our reproducibility experiment for the CIFAR-10 image classification dataset \cite{krizhevsky2009learning}.
There are multiple reasons why CIFAR-10 is an important example for measuring how well current models generalize to unseen data.
\begin{itemize}
\item CIFAR-10 is one of the most widely used datasets in machine learning and serves as a test ground for many image classification methods.
  A concrete measure of popularity is the fact that CIFAR-10 was the second most common dataset in NIPS 2017 (after MNIST) \cite{hamnerpopular}.
\item The dataset creation process for CIFAR-10 is transparent and well documented \cite{krizhevsky2009learning}.
  Importantly, CIFAR-10 draws from the larger Tiny Images repository that has more fine-grained labels than the ten CIFAR-10 classes \cite{tinyimages}.
This enables us to minimize various forms of distribution shift between the original and new test set.
\item CIFAR-10 poses a difficult enough problem so that the dataset is still the subject of active research (e.g., see \cite{cutout,shakeshake,shakedrop,nas,RAHL18,autoaugment}).
Moreover, there is a wide range of classification models that achieve significantly different accuracy scores.
Since code for these models has been published in various open source repositories, they can be treated as independent of our new test set.
\end{itemize}

Compared to ImageNet, CIFAR-10 is significantly smaller both in the number of images and in the size of each image.
This makes it easier to conduct various follow-up experiments that require training new classification models.
Moreover, the smaller size of CIFAR-10 also means that the dataset has been accessible to more researchers for a longer time.
Hence it is plausible that CIFAR-10 experienced more test set adaptivity than ImageNet, where it is much more costly to tune hyperparameters.

Before we describe how we created our new test set, we briefly review relevant background on CIFAR-10 and Tiny Images.

\paragraph{Tiny Images.} 
The dataset contains 80 million RGB color images with resolution 32 $\times$ 32 pixels and was released in 2007 \cite{tinyimages}.
The images are organized by roughly 75,000 \emph{keywords} that correspond to the non-abstract nouns from the WordNet database \cite{wordnet}
Each keyword was entered into multiple Internet search engines to collect roughly 1,000 to 2,500 images per keyword.
It is important to note that Tiny Images is a fairly noisy dataset.
Many of the images filed under a certain keyword do not clearly (or not at all) correspond to the respective keyword.

\paragraph{CIFAR-10.} The CIFAR-10 dataset was created as a cleanly labeled subset of Tiny Images for experiments with multi-layer networks.
To this end, the researchers assembled a dataset consisting of ten classes with 6,000 images per class, which was published in 2009 \cite{krizhevsky2009learning}.
These classes are \keyword{airplane}, \keyword{automobile}, \keyword{bird}, \keyword{cat}, \keyword{deer}, \keyword{dog}, \keyword{frog}, \keyword{horse}, \keyword{ship},  and \keyword{truck}.
The standard train / test split is class-balanced and contains 50,000 training images and 10,000 test images.

The CIFAR-10 creation process is well-documented \cite{krizhevsky2009learning}.
First, the researchers assembled a set of relevant keywords for each class by using the hyponym relations in WordNet \cite{wordnet} (for instance, ``Chihuahua'' is a hyponym of ``dog'').
Since directly using the corresponding images from Tiny Images would not give a high quality dataset, the researchers paid student annotators to label the images from Tiny Images.
The labeler instructions can be found in Appendix C of \cite{krizhevsky2009learning} and include a set of specific guidelines (e.g., an image should not contain two object of the corresponding class).
The researchers then verified the labels of the images selected by the annotators and removed near-duplicates from the dataset via an $\ell_2$ nearest neighbor search.

\subsection{Dataset Creation Methodology}
\label{apx:cifar10_dataset_creation}
Our overall goal was to create a new test set that is as close as possible to being drawn from the same distribution as the original CIFAR-10 dataset.
One crucial aspect here is that the CIFAR-10 dataset did not exhaust any of the Tiny Image keywords it is drawn from.
So by collecting new images from the same keywords as CIFAR-10, our new test set can match the sub-class distribution of the original dataset.

\paragraph{Understanding the Sub-Class Distribution.} As the first step, we determined the Tiny Image keyword for every image in the CIFAR-10 dataset.
A simple nearest-neighbor search sufficed since every image in CIFAR-10 had an exact duplicate ($\ell_2$-distance $0$) in Tiny Images.
Based on this information, we then assembled a list of the 25 most common keywords for each class.
We decided on 25 keywords per class since the 250 total keywords make up more than 95\% of CIFAR-10.
Moreover, we wanted to avoid accidentally creating a harder dataset with infrequent keywords that the classifiers had little incentive to learn based on the original CIFAR-10 dataset.

The keyword distribution can be found in Appendix \ref{apx:v4_keywords}.
Inspecting this list reveals the importance of matching the sub-class distribution.
For instance, the most common keyword in the \airplane{} class is \keyword{stealth\_bomber} and not a more common civilian type of airplane.
In addition, the third most common keyword for the \airplane{} class is \keyword{stealth\_fighter}.
Both types of planes are highly distinctive.
There are more examples where certain sub-classes are considerably different.
For instance, trucks from the keyword \keyword{fire\_truck} are mostly red, which is quite different from pictures for \keyword{dump\_truck} or other keywords.

\paragraph{Collecting New Images.} After determining the keywords, we collected corresponding images.
To simulate the student / researcher split in the original CIFAR-10 collection procedure, we introduced a similar split among two authors of this paper.
Author A took the role of the original student annotators and selected new suitable images for the 250 keywords.
In order to ensure a close match between the original and new images for each keyword, we built a user interface that allowed Author A to first look through existing CIFAR-10 images for a given keyword and then select new candidates from the remaining pictures in Tiny Images.
Author A followed the labeling guidelines in the original instruction sheet \cite{krizhevsky2009learning}.
The number of images Author A selected per keyword was so that our final dataset would contain between 2,000 and 4,000 images.
We decided on 2,000 images as a target number for two reasons:
\begin{itemize}
\item While the original CIFAR-10 test set contains 10,000 images, a test set of size 2,000 is already sufficient for a fairly small confidence interval.
In particular, a conservative confidence interval (Clopper-Pearson at confidence level 95\%) for accuracy 90\% has size about $\pm 1\%$ with $n =$ 2,000 (to be precise, $[88.6\%, \, 91.3\%]$).
Since we considered a potential discrepancy between original and new test accuracy only interesting if it is significantly larger than 1\%, we decided that a new test set of size 2,000 was large enough for our study.
\item As with very infrequent keywords, our goal was to avoid accidentally creating a harder test set.
Since some of the Tiny Image keywords have only a limited supply of remaining adequate images, we decided that a smaller target size for the new dataset would reduce bias to include images of more questionable difficulty.
\end{itemize}
After Author A had selected a set of about 9,000 candidate images, Author B adopted the role of the researchers in the original CIFAR-10 dataset creation process.
In particular, Author B reviewed all candidate images and removed images that were unclear to Author B or did not conform to the labeling instructions in their opinion (some of the criteria are subjective).
In the process, a small number of keywords did not have enough images remaining to reach the $n =$ 2,000 threshold.
Author B then notified Author A about the respective keywords and Author A selected a further set of images for these keywords.
In this process, there was only one keyword where Author A had to carefully examine all available images in Tiny Images.
This keyword was \keyword{alley\_cat} and comprises less than 0.3\% of the overall CIFAR-10 dataset.

\paragraph{Final Assembly.}
After collecting a sufficient number of high-quality images for each keyword, we sampled a random subset from our pruned candidate set.
The sampling procedure was such that the keyword-level distribution of our new dataset matches the
keyword-level distribution of CIFAR-10 (see Appendix \ref{apx:v4_keywords}).
In the final stage, we again proceeded similar to the original CIFAR-10 dataset creation process and used $\ell_2$-nearest neighbors to filter out near duplicates.
In particular, we removed near-duplicates within our new dataset and also images that had a near duplicate in the original CIFAR-10 dataset (train or test).
The latter aspect is particularly important since our reproducibility study is only interesting if we evaluate on truly unseen data.
Hence we manually reviewed the top-10 nearest neighbors for each image in our new test set.
After removing near-duplicates in our dataset, we re-sampled the respective keywords until this process converged to our final dataset.

Figure \ref{fig:original_test} shows a random subset of images from the original and our new test set.

We remark that we did not run any classifiers on our new dataset during the data collection phase of our study.
In order to ensure that the new data does not depend on the existing classifiers, it is important to strictly separate the data collection phase from the following evaluation phase.

\begin{figure*}
   \centering
   \newlength{\imagedim}
   \setlength{\imagedim}{1.2cm}
   \newlength{\imagexspacing}
   \setlength{\imagexspacing}{0.1cm}
   \newlength{\imageyspacing}
   \setlength{\imageyspacing}{0.1cm}
   \begin{subfigure}[t]{0.49\textwidth}
   \centering
   \begin{tikzpicture}
   \tikzstyle{img}=[inner sep=0pt,outer sep=0pt];
   \node [img] (image0) {\includegraphics[width=\imagedim]{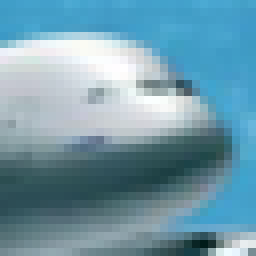}};
   \node [img,anchor=west,at=(image0.east),xshift=\imagexspacing] (image1)
   {\includegraphics[width=\imagedim]{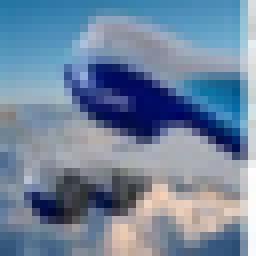}};
   \node [img,anchor=west,at=(image1.east),xshift=\imagexspacing] (image2)
   {\includegraphics[width=\imagedim]{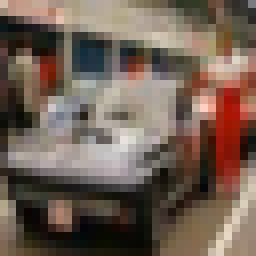}};
   \node [img,anchor=west,at=(image2.east),xshift=\imagexspacing] (image3)
   {\includegraphics[width=\imagedim]{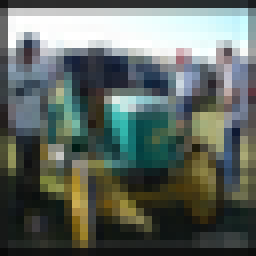}};
   \node [img,anchor=west,at=(image3.east),xshift=\imagexspacing] (image4)
   {\includegraphics[width=\imagedim]{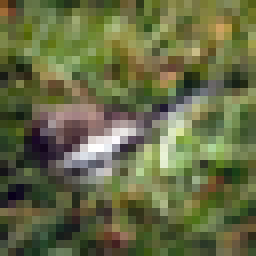}};
   \node [img,anchor=north,at=(image0.south),yshift=-\imageyspacing] (image5)
   {\includegraphics[width=\imagedim]{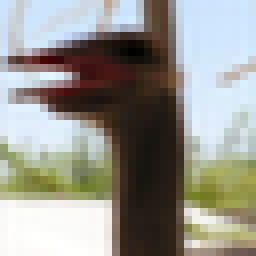}};
   \node [img,anchor=west,at=(image5.east),xshift=\imagexspacing] (image6)
   {\includegraphics[width=\imagedim]{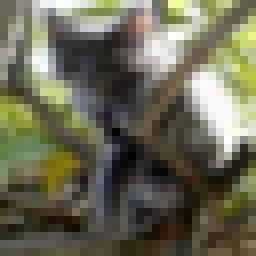}};
   \node [img,anchor=west,at=(image6.east),xshift=\imagexspacing] (image7)
   {\includegraphics[width=\imagedim]{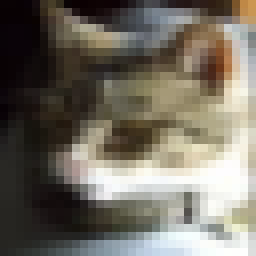}};
   \node [img,anchor=west,at=(image7.east),xshift=\imagexspacing] (image8)
   {\includegraphics[width=\imagedim]{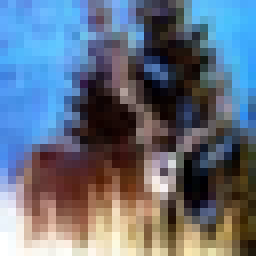}};
   \node [img,anchor=west,at=(image8.east),xshift=\imagexspacing] (image9)
   {\includegraphics[width=\imagedim]{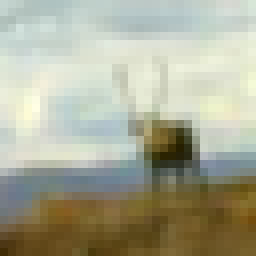}};
   \node [img,anchor=north,at=(image5.south),yshift=-\imageyspacing] (image10)
   {\includegraphics[width=\imagedim]{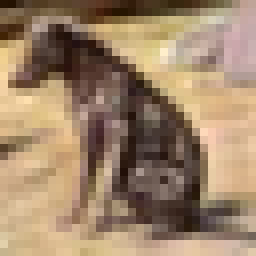}};
   \node [img,anchor=west,at=(image10.east),xshift=\imagexspacing] (image11)
   {\includegraphics[width=\imagedim]{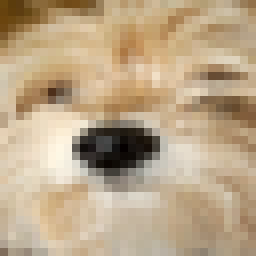}};
   \node [img,anchor=west,at=(image11.east),xshift=\imagexspacing] (image12)
   {\includegraphics[width=\imagedim]{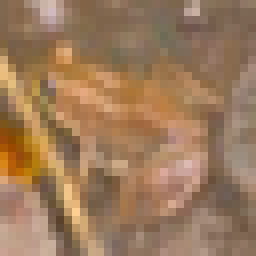}};
   \node [img,anchor=west,at=(image12.east),xshift=\imagexspacing] (image13)
   {\includegraphics[width=\imagedim]{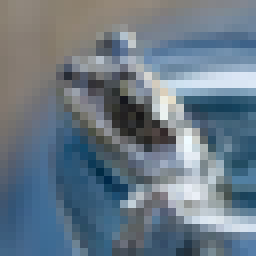}};
   \node [img,anchor=west,at=(image13.east),xshift=\imagexspacing] (image14)
   {\includegraphics[width=\imagedim]{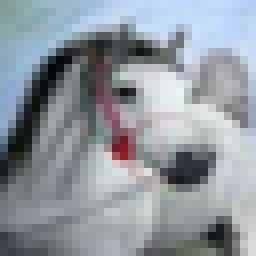}};
   \node [img,anchor=north,at=(image10.south),yshift=-\imageyspacing] (image15)
   {\includegraphics[width=\imagedim]{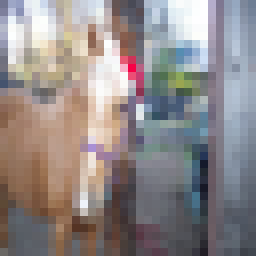}};
   \node [img,anchor=west,at=(image15.east),xshift=\imagexspacing] (image16)
   {\includegraphics[width=\imagedim]{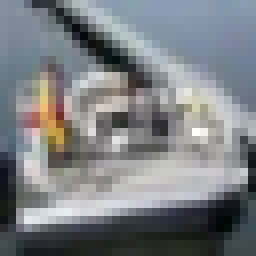}};
   \node [img,anchor=west,at=(image16.east),xshift=\imagexspacing] (image17)
   {\includegraphics[width=\imagedim]{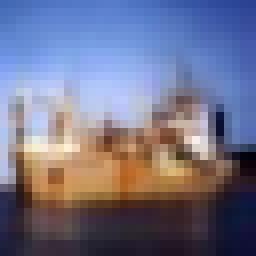}};
   \node [img,anchor=west,at=(image17.east),xshift=\imagexspacing] (image18)
   {\includegraphics[width=\imagedim]{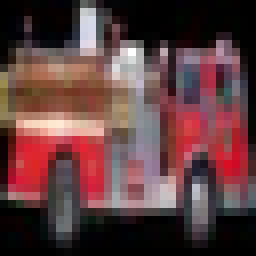}};
   \node [img,anchor=west,at=(image18.east),xshift=\imagexspacing] (image19)
   {\includegraphics[width=\imagedim]{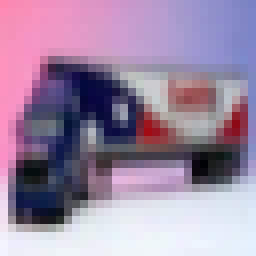}};
   \end{tikzpicture}
     \caption{Test set A}
     \label{fig:new_test}
   \end{subfigure}
   \begin{subfigure}[t]{0.49\textwidth}
   \centering
   \begin{tikzpicture}
   \tikzstyle{img}=[inner sep=0pt,outer sep=0pt];
   \node [img] (image0) {\includegraphics[width=\imagedim]{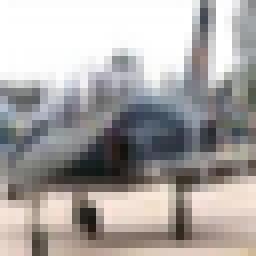}};
   \node [img,anchor=west,at=(image0.east),xshift=\imagexspacing] (image1)
   {\includegraphics[width=\imagedim]{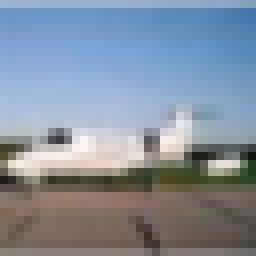}};
   \node [img,anchor=west,at=(image1.east),xshift=\imagexspacing] (image2)
   {\includegraphics[width=\imagedim]{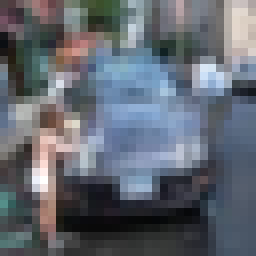}};
   \node [img,anchor=west,at=(image2.east),xshift=\imagexspacing] (image3)
   {\includegraphics[width=\imagedim]{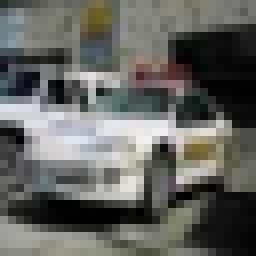}};
   \node [img,anchor=west,at=(image3.east),xshift=\imagexspacing] (image4)
   {\includegraphics[width=\imagedim]{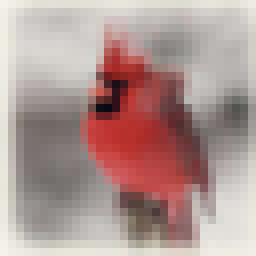}};
   \node [img,anchor=north,at=(image0.south),yshift=-\imageyspacing] (image5)
   {\includegraphics[width=\imagedim]{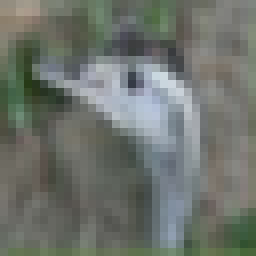}};
   \node [img,anchor=west,at=(image5.east),xshift=\imagexspacing] (image6)
   {\includegraphics[width=\imagedim]{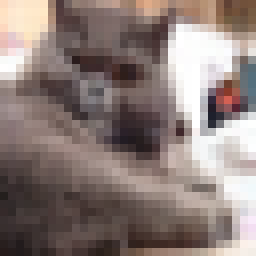}};
   \node [img,anchor=west,at=(image6.east),xshift=\imagexspacing] (image7)
   {\includegraphics[width=\imagedim]{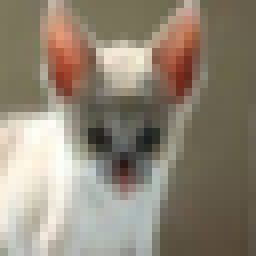}};
   \node [img,anchor=west,at=(image7.east),xshift=\imagexspacing] (image8)
   {\includegraphics[width=\imagedim]{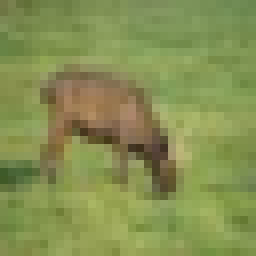}};
   \node [img,anchor=west,at=(image8.east),xshift=\imagexspacing] (image9)
   {\includegraphics[width=\imagedim]{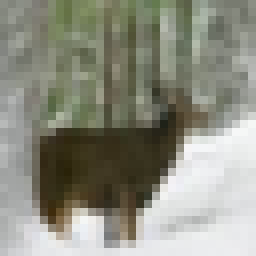}};
   \node [img,anchor=north,at=(image5.south),yshift=-\imageyspacing] (image10)
   {\includegraphics[width=\imagedim]{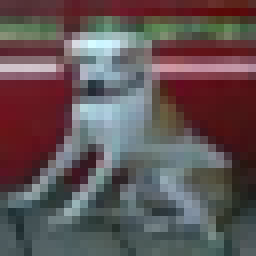}};
   \node [img,anchor=west,at=(image10.east),xshift=\imagexspacing] (image11)
   {\includegraphics[width=\imagedim]{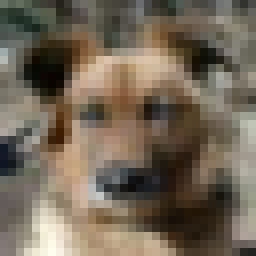}};
   \node [img,anchor=west,at=(image11.east),xshift=\imagexspacing] (image12)
   {\includegraphics[width=\imagedim]{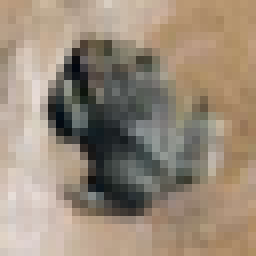}};
   \node [img,anchor=west,at=(image12.east),xshift=\imagexspacing] (image13)
   {\includegraphics[width=\imagedim]{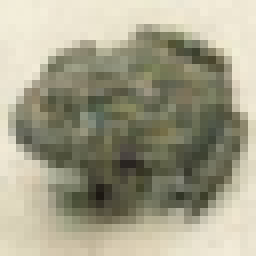}};
   \node [img,anchor=west,at=(image13.east),xshift=\imagexspacing] (image14)
   {\includegraphics[width=\imagedim]{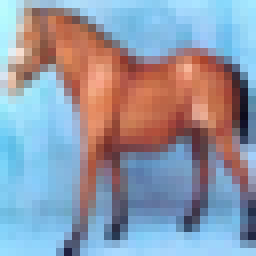}};
   \node [img,anchor=north,at=(image10.south),yshift=-\imageyspacing] (image15)
   {\includegraphics[width=\imagedim]{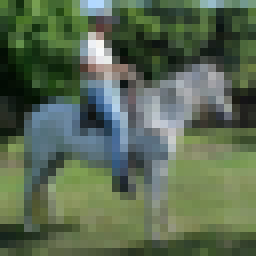}};
   \node [img,anchor=west,at=(image15.east),xshift=\imagexspacing] (image16)
   {\includegraphics[width=\imagedim]{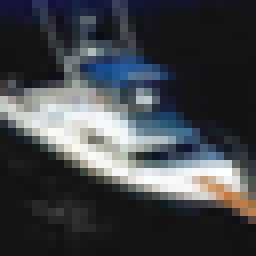}};
   \node [img,anchor=west,at=(image16.east),xshift=\imagexspacing] (image17)
   {\includegraphics[width=\imagedim]{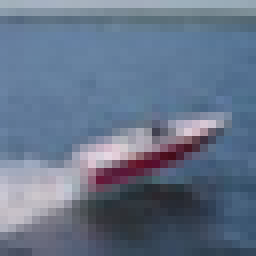}};
   \node [img,anchor=west,at=(image17.east),xshift=\imagexspacing] (image18)
   {\includegraphics[width=\imagedim]{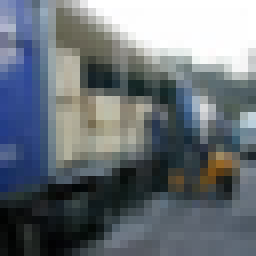}};
   \node [img,anchor=west,at=(image18.east),xshift=\imagexspacing] (image19)
   {\includegraphics[width=\imagedim]{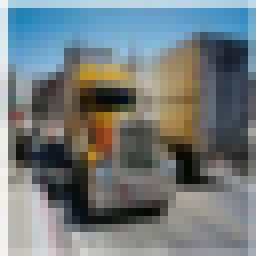}};
   \end{tikzpicture}
     \caption{Test set B}
     \label{fig:original_test}
   \end{subfigure}
   \caption{Randomly selected images from the original and new CIFAR-10 test sets.
   Each grid contains two images for each of the ten classes.
   The following footnote reveals which of the two grids corresponds to the new test set.\protect \footnotemark}
   \label{fig:testexamples}
   \end{figure*}

\subsection{Follow-up Hypotheses}
\label{apx:explain_gap_cifar}
Since the gap between original and new accuracy is concerningly large, we investigated multiple hypotheses for explaining this gap.

\subsubsection{Statistical Error}
A first natural guess is that the gap is simply due to statistical fluctuations.
But as noted before, the sample size of our new test set is large enough so that a 95\% confidence interval has size about $\pm 1.2\%$.
Since a 95\% confidence interval for the original CIFAR-10 test accuracy is even smaller (roughly $\pm 0.6\%$ for 90\% classification accuracy and $\pm 0.3\%$ for 97\% classification accuracy), we can rule out statistical error as the main explanation.

\subsubsection{Differences in Near-Duplicate Removal}
As mentioned in Section \ref{apx:cifar10_dataset_creation}, the final step of both the original CIFAR-10 and our dataset creation procedure is to remove near-duplicates.
While removing near-duplicates between our new test set and the original CIFAR-10 dataset, we noticed that the original test set contained images that we would have ruled out as near-duplicates.
A large number of near-duplicates between CIFAR-10 train and test, combined with our more stringent near-duplicate removal, could explain some of the accuracy drop.
Indeed, we found about 800 images in the original CIFAR-10 test set that we would classify as near-duplicates (8\% of the entire test set).
Moreover, most classifiers have accuracy between 99\% and 100\% on these near-duplicates (recall that most models achieve 100\% training error).
However, the following calculation shows that the near-duplicates can explain at most 1\% of the observed difference.

For concreteness, we consider a model with 93\% original test set accuracy such as a common VGG or ResNet architecture.
Let $\acc_{\text{true}}$ be the ``true'' accuracy of the model on test images that are not near-duplicates, and let $\acc_{\text{nd}}$ be the accuracy on near-duplicates.
Then for 8\% near-duplicates, the overall accuracy is given by
\[
  \acc \; = \; 0.92 \cdot \acc_{\text{true}} + 0.08 \cdot \acc_{\text{nd}} \; .
\]
Using $\acc = 0.93$, $\acc_{\text{nd}} = 1.0$, and solving for $\acc_{\text{true}}$ then yields $\acc_{\text{true}} \approx 0.924$.
So the accuracy on original test images that are not near-duplicates is indeed lower, but only by a small amount (0.6\%).
This is in contrast to the 8\% - 9\% accuracy drop that VGG and ResNet models with 93\% original accuracy see in our experiments.

For completeness, we describe our process for finding near duplicates in detail.
For every test image, we visually inspected the top-10 nearest neighbors in both $\ell_2$-distance and the SSIM (structural similarity) metric.
We compared the original test set to the CIFAR-10 training set, and our new test set to both the original training and test sets. 
We consider an image pair as near-duplicates if both images have the same object in the same pose. 
We include images that have different zoom, color scale, stretch in the horizontal or vertical direction, or small shifts in vertical or horizontal position.  
If the object was rotated or in a different pose, we did not include it as a near-duplicate.

\subsubsection{Hyperparameter Tuning}
Another conjecture is that we can recover some of the missing accuracy by re-tuning hyperparameters of a model.
To this end, we performed a grid search over multiple parameters of a VGG model.
We selected three standard hyperparameters known to strongly influence test set performance: initial learning rate, dropout, and weight decay. 
The \model{vgg16\_keras} architecture uses different amounts of dropout across different layers of the network, so we chose to tune a multiplicative scaling factor for the amount of dropout.
This keeps the ratio of dropout across different layers constant. 
\footnotetext{Test Set A is the new test set and Test Set B is the original test set.}

We initialized a hyperparameter configuration from values tuned to the original test set (learning
rate $0.1$, dropout ratio $1$, weight decay $\num{5e-4}$), and performed a grid search across the following values:
\begin{itemize}
\item Learning rate in $ \{0.0125, 0.025, 0.05, 0.1, 0.2, 0.4, 0.8\}$.
\item Dropout ratio in $\{0.5, 0.75, 1, 1.25, 1.75\}$.
\item Weight decay in $\{\num{5e-5}$, $\num{1e-4}$, $\num{5e-4}$, $\num{1e-3}$, $\num{5e-3} \}$.
\end{itemize}

We ensured that the best performance was never at an extreme point of any range we tested for an individual hyperparameter. 
Overall, we did not find a hyperparameter setting with a significantly better accuracy on the new test set (the biggest improvement was from 85.3\% to 85.8\%).

\subsubsection{Visually Inspecting Hard Images}
It is also possible that we accidentally created a more difficult test set by including a set of ``harder'' images.  
To explore this question, we visually inspected the set of images that most models incorrectly classified.
Figure \ref{fig:hardtest} in Appendix \ref{app:cifar_hard_images} shows examples of the hard images in our new test set that no model correctly classified.
We find that all the new images are valid images that are recognizable to humans.

\subsubsection{Human Accuracy Comparison}
\label{app:cifar_human}
The visual inspection of hard images in the previous section is one way to compare the original and new test sets.
However, our conclusion may be biased since we have created the new test set ourselves.
To compare the relative hardness of the two test sets  more objectively, we also conducted a small experiment to measure human accuray on the two test sets.\footnote{Use of this data was permitted by the Berkelely Committee for Protection of Human Subjects (CPHS).}
The goal of the experiment was to measure if human accuracy is significantly different on the original and new test sets.

Since we conjectured that our new test set included particularly hard images, we focused our experiment on the approximately 5\% hardest images in both test sets.
Here, ``hardness'' is defined by how many models correctly classified an image.
After rounding to include all images that were classified by the same number of models, we obtained 500 images from the original test set and 115 images from our new test set.

We recruited nine graduate students from three different research groups in the Electrical Engineering \& Computer Sciences Department at UC Berkeley.
We wrote a simple user interface that allowed the participants to label images with one of the ten CIFAR-10 classes.
To ensure that the participants did not know which dataset an image came from, we presented the images in random order.

Table \ref{tab:cifar10_human} shows the results of our experiment.
We find that four participants performed better on the original test set and five participants were better on our new test set.
The average difference is -0.8\%, i.e., the participants do not see a drop in average accuracy on this subset of original and new test images.
This suggests that our new test set is not significantly harder for humans.
However, we remark that our results here should only be seen as a preliminary study.
Understanding human accuracy on CIFAR-10 in more detail will require further experiments.

\begin{table*}[ht!]
\centering
\rowcolors{3}{white}{gray!15}
\begin{tabular}{cccc}
\toprule
& \multicolumn{3}{c}{Human Accuracy ($\%$)} \\
\midrule
& Original Test Set & New Test Set & Gap \\
\midrule
Participant 1 & 85 {\footnotesize \textcolor{gray}{[81.6, 88.0]}} & 83 {\footnotesize \textcolor{gray}{[74.2, 89.8]}} & 2 \\
Participant 2 & 83 {\footnotesize \textcolor{gray}{[79.4, 86.2]}} & 81 {\footnotesize \textcolor{gray}{[71.9, 88.2]}} & 2 \\
Participant 3 & 82 {\footnotesize \textcolor{gray}{[78.3, 85.3]}} & 78 {\footnotesize \textcolor{gray}{[68.6, 85.7]}} & 4 \\
Participant 4 & 79 {\footnotesize \textcolor{gray}{[75.2, 82.5]}} & 84 {\footnotesize \textcolor{gray}{[75.3, 90.6]}} & -5 \\
Participant 5 & 76 {\footnotesize \textcolor{gray}{[72.0, 79.7]}} & 77 {\footnotesize \textcolor{gray}{[67.5, 84.8]}} & -1 \\
Participant 6 & 75 {\footnotesize \textcolor{gray}{[71.0, 78.7]}} & 73 {\footnotesize \textcolor{gray}{[63.2, 81.4]}} & 2 \\
Participant 7 & 74 {\footnotesize \textcolor{gray}{[69.9, 77.8]}} & 79 {\footnotesize \textcolor{gray}{[69.7, 86.5]}} & -5 \\
Participant 8 & 74 {\footnotesize \textcolor{gray}{[69.9, 77.8]}} & 76 {\footnotesize \textcolor{gray}{[66.4, 84.0]}} & -2 \\
Participant 9 & 67 {\footnotesize \textcolor{gray}{[62.7, 71.1]}} & 71 {\footnotesize \textcolor{gray}{[61.1, 79.6]}} & -4 \\
\bottomrule
\end{tabular}
\caption{Human accuracy on the ``hardest'' images in the original and our new CIFAR-10 test set.
  We ordered the images by number of incorrect classifications from models in our testbed and then selected the top 5\% images from the original and new test set (500 images from the original test set, 115 images from our new test set).
  The results show that on average humans do not see a drop in accuracy on this subset of images.
}
\label{tab:cifar10_human}
\end{table*}

\subsubsection{Training on Part of Our New Test Set}
If our new test set distribution is significantly different from the original CIFAR-10 distribution, retraining on part of our new test set (plus the original training data) may improve the accuracy on the held-out fraction of our new test set.  

We conducted this experiment by randomly drawing a class-balanced split containing about 1,000 images from the new test set.
We then added these images to the full CIFAR-10 training set and retrained the \model{vgg16\_keras} model.
After training, we tested the model on the remaining half of the new test set.
We repeated this experiment twice with different randomly selected splits from our test set, obtaining accuracies of 85.1\% and 85.4\% (compared to 84.9\% without the extra training data\footnote{This number is slightly lower than the accuracy of \model{vgg16\_keras} on our new test set in Table \ref{tab:v4_results}, but still within the 95\% confidence interval $[83.6, 86.8]$. Hence we conjecture that the difference is due to the random fluctuation arising from randomly initializing the model.}).
This provides evidence that there is no large distribution shift between our new test set and the original CIFAR-10 dataset, or that the model is unable to learn the modified distribution.

\subsubsection{Cross-validation}
Cross-validation can be a more reliable way of measuring a model's generalization ability than using only a single train / test split.
Hence we tested if cross-validation on the original CIFAR-10 dataset could predict a model's error on our new test set. 
We created cross-validation data by randomly dividing the training set into 5 class-balanced splits.
We then randomly shuffled together 4 out of the 5 training splits with the original test set.
The leftover held-out split from the training set then became the new test set.

We retrained the models \model{vgg\_15\_BN\_64}, \model{wide\_resnet\_28\_10}, and \model{shake\_shake\_64d\_cutout} on each of the 5 new datasets we created.  
The accuracies are reported in Table \ref{tab:cifar10_cross_validation}.
The accuracies on the cross-validation splits did not differ much from the accuracy on the original test set.
The variation among the cross-validation splits is significantly smaller than the drop on our new test set.

\begin{table*}[ht!]
\centering
\rowcolors{4}{white}{gray!15}
\begin{tabular}{cccc}
\toprule
& \multicolumn{3}{c}{Model Accuracy ($\%$)} \\
\midrule
Dataset & \model{vgg\_15\_BN\_64} & \model{wide\_resnet\_28\_10} & \model{shake\_shake\_64d\_cutout} \\
\midrule
Original Test Set & 
93.6 {\footnotesize \textcolor{gray}{[93.1, 94.1]}} & 
95.7 {\footnotesize \textcolor{gray}{[95.3, 96.1]}} & 
97.1 {\footnotesize \textcolor{gray}{[96.8, 97.4]}} \\
\midrule
Split 1 & 
93.9 {\footnotesize \textcolor{gray}{[93.4, 94.3]}} & 
96.2 {\footnotesize \textcolor{gray}{[95.8, 96.6]}} &	
97.2 {\footnotesize \textcolor{gray}{[96.9, 97.5]}}\\
Split 2 & 
93.8 {\footnotesize \textcolor{gray}{[93.3, 94.3]}}	& 
96.0 {\footnotesize \textcolor{gray}{[95.6, 96.4]}}& 
97.3 {\footnotesize \textcolor{gray}{[97.0, 97.6]}}\\
Split 3 & 
94.0 {\footnotesize \textcolor{gray}{[93.5, 94.5]}}	& 
96.4 {\footnotesize \textcolor{gray}{[96.0, 96.8]}}& 
97.4 {\footnotesize \textcolor{gray}{[97.1, 97.7]}}\\
Split 4 & 
94.0 {\footnotesize \textcolor{gray}{[93.5, 94.5]}}	& 
96.2 {\footnotesize \textcolor{gray}{[95.8, 96.6]}}& 
97.4 {\footnotesize \textcolor{gray}{[97.1, 97.7]}}\\
Split 5 & 
93.5 {\footnotesize \textcolor{gray}{[93.0, 94.0]}}	& 
96.5 {\footnotesize \textcolor{gray}{[96.1, 96.9]}}& 
97.4 {\footnotesize \textcolor{gray}{[97.1, 97.7]}}\\
\midrule
New Test Set & 
84.9 {\footnotesize \textcolor{gray}{[83.2, 86.4]}} & 
89.7 {\footnotesize \textcolor{gray}{[88.3, 91.0]}} & 
93.0 {\footnotesize \textcolor{gray}{[91.8, 94.1]}} \\
\bottomrule
\end{tabular}
\caption{Model accuracies on cross-validation splits for the original CIFAR-10 data.
The difference in cross-validation accuracies is significantly smaller than the drop to the new test set.}
\label{tab:cifar10_cross_validation}
\end{table*}

\subsubsection{Training a Discriminator for Original vs.\ New Test Set}
Our main hypothesis for the accuracy drop is that small variations in the test set creation process suffice to significantly reduce a model's accuracy.
To test whether these variations could be detected by a convolutional network, we investigated whether a discriminator model could distinguish between the two test sets. 

We first created a training set consisting of $3,200$ images (1,600 from the original test set and 1,600 from our new test set) and a test set of $800$ images (consisting of 400 images from original and new test set each).
Each image had a binary label indicating whether it came from the original or new test set.
Additionally, we ensured that that both datasets were class balanced. 

We then trained \model{resnet\_32} and \model{resnet\_110} models for 160 epochs using a standard SGD optimizer to learn a binary classifier between the two datasets.
We conducted two variants of this experiment: in one variant, we traind the model from scratch.
In the other variant, we started with a model pre-trained on the regular CIFAR-10 classification task.

Our results are summarized in Table~\ref{tab:cifar_discrim}.
Overall we found that the resulting models could not discriminate well between the original and our new test set: the best accuracy we obtained is 53.1\%.

\begin{table*}[ht!]
\centering
\rowcolors{3}{gray!15}{white}
\begin{tabular}{c c c}
\toprule
Model & Discriminator Accuracy ($\%$)  & Discriminator Accuracy ($\%$)  \\
      & random initialization          & pre-trained \\
\midrule
\model{resnet\_32} & $50.1$ {\footnotesize \textcolor{gray}{[46.6, 53.6]}} & $52.9$ {\footnotesize \textcolor{gray}{[49.4, 56.4]}} \\
\model{resnet\_110} & $50.3$ {\footnotesize \textcolor{gray}{[46.7, 53.8]}} & $53.1$ {\footnotesize \textcolor{gray}{[49.6, 56.6]}} \\
\bottomrule
\end{tabular}
\caption{Accuracies for discriminator models trained to distinugish between the original and new CIFAR-10 test sets. 
The models were initialized either randomly or using a model pre-trained on the original CIFAR-10 dataset. 
Although the models performed slightly better than random chance, the confidence intervals (95\% Clopper Pearson) still overlap with 50\% accuracy. \label{tab:cifar_discrim}}
\end{table*}

\subsubsection{An Exactly Class-balanced Test Set}
The top 25 keywords of each class in CIFAR-10 capture approximately 95\% of the dataset.  
However, the remaining 5\% of the dataset are skewed towards the class \class{ship}. 
As a result, our new dataset was not exactly class-balanced and contained only 8\% images of class \class{ship} (as opposed to 10\% in the original test set).
 
To measure whether this imbalance affected the acccuracy scores, we created an exactly class-balanced version of our new test set with 2,000 images (200 per class).
In this version, we selected the top 50 keywords in each class and computed a fractional number of images for each keyword.
We then rounded these numbers so that images for keywords with the largest fractional part were added first.
The resulting model accuracies can be found in Table \ref{tab:v6_results} (Appendix \ref{apx:cifar10_model_accuracies_balanced}).
Models with lower original accuracies achieve a small accuracy improvement on the exactly class-balanced test set (around 0.3\%), but the accuracy drop of the best-performing model remains unchanged.

\subsection{Additional Figures, Tables, and Lists}
In this appendix we provide large figures etc.\ that did not fit into the preceding sections about our CIFAR-10 experiments.

\subsubsection{Keyword Distribution in CIFAR-10}
\label{apx:v4_keywords}
The sub-tables in Table \ref{tab:keywords} show the keyword distribution for each of the ten classes in the original CIFAR-10 test set and our new test set.

\captionsetup[subtable]{labelformat=empty}
\captionsetup[subtable]{position=top}
\begin{table*}[h!]
  \caption{Distribution of the top 25 keywords in each class for the new and original test set.}
  \label{tab:keywords}
  \begin{subtable}{0.3\linewidth}
     \centering
     \kwfontsize
     \begin{tabular}{L{4cm}R{1cm}R{1cm}}
\toprule
\multicolumn{3}{c}{\textbf{Frog}} \\
\midrule
{} &   New &  Original \\
\midrule
bufo\_bufo           & 0.64\% &     0.63\% \\
leopard\_frog        & 0.64\% &     0.64\% \\
bufo\_viridis        & 0.59\% &     0.57\% \\
rana\_temporaria     & 0.54\% &     0.53\% \\
bufo                & 0.49\% &     0.47\% \\
bufo\_americanus     & 0.49\% &     0.46\% \\
toad                & 0.49\% &     0.46\% \\
green\_frog          & 0.45\% &     0.44\% \\
rana\_catesbeiana    & 0.45\% &     0.43\% \\
bufo\_marinus        & 0.45\% &     0.43\% \\
bullfrog            & 0.45\% &     0.42\% \\
american\_toad       & 0.45\% &     0.43\% \\
frog                & 0.35\% &     0.35\% \\
rana\_pipiens        & 0.35\% &     0.32\% \\
toad\_frog           & 0.30\% &     0.30\% \\
spadefoot           & 0.30\% &     0.27\% \\
western\_toad        & 0.30\% &     0.26\% \\
grass\_frog          & 0.30\% &     0.27\% \\
pickerel\_frog       & 0.25\% &     0.24\% \\
spring\_frog         & 0.25\% &     0.22\% \\
rana\_clamitans      & 0.20\% &     0.20\% \\
natterjack          & 0.20\% &     0.17\% \\
crapaud             & 0.20\% &     0.18\% \\
bufo\_calamita       & 0.20\% &     0.18\% \\
alytes\_obstetricans & 0.20\% &     0.16\% \\
\bottomrule
\end{tabular}

     \label{tab:frog}
  \end{subtable}%
  \hspace*{10em}
  \begin{subtable}{0.3\linewidth}
     \centering
     \kwfontsize
     \begin{tabular}{L{4cm}R{1cm}R{1cm}}
\toprule
\multicolumn{3}{c}{\textbf{Cat}} \\
\midrule
{} &   New &  Original \\
\midrule
tabby\_cat         & 1.78\% &     1.78\% \\
tabby             & 1.53\% &     1.52\% \\
domestic\_cat      & 1.34\% &     1.33\% \\
cat               & 1.24\% &     1.25\% \\
house\_cat         & 0.79\% &     0.79\% \\
felis\_catus       & 0.69\% &     0.69\% \\
mouser            & 0.64\% &     0.63\% \\
felis\_domesticus  & 0.54\% &     0.50\% \\
true\_cat          & 0.49\% &     0.47\% \\
tomcat            & 0.49\% &     0.49\% \\
alley\_cat         & 0.30\% &     0.30\% \\
felis\_bengalensis & 0.15\% &     0.11\% \\
nougat            & 0.10\% &     0.05\% \\
gray              & 0.05\% &     0.03\% \\
manx\_cat          & 0.05\% &     0.04\% \\
fissiped          & 0.05\% &     0.03\% \\
persian\_cat       & 0.05\% &     0.03\% \\
puss              & 0.05\% &     0.05\% \\
catnap            & 0.05\% &     0.03\% \\
tiger\_cat         & 0.05\% &     0.03\% \\
black\_cat         & 0.05\% &     0.04\% \\
bedspread         & 0.00\% &     0.02\% \\
siamese\_cat       & 0.00\% &     0.02\% \\
tortoiseshell     & 0.00\% &     0.02\% \\
kitty-cat         & 0.00\% &     0.02\% \\
\bottomrule
\end{tabular}

     \label{tab:cat} 
  \end{subtable}
\end{table*}

\begin{table*}[h!]
  \begin{subtable}{0.3\linewidth}
     \centering
     \kwfontsize
     \begin{tabular}{L{4cm}R{1cm}R{1cm}}
\toprule
\multicolumn{3}{c}{\textbf{Dog}} \\
\midrule
\toprule
{} &   New &  Original \\
\midrule
pekingese            & 1.24\% &     1.22\% \\
maltese              & 0.94\% &     0.93\% \\
puppy                & 0.89\% &     0.87\% \\
chihuahua            & 0.84\% &     0.81\% \\
dog                  & 0.69\% &     0.67\% \\
pekinese             & 0.69\% &     0.66\% \\
toy\_spaniel          & 0.59\% &     0.60\% \\
mutt                 & 0.49\% &     0.47\% \\
mongrel              & 0.49\% &     0.49\% \\
maltese\_dog          & 0.45\% &     0.43\% \\
toy\_dog              & 0.40\% &     0.36\% \\
japanese\_spaniel     & 0.40\% &     0.38\% \\
blenheim\_spaniel     & 0.35\% &     0.35\% \\
english\_toy\_spaniel  & 0.35\% &     0.31\% \\
domestic\_dog         & 0.35\% &     0.32\% \\
peke                 & 0.30\% &     0.28\% \\
canis\_familiaris     & 0.30\% &     0.27\% \\
lapdog               & 0.30\% &     0.30\% \\
king\_charles\_spaniel & 0.20\% &     0.17\% \\
toy                  & 0.15\% &     0.13\% \\
feist                & 0.10\% &     0.06\% \\
pet                  & 0.10\% &     0.07\% \\
cavalier             & 0.10\% &     0.05\% \\
canine               & 0.05\% &     0.04\% \\
cur                  & 0.05\% &     0.04\% \\
\bottomrule
\end{tabular}

     \label{tab:dog}
  \end{subtable}
  \hspace*{10em}
  \begin{subtable}{0.3\linewidth}
     \centering
     \kwfontsize
     \begin{tabular}{L{4cm}R{1cm}R{1cm}}
\toprule
\multicolumn{3}{c}{\textbf{Deer}} \\
\midrule
{} &   New &  Original \\
\midrule
elk                 & 0.79\% &     0.77\% \\
capreolus\_capreolus & 0.74\% &     0.71\% \\
cervus\_elaphus      & 0.64\% &     0.61\% \\
fallow\_deer         & 0.64\% &     0.63\% \\
roe\_deer            & 0.59\% &     0.60\% \\
deer                & 0.59\% &     0.60\% \\
muntjac             & 0.54\% &     0.51\% \\
mule\_deer           & 0.54\% &     0.51\% \\
odocoileus\_hemionus & 0.49\% &     0.50\% \\
fawn                & 0.49\% &     0.49\% \\
alces\_alces         & 0.40\% &     0.36\% \\
wapiti              & 0.40\% &     0.36\% \\
american\_elk        & 0.40\% &     0.35\% \\
red\_deer            & 0.35\% &     0.33\% \\
moose               & 0.35\% &     0.35\% \\
rangifer\_caribou    & 0.25\% &     0.24\% \\
rangifer\_tarandus   & 0.25\% &     0.24\% \\
caribou             & 0.25\% &     0.23\% \\
sika                & 0.25\% &     0.22\% \\
woodland\_caribou    & 0.25\% &     0.21\% \\
dama\_dama           & 0.20\% &     0.19\% \\
cervus\_sika         & 0.20\% &     0.16\% \\
barking\_deer        & 0.20\% &     0.18\% \\
sambar              & 0.15\% &     0.15\% \\
stag                & 0.15\% &     0.13\% \\
\bottomrule
\end{tabular}

     \label{tab:deer}
  \end{subtable}
\end{table*} 

\begin{table*}  
  \begin{subtable}{0.3\linewidth}
     \centering
     \kwfontsize
     \begin{tabular}{L{4cm}R{1cm}R{1cm}}
\toprule
\multicolumn{3}{c}{\textbf{Bird}} \\
\midrule
{} &   New &  Original \\
\midrule
cassowary                & 0.89\% &     0.85\% \\
bird                     & 0.84\% &     0.84\% \\
wagtail                  & 0.74\% &     0.74\% \\
ostrich                  & 0.69\% &     0.68\% \\
struthio\_camelus         & 0.54\% &     0.51\% \\
sparrow                  & 0.54\% &     0.52\% \\
emu                      & 0.54\% &     0.51\% \\
pipit                    & 0.49\% &     0.47\% \\
passerine                & 0.49\% &     0.50\% \\
accentor                 & 0.49\% &     0.49\% \\
honey\_eater              & 0.40\% &     0.37\% \\
dunnock                  & 0.40\% &     0.37\% \\
alauda\_arvensis          & 0.30\% &     0.26\% \\
nandu                    & 0.30\% &     0.27\% \\
prunella\_modularis       & 0.30\% &     0.30\% \\
anthus\_pratensis         & 0.30\% &     0.28\% \\
finch                    & 0.25\% &     0.24\% \\
lark                     & 0.25\% &     0.20\% \\
meadow\_pipit             & 0.25\% &     0.20\% \\
rhea\_americana           & 0.25\% &     0.21\% \\
flightless\_bird          & 0.15\% &     0.10\% \\
emu\_novaehollandiae      & 0.15\% &     0.12\% \\
dromaius\_novaehollandiae & 0.15\% &     0.14\% \\
apteryx                  & 0.15\% &     0.10\% \\
flying\_bird              & 0.15\% &     0.13\% \\
\bottomrule
\end{tabular}

     \label{tab:bird}
  \end{subtable}
  \hspace*{10em}
  \begin{subtable}{0.3\linewidth}
     \centering
     \kwfontsize
     \begin{tabular}{L{4cm}R{1cm}R{1cm}}
\toprule
\multicolumn{3}{c}{\textbf{Ship}} \\
\midrule
{} &   New &  Original \\
\midrule
passenger\_ship   & 0.79\% &     0.78\% \\
boat             & 0.64\% &     0.64\% \\
cargo\_ship       & 0.40\% &     0.37\% \\
cargo\_vessel     & 0.40\% &     0.39\% \\
pontoon          & 0.35\% &     0.31\% \\
container\_ship   & 0.35\% &     0.31\% \\
speedboat        & 0.35\% &     0.32\% \\
freighter        & 0.35\% &     0.32\% \\
pilot\_boat       & 0.35\% &     0.31\% \\
ship             & 0.35\% &     0.31\% \\
cabin\_cruiser    & 0.30\% &     0.29\% \\
police\_boat      & 0.30\% &     0.25\% \\
sea\_boat         & 0.30\% &     0.29\% \\
oil\_tanker       & 0.30\% &     0.29\% \\
pleasure\_boat    & 0.25\% &     0.21\% \\
lightship        & 0.25\% &     0.22\% \\
powerboat        & 0.25\% &     0.25\% \\
guard\_boat       & 0.25\% &     0.20\% \\
dredger          & 0.25\% &     0.20\% \\
hospital\_ship    & 0.25\% &     0.21\% \\
banana\_boat      & 0.20\% &     0.19\% \\
merchant\_ship    & 0.20\% &     0.17\% \\
liberty\_ship     & 0.20\% &     0.15\% \\
container\_vessel & 0.20\% &     0.19\% \\
tanker           & 0.20\% &     0.18\% \\
\bottomrule
\end{tabular}

     \label{tab:ship}
  \end{subtable}
\end{table*}

\begin{table*}  
  \begin{subtable}{0.3\linewidth}
     \centering
     \kwfontsize
     \begin{tabular}{L{4cm}R{1cm}R{1cm}}
\toprule
\multicolumn{3}{c}{\textbf{Truck}} \\
\midrule
{} &   New &  Original \\
\midrule
dump\_truck          & 0.89\% &     0.89\% \\
trucking\_rig        & 0.79\% &     0.76\% \\
delivery\_truck      & 0.64\% &     0.61\% \\
truck               & 0.64\% &     0.65\% \\
tipper\_truck        & 0.64\% &     0.60\% \\
camion              & 0.59\% &     0.58\% \\
fire\_truck          & 0.59\% &     0.55\% \\
lorry               & 0.54\% &     0.53\% \\
garbage\_truck       & 0.54\% &     0.53\% \\
moving\_van          & 0.35\% &     0.32\% \\
tractor\_trailer     & 0.35\% &     0.34\% \\
tipper              & 0.35\% &     0.30\% \\
aerial\_ladder\_truck & 0.35\% &     0.34\% \\
ladder\_truck        & 0.30\% &     0.26\% \\
fire\_engine         & 0.30\% &     0.27\% \\
dumper              & 0.30\% &     0.28\% \\
trailer\_truck       & 0.30\% &     0.28\% \\
wrecker             & 0.30\% &     0.27\% \\
articulated\_lorry   & 0.25\% &     0.24\% \\
tipper\_lorry        & 0.25\% &     0.25\% \\
semi                & 0.20\% &     0.18\% \\
sound\_truck         & 0.15\% &     0.12\% \\
tow\_truck           & 0.15\% &     0.12\% \\
delivery\_van        & 0.15\% &     0.11\% \\
bookmobile          & 0.10\% &     0.10\% \\
\bottomrule
\end{tabular}

     \label{tab:truck}  
  \end{subtable}
  \hspace*{10em}
  \begin{subtable}{0.3\linewidth}
     \centering
     \kwfontsize
     \begin{tabular}{L{4cm}R{1cm}R{1cm}}
\toprule
\multicolumn{3}{c}{\textbf{Horse}} \\
\midrule
{} &   New &  Original \\
\midrule
arabian                 & 1.14\% &     1.12\% \\
lipizzan                & 1.04\% &     1.02\% \\
broodmare               & 0.99\% &     0.97\% \\
gelding                 & 0.74\% &     0.73\% \\
quarter\_horse           & 0.74\% &     0.72\% \\
stud\_mare               & 0.69\% &     0.69\% \\
lippizaner              & 0.54\% &     0.52\% \\
appaloosa               & 0.49\% &     0.45\% \\
lippizan                & 0.49\% &     0.46\% \\
dawn\_horse              & 0.45\% &     0.42\% \\
stallion                & 0.45\% &     0.43\% \\
tennessee\_walker        & 0.45\% &     0.45\% \\
tennessee\_walking\_horse & 0.40\% &     0.38\% \\
walking\_horse           & 0.30\% &     0.28\% \\
riding\_horse            & 0.20\% &     0.20\% \\
saddle\_horse            & 0.20\% &     0.18\% \\
female\_horse            & 0.15\% &     0.11\% \\
cow\_pony                & 0.15\% &     0.11\% \\
male\_horse              & 0.15\% &     0.14\% \\
buckskin                & 0.15\% &     0.13\% \\
horse                   & 0.10\% &     0.08\% \\
equine                  & 0.10\% &     0.08\% \\
quarter                 & 0.10\% &     0.07\% \\
cavalry\_horse           & 0.10\% &     0.09\% \\
thoroughbred            & 0.10\% &     0.06\% \\
\bottomrule
\end{tabular}

     \label{tab:horse}
  \end{subtable}
\end{table*}

\begin{table*}
  \begin{subtable}{0.3\linewidth}
     \centering
     \kwfontsize
     \begin{tabular}{L{4cm}R{1cm}R{1cm}}
\toprule
\multicolumn{3}{c}{\textbf{Airplane}} \\
\midrule
{} &   New &  Original \\
\midrule
stealth\_bomber       & 0.94\% &     0.92\% \\
airbus               & 0.89\% &     0.89\% \\
stealth\_fighter      & 0.79\% &     0.80\% \\
fighter\_aircraft     & 0.79\% &     0.76\% \\
biplane              & 0.74\% &     0.74\% \\
attack\_aircraft      & 0.69\% &     0.67\% \\
airliner             & 0.64\% &     0.61\% \\
jetliner             & 0.59\% &     0.56\% \\
monoplane            & 0.54\% &     0.55\% \\
twinjet              & 0.54\% &     0.52\% \\
dive\_bomber          & 0.54\% &     0.52\% \\
jumbo\_jet            & 0.49\% &     0.47\% \\
jumbojet             & 0.35\% &     0.35\% \\
propeller\_plane      & 0.30\% &     0.28\% \\
fighter              & 0.20\% &     0.20\% \\
plane                & 0.20\% &     0.15\% \\
amphibious\_aircraft  & 0.20\% &     0.20\% \\
multiengine\_airplane & 0.15\% &     0.14\% \\
seaplane             & 0.15\% &     0.14\% \\
floatplane           & 0.10\% &     0.05\% \\
multiengine\_plane    & 0.10\% &     0.06\% \\
reconnaissance\_plane & 0.10\% &     0.09\% \\
airplane             & 0.10\% &     0.08\% \\
tail                 & 0.10\% &     0.05\% \\
joint                & 0.05\% &     0.04\% \\
\bottomrule
\end{tabular}

     \label{tab:airplane}
  \end{subtable}
  \hspace*{10em}
  \begin{subtable}{0.3\linewidth}
     \centering
     \kwfontsize
     \begin{tabular}{L{4cm}R{1cm}R{1cm}}
\toprule
\multicolumn{3}{c}{\textbf{Automobile}} \\
\midrule
{} &   New &  Original \\
\midrule
coupe          & 1.29\% &     1.26\% \\
convertible    & 1.19\% &     1.18\% \\
station\_wagon  & 0.99\% &     0.98\% \\
automobile     & 0.89\% &     0.90\% \\
car            & 0.84\% &     0.81\% \\
auto           & 0.84\% &     0.83\% \\
compact\_car    & 0.79\% &     0.76\% \\
shooting\_brake & 0.64\% &     0.63\% \\
estate\_car     & 0.59\% &     0.59\% \\
wagon          & 0.54\% &     0.51\% \\
police\_cruiser & 0.45\% &     0.45\% \\
motorcar       & 0.40\% &     0.40\% \\
taxi           & 0.20\% &     0.17\% \\
cruiser        & 0.15\% &     0.13\% \\
compact        & 0.15\% &     0.11\% \\
beach\_wagon    & 0.15\% &     0.13\% \\
funny\_wagon    & 0.10\% &     0.05\% \\
gallery        & 0.10\% &     0.07\% \\
cab            & 0.10\% &     0.07\% \\
ambulance      & 0.10\% &     0.07\% \\
door           & 0.00\% &     0.03\% \\
ford           & 0.00\% &     0.03\% \\
opel           & 0.00\% &     0.03\% \\
sport\_car      & 0.00\% &     0.03\% \\
sports\_car     & 0.00\% &     0.03\% \\
\bottomrule
\end{tabular}

     \label{tab:Automobile}
  \end{subtable}
\end{table*}

\clearpage

\subsubsection{Full List of Models Evaluated on CIFAR-10}
\label{apx:cifar10_model_descriptions}
The following list contains all models we evaluated on CIFAR-10 with references and links to the corresponding source code.

\begin{enumerate}
\item \model{autoaug\_pyramid\_net} \cite{autoaugment, pyramidnet} \url{https://github.com/tensorflow/models/tree/master/research/autoaugment}
\item \model{autoaug\_shake\_shake\_112} \cite{autoaugment, shakeshake} \url{https://github.com/tensorflow/models/tree/master/research/autoaugment}
\item \model{autoaug\_shake\_shake\_32} \cite{autoaugment, shakeshake} \url{https://github.com/tensorflow/models/tree/master/research/autoaugment}
\item \model{autoaug\_shake\_shake\_96} \cite{autoaugment, shakeshake} \url{https://github.com/tensorflow/models/tree/master/research/autoaugment}
\item \model{autoaug\_wrn} \cite{autoaugment, wrn} \url{https://github.com/tensorflow/models/tree/master/research/autoaugment}
\item \model{cudaconvnet}   \cite{alexnet} \url{https://github.com/akrizhevsky/cuda-convnet2}
\item \model{darc}   \cite{darc} \url{http://lis.csail.mit.edu/code/gdl.html}
\item \model{densenet\_BC\_100\_12} \cite{densenet} \url{https://github.com/hysts/pytorch\_image\_classification/} 
\item \model{nas} \cite{nas} \url{https://github.com/tensorflow/models/blob/master/research/slim/nets/nasnet/nasnet.py#L32}
\item \model{pyramidnet\_basic\_110\_270} \cite{pyramidnet} \url{https://github.com/hysts/pytorch\_image\_classification/} 
\item \model{pyramidnet\_basic\_110\_84}  \cite{pyramidnet} \url{https://github.com/hysts/pytorch\_image\_classification/} 
\item \model{random\_features\_256k\_aug} \cite{rf} \url{https://github.com/modestyachts/nondeep} Random 1 layer convolutional network with 256k filters sampled from image patches, patch size = $6$, pool size $15$, pool stride $6$, and horizontal flip data augmentation.
\item \model{random\_features\_256k}   \cite{rf} \url{https://github.com/modestyachts/nondeep} Random 1 layer convolutional network with 256k filters sampled from image patches, patch size = $6$, pool size $15$, pool stride $6$.
\item \model{random\_features\_32k\_aug} \cite{rf} \url{https://github.com/modestyachts/nondeep} Random 1 layer convolutional network with 32k filters sampled from image patches, patch size = $6$, pool size $15$, pool stride $6$, and horizontal flip data augmentation.
\item \model{random\_features\_32k}   \cite{rf} Random 1 layer convolutional network with 32k filters sampled from image patches, patch size = $6$, pool size $15$, pool stride $16$.
\item \model{resnet\_basic\_32} \cite{resnet} \url{https://github.com/hysts/pytorch\_image\_classification/} 
\item \model{resnet\_basic\_44} \cite{resnet} \url{https://github.com/hysts/pytorch\_image\_classification/} 
\item \model{resnet\_basic\_56} \cite{resnet} \url{https://github.com/hysts/pytorch\_image\_classification/} 
\item \model{resnet\_basic\_110} \cite{resnet} \url{https://github.com/hysts/pytorch\_image\_classification/} 
\item \model{resnet\_preact\_basic\_110}  \cite{resnet_preact} \url{https://github.com/hysts/pytorch\_image\_classification/} 
\item \model{resnet\_preact\_bottleneck\_164} \cite{resnet_preact} \url{https://github.com/hysts/pytorch\_image\_classification/} 
\item \model{resnet\_preact\_tf} \cite{resnet_preact} \url{https://github.com/tensorflow/models/tree/b871670b5ae29aaa6cad1b2d4e004882f716c466/resnet}
\item \model{resnext\_29\_4x64d} \cite{resnext} \url{https://github.com/hysts/pytorch\_image\_classification/} 
\item \model{resnext\_29\_8x64d} \cite{resnext} \url{https://github.com/hysts/pytorch\_image\_classification/} 
\item \model{shake\_drop} \cite{shakedrop} \url{https://github.com/imenurok/ShakeDrop}
\item \model{shake\_shake\_32d} \cite{shakeshake} \url{https://github.com/hysts/pytorch\_image\_classification/} 
\item \model{shake\_shake\_64d} \cite{shakeshake} \url{https://github.com/hysts/pytorch\_image\_classification/}
\item \model{shake\_shake\_96d} \cite{shakeshake} \url{https://github.com/hysts/pytorch\_image\_classification/} 
\item \model{shake\_shake\_64d\_cutout}  \cite{shakeshake,cutout} \url{https://github.com/hysts/pytorch\_image\_classification/} 
\item \model{vgg16\_keras} \cite{vgg, vgg_cifar} \url{https://github.com/geifmany/cifar-vgg}
\item \model{vgg\_15\_BN\_64} \cite{vgg, vgg_cifar} \url{https://github.com/hysts/pytorch\_image\_classification/} 
\item \model{wide\_resnet\_tf} \cite{wrn} \url{https://github.com/tensorflow/models/tree/b871670b5ae29aaa6cad1b2d4e004882f716c466/resnet}
\item \model{wide\_resnet\_28\_10} \cite{wrn} \url{https://github.com/hysts/pytorch\_image\_classification/}
\item \model{wide\_resnet\_28\_10\_cutout} \cite{wrn,cutout} \url{https://github.com/hysts/pytorch\_image\_classification/}
\end{enumerate}

\subsubsection{Full Results Table}
\label{apx:cifar10_model_accuracies}
Table \ref{tab:v4_results} contains the detailed accuracy scores for the original CIFAR-10 test set and our new test set.

\begin{table*}[ht!]
  \caption{Model accuracy on the original CIFAR-10 test set and our new test set.
  $\Delta$ Rank is the relative difference in the ranking from the original test set to the new test set.
  For example, $\Delta \text{Rank} = -2$ means that a model dropped by two places on the new test set compared to the original test set.
  The confidence intervals are 95\% Clopper-Pearson intervals.
  Due to space constraints, references for the models can be found in Appendix \ref{apx:cifar10_model_descriptions}.
 }
   \label{tab:v4_results}
   \centering
   \rowcolors{3}{white}{gray!15}
   \begin{tabular}{rp{4.75cm}rrrrr}
\toprule 
\multicolumn{7}{c}{\textbf{CIFAR-10}} \\ 
\midrule
\multicolumn{1}{l}{Orig.} &                                    &                                        &                                          &    & \multicolumn{1}{l}{New} &  \\ 
 \multicolumn{1}{l}{Rank} & Model & Orig. Accuracy & New Accuracy & Gap & \multicolumn{1}{l}{Rank} & $\Delta$ Rank \\
\midrule
 1 &  \model{autoaug\_pyramid\_net\_tf} &  98.4 {\footnotesize \textcolor{gray}{[98.1, 98.6]}} &  95.5 {\footnotesize \textcolor{gray}{[94.5, 96.4]}} &  2.9 &  1 &  0 \\
 2 &  \model{autoaug\_shake\_shake\_112\_tf} &  98.1 {\footnotesize \textcolor{gray}{[97.8, 98.4]}} &  93.9 {\footnotesize \textcolor{gray}{[92.7, 94.9]}} &  4.3 &  2 &  0 \\
 3 &  \model{autoaug\_shake\_shake\_96\_tf} &  98.0 {\footnotesize \textcolor{gray}{[97.7, 98.3]}} &  93.7 {\footnotesize \textcolor{gray}{[92.6, 94.7]}} &  4.3 &  3 &  0 \\
 4 &  \model{autoaug\_wrn\_tf} &  97.5 {\footnotesize \textcolor{gray}{[97.1, 97.8]}} &  93.0 {\footnotesize \textcolor{gray}{[91.8, 94.1]}} &  4.4 &  4 &  0 \\
 5 &  \model{autoaug\_shake\_shake\_32\_tf} &  97.3 {\footnotesize \textcolor{gray}{[97.0, 97.6]}} &  92.9 {\footnotesize \textcolor{gray}{[91.7, 94.0]}} &  4.4 &  6 &  -1 \\
 6 &  \model{shake\_shake\_64d\_cutout} &  97.1 {\footnotesize \textcolor{gray}{[96.8, 97.4]}} &  93.0 {\footnotesize \textcolor{gray}{[91.8, 94.1]}} &  4.1 &  5 &  1 \\
 7 &  \model{shake\_shake\_26\_2x96d\_SSI} &  97.1 {\footnotesize \textcolor{gray}{[96.7, 97.4]}} &  91.9 {\footnotesize \textcolor{gray}{[90.7, 93.1]}} &  5.1 &  9 &  -2 \\
 8 &  \model{shake\_shake\_64d} &  97.0 {\footnotesize \textcolor{gray}{[96.6, 97.3]}} &  91.4 {\footnotesize \textcolor{gray}{[90.1, 92.6]}} &  5.6 &  10 &  -2 \\
 9 &  \model{wrn\_28\_10\_cutout16} &  97.0 {\footnotesize \textcolor{gray}{[96.6, 97.3]}} &  92.0 {\footnotesize \textcolor{gray}{[90.7, 93.1]}} &  5.0 &  8 &  1 \\
 10 &  \model{shake\_drop} &  96.9 {\footnotesize \textcolor{gray}{[96.5, 97.2]}} &  92.3 {\footnotesize \textcolor{gray}{[91.0, 93.4]}} &  4.6 &  7 &  3 \\
 11 &  \model{shake\_shake\_32d} &  96.6 {\footnotesize \textcolor{gray}{[96.2, 96.9]}} &  89.8 {\footnotesize \textcolor{gray}{[88.4, 91.1]}} &  6.8 &  13 &  -2 \\
 12 &  \model{darc} &  96.6 {\footnotesize \textcolor{gray}{[96.2, 96.9]}} &  89.5 {\footnotesize \textcolor{gray}{[88.1, 90.8]}} &  7.1 &  16 &  -4 \\
 13 &  \model{resnext\_29\_4x64d} &  96.4 {\footnotesize \textcolor{gray}{[96.0, 96.7]}} &  89.6 {\footnotesize \textcolor{gray}{[88.2, 90.9]}} &  6.8 &  15 &  -2 \\
 14 &  \model{pyramidnet\_basic\_110\_270} &  96.3 {\footnotesize \textcolor{gray}{[96.0, 96.7]}} &  90.5 {\footnotesize \textcolor{gray}{[89.1, 91.7]}} &  5.9 &  11 &  3 \\
 15 &  \model{resnext\_29\_8x64d} &  96.2 {\footnotesize \textcolor{gray}{[95.8, 96.6]}} &  90.0 {\footnotesize \textcolor{gray}{[88.6, 91.2]}} &  6.3 &  12 &  3 \\
 16 &  \model{wrn\_28\_10} &  95.9 {\footnotesize \textcolor{gray}{[95.5, 96.3]}} &  89.7 {\footnotesize \textcolor{gray}{[88.3, 91.0]}} &  6.2 &  14 &  2 \\
 17 &  \model{pyramidnet\_basic\_110\_84} &  95.7 {\footnotesize \textcolor{gray}{[95.3, 96.1]}} &  89.3 {\footnotesize \textcolor{gray}{[87.8, 90.6]}} &  6.5 &  17 &  0 \\
 18 &  \model{densenet\_BC\_100\_12} &  95.5 {\footnotesize \textcolor{gray}{[95.1, 95.9]}} &  87.6 {\footnotesize \textcolor{gray}{[86.1, 89.0]}} &  8.0 &  20 &  -2 \\
 19 &  \model{nas} &  95.4 {\footnotesize \textcolor{gray}{[95.0, 95.8]}} &  88.8 {\footnotesize \textcolor{gray}{[87.4, 90.2]}} &  6.6 &  18 &  1 \\
 20 &  \model{wide\_resnet\_tf\_28\_10} &  95.0 {\footnotesize \textcolor{gray}{[94.6, 95.4]}} &  88.5 {\footnotesize \textcolor{gray}{[87.0, 89.9]}} &  6.5 &  19 &  1 \\
 21 &  \model{resnet\_v2\_bottleneck\_164} &  94.2 {\footnotesize \textcolor{gray}{[93.7, 94.6]}} &  85.9 {\footnotesize \textcolor{gray}{[84.3, 87.4]}} &  8.3 &  22 &  -1 \\
 22 &  \model{vgg16\_keras} &  93.6 {\footnotesize \textcolor{gray}{[93.1, 94.1]}} &  85.3 {\footnotesize \textcolor{gray}{[83.6, 86.8]}} &  8.3 &  23 &  -1 \\
 23 &  \model{resnet\_basic\_110} &  93.5 {\footnotesize \textcolor{gray}{[93.0, 93.9]}} &  85.2 {\footnotesize \textcolor{gray}{[83.5, 86.7]}} &  8.3 &  24 &  -1 \\
 24 &  \model{resnet\_v2\_basic\_110} &  93.4 {\footnotesize \textcolor{gray}{[92.9, 93.9]}} &  86.5 {\footnotesize \textcolor{gray}{[84.9, 88.0]}} &  6.9 &  21 &  3 \\
 25 &  \model{resnet\_basic\_56} &  93.3 {\footnotesize \textcolor{gray}{[92.8, 93.8]}} &  85.0 {\footnotesize \textcolor{gray}{[83.3, 86.5]}} &  8.3 &  25 &  0 \\
 26 &  \model{resnet\_basic\_44} &  93.0 {\footnotesize \textcolor{gray}{[92.5, 93.5]}} &  84.2 {\footnotesize \textcolor{gray}{[82.6, 85.8]}} &  8.8 &  29 &  -3 \\
 27 &  \model{vgg\_15\_BN\_64} &  93.0 {\footnotesize \textcolor{gray}{[92.5, 93.5]}} &  84.9 {\footnotesize \textcolor{gray}{[83.2, 86.4]}} &  8.1 &  27 &  0 \\
 28 &  \model{resnetv2\_tf\_32} &  92.7 {\footnotesize \textcolor{gray}{[92.2, 93.2]}} &  84.4 {\footnotesize \textcolor{gray}{[82.7, 85.9]}} &  8.3 &  28 &  0 \\
 29 &  \model{resnet\_basic\_32} &  92.5 {\footnotesize \textcolor{gray}{[92.0, 93.0]}} &  84.9 {\footnotesize \textcolor{gray}{[83.2, 86.4]}} &  7.7 &  26 &  3 \\
 30 &  \model{cudaconvnet} &  88.5 {\footnotesize \textcolor{gray}{[87.9, 89.2]}} &  77.5 {\footnotesize \textcolor{gray}{[75.7, 79.3]}} &  11.0 &  30 &  0 \\
 31 &  \model{random\_features\_256k\_aug} &  85.6 {\footnotesize \textcolor{gray}{[84.9, 86.3]}} &  73.1 {\footnotesize \textcolor{gray}{[71.1, 75.1]}} &  12.5 &  31 &  0 \\
 32 &  \model{random\_features\_32k\_aug} &  85.0 {\footnotesize \textcolor{gray}{[84.3, 85.7]}} &  71.9 {\footnotesize \textcolor{gray}{[69.9, 73.9]}} &  13.0 &  32 &  0 \\
 33 &  \model{random\_features\_256k} &  84.2 {\footnotesize \textcolor{gray}{[83.5, 84.9]}} &  69.9 {\footnotesize \textcolor{gray}{[67.8, 71.9]}} &  14.3 &  33 &  0 \\
 34 &  \model{random\_features\_32k} &  83.3 {\footnotesize \textcolor{gray}{[82.6, 84.0]}} &  67.9 {\footnotesize \textcolor{gray}{[65.9, 70.0]}} &  15.4 &  34 &  0 \\
\bottomrule
\end{tabular}

\end{table*}

\subsubsection{Full Results Table for the Exactly Class-Balanced Test Set}
\label{apx:cifar10_model_accuracies_balanced}
Table \ref{tab:v6_results} contains the detailed accuracy scores for the original CIFAR-10 test set and the exactly class-balanced variant of our new test set.
\begin{table*}[h!]
  \caption{Model accuracy on the original CIFAR-10 test set and the exactly class-balanced variant of our new test set.
  $\Delta$ Rank is the relative difference in the ranking from the original test set to the new test set.
  For example, $\Delta \text{Rank} = -2$ means that a model dropped by two places on the new test set compared to the original test set.
  The confidence intervals are 95\% Clopper-Pearson intervals.
  Due to space constraints, references for the models can be found in Appendix \ref{apx:cifar10_model_descriptions}.
}
  \label{tab:v6_results}
  \centering
  \rowcolors{3}{white}{gray!15}
  \begin{tabular}{rp{4.75cm}rrrrr}
\toprule 
\multicolumn{7}{c}{\textbf{CIFAR-10}} \\ 
\midrule
\multicolumn{1}{l}{Orig.} &                                    &                                        &                                          &    & \multicolumn{1}{l}{New} &  \\ 
 \multicolumn{1}{l}{Rank} & Model & Orig. Accuracy & New Accuracy & Gap & \multicolumn{1}{l}{Rank} & $\Delta$ Rank \\
\midrule
 1 &  \model{autoaug\_pyramid\_net\_tf} &  98.4 {\footnotesize \textcolor{gray}{[98.1, 98.6]}} &  95.5 {\footnotesize \textcolor{gray}{[94.5, 96.4]}} &  2.9 &  1 &  0 \\
 2 &  \model{autoaug\_shake\_shake\_112\_tf} &  98.1 {\footnotesize \textcolor{gray}{[97.8, 98.4]}} &  94.0 {\footnotesize \textcolor{gray}{[92.9, 95.0]}} &  4.1 &  2 &  0 \\
 3 &  \model{autoaug\_shake\_shake\_96\_tf} &  98.0 {\footnotesize \textcolor{gray}{[97.7, 98.3]}} &  93.9 {\footnotesize \textcolor{gray}{[92.8, 94.9]}} &  4.1 &  3 &  0 \\
 4 &  \model{autoaug\_wrn\_tf} &  97.5 {\footnotesize \textcolor{gray}{[97.1, 97.8]}} &  93.0 {\footnotesize \textcolor{gray}{[91.8, 94.1]}} &  4.5 &  6 &  -2 \\
 5 &  \model{autoaug\_shake\_shake\_32\_tf} &  97.3 {\footnotesize \textcolor{gray}{[97.0, 97.6]}} &  93.2 {\footnotesize \textcolor{gray}{[92.0, 94.2]}} &  4.2 &  4 &  1 \\
 6 &  \model{shake\_shake\_64d\_cutout} &  97.1 {\footnotesize \textcolor{gray}{[96.8, 97.4]}} &  93.1 {\footnotesize \textcolor{gray}{[91.9, 94.2]}} &  4.0 &  5 &  1 \\
 7 &  \model{shake\_shake\_26\_2x96d\_SSI} &  97.1 {\footnotesize \textcolor{gray}{[96.7, 97.4]}} &  92.0 {\footnotesize \textcolor{gray}{[90.7, 93.1]}} &  5.1 &  9 &  -2 \\
 8 &  \model{shake\_shake\_64d} &  97.0 {\footnotesize \textcolor{gray}{[96.6, 97.3]}} &  91.9 {\footnotesize \textcolor{gray}{[90.6, 93.1]}} &  5.1 &  10 &  -2 \\
 9 &  \model{wrn\_28\_10\_cutout16} &  97.0 {\footnotesize \textcolor{gray}{[96.6, 97.3]}} &  92.1 {\footnotesize \textcolor{gray}{[90.8, 93.2]}} &  4.9 &  8 &  1 \\
 10 &  \model{shake\_drop} &  96.9 {\footnotesize \textcolor{gray}{[96.5, 97.2]}} &  92.3 {\footnotesize \textcolor{gray}{[91.1, 93.4]}} &  4.6 &  7 &  3 \\
 11 &  \model{shake\_shake\_32d} &  96.6 {\footnotesize \textcolor{gray}{[96.2, 96.9]}} &  90.0 {\footnotesize \textcolor{gray}{[88.6, 91.3]}} &  6.6 &  15 &  -4 \\
 12 &  \model{darc} &  96.6 {\footnotesize \textcolor{gray}{[96.2, 96.9]}} &  89.9 {\footnotesize \textcolor{gray}{[88.5, 91.2]}} &  6.7 &  16 &  -4 \\
 13 &  \model{resnext\_29\_4x64d} &  96.4 {\footnotesize \textcolor{gray}{[96.0, 96.7]}} &  90.1 {\footnotesize \textcolor{gray}{[88.8, 91.4]}} &  6.2 &  12 &  1 \\
 14 &  \model{pyramidnet\_basic\_110\_270} &  96.3 {\footnotesize \textcolor{gray}{[96.0, 96.7]}} &  90.5 {\footnotesize \textcolor{gray}{[89.1, 91.7]}} &  5.8 &  11 &  3 \\
 15 &  \model{resnext\_29\_8x64d} &  96.2 {\footnotesize \textcolor{gray}{[95.8, 96.6]}} &  90.1 {\footnotesize \textcolor{gray}{[88.7, 91.4]}} &  6.1 &  14 &  1 \\
 16 &  \model{wrn\_28\_10} &  95.9 {\footnotesize \textcolor{gray}{[95.5, 96.3]}} &  90.1 {\footnotesize \textcolor{gray}{[88.8, 91.4]}} &  5.8 &  13 &  3 \\
 17 &  \model{pyramidnet\_basic\_110\_84} &  95.7 {\footnotesize \textcolor{gray}{[95.3, 96.1]}} &  89.6 {\footnotesize \textcolor{gray}{[88.2, 90.9]}} &  6.1 &  17 &  0 \\
 18 &  \model{densenet\_BC\_100\_12} &  95.5 {\footnotesize \textcolor{gray}{[95.1, 95.9]}} &  87.9 {\footnotesize \textcolor{gray}{[86.4, 89.3]}} &  7.6 &  20 &  -2 \\
 19 &  \model{nas} &  95.4 {\footnotesize \textcolor{gray}{[95.0, 95.8]}} &  89.2 {\footnotesize \textcolor{gray}{[87.8, 90.5]}} &  6.2 &  18 &  1 \\
 20 &  \model{wide\_resnet\_tf\_28\_10} &  95.0 {\footnotesize \textcolor{gray}{[94.6, 95.4]}} &  88.8 {\footnotesize \textcolor{gray}{[87.4, 90.2]}} &  6.2 &  19 &  1 \\
 21 &  \model{resnet\_v2\_bottleneck\_164} &  94.2 {\footnotesize \textcolor{gray}{[93.7, 94.6]}} &  86.1 {\footnotesize \textcolor{gray}{[84.5, 87.6]}} &  8.1 &  22 &  -1 \\
 22 &  \model{vgg16\_keras} &  93.6 {\footnotesize \textcolor{gray}{[93.1, 94.1]}} &  85.6 {\footnotesize \textcolor{gray}{[84.0, 87.1]}} &  8.0 &  23 &  -1 \\
 23 &  \model{resnet\_basic\_110} &  93.5 {\footnotesize \textcolor{gray}{[93.0, 93.9]}} &  85.4 {\footnotesize \textcolor{gray}{[83.8, 86.9]}} &  8.1 &  24 &  -1 \\
 24 &  \model{resnet\_v2\_basic\_110} &  93.4 {\footnotesize \textcolor{gray}{[92.9, 93.9]}} &  86.9 {\footnotesize \textcolor{gray}{[85.4, 88.3]}} &  6.5 &  21 &  3 \\
 25 &  \model{resnet\_basic\_56} &  93.3 {\footnotesize \textcolor{gray}{[92.8, 93.8]}} &  84.9 {\footnotesize \textcolor{gray}{[83.2, 86.4]}} &  8.5 &  28 &  -3 \\
 26 &  \model{resnet\_basic\_44} &  93.0 {\footnotesize \textcolor{gray}{[92.5, 93.5]}} &  84.8 {\footnotesize \textcolor{gray}{[83.2, 86.3]}} &  8.2 &  29 &  -3 \\
 27 &  \model{vgg\_15\_BN\_64} &  93.0 {\footnotesize \textcolor{gray}{[92.5, 93.5]}} &  85.0 {\footnotesize \textcolor{gray}{[83.4, 86.6]}} &  7.9 &  27 &  0 \\
 28 &  \model{resnetv2\_tf\_32} &  92.7 {\footnotesize \textcolor{gray}{[92.2, 93.2]}} &  85.1 {\footnotesize \textcolor{gray}{[83.5, 86.6]}} &  7.6 &  26 &  2 \\
 29 &  \model{resnet\_basic\_32} &  92.5 {\footnotesize \textcolor{gray}{[92.0, 93.0]}} &  85.2 {\footnotesize \textcolor{gray}{[83.6, 86.7]}} &  7.3 &  25 &  4 \\
 30 &  \model{cudaconvnet} &  88.5 {\footnotesize \textcolor{gray}{[87.9, 89.2]}} &  78.2 {\footnotesize \textcolor{gray}{[76.3, 80.0]}} &  10.3 &  30 &  0 \\
 31 &  \model{random\_features\_256k\_aug} &  85.6 {\footnotesize \textcolor{gray}{[84.9, 86.3]}} &  73.6 {\footnotesize \textcolor{gray}{[71.6, 75.5]}} &  12.0 &  31 &  0 \\
 32 &  \model{random\_features\_32k\_aug} &  85.0 {\footnotesize \textcolor{gray}{[84.3, 85.7]}} &  72.2 {\footnotesize \textcolor{gray}{[70.2, 74.1]}} &  12.8 &  32 &  0 \\
 33 &  \model{random\_features\_256k} &  84.2 {\footnotesize \textcolor{gray}{[83.5, 84.9]}} &  70.5 {\footnotesize \textcolor{gray}{[68.4, 72.4]}} &  13.8 &  33 &  0 \\
 34 &  \model{random\_features\_32k} &  83.3 {\footnotesize \textcolor{gray}{[82.6, 84.0]}} &  68.7 {\footnotesize \textcolor{gray}{[66.6, 70.7]}} &  14.6 &  34 &  0 \\
\bottomrule
\end{tabular}

\end{table*}

\subsubsection{Hard Images}
\label{app:cifar_hard_images}
Figure \ref{fig:hardtest} shows the images in our new CIFAR-10 test set that were misclassified by all models in our testbed.
As can be seen in the figure, the class labels for these images are correct.

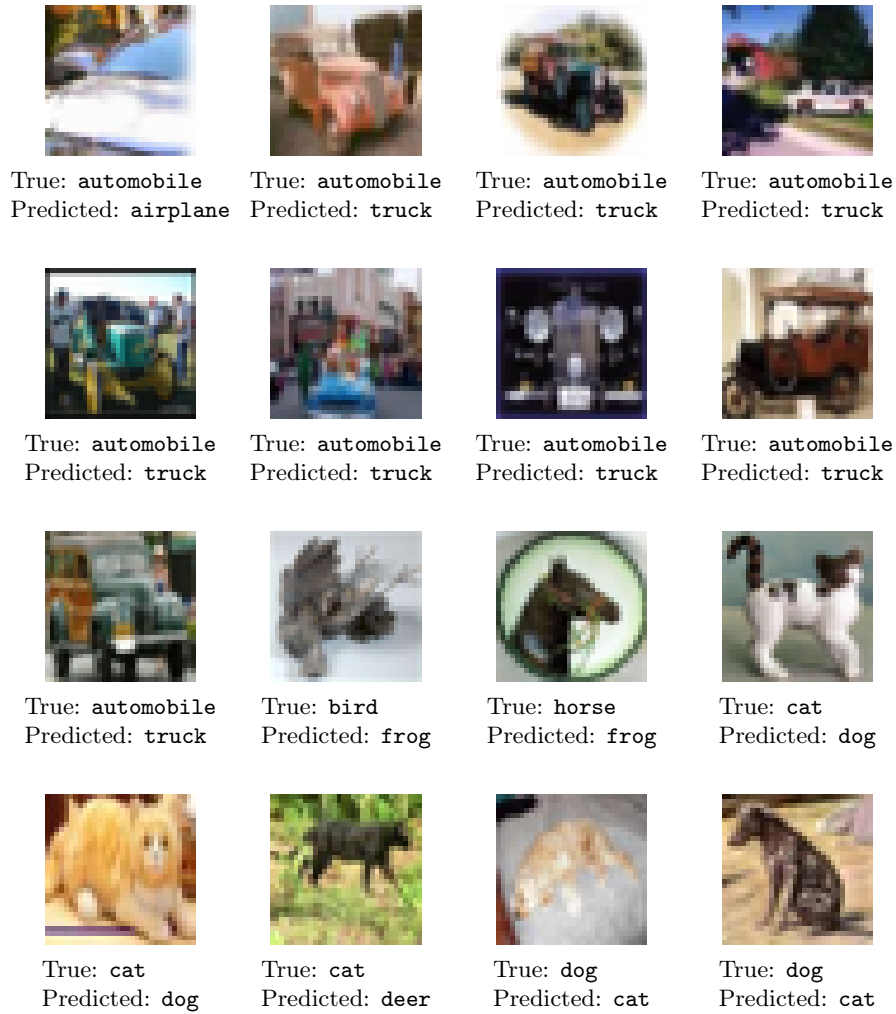
\begin{figure*}[htb]
\centering
\setlength{\imagedim}{2cm}
\setlength{\imagexspacing}{1cm}
\setlength{\imageyspacing}{1.5cm}
\newlength{\labelspacingtwo}
\setlength{\labelspacingtwo}{.2cm}
\centering
\begin{tikzpicture}
\footnotesize
\tikzstyle{img}=[inner sep=0pt,outer sep=0pt];
\tikzstyle{imglabel}=[anchor=north,inner sep=0pt,yshift=-\labelspacingtwo];
\node [img] (image0) {\includegraphics[width=\imagedim]{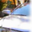}};
\node [img,anchor=west,at=(image0.east),xshift=\imagexspacing] (image1)
{\includegraphics[width=\imagedim]{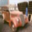}};
\node [img,anchor=west,at=(image1.east),xshift=\imagexspacing] (image2)
{\includegraphics[width=\imagedim]{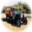}};
\node [img,anchor=west,at=(image2.east),xshift=\imagexspacing] (image3)
{\includegraphics[width=\imagedim]{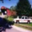}};

\node [img,anchor=north,at=(image0.south),yshift=-\imageyspacing] (image4)
{\includegraphics[width=\imagedim]{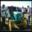}};
\node [img,anchor=west,at=(image4.east),xshift=\imagexspacing] (image5)
{\includegraphics[width=\imagedim]{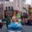}};
\node [img,anchor=west,at=(image5.east),xshift=\imagexspacing] (image6)
{\includegraphics[width=\imagedim]{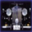}};
\node [img,anchor=west,at=(image6.east),xshift=\imagexspacing] (image7)
{\includegraphics[width=\imagedim]{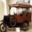}};

\node [img,anchor=north,at=(image4.south),yshift=-\imageyspacing] (image8)
{\includegraphics[width=\imagedim]{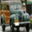}};
\node [img,anchor=west,at=(image8.east),xshift=\imagexspacing] (image9)
{\includegraphics[width=\imagedim]{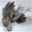}};
\node [img,anchor=west,at=(image9.east),xshift=\imagexspacing] (image10)
{\includegraphics[width=\imagedim]{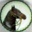}};
\node [img,anchor=west,at=(image10.east),xshift=\imagexspacing] (image11)
{\includegraphics[width=\imagedim]{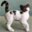}};

\node [img,anchor=north,at=(image8.south),yshift=-\imageyspacing] (image12)
{\includegraphics[width=\imagedim]{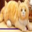}};
\node [img,anchor=west,at=(image12.east),xshift=\imagexspacing] (image13)
{\includegraphics[width=\imagedim]{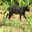}};
\node [img,anchor=west,at=(image13.east),xshift=\imagexspacing] (image14)
{\includegraphics[width=\imagedim]{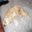}};
\node [img,anchor=west,at=(image14.east),xshift=\imagexspacing] (image15)
{\includegraphics[width=\imagedim]{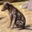}};

\node [imglabel] (label0) at (image0.south) [align=left]{True: \class{automobile}\\ Predicted: \class{airplane}};
\node [imglabel] (label1) at (image1.south) [align=left]{True: \class{automobile}\\ Predicted: \class{truck}};
\node [imglabel] (label2) at (image2.south) [align=left]{True: \class{automobile}\\ Predicted: \class{truck}};
\node [imglabel] (label3) at (image3.south) [align=left]{True: \class{automobile}\\ Predicted: \class{truck}};
\node [imglabel] (label4) at (image4.south) [align=left]{True: \class{automobile}\\ Predicted: \class{truck}};
\node [imglabel] (label5) at (image5.south) [align=left]{True: \class{automobile}\\ Predicted: \class{truck}};
\node [imglabel] (label6) at (image6.south) [align=left]{True: \class{automobile}\\ Predicted: \class{truck}};
\node [imglabel] (label7) at (image7.south) [align=left]{True: \class{automobile}\\ Predicted: \class{truck}};
\node [imglabel] (label8) at (image8.south) [align=left]{True: \class{automobile}\\ Predicted: \class{truck}};

\node [imglabel] (label9) at (image9.south) [align=left]{True: \class{bird}\\ Predicted: \class{frog}};
\node [imglabel] (label10) at (image10.south) [align=left]{True: \class{horse}\\ Predicted: \class{frog}};
\node [imglabel] (label11) at (image11.south) [align=left]{True: \class{cat}\\ Predicted: \class{dog}};
\node [imglabel] (label12) at (image12.south) [align=left]{True: \class{cat}\\ Predicted: \class{dog}};
\node [imglabel] (label13) at (image13.south) [align=left]{True: \class{cat}\\ Predicted: \class{deer}};
\node [imglabel] (label14) at (image14.south) [align=left]{True: \class{dog}\\ Predicted: \class{cat}};
\node [imglabel] (label15) at (image15.south) [align=left]{True: \class{dog}\\ Predicted: \class{cat}};

\end{tikzpicture}
  \caption{Hard images from our new test set that no model correctly.
    The caption of each image states the correct class label (``True'') and the label predicted by most models (``Predicted'').}
  \label{fig:hardtest}
\end{figure*}

\newpage
\clearpage

\section{Details of the ImageNet Experiments}
\label{app:imagenet}
Our results on CIFAR-10 show that current models fail to reliably generalize in the presence of small variations in the data distribution.
One hypothesis is that the accuracy drop stems from the limited nature of the CIFAR-10 dataset.
Compared to other datasets, CIFAR-10 is relatively small, both in terms of image resolution and the number of images in the dataset.
Since the CIFAR-10 models are only exposed to a constrained visual environment, they may be unable to learn a more reliable representation.

To investigate whether ImageNet models generalize more reliably, we assemble a new test set for ImageNet.
ImageNet captures a much broader variety of natural images: it contains about $24\times$ more training images than CIFAR-10 with roughly $100 \times$ more pixels per image.
As a result, ImageNet poses a significantly harder problem and is among the most prestigious machine learning benchmarks.
The steadily improving accuracy numbers have also been cited as an important breakthrough in machine learning \cite{MalikCACM}.
If popular ImageNet models are indeed more robust to natural variations in the data (and there is again no adaptive overfitting), the accuracies on our new test set should roughly match the existing accuracies.

Before we proceed to our experiments, we briefly describe the relevant background concerning the ImageNet dataset.
For more details, we refer the reader to the original ImageNet publications \cite{imagenet,RDSKSMHKKBBL15}.

\paragraph{ImageNet.} ImageNet \cite{imagenet,RDSKSMHKKBBL15} is a large image database consisting of more than 14 million human-annotated images depicting almost 22,000 classes.
The images do not have a uniform size, but most of them are stored as RGB color images with a resolution around $500 \times 400$ pixels.
The classes are derived from the WordNet hierarchy \cite{wordnet}, which represents each class by a set of synonyms (``synset'') and is organized into semantically meaningful relations.
Each class has an associated definition (``gloss'') and a unique WordNet ID (``wnid'').

The ImageNet team populated the classes with images downloaded from various image search engines, using the WordNet synonyms as queries.
The researchers then annotated the images via Amazon Mechanical Turk (MTurk).
A class-specific threshold decided how many agreements among the MTurk workers were necessary for an image to be considered valid.
Overall, the researchers employed over 49,000 workers from 167 countries \cite{LD17}.

Since 2010, the ImageNet team has run the yearly ImageNet Large Scale Visual Recognition Challenge (ILSVRC), which consists of separate tracks for object classification, localization, and detection.
All three tracks are based on subsets of the ImageNet data.
The classification track has received the most attention and is also the focus of our paper.

The ILSVRC2012 competition data has become the de facto benchmark version of the dataset and comprises 1.2 million training images, 50,000 validation images, and 100,000 test images depicting 1,000 categories.
We generally refer to this data as the ImageNet training, validation, and test set.
The labels for the ImageNet test set were never publicly released in order to minimize adaptive overfitting.
Instead, teams could submit a limited number of requests to an evaluation server in order to obtain accuracy scores.
There were no similar limitations in place for the validation set.
Most publications report accuracy numbers on the validation set.

The training, validation, and test sets were not drawn strictly i.i.d.\ from the same distribution (i.e., there was not a single data collection run with the result split randomly into training, validation, and test).
Instead, the data collection was an ongoing process and both the validation and test sets were refreshed in various years of the ILSVRC.
One notable difference is that the ImageNet training and validation sets do not have the same data source: while the ImageNet training set consists of images from several search engines (e.g., Google, MSN, Yahoo, and Flickr), the validation set consists almost entirely of images from Flickr \cite{bergpersonal}.

\subsection{Dataset Creation Methodology}
    \label{sec:imagenet_building_new_test_set}

    Since the existing training, validation, and test sets are not strictly i.i.d.\ (see above), the first question was which dataset part to replicate.
    For our experiment, we decided to match the distribution of the \emph{validation set}.
    There are multiple reasons for this choice:
    \begin{itemize}
      \item In contrast to the training set, the validation set comes from only one data source (Flickr).
        Moreover, the Flickr API allows fine-grained searches, which makes it easier to control the data source and match the original distribution.
      \item In contrast to the original test set, the validation set comes with label information.
        This makes it easier to inspect the existing image distribution for each class, which is important to ensure that we match various intricacies of the dataset (e.g., see Appendix \ref{app:ambiguous_imagenet} for examples of ambiguous classes). 
      \item Most papers report accuracy numbers on the validation set.
        Hence comparing new vs.\ existing accuracies is most relevant for the validation set.
      \item The validation set is commonly used to develop new architectures and tune hyperparameters, which leads to the possibility of adaptive overfitting.
        If we again observe no diminishing returns in accuracy on our new test set, this indicates that even the validation set is resilient to adaptive overfitting.
    \end{itemize}

    Therefore, our goal was to replicate the distribution of the original validation set as closely as possible.
    We aimed for a new test set of size 10,000 since this would already result in accuracy scores with small confidence intervals (see Section \ref{sec:formal}).
    While a larger dataset would result in even smaller confidence intervals, we were also concerned that searching for more images might lead to a larger distribution shift.
    In particular, we decided to use a time range for our Flickr queries \emph{after} the original ImageNet collection period (see below for the corresponding considerations).
    Since a given time period only has a limited supply of high quality images, a larger test set would have required a longer time range.
    This in turn may create a larger temporal distribution shift.
    To balance these two concerns, we decided on a size of 10,000 images for the new test set.

    Figure \ref{fig:imagenet_dataset_pipeline} presents a visual overview of our dataset creation pipeline.
    It consists of two parts: creating a pool of candidate images and sampling a clean dataset from this candidate pool.
    We now describe each part in detail to give the reader insights into the design choices potentially affecting the final distribution.

    \begin{figure*}[h!]
        \centering
        \includegraphics[width=\textwidth]{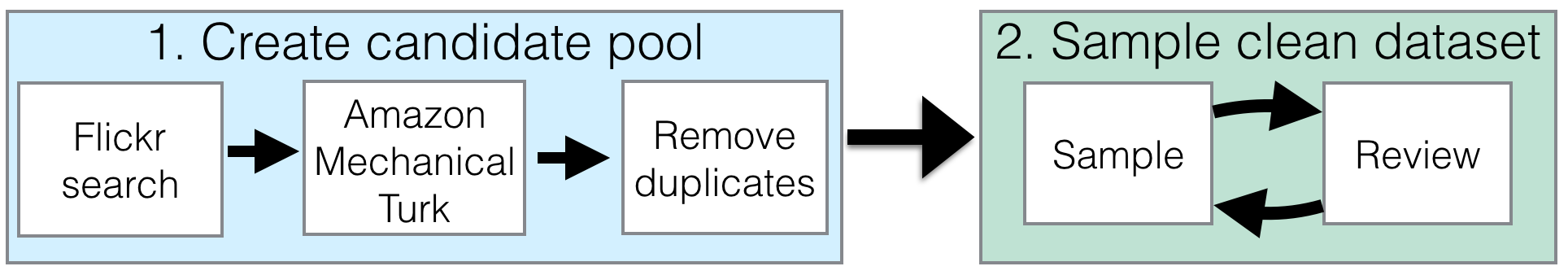}
        \caption{The pipeline for the new ImageNet test set. It consists of two parts: creating the candidate pool and sampling the final dataset from this candidate pool.}
        \label{fig:imagenet_dataset_pipeline}
    \end{figure*}

    \subsubsection{Creating a Candidate Pool}
    Similar to the creation procedure for the original ImageNet validation set, we collected candidate images from the Flickr image hosting service and then annotated them with Amazon Mechanical Turk (MTurk).

    \paragraph{Downloading images from Flickr.}
    The Flickr API has a range of parameters for image searches such as the query terms, an allowed time range, a maximum number of returned images, and a sorting order.
    We summarize the main points here:
    \begin{itemize}
      \item \textbf{Query terms:} For each class, we used each of the WordNet synonyms as a search term in separate queries.
      \item \textbf{Date range:} There were two main options for the date range associated with our queries to Flickr: either the same date range as the original ImageNet data collection, or a date range directly after ImageNet.
        The advantage of using the ImageNet date range is that it avoids a distribution shift due to the time the images were taken.
        However, this option also comes with two important caveats:
        First, the pool of high quality images in the original ImageNet date range could have been largely exhausted by ImageNet.
        Second, the new dataset could end up with near-duplicates of images in the original validation or training set that are hard to detect.
        Especially the first issue is difficult to quantify, so we decided on a time range directly after the ImageNet collection period.

        In particular, we initially searched for images taken and uploaded to Flickr between July 11, 2012 and July 11, 2013 because the final ILSVRC2012 public data release was on July 10, 2012.
        Since we used a period of only one year (significantly shorter than the ImageNet collection period), we believe that the temporal component of the distribution shift is small.
      \item \textbf{Result size:} We initially downloaded up to 100 images for each class.
     If a class has $k$ synonyms associated with it, we requested 100/$k$ images for each synonym.
     We decided on 100 images per class since we aimed for 10,000 images overall and estimated that 10\% of the candidate images would be of sufficiently high quality (similar to ImageNet \cite{imagenet}).
     \item \textbf{Result order:} Flickr offers the sorting options ``relevance'', ``interestingness'', and various temporal orderings.
       Note that the ``relevance'' and ``interestingness'' orderings may rely on machine learning models trained on ImageNet.
       Since these orderings may introduce a significant bias (e.g., by mainly showing images that current ImageNet models recognize for the respective search term), we chose to order the images by their upload date.
       This helps to ensure that our new test set is independent of current classifiers.
    \end{itemize}

    After our first data collection, we found it necessary to expand the initial candidate pool for particular classes in order to reach a sufficient number of valid images.
    This is similar to the original ImageNet creation process, where the authors expanded the set of queries using two methods \cite{imagenet,RDSKSMHKKBBL15}.
    The first method appended a word from the parent class to the queries if this word also appeared in the gloss of the target class.
    The second method included translations of the queries into other languages such as Chinese, Spanish, Dutch, and Italian.

    We took the following steps to expand our search queries, only proceeding to the next step for a given class when in need of more images.
    \begin{enumerate}
    \item Append a word from the parent class if the word appears in the gloss of the target class.
    \item Expand the maximum number of images to 200 for this class.
    \item Expand the search range to include photos taken or uploaded before July 11, 2014 (i.e., a time span of two years instead of one).
    \item Concatenate compound queries, i.e., search for ``dialphone'' instead of ``dial phone''.
    \item Manually pick alternative query words, including translations of the queries.
    \end{enumerate}

    In total, we obtained 208,145 candidate images from Flickr.

    \paragraph{Amazon Mechanical Turk.}
    While the candidate images from Flickr are correlated with their corresponding class, a large number of images are still unsuitable for an image classification dataset.
    For instance, the images may be of low quality (blurry, unclear object presence, etc.), violate dataset rules (e.g., no paintings), or be simply unrelated to the target class.
    So similar to ImageNet, we utilized MTurk to filter our pool of candidate images.

    We designed our MTurk tasks and UI to be close to those used in ImageNet.
    As in ImageNet, we showed each MTurk worker a grid of 48 candidate images for a given target class.
    The task description was derived from the original ImageNet instructions and included the definition of the target class with a link to a corresponding Wikipedia page.
    We asked the MTurk workers to select images belonging to the target class regardless of ``occlusions, other objects, and clutter or text in the scene'' and to avoid drawings or paintings (both as in ImageNet).
    Appendix \ref{apx:mturk_ui} shows a screenshot of our UI and a screenshot of the original UI for comparison.

    For quality control, we embedded at least six randomly selected images from the original validation set in each MTurk task (three from the same class, three from a class that is nearby in the WordNet hierarchy).
    These images appeared in random locations of the image grid for each task.
    We obfuscated all image URLs and resized our images to match the most common size of the existing validation images so that the original validation images were not easy to spot.

    The main outcome of the MTurk tasks is a \emph{selection frequency} for each image, i.e., what fraction of MTurk workers selected the image in a task for its target class.
    We recruited at least ten MTurk workers for each task (and hence for each image), which is similar to ImageNet.
    Since each task contained original validation images, we could also estimate how often images from the original dataset were selected by our MTurk workers.

    \paragraph{Removing near-duplicate images.}
    The final step in creating the candidate pool was to remove near-duplicates, both within our new test set and between our new test set and the original ImageNet dataset.
    Both types of near-duplicates could harm the quality of our dataset.

    Since we obtained results from Flickr in a temporal ordering, certain events (e.g., the 2012 Olympics) led to a large number of similar images depicting the same scene (e.g., in the class for the ``horizontal bar`` gymnastics instrument).
    Inspecting the ImageNet validation set revealed only very few sets of images from a single event.
    Moreover, the ImageNet paper also remarks that they removed near-duplicates \cite{imagenet}.
    Hence we decided to remove near-duplicates within our new test set.

    Near-duplicates between our dataset and the original test set are also problematic.
    Since the models typically achieve high accuracy on the training set, testing on a near-duplicate of a training image checks for memorization more than generalization.
    A near-duplicate between the existing validation set and our new test set also defeats the purpose of measuring generalization to previously unseen data (as opposed to data that may already have been the victim of adaptive overfitting).

    To find near-duplicates, we computed the 30 nearest neighbors for each candidate image in three different metrics:
    $\ell_2$-distance on raw pixels, $\ell_2$-distance on features extracted from a pre-trained VGG \cite{vgg} model (fc7), and SSIM (structural similarity) \cite{WBSS04}, which is a popular image similarity metric.
    For metrics that were cheap to evaluate ($\ell_2$-distance on pixels and $\ell_2$-distance on fc7), we computed nearest neighbor distances to all candidate images and all of the original ImageNet data.
    For the more compute-intensive SSIM metric, we restricted the set of reference images to include all candidate images and the five closest ImageNet classes based on the tree distance in the WordNet hierarchy.
    We then manually reviewed nearest neighbor pairs below certain thresholds for each metric and removed any duplicates we discovered.

    To the best of our knowledge, ImageNet used only nearest neighbors in the $\ell_2$-distance to find near-duplicates \cite{bergpersonal}.
    While this difference may lead to a small change in distribution, we still decided to use multiple metrics since including images that have near-duplicates in ImageNet would be contrary to the main goal of our experiment.
    Moreover, a manual inspection of the original validation set revealed only a very small number of near-duplicates within the existing dataset.

    \subsubsection{Sampling a Clean Dataset}
    \label{sec:imagenetsampling}
    The result of collecting a candidate pool was a set of images with annotations from MTurk, most importantly the selection frequency of each image.
    In the next step, we used this candidate pool to sample a new test set that closely resembles the distribution of the existing validation set.
    There were two main difficulties in this process.

    First, the ImageNet publications do not provide the agreement thresholds for each class that were used to determine which images were valid.
    One possibility could be to run the algorithm the ImageNet authors designed to compute the agreement thresholds.
    However, this algorithm would need to be exactly specified, which is unfortunately not the case to the best of our knowledge.\footnote{To be precise: Jia Deng's PhD thesis \cite{jiadengthesis} provides a clear high-level description of their algorithm for computing agreement thresholds. However -- as is commonly the case in synopses of algorithms -- the description still omits some details such as the binning procedure or the number of images used to compute the thresholds. Since it is usually hard to exactly reconstruct a non-trivial algorithm from an informal summary, we instead decided to implement three different sampling strategies and compare their outcomes. Potential deviations from the ImageNet sampling procedure are also alleviated by the fact that our MTurk tasks always included at least a few images from the original validation set, which allowed us to calibrate our sampling strategies to match the existing ImageNet data.}

    Second, and more fundamentally, it is impossible to exactly replicate the MTurk worker population from 2010 -- 2012 with a reproducibility experiment in 2018.
    Even if we had access to the original agreement thresholds, it is unclear if they would be meaningful for our MTurk data collection (e.g., because the judgments of our annotations could be different).
    Similarly, re-running the algorithm for computing agreement thresholds could give different results with our MTurk worker population.

    So instead of attempting to directly replicate the original agreement thresholds, we instead explored three different sampling strategies.
    An important asset in this part of our experiment was that we had inserted original validation images into the MTurk tasks (see the previous subsection).
    So at least for \emph{our} MTurk worker population, we could estimate how frequently the MTurk workers select the original validation images.

    In this subsection, we describe our sampling strategy that closely matches the selection frequency distribution of the original validation set.
    The follow-up experiments in Section \ref{sec:imagenet_details} then explore the impact of this design choice in more detail.
    As we will see, the sampling strategy has significant influence on the model accuracies.

    \paragraph{Matching the Per-class Selection Frequency.} A simple approach to matching the selection frequency of the existing validation set would be to sample new images so that the mean selection frequency is the same as for the original dataset.
    However, a closer inspection of the selection frequencies reveals significant differences between the various classes.
  For instance, well-defined and well-known classes such as ``African elephant'' tend to have high selection frequencies ranging from 0.8 to 1.0.
    At the other end of the spectrum are classes with an unclear definition or easily confused alternative classes.
    For instance, the MTurk workers in our experiment often confused the class ``nail'' (the fastener) with fingernails, which led to significantly lower selection frequencies for the original validation images belonging to this class.
    In order to match these class-level details, we designed a sampling process that approximately matches the selection frequency distribution for each class.

    As a first step, we built an estimate of the per-class distribution of selection frequencies.
    For each class, we divided the annotated validation images into five histogram bins based on their selection frequency.
    These frequency bins were $[0.0, 0.2)$, $[0.2, 0.4)$, $[0.4, 0.6)$, $[0.6, 0.8)$, and $[0.8, 1.0]$.
    Intuitively, these bins correspond to a notion of image quality assessed by the MTurk workers, with the $[0.0, 0.2)$ bin containing the worst images and the $[0.8, 1.0]$ bin containing the best images.
    Normalizing the resulting histograms then yielded a distribution over these selection frequency bins for each class.

    Next, we sampled ten images for each class from our candidate pool, following the distribution given by the class-specific selection frequency histograms.
    More precisely, we first computed the target number of images for each histogram bin, and then sampled from the candidates images falling into this histogram bin uniformly at random.
    Since we only had a limited number of images for each class, this process ran out of images for a small number of classes.
    In these cases, we then sampled candidate images from the next higher bin until we found a histogram bin that still had images remaining.
    While this slightly changes the distribution, we remark that it makes our new test set easier and only affected 0.8\% of the images in the new test set.

    At the end of this sampling process, we had a test set with $10,000$ images and an average sampling frequency of $0.73$.
    This is close to the average sampling frequency of the annotated validation images~($0.71$).

    \paragraph{Final Reviewing.}
    While the methodology outlined so far closely matches the original ImageNet distribution, it is still hard to ensure that no unintended biases crept into the dataset (e.g., our MTurk workers could interpret some of the class definitions differently and select different images).
    So for quality control, we added a final reviewing step to our dataset creation pipeline.
    Its purpose was to rule out obvious biases and ensure that the dataset satisfies our quality expectations \emph{before} we ran any models on the new dataset.
    This minimizes dependencies between the new test set and the existing models.

    In the final reviewing step, the authors of this paper manually reviewed every image in the dataset.
    Appendix \ref{apx:reviewing_ui} includes a screenshot and brief description of the user interface.
    When we found an incorrect image or a near-duplicate, we removed it from the dataset.
    After a pass through the dataset, we then re-sampled new images from our candidate pool.
    In some cases, this also required new targeted Flickr searches for certain classes.
    We repeated this process until the dataset converged after 33 iterations.
    We remark that the majority of iterations only changed a small number of images.

    One potential downside of the final reviewing step is that it may lead to a distribution shift.
    However, we accepted this possibility since we view dataset correctness to be more important than minimizing distribution shift.
    In the end, a test set is only interesting if it has correct labels.
    Note also that removing incorrect images from the dataset makes it easier, which goes \emph{against} the main trend we observe (a drop in accuracy).
    Finally, we kept track of all intermediate iterations of our dataset so that we could measure the impact of this final reviewing step (see Section \ref{apx:impact_of_dataset_revisions}).
    This analysis shows that the main trends (a significant accuracy drop and an approximately linear relationship between original and new accuracy) also hold for the first iteration of the dataset without any additional reviewing.

\subsection{Model Performance Results}
\label{app:imagenetresults}
After assembling our new test sets, we evaluated a broad range of models on both the original validation set and our new test sets.
Section \ref{apx:imagenet_model_descriptions} contains a list of all models we evaluated with corresponding references and links to source code repositories.
Tables \ref{tab:imagenetv2-b-33_top1_full_results} and \ref{tab:imagenetv2-b-33_top5_full_results} show the top-1 and top-5 accuracies for our main test set \datasetb{}.
Figure \ref{fig:imagenet_plotpage} visualizes the top-1 and top-5 accuracies on all three test sets.

In the main text of the paper and Figure \ref{fig:imagenet_plotpage}, we have chosen to exclude the Fisher Vector models and show accuracies only for the convolutional neural networks (convnets).
Since the Fisher Vector models achieve significantly lower accuracy, a plot involving both model families would have sacrificed resolution among the convnets.
We decided to focus on convnets in the main text because they have become the most widely used model family on ImageNet.

Moreover, a linear model of accuracies (as shown in previous plots) is often not a good fit when the accuracies span a wide range.
Instead, a non-linear model such as a logistic or probit model can sometimes describe the data better.
Indeed, this is also the case for our data on ImageNet.
Figure \ref{fig:linear_vs_probit} shows the accuracies both on a linear scale as in the previous plots, and on a \emph{probit} scale, i.e., after applying the inverse of the Gaussian CDF to all accuracy scores.
As can be seen by comparing the two plots in Figure \ref{fig:linear_vs_probit}, the probit model is a better fit for our data.
It accurately summarizes the relationship between original and new test set accuracy for all models from both model families in our testbed.

Similar to Figure \ref{fig:imagenet_plotpage}, we also show the top-1 and top-5 accuracies for all three datasets in the probit domain in Figure \ref{fig:imagenet_probit_plotpage}.
Section \ref{sec:probitmodel} describes a possible generative model that leads to a linear fit in the probit domain as exhibited by the plots in Figures \ref{fig:linear_vs_probit} and \ref{fig:imagenet_probit_plotpage}.

\begin{figure*}[tb!]
  \centering
  \begin{subfigure}{0.48\textwidth}
    \includegraphics[width=\linewidth]{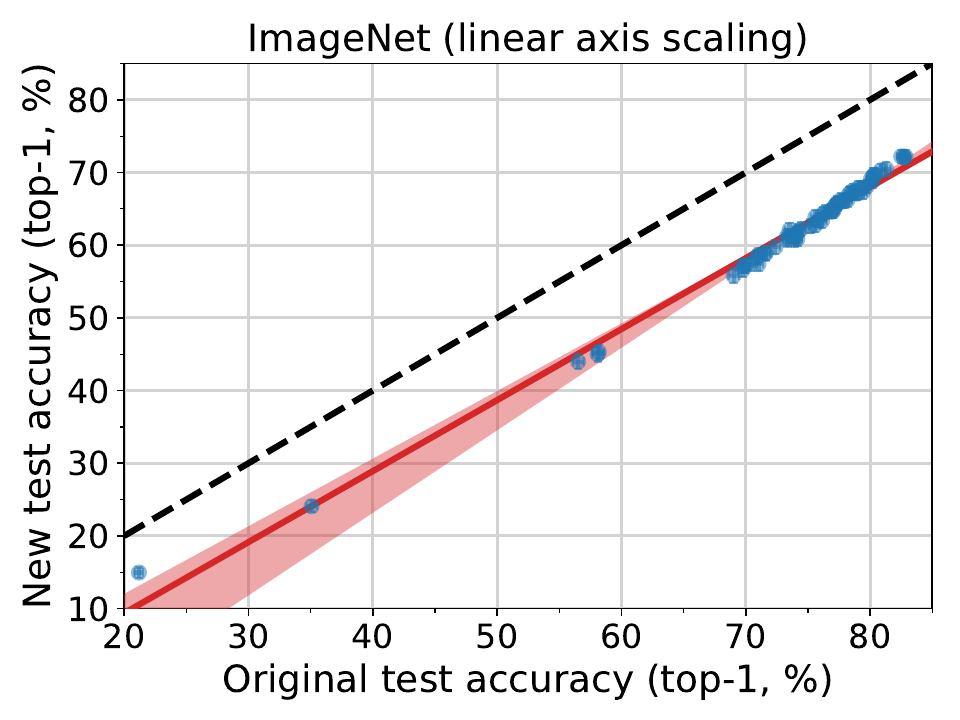}
  \end{subfigure}
  \hfill
  \begin{subfigure}{0.48\textwidth}
    \includegraphics[width=\linewidth]{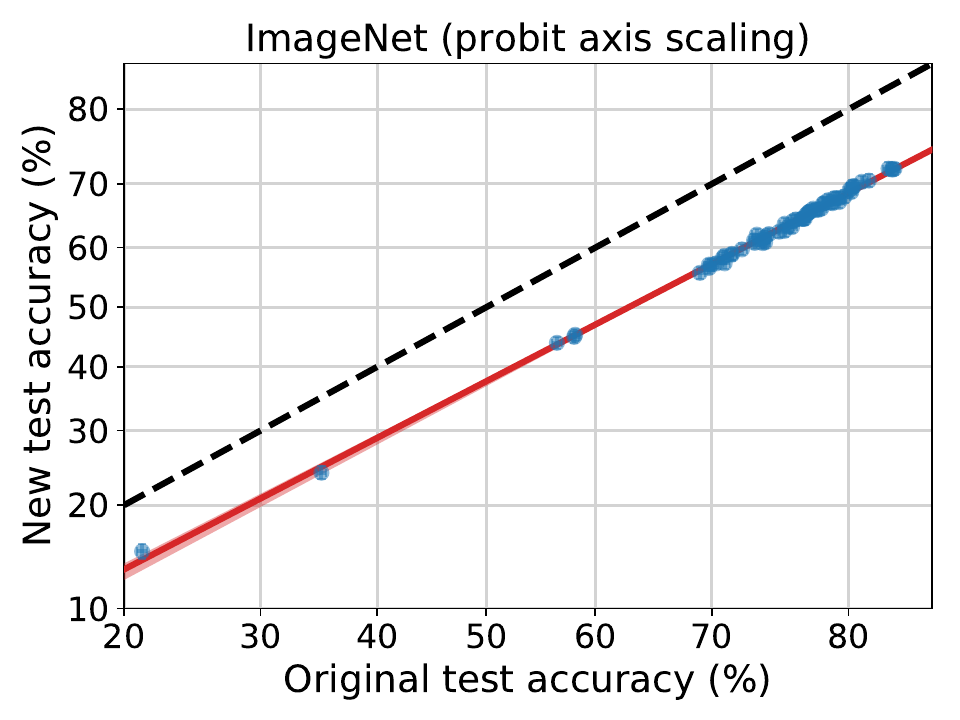}
  \end{subfigure}
  \begin{subfigure}{\textwidth}
    \vspace{-.15cm}
    \centering
    \includegraphics[width=.75\linewidth]{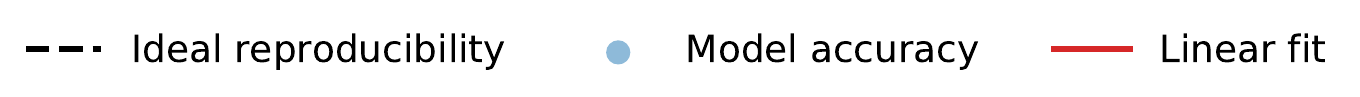}
  \end{subfigure}
  \vspace{-.6cm}
  \caption{
    Model accuracy on the original ImageNet validation set vs.\ our new test set \datasetb{}.
    Each data point corresponds to one model in our testbed (shown with 95\% Clopper-Pearson confidence intervals), and we now also include the Fisher Vector models.
    The left plot shows the model accuracies with a linear scale on the axes.
    The right plot instead uses a \emph{probit} scale, i.e., accuracy $\alpha$ appears at $\Phi^{-1}(\alpha)$, where $\Phi$ is the Gaussian CDF.
    Comparing the two plot provides evidence that the probit model is a better fit for the accuracy scores.
    Over a range of 60 percentage points, the linear fit in the probit domain accurately describes the relationship between original and new test set accuracy.
    The shaded region around the linear fit is a 95\% confidence region from 100,000 bootstrap samples.
    The confidence region is present in both plots but is significantly smaller in the right plot due to the better fit in the probit domain.
    \vspace{-.3cm}
  }
  \label{fig:linear_vs_probit}
\end{figure*}

    \subsection{Follow-up Hypotheses}
    \label{apx:imagenetfollowup}
    As for CIFAR-10, the gap between original and new accuracy is concerningly large.
    Hence we investigated multiple hypotheses for explaining this gap.
    
    \subsubsection{Cross Validation}
    \label{apx:imagenet_cross_validation}

A natural question is whether cross-validation with the existing ImageNet data could have pointed towards a significant drop in accuracy.
If adaptive overfitting to the images in the validation set is a cause for the accuracy drop, testing on different images from another cross-validation fold could produce lower accuracies.\footnote{Note however that the training images may also be affected by adaptive overfitting since the model hyperparameters are often tuned for fast training speed and high training accuracy.}
Moreover, recall that the ImageNet validation set is not a strictly i.i.d.\ sample from the same distribution as the training set (see the beginning of Section \ref{sec:imagenet_details}).
This also raises the question of how well a model would perform on a cross-validation fold from the training data.

To investigate these two effects, we conducted a cross-validation experiment with the ImageNet training and validation sets.
In order to ensure that the new cross-validation folds contain only training images, we treated the existing validation set as one fold and created five additional folds with 50,000 images each.
To this end, we drew a class-balanced sample of 250,000 images from the training set and then randomly partitioned this sample into five cross-validation folds (again class-balanced).
For each of these five folds, we added the validation set (and the other training folds) to the training data so that the size of the training set was unchanged.
We then trained one \model{resnet50} model\footnote{To save computational resources, we used the optimized training code from \url{https://github.com/fastai/imagenet-fast}. Hence the top-5 accuracy on the original validation set is 0.4\% lower than in Table \ref{tab:imagenetv2-b-33_top5_full_results}.} \cite{resnet} for each of the five training sets and evaluated them on the corresponding held-out data.
Table \ref{tab:imagenet_cross_validation} shows the resulting accuracies for each split.

\begin{table*}[h!]
  \centering
  \rowcolors{3}{gray!15}{white}
  \begin{tabular}{c  c  c  c  c  c}
  \toprule
  Dataset & \model{resnet50} Top-5 Accuracy ($\%$) \\
  \midrule
  Original validation set &  92.5 {\footnotesize \textcolor{gray}{[92.3, 92.8]}} \\
  \midrule
  Split 1 & 92.60 {\footnotesize \textcolor{gray}{[92.4, 92.8]}} \\
  Split 2 & 92.59 {\footnotesize \textcolor{gray}{[92.4, 92.8]}} \\
  Split 3 & 92.61 {\footnotesize \textcolor{gray}{[92.4, 92.8]}} \\
  Split 4 & 92.55 {\footnotesize \textcolor{gray}{[92.3, 92.8]}} \\
  Split 5 & 92.62 {\footnotesize \textcolor{gray}{[92.4, 92.9]}} \\ 
  \midrule
  New test set (\datasetb{}) & 84.7 {\footnotesize \textcolor{gray}{[83.9, 85.4]}} \\
  \bottomrule
  \end{tabular}
  \caption{\model{resnet50} accuracy on cross-validation splits created from the original ImageNet train and validation sets.
  The accuracy increase is likely caused by a small shift in distribution between the ImageNet training and validation sets.}
  \label{tab:imagenet_cross_validation}
\end{table*}

Overall, we do not see a large difference in accuracy on the new cross validation splits: all differences fall within the 95\% confidence intervals around the accuracy scores.
This is in contrast to the significantly larger accuracy drops on our new test sets.

\subsubsection{Impact of Dataset Revisions}
    \label{apx:impact_of_dataset_revisions}
    As mentioned in Section \ref{sec:imagenetsampling}, our final reviewing pass may have led to a distribution shift compared to the original ImageNet validation set.
    In general, our reviewing criterion was to blacklist images that were
    \begin{itemize}
    \item not representative of the target class,
    \item cartoons, paintings, drawings, or renderings,
    \item significantly different in distribution from the original ImageNet validation set,
    \item unclear, blurry, severely occluded, overly edited, or including only a small target object.
    \end{itemize}
    For each class, our reviewing UI (screenshot in Appendix \ref{apx:reviewing_ui}) displayed a random sample of ten original validation images directly next to the ten new candidate images currently chosen.
    At least to some extent, this allowed us to detect and correct distribution shifts between the original validation set and our candidate pool.
    As a concrete example, we noted in one revision of our dataset that approximately half of the images for ``great white shark'' were not live sharks in the water but models in museums or statues outside.
    In contrast, the ImageNet validation set had fewer examples of such artificial sharks.
    Hence we decided to remove some non-live sharks from our candidate pool and sampled new shark images as a replacement in the dataset.

    Unfortunately, some of these reviewing choices are subjective.
    However, such choices are often an inherent part of creating a dataset and it is unclear whether a more ``hands-off'' approach would lead to more meaningful conclusions.
    For instance, if the drop in accuracy was mainly caused by a distribution shift that is easy to identify and correct (e.g., an increase in black-and-white images), the resulting drop may not be an interesting phenomenon (beyond counting black-and-white images).
    Hence we decided to \emph{both} remove distribution shifts that we found easy to identify visually, and also to measure the effect of these interventions.

    Our reviewing process was iterative, i.e., we made a full pass over every incomplete class in a given dataset revision before sampling new images to fill the resulting gaps.
    Each time we re-sampled our dataset, we saved the current list of images in our version control system.
    This allowed us to track the datasets over time and later measure the model accuracy for each dataset revision.
    We remark that we only computed model accuracies on intermediate revisions after we had arrived at the final revision of the corresponding dataset.

    Figure \ref{fig:imagenetv2-b-reviewing} plots the top-1 accuracy of a \model{resnet50} model versus the dataset revision for our new \datasetb{} test set.
    Overall, reviewing improved model accuracy by about 4\% for this dataset.
    This is evidence that our manual reviewing did not cause the drop in accuracy between the original and new dataset.

    \begin{figure*}[h!]
      \centering
      \includegraphics[height=0.3\textheight]{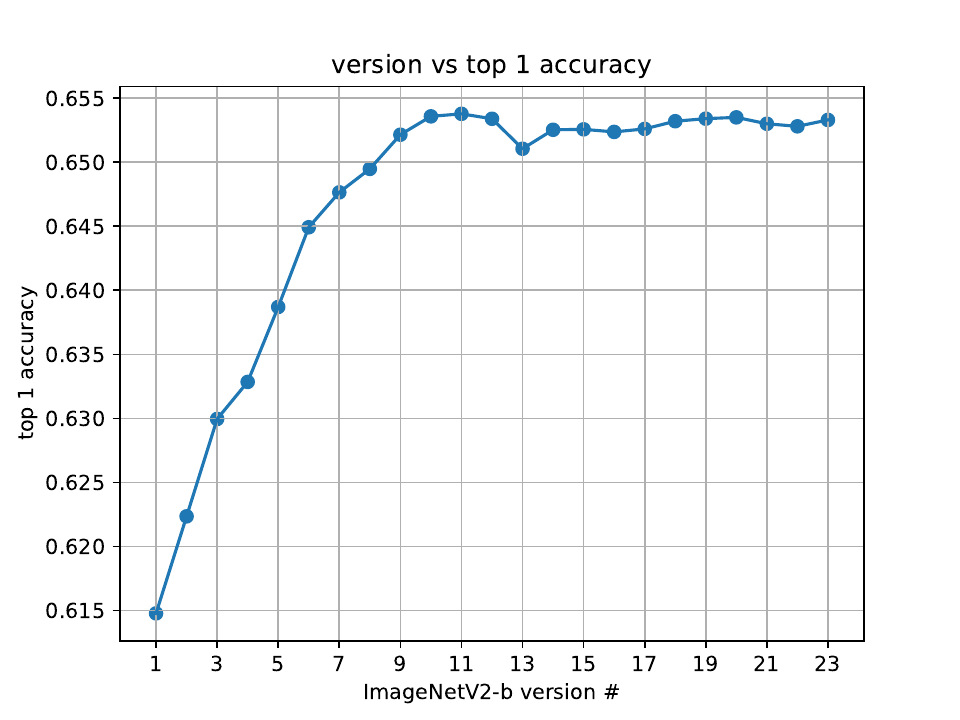}
      \includegraphics[height=0.3\textheight]{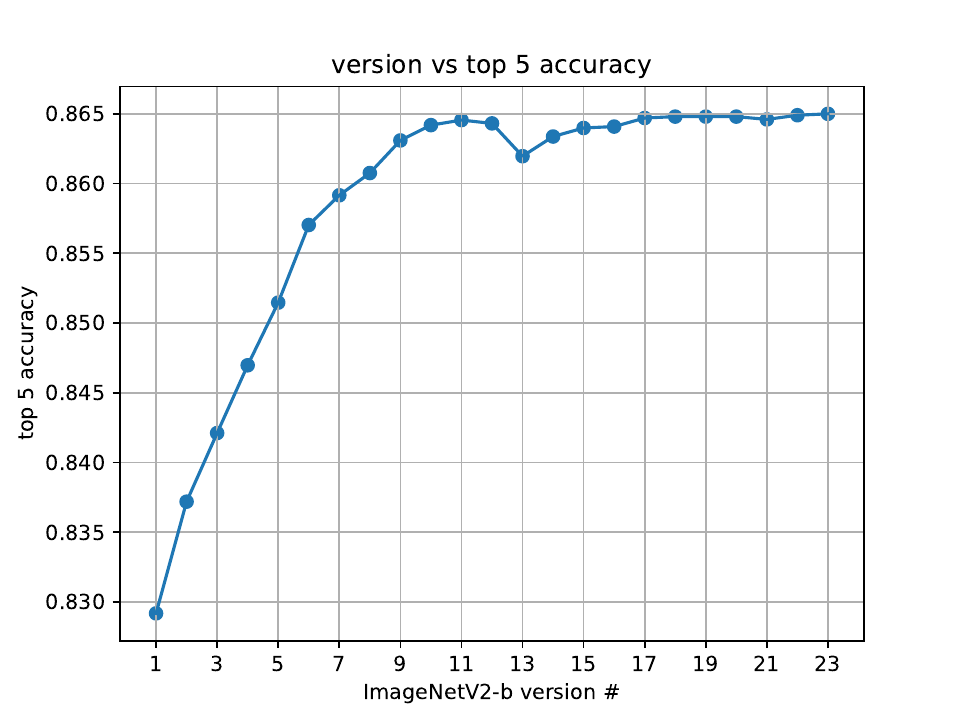}
      \includegraphics[height=0.3\textheight]{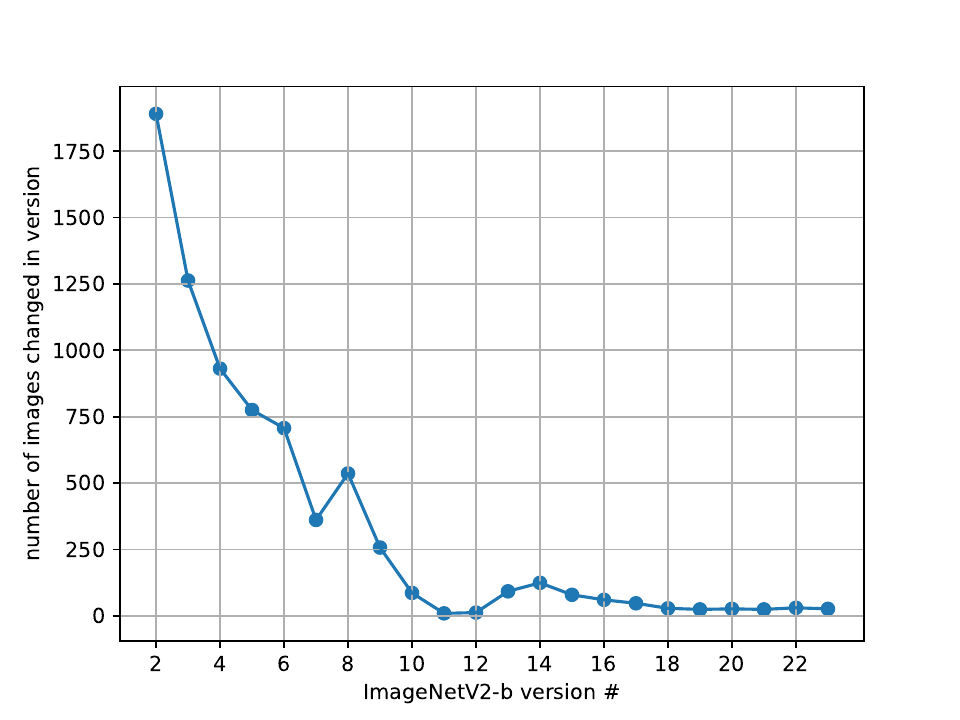}
      \caption{Impact of the reviewing passes on the accuracy of a \model{resnet152} on our new \datasetb{} test set.
        The revision numbers correspond to the chronological ordering in which we created the dataset revisions \label{fig:imagenetv2-b-reviewing}}
    \end{figure*}
    
    In addition, we also investigated whether the linear relationship between original and new test accuracy was affected by our final reviewing passes.
    To this end, we evaluated our model testbed on the first revision of our \datasetb{} test set.
    As can be seen in Figure \ref{fig:imagenet_v1}, the resulting accuracies still show a good linear fit that is of similar quality as in Figure \ref{fig:imagenet_plotpage}.
    This shows that the linear relationship between the test accuracies is not a result of our reviewing intervention.

  \begin{figure*}[ht!]
  \centering
  \begin{subfigure}{0.48\textwidth}
    \includegraphics[width=\linewidth]{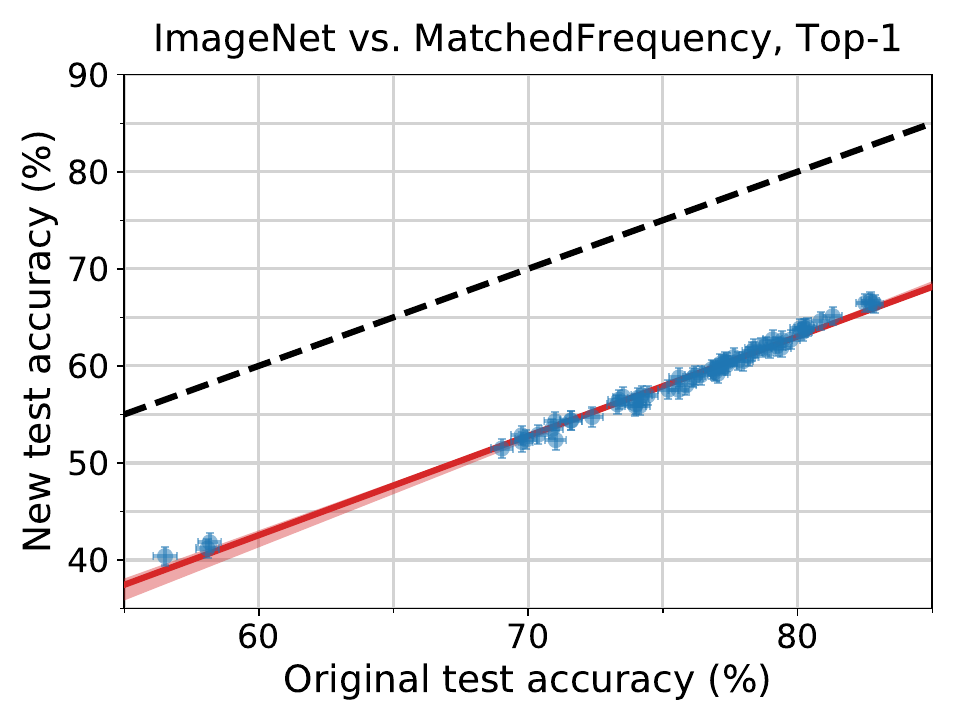}
  \end{subfigure}
  \hfill
  \begin{subfigure}{0.48\textwidth}
    \includegraphics[width=\linewidth]{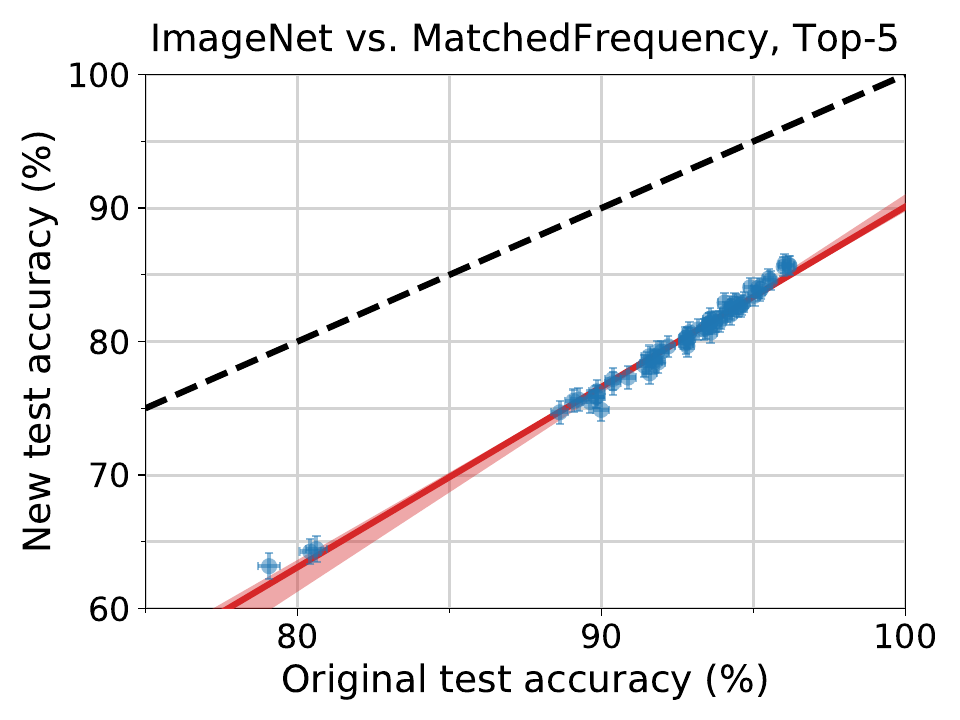}
  \end{subfigure}
  \begin{subfigure}{\textwidth}
    \vspace{-.15cm}
    \centering
    \includegraphics[width=.75\linewidth]{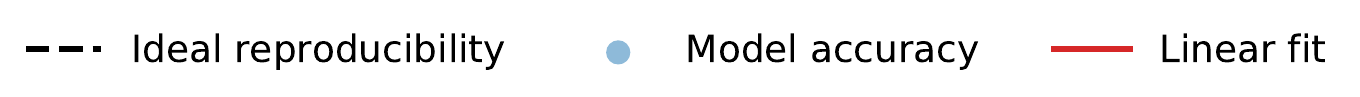}
  \end{subfigure}
  \vspace{-.6cm}
  \caption{Model accuracy on the original ImageNet validation set vs.\ accuracy on \emph{the first revision} of our \datasetb{} test set.
    Each data point corresponds to one model in our testbed (shown with 95\% Clopper-Pearson confidence intervals).
    The red shaded region is a 95\% confidence region for the linear fit from 100,000 bootstrap samples.
    The plots show that the linear relationship between original and new test accuracy also occurs without our final dataset reviewing step.
  The accuracy plots for the final revision of \datasetb{} can be found in Figure \ref{fig:imagenet_plotpage}.}
  \label{fig:imagenet_v1}
\end{figure*}

\subsection{Additional Figures, Tables, and Lists}
In this appendix we provide large figures etc.\ that did not fit into the preceding sections about our ImageNet experiments.

\subsubsection{MTurk User Interfaces}
\label{apx:mturk_ui}
For comparison, we include the original ImageNet MTurk user interface (UI) in Figure \ref{fig:imagenet_mturk_ui} and the MTurk UI we used in our experiments in Figure \ref{fig:imagenetv2_mturk_ui}.
Each UI corresponds to one task for the MTurk workers, which consists of 48 images in both cases.
In contrast to the original ImageNet UI, our UI takes up more than one screen.
This requires the MTurk workers to scroll but also provides more details in the images.
While the task descriptions are not exactly the same, they are very similar and contain the same directions (e.g., both descriptions ask the MTurk workers to exclude drawings or paintings).

\begin{figure*}[h!]
    \centering
    \includegraphics[width=\textwidth]{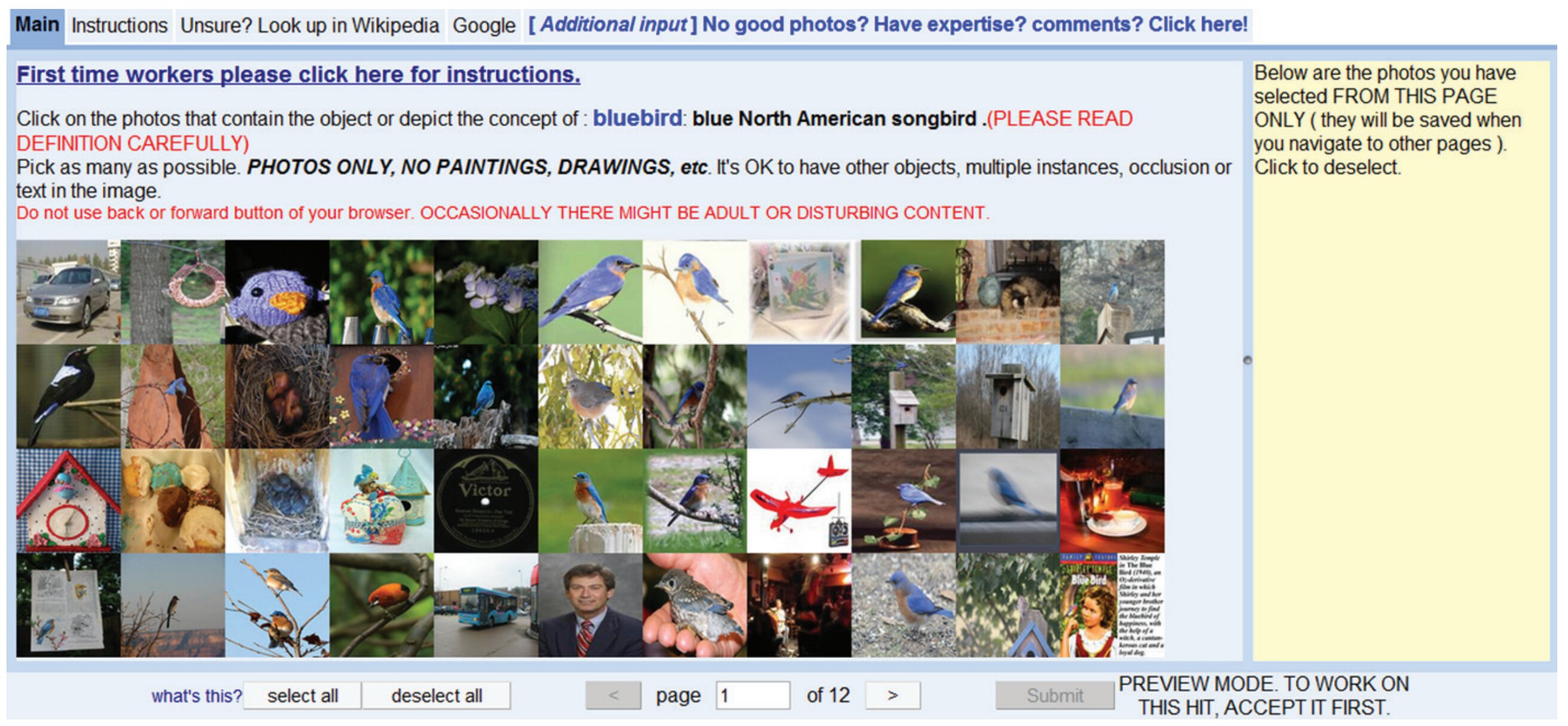}
    \caption{The user interface employed in the original ImageNet collection process for the labeling tasks on Amazon Mechanical Turk.}
    \label{fig:imagenet_mturk_ui}
\end{figure*}

\begin{figure*}[h!]
    \centering
    \frame{\includegraphics[width=.9\textwidth]{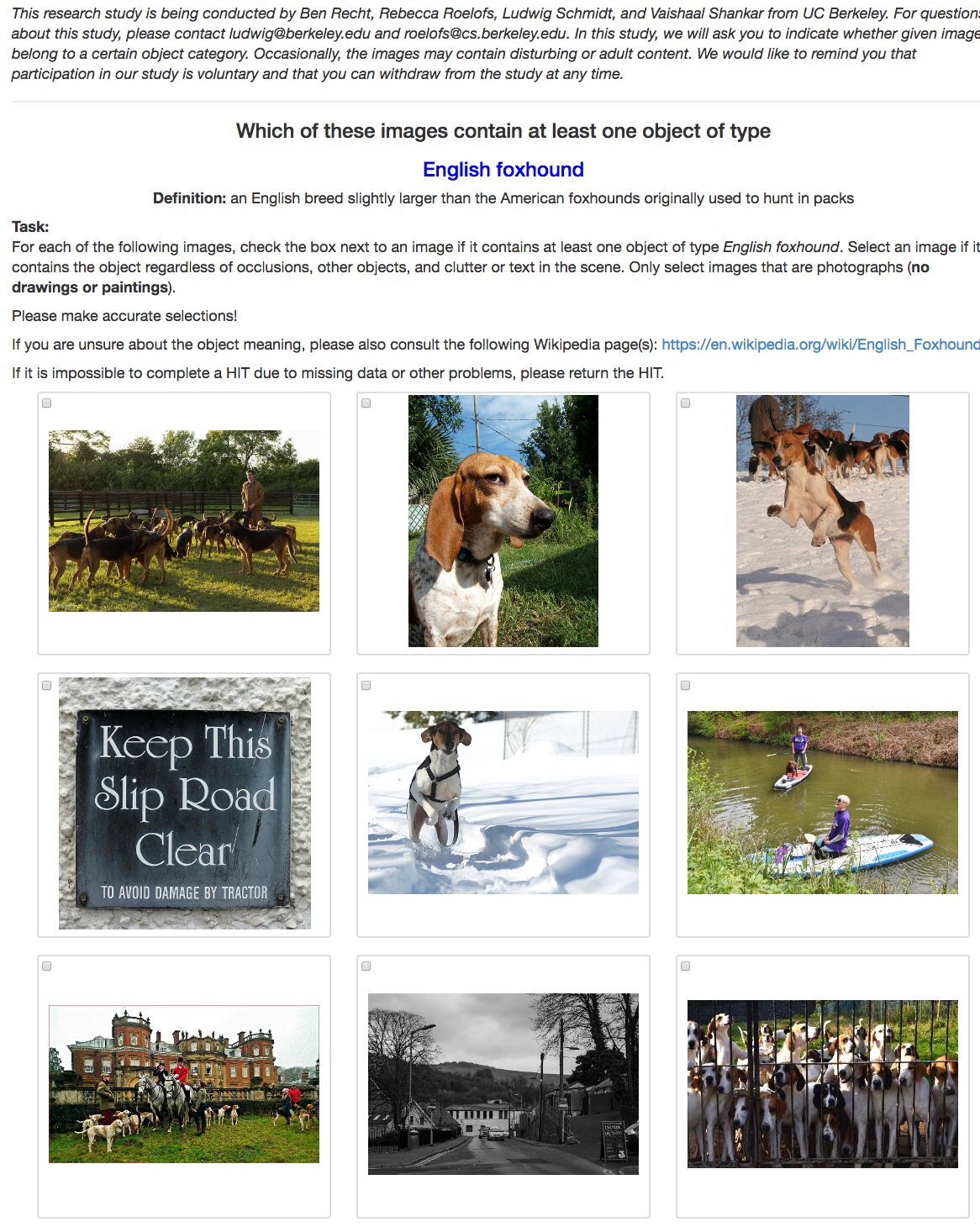}}
    \caption{Our user interface for labeling tasks on Amazon Mechanical Turk.
    The screenshot here omits the scroll bar and shows only a subset of the entire MTurk task.
    As in the ImageNet UI, the full task involves a grid of 48 images.}
    \label{fig:imagenetv2_mturk_ui}
\end{figure*}

\subsubsection{User Interface for our Final Reviewing Process}
\label{apx:reviewing_ui}
Figure \ref{fig:imagenetv2_review_ui} shows a screenshot of the reviewing UI that the paper authors used to manually review the new ImageNet datasets.
At the top, the UI displays the wnid (``n01667114''), the synset (\textbf{mud turtle}), and the gloss.
Next, a grid of 20 images is shown in 4 rows.

The top two rows correspond to the candidate images currently sampled for the new dataset.
Below each image, our UI shows a unique identifier for the image and the date the image was taken.
There is also a check box to blacklist any incorrect images.
In addition, there is a check box for each image to add it to the blacklist of incorrect images.
If an image is added to the blacklist, it will be removed in the next revision of the dataset and replaced with a new image from the candidate pools.
The candidate images are sorted by the date they were taken, which makes it easier to spot and remove near-duplicates.
Images are marked as near-duplicates by adding their identifier to the ``Near-duplicate set'' text field.

The bottom two rows correspond to a random sample of images from the validation set that belong to the target class.
We display these images to make it easier to detect and correct for distribution shifts between our new test sets and the original ImageNet validation dataset.

\begin{figure*}[ht!]
    \centering
    \frame{\includegraphics[width=\textwidth]{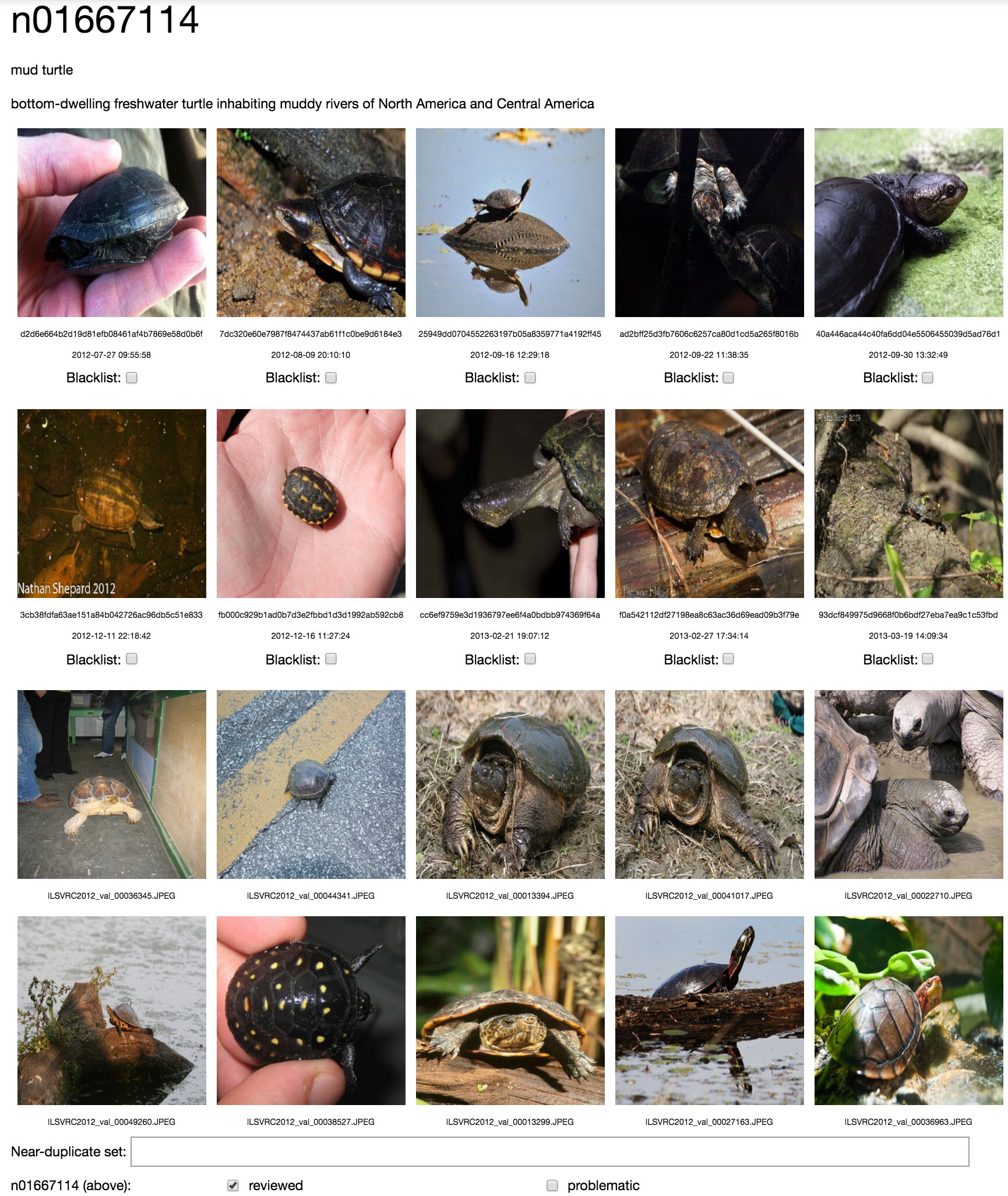}}
    \caption{The user interface we built to review dataset revisions and remove incorrect or near duplicate images.
    This user interface was not used for MTurk but only in the final dataset review step conducted by the authors of this paper.}
    \label{fig:imagenetv2_review_ui}
\end{figure*}

\subsubsection{Full List of Models Evaluated on ImageNet}
\label{apx:imagenet_model_descriptions}
The following list contains all models we evaluated on ImageNet with references and links to the corresponding source code.
\begin{enumerate}
\item \model{alexnet} \cite{alexnet} \url{https://github.com/Cadene/pretrained-models.pytorch}
\item \model{bninception} \cite{inceptionv2}  \url{https://github.com/Cadene/pretrained-models.pytorch}
\item \model{cafferesnet101} \cite{resnet} \url{https://github.com/Cadene/pretrained-models.pytorch}
\item \model{densenet121} \cite{densenet} \url{https://github.com/Cadene/pretrained-models.pytorch}
\item \model{densenet161} \cite{densenet}\url{https://github.com/Cadene/pretrained-models.pytorch}
\item \model{densenet169} \cite{densenet} \url{https://github.com/Cadene/pretrained-models.pytorch}
\item \model{densenet201} \cite{densenet} \url{https://github.com/Cadene/pretrained-models.pytorch}
\item \model{dpn107} \cite{dpn} \url{https://github.com/Cadene/pretrained-models.pytorch}
\item \model{dpn131} \cite{dpn} \url{https://github.com/Cadene/pretrained-models.pytorch}
\item \model{dpn68b} \cite{dpn}\url{https://github.com/Cadene/pretrained-models.pytorch}
\item \model{dpn68} \cite{dpn} \url{https://github.com/Cadene/pretrained-models.pytorch}
\item \model{dpn92} \cite{dpn} \url{https://github.com/Cadene/pretrained-models.pytorch}
\item \model{dpn98} \cite{dpn} \url{https://github.com/Cadene/pretrained-models.pytorch}
\item \model{fbresnet152} \cite{resnet} \url{https://github.com/tensorflow/models/tree/master/research/slim/}
\item \model{fv\_4k} \cite{fishervectors,xrce} \url{https://github.com/modestyachts/nondeep} FisherVector model using SIFT, local color statistic features, and 16 GMM centers.
\item \model{fv\_16k} \cite{fishervectors,xrce} \url{https://github.com/modestyachts/nondeep} FisherVector model using SIFT, local color statistic features, and 64 GMM centers.
\item \model{fv\_64k} \cite{fishervectors,xrce} \url{https://github.com/modestyachts/nondeep} FisherVector model using SIFT, local color statistic features, and 256 GMM centers.
\item \model{inception\_resnet\_v2\_tf} \cite{inceptionv4} \url{https://github.com/tensorflow/models/tree/master/research/slim/}
\item \model{inception\_v1\_tf} \cite{inceptionv1} \url{https://github.com/tensorflow/models/tree/master/research/slim/}
\item \model{inception\_v2\_tf} \cite{inceptionv2} \url{https://github.com/tensorflow/models/tree/master/research/slim/}
\item \model{inception\_v3\_tf} \cite{inceptionv3} \url{https://github.com/tensorflow/models/tree/master/research/slim/}
\item \model{inception\_v3} \cite{inceptionv3} \url{https://github.com/Cadene/pretrained-models.pytorch}
\item \model{inception\_v4\_tf} \cite{inceptionv4} \url{https://github.com/tensorflow/models/tree/master/research/slim/}
\item \model{inceptionresnetv2} \cite{inceptionv2}  \url{https://github.com/Cadene/pretrained-models.pytorch}
\item \model{inceptionv3} \cite{inceptionv3} \url{https://github.com/Cadene/pretrained-models.pytorch}
\item \model{inceptionv4} \cite{inceptionv4} \url{https://github.com/Cadene/pretrained-models.pytorch}
\item \model{mobilenet\_v1\_tf} \cite{mobilenet} \url{https://github.com/tensorflow/models/tree/master/research/slim/}
\item \model{nasnet\_large\_tf} \cite{nas} \url{https://github.com/tensorflow/models/tree/master/research/slim/}
\item \model{nasnet\_mobile\_tf} \cite{nas} \url{https://github.com/tensorflow/models/tree/master/research/slim/}
\item \model{nasnetalarge} \cite{nas} \url{https://github.com/Cadene/pretrained-models.pytorch}
\item \model{nasnetamobile} \cite{nas} \url{https://github.com/Cadene/pretrained-models.pytorch}
\item \model{pnasnet5large} \cite{pnasnet} \url{https://github.com/Cadene/pretrained-models.pytorch}
\item \model{pnasnet\_large\_tf} \cite{pnasnet} \url{https://github.com/tensorflow/models/tree/master/research/slim/}
\item \model{pnasnet\_mobile\_tf} \cite{pnasnet} \url{https://github.com/tensorflow/models/tree/master/research/slim/}
\item \model{polynet} \cite{polynet}  \url{https://github.com/Cadene/pretrained-models.pytorch}
\item \model{resnet101} \cite{resnet} \url{https://github.com/Cadene/pretrained-models.pytorch}
\item \model{resnet152} \cite{resnet} \url{https://github.com/Cadene/pretrained-models.pytorch}
\item \model{resnet18} \cite{resnet} \url{https://github.com/Cadene/pretrained-models.pytorch}
\item \model{resnet34} \cite{resnet} \url{https://github.com/Cadene/pretrained-models.pytorch}
\item \model{resnet50} \cite{resnet} \url{https://github.com/Cadene/pretrained-models.pytorch}
\item \model{resnet\_v1\_101\_tf} \cite{resnet} \url{https://github.com/tensorflow/models/tree/master/research/slim/}
\item \model{resnet\_v1\_152\_tf} \cite{resnet} \url{https://github.com/tensorflow/models/tree/master/research/slim/}
\item \model{resnet\_v1\_50\_tf} \cite{resnet} \url{https://github.com/tensorflow/models/tree/master/research/slim/}
\item \model{resnet\_v2\_101\_tf} \cite{resnet_preact} \url{https://github.com/tensorflow/models/tree/master/research/slim/}
\item \model{resnet\_v2\_152\_tf} \cite{resnet_preact} \url{https://github.com/tensorflow/models/tree/master/research/slim/}
\item \model{resnet\_v2\_50\_tf} \cite{resnet_preact} \url{https://github.com/tensorflow/models/tree/master/research/slim/}
\item \model{resnext101\_32x4d} \cite{resnext} \url{https://github.com/Cadene/pretrained-models.pytorch}
\item \model{resnext101\_64x4d} \cite{resnext} \url{https://github.com/Cadene/pretrained-models.pytorch}
\item \model{se\_resnet101} \cite{senet} \url{https://github.com/Cadene/pretrained-models.pytorch}
\item \model{se\_resnet152} \cite{senet} \url{https://github.com/Cadene/pretrained-models.pytorch}
\item \model{se\_resnet50} \cite{senet} \url{https://github.com/Cadene/pretrained-models.pytorch}
\item \model{se\_resnext101\_32x4d} \cite{senet} \url{https://github.com/Cadene/pretrained-models.pytorch}
\item \model{se\_resnext50\_32x4d} \cite{senet} \url{https://github.com/Cadene/pretrained-models.pytorch}
\item \model{senet154} \cite{senet} \url{https://github.com/Cadene/pretrained-models.pytorch}
\item \model{squeezenet1\_0} \cite{squeezenet} \url{https://github.com/Cadene/pretrained-models.pytorch}
\item \model{squeezenet1\_1} \cite{squeezenet} \url{https://github.com/Cadene/pretrained-models.pytorch}
\item \model{vgg11\_bn} \cite{inceptionv2} \url{https://github.com/Cadene/pretrained-models.pytorch}
\item \model{vgg11} \cite{vgg} \url{https://github.com/Cadene/pretrained-models.pytorch}
\item \model{vgg13\_bn} \cite{inceptionv2} \url{https://github.com/Cadene/pretrained-models.pytorch}
\item \model{vgg13} \cite{vgg} \url{https://github.com/Cadene/pretrained-models.pytorch}
\item \model{vgg16\_bn} \cite{inceptionv2} \url{https://github.com/Cadene/pretrained-models.pytorch}
\item \model{vgg16} \cite{vgg} \url{https://github.com/Cadene/pretrained-models.pytorch}
\item \model{vgg19\_bn} \cite{inceptionv2} \url{https://github.com/Cadene/pretrained-models.pytorch}
\item \model{vgg19} \cite{vgg} \url{https://github.com/Cadene/pretrained-models.pytorch}
\item \model{vgg\_16\_tf} \cite{vgg} \url{https://github.com/tensorflow/models/tree/master/research/slim/}
\item \model{vgg\_19\_tf} \cite{vgg} \url{https://github.com/tensorflow/models/tree/master/research/slim/}
\item \model{xception} \cite{xception} \url{https://github.com/Cadene/pretrained-models.pytorch}
\end{enumerate}

\subsubsection{Full Results Tables}
\label{sec:imagenettable}
Tables \ref{tab:imagenetv2-b-33_top1_full_results} and \ref{tab:imagenetv2-b-33_top5_full_results} contain the detailed accuracy scores (top-1 and top-5, respectively) for the original ImageNet validation set and our main new test set \datasetb{}.
Tables \ref{tab:imagenetv2-a-44_top1_full_results} -- \ref{tab:imagenetv2-c-12_top5_full_results} contain the accuracy scores for our \dataseta{} and \datasetc{} test sets.

\newpage
\clearpage

\begin{table*}[ht!]
  \rowcolors{3}{white}{gray!15}
  \caption{Top-1 model accuracy on the original ImageNet validation set and our new test set \datasetb{}.
  $\Delta$ Rank is the relative difference in the ranking from the original test set to the new test set.
  For example, $\Delta \text{Rank} = -2$ means that a model dropped by two places on the new test set compared to the original test set.
  The confidence intervals are 95\% Clopper-Pearson intervals.
  Due to space constraints, references for the models can be found in Appendix \ref{apx:imagenet_model_descriptions}.
  The second part of the table can be found on the following page.
  }
  \label{tab:imagenetv2-b-33_top1_full_results}
  \begin{tabular}{rp{4.75cm}rrrrr}
\toprule 
\multicolumn{7}{c}{\textbf{ImageNet Top-1 \datasetb}} \\ 
\midrule
\multicolumn{1}{l}{Orig.} &                               &                                        &                                          &    & \multicolumn{1}{l}{New} &  \\ 
 \multicolumn{1}{l}{Rank} & Model & Orig. Accuracy & New Accuracy & Gap & \multicolumn{1}{l}{Rank} & $\Delta$ Rank \\
\midrule
 1 &  \model{pnasnet\_large\_tf} &  82.9 {\footnotesize \textcolor{gray}{[82.5, 83.2]}} &  72.2 {\footnotesize \textcolor{gray}{[71.3, 73.1]}} &  10.7 &  3 &  -2 \\
 2 &  \model{pnasnet5large} &  82.7 {\footnotesize \textcolor{gray}{[82.4, 83.1]}} &  72.1 {\footnotesize \textcolor{gray}{[71.2, 73.0]}} &  10.7 &  4 &  -2 \\
 3 &  \model{nasnet\_large\_tf} &  82.7 {\footnotesize \textcolor{gray}{[82.4, 83.0]}} &  72.2 {\footnotesize \textcolor{gray}{[71.3, 73.1]}} &  10.5 &  2 &  1 \\
 4 &  \model{nasnetalarge} &  82.5 {\footnotesize \textcolor{gray}{[82.2, 82.8]}} &  72.2 {\footnotesize \textcolor{gray}{[71.3, 73.1]}} &  10.3 &  1 &  3 \\
 5 &  \model{senet154} &  81.3 {\footnotesize \textcolor{gray}{[81.0, 81.6]}} &  70.5 {\footnotesize \textcolor{gray}{[69.6, 71.4]}} &  10.8 &  5 &  0 \\
 6 &  \model{polynet} &  80.9 {\footnotesize \textcolor{gray}{[80.5, 81.2]}} &  70.3 {\footnotesize \textcolor{gray}{[69.4, 71.2]}} &  10.5 &  6 &  0 \\
 7 &  \model{inception\_resnet\_v2\_tf} &  80.4 {\footnotesize \textcolor{gray}{[80.0, 80.7]}} &  69.7 {\footnotesize \textcolor{gray}{[68.7, 70.6]}} &  10.7 &  7 &  0 \\
 8 &  \model{inceptionresnetv2} &  80.3 {\footnotesize \textcolor{gray}{[79.9, 80.6]}} &  69.6 {\footnotesize \textcolor{gray}{[68.7, 70.5]}} &  10.6 &  8 &  0 \\
 9 &  \model{se\_resnext101\_32x4d} &  80.2 {\footnotesize \textcolor{gray}{[79.9, 80.6]}} &  69.3 {\footnotesize \textcolor{gray}{[68.4, 70.2]}} &  10.9 &  9 &  0 \\
 10 &  \model{inception\_v4\_tf} &  80.2 {\footnotesize \textcolor{gray}{[79.8, 80.5]}} &  68.8 {\footnotesize \textcolor{gray}{[67.9, 69.7]}} &  11.4 &  11 &  -1 \\
 11 &  \model{inceptionv4} &  80.1 {\footnotesize \textcolor{gray}{[79.7, 80.4]}} &  69.1 {\footnotesize \textcolor{gray}{[68.2, 70.0]}} &  10.9 &  10 &  1 \\
 12 &  \model{dpn107} &  79.7 {\footnotesize \textcolor{gray}{[79.4, 80.1]}} &  68.1 {\footnotesize \textcolor{gray}{[67.2, 69.0]}} &  11.7 &  12 &  0 \\
 13 &  \model{dpn131} &  79.4 {\footnotesize \textcolor{gray}{[79.1, 79.8]}} &  67.9 {\footnotesize \textcolor{gray}{[67.0, 68.8]}} &  11.5 &  13 &  0 \\
 14 &  \model{dpn92} &  79.4 {\footnotesize \textcolor{gray}{[79.0, 79.8]}} &  67.3 {\footnotesize \textcolor{gray}{[66.3, 68.2]}} &  12.1 &  17 &  -3 \\
 15 &  \model{dpn98} &  79.2 {\footnotesize \textcolor{gray}{[78.9, 79.6]}} &  67.8 {\footnotesize \textcolor{gray}{[66.9, 68.8]}} &  11.4 &  15 &  0 \\
 16 &  \model{se\_resnext50\_32x4d} &  79.1 {\footnotesize \textcolor{gray}{[78.7, 79.4]}} &  67.9 {\footnotesize \textcolor{gray}{[66.9, 68.8]}} &  11.2 &  14 &  2 \\
 17 &  \model{resnext101\_64x4d} &  79.0 {\footnotesize \textcolor{gray}{[78.6, 79.3]}} &  67.1 {\footnotesize \textcolor{gray}{[66.2, 68.0]}} &  11.9 &  20 &  -3 \\
 18 &  \model{xception} &  78.8 {\footnotesize \textcolor{gray}{[78.5, 79.2]}} &  67.2 {\footnotesize \textcolor{gray}{[66.2, 68.1]}} &  11.7 &  18 &  0 \\
 19 &  \model{se\_resnet152} &  78.7 {\footnotesize \textcolor{gray}{[78.3, 79.0]}} &  67.5 {\footnotesize \textcolor{gray}{[66.6, 68.5]}} &  11.1 &  16 &  3 \\
 20 &  \model{se\_resnet101} &  78.4 {\footnotesize \textcolor{gray}{[78.0, 78.8]}} &  67.2 {\footnotesize \textcolor{gray}{[66.2, 68.1]}} &  11.2 &  19 &  1 \\
 21 &  \model{resnet152} &  78.3 {\footnotesize \textcolor{gray}{[77.9, 78.7]}} &  67.0 {\footnotesize \textcolor{gray}{[66.1, 67.9]}} &  11.3 &  21 &  0 \\
 22 &  \model{resnext101\_32x4d} &  78.2 {\footnotesize \textcolor{gray}{[77.8, 78.5]}} &  66.2 {\footnotesize \textcolor{gray}{[65.3, 67.2]}} &  11.9 &  22 &  0 \\
 23 &  \model{inception\_v3\_tf} &  78.0 {\footnotesize \textcolor{gray}{[77.6, 78.3]}} &  66.1 {\footnotesize \textcolor{gray}{[65.1, 67.0]}} &  11.9 &  24 &  -1 \\
 24 &  \model{resnet\_v2\_152\_tf} &  77.8 {\footnotesize \textcolor{gray}{[77.4, 78.1]}} &  66.1 {\footnotesize \textcolor{gray}{[65.1, 67.0]}} &  11.7 &  25 &  -1 \\
 25 &  \model{se\_resnet50} &  77.6 {\footnotesize \textcolor{gray}{[77.3, 78.0]}} &  66.2 {\footnotesize \textcolor{gray}{[65.3, 67.2]}} &  11.4 &  23 &  2 \\
 26 &  \model{fbresnet152} &  77.4 {\footnotesize \textcolor{gray}{[77.0, 77.8]}} &  65.8 {\footnotesize \textcolor{gray}{[64.9, 66.7]}} &  11.6 &  26 &  0 \\
 27 &  \model{resnet101} &  77.4 {\footnotesize \textcolor{gray}{[77.0, 77.7]}} &  65.7 {\footnotesize \textcolor{gray}{[64.7, 66.6]}} &  11.7 &  28 &  -1 \\
 28 &  \model{inceptionv3} &  77.3 {\footnotesize \textcolor{gray}{[77.0, 77.7]}} &  65.7 {\footnotesize \textcolor{gray}{[64.8, 66.7]}} &  11.6 &  27 &  1 \\
 29 &  \model{inception\_v3} &  77.2 {\footnotesize \textcolor{gray}{[76.8, 77.6]}} &  65.4 {\footnotesize \textcolor{gray}{[64.5, 66.4]}} &  11.8 &  29 &  0 \\
 30 &  \model{densenet161} &  77.1 {\footnotesize \textcolor{gray}{[76.8, 77.5]}} &  65.3 {\footnotesize \textcolor{gray}{[64.4, 66.2]}} &  11.8 &  30 &  0 \\
 31 &  \model{dpn68b} &  77.0 {\footnotesize \textcolor{gray}{[76.7, 77.4]}} &  64.7 {\footnotesize \textcolor{gray}{[63.7, 65.6]}} &  12.4 &  32 &  -1 \\
 32 &  \model{resnet\_v2\_101\_tf} &  77.0 {\footnotesize \textcolor{gray}{[76.6, 77.3]}} &  64.6 {\footnotesize \textcolor{gray}{[63.7, 65.6]}} &  12.3 &  34 &  -2 \\
 33 &  \model{densenet201} &  76.9 {\footnotesize \textcolor{gray}{[76.5, 77.3]}} &  64.7 {\footnotesize \textcolor{gray}{[63.7, 65.6]}} &  12.2 &  31 &  2 \\
\bottomrule
\end{tabular}

  \centering
\end{table*}
\begin{table*}[ht!]
  \rowcolors{3}{white}{gray!15}
  \begin{tabular}{rp{4.75cm}rrrrr}
\toprule 
\multicolumn{7}{c}{\textbf{ImageNet Top-1 \datasetb}} \\ 
\midrule
\multicolumn{1}{l}{Orig.} &                               &                                        &                                          &    & \multicolumn{1}{l}{New} &  \\ 
 \multicolumn{1}{l}{Rank} & Model & Orig. Accuracy & New Accuracy & Gap & \multicolumn{1}{l}{Rank} & $\Delta$ Rank \\
\midrule
 34 &  \model{resnet\_v1\_152\_tf} &  76.8 {\footnotesize \textcolor{gray}{[76.4, 77.2]}} &  64.6 {\footnotesize \textcolor{gray}{[63.7, 65.6]}} &  12.2 &  33 &  1 \\
 35 &  \model{resnet\_v1\_101\_tf} &  76.4 {\footnotesize \textcolor{gray}{[76.0, 76.8]}} &  64.5 {\footnotesize \textcolor{gray}{[63.6, 65.5]}} &  11.9 &  35 &  0 \\
 36 &  \model{cafferesnet101} &  76.2 {\footnotesize \textcolor{gray}{[75.8, 76.6]}} &  64.3 {\footnotesize \textcolor{gray}{[63.4, 65.2]}} &  11.9 &  36 &  0 \\
 37 &  \model{resnet50} &  76.1 {\footnotesize \textcolor{gray}{[75.8, 76.5]}} &  63.3 {\footnotesize \textcolor{gray}{[62.4, 64.3]}} &  12.8 &  39 &  -2 \\
 38 &  \model{dpn68} &  75.9 {\footnotesize \textcolor{gray}{[75.5, 76.2]}} &  63.4 {\footnotesize \textcolor{gray}{[62.5, 64.4]}} &  12.4 &  38 &  0 \\
 39 &  \model{densenet169} &  75.6 {\footnotesize \textcolor{gray}{[75.2, 76.0]}} &  63.9 {\footnotesize \textcolor{gray}{[62.9, 64.8]}} &  11.7 &  37 &  2 \\
 40 &  \model{resnet\_v2\_50\_tf} &  75.6 {\footnotesize \textcolor{gray}{[75.2, 76.0]}} &  62.7 {\footnotesize \textcolor{gray}{[61.8, 63.7]}} &  12.9 &  40 &  0 \\
 41 &  \model{resnet\_v1\_50\_tf} &  75.2 {\footnotesize \textcolor{gray}{[74.8, 75.6]}} &  62.6 {\footnotesize \textcolor{gray}{[61.6, 63.5]}} &  12.6 &  41 &  0 \\
 42 &  \model{densenet121} &  74.4 {\footnotesize \textcolor{gray}{[74.0, 74.8]}} &  62.2 {\footnotesize \textcolor{gray}{[61.3, 63.2]}} &  12.2 &  42 &  0 \\
 43 &  \model{vgg19\_bn} &  74.2 {\footnotesize \textcolor{gray}{[73.8, 74.6]}} &  61.9 {\footnotesize \textcolor{gray}{[60.9, 62.8]}} &  12.3 &  44 &  -1 \\
 44 &  \model{pnasnet\_mobile\_tf} &  74.1 {\footnotesize \textcolor{gray}{[73.8, 74.5]}} &  60.9 {\footnotesize \textcolor{gray}{[59.9, 61.8]}} &  13.3 &  48 &  -4 \\
 45 &  \model{nasnetamobile} &  74.1 {\footnotesize \textcolor{gray}{[73.7, 74.5]}} &  61.6 {\footnotesize \textcolor{gray}{[60.6, 62.5]}} &  12.5 &  45 &  0 \\
 46 &  \model{inception\_v2\_tf} &  74.0 {\footnotesize \textcolor{gray}{[73.6, 74.4]}} &  61.2 {\footnotesize \textcolor{gray}{[60.2, 62.2]}} &  12.8 &  46 &  0 \\
 47 &  \model{nasnet\_mobile\_tf} &  74.0 {\footnotesize \textcolor{gray}{[73.6, 74.4]}} &  60.8 {\footnotesize \textcolor{gray}{[59.8, 61.7]}} &  13.2 &  50 &  -3 \\
 48 &  \model{bninception} &  73.5 {\footnotesize \textcolor{gray}{[73.1, 73.9]}} &  62.1 {\footnotesize \textcolor{gray}{[61.2, 63.1]}} &  11.4 &  43 &  5 \\
 49 &  \model{vgg16\_bn} &  73.4 {\footnotesize \textcolor{gray}{[73.0, 73.7]}} &  60.8 {\footnotesize \textcolor{gray}{[59.8, 61.7]}} &  12.6 &  49 &  0 \\
 50 &  \model{resnet34} &  73.3 {\footnotesize \textcolor{gray}{[72.9, 73.7]}} &  61.2 {\footnotesize \textcolor{gray}{[60.2, 62.2]}} &  12.1 &  47 &  3 \\
 51 &  \model{vgg19} &  72.4 {\footnotesize \textcolor{gray}{[72.0, 72.8]}} &  59.7 {\footnotesize \textcolor{gray}{[58.7, 60.7]}} &  12.7 &  51 &  0 \\
 52 &  \model{vgg16} &  71.6 {\footnotesize \textcolor{gray}{[71.2, 72.0]}} &  58.8 {\footnotesize \textcolor{gray}{[57.9, 59.8]}} &  12.8 &  53 &  -1 \\
 53 &  \model{vgg13\_bn} &  71.6 {\footnotesize \textcolor{gray}{[71.2, 72.0]}} &  59.0 {\footnotesize \textcolor{gray}{[58.0, 59.9]}} &  12.6 &  52 &  1 \\
 54 &  \model{mobilenet\_v1\_tf} &  71.0 {\footnotesize \textcolor{gray}{[70.6, 71.4]}} &  57.4 {\footnotesize \textcolor{gray}{[56.4, 58.4]}} &  13.6 &  56 &  -2 \\
 55 &  \model{vgg\_19\_tf} &  71.0 {\footnotesize \textcolor{gray}{[70.6, 71.4]}} &  58.6 {\footnotesize \textcolor{gray}{[57.7, 59.6]}} &  12.4 &  54 &  1 \\
 56 &  \model{vgg\_16\_tf} &  70.9 {\footnotesize \textcolor{gray}{[70.5, 71.3]}} &  58.4 {\footnotesize \textcolor{gray}{[57.4, 59.3]}} &  12.5 &  55 &  1 \\
 57 &  \model{vgg11\_bn} &  70.4 {\footnotesize \textcolor{gray}{[70.0, 70.8]}} &  57.4 {\footnotesize \textcolor{gray}{[56.4, 58.4]}} &  13.0 &  57 &  0 \\
 58 &  \model{vgg13} &  69.9 {\footnotesize \textcolor{gray}{[69.5, 70.3]}} &  57.1 {\footnotesize \textcolor{gray}{[56.2, 58.1]}} &  12.8 &  59 &  -1 \\
 59 &  \model{inception\_v1\_tf} &  69.8 {\footnotesize \textcolor{gray}{[69.4, 70.2]}} &  56.6 {\footnotesize \textcolor{gray}{[55.7, 57.6]}} &  13.1 &  60 &  -1 \\
 60 &  \model{resnet18} &  69.8 {\footnotesize \textcolor{gray}{[69.4, 70.2]}} &  57.2 {\footnotesize \textcolor{gray}{[56.2, 58.2]}} &  12.6 &  58 &  2 \\
 61 &  \model{vgg11} &  69.0 {\footnotesize \textcolor{gray}{[68.6, 69.4]}} &  55.8 {\footnotesize \textcolor{gray}{[54.8, 56.8]}} &  13.2 &  61 &  0 \\
 62 &  \model{squeezenet1\_1} &  58.2 {\footnotesize \textcolor{gray}{[57.7, 58.6]}} &  45.3 {\footnotesize \textcolor{gray}{[44.4, 46.3]}} &  12.8 &  62 &  0 \\
 63 &  \model{squeezenet1\_0} &  58.1 {\footnotesize \textcolor{gray}{[57.7, 58.5]}} &  45.0 {\footnotesize \textcolor{gray}{[44.0, 46.0]}} &  13.1 &  63 &  0 \\
 64 &  \model{alexnet} &  56.5 {\footnotesize \textcolor{gray}{[56.1, 57.0]}} &  44.0 {\footnotesize \textcolor{gray}{[43.0, 45.0]}} &  12.5 &  64 &  0 \\
 65 &  \model{fv\_64k} &  35.1 {\footnotesize \textcolor{gray}{[34.7, 35.5]}} &  24.1 {\footnotesize \textcolor{gray}{[23.2, 24.9]}} &  11.0 &  65 &  0 \\
 66 &  \model{fv\_16k} &  28.3 {\footnotesize \textcolor{gray}{[27.9, 28.7]}} &  19.2 {\footnotesize \textcolor{gray}{[18.5, 20.0]}} &  9.1 &  66 &  0 \\
 67 &  \model{fv\_4k} &  21.2 {\footnotesize \textcolor{gray}{[20.8, 21.5]}} &  15.0 {\footnotesize \textcolor{gray}{[14.3, 15.7]}} &  6.2 &  67 &  0 \\
\bottomrule
\end{tabular}

  \centering
\end{table*}

\begin{table*}[ht!]
  \rowcolors{3}{white}{gray!15}
  \caption{Top-5 model accuracy on the original ImageNet validation set and our new test set \datasetb{}.
  $\Delta$ Rank is the relative difference in the ranking from the original test set to the new test set.
  For example, $\Delta \text{Rank} = -2$ means that a model dropped by two places on the new test set compared to the original test set.
  The confidence intervals are 95\% Clopper-Pearson intervals.
  Due to space constraints, references for the models can be found in Appendix \ref{apx:imagenet_model_descriptions}.
  The second part of the table can be found on the following page.
  }
  \begin{tabular}{rp{4.75cm}rrrrr}
\toprule 
\multicolumn{7}{c}{\textbf{ImageNet Top-5 \datasetb}} \\ 
\midrule
\multicolumn{1}{l}{Orig.} &                               &                                        &                                          &    & \multicolumn{1}{l}{New} &  \\ 
 \multicolumn{1}{l}{Rank} & Model & Orig. Accuracy & New Accuracy & Gap & \multicolumn{1}{l}{Rank} & $\Delta$ Rank \\
\midrule
 1 &  \model{pnasnet\_large\_tf} &  96.2 {\footnotesize \textcolor{gray}{[96.0, 96.3]}} &  90.1 {\footnotesize \textcolor{gray}{[89.5, 90.7]}} &  6.1 &  3 &  -2 \\
 2 &  \model{nasnet\_large\_tf} &  96.2 {\footnotesize \textcolor{gray}{[96.0, 96.3]}} &  90.1 {\footnotesize \textcolor{gray}{[89.5, 90.6]}} &  6.1 &  4 &  -2 \\
 3 &  \model{nasnetalarge} &  96.0 {\footnotesize \textcolor{gray}{[95.8, 96.2]}} &  90.4 {\footnotesize \textcolor{gray}{[89.8, 91.0]}} &  5.6 &  1 &  2 \\
 4 &  \model{pnasnet5large} &  96.0 {\footnotesize \textcolor{gray}{[95.8, 96.2]}} &  90.2 {\footnotesize \textcolor{gray}{[89.6, 90.8]}} &  5.8 &  2 &  2 \\
 5 &  \model{polynet} &  95.6 {\footnotesize \textcolor{gray}{[95.4, 95.7]}} &  89.1 {\footnotesize \textcolor{gray}{[88.5, 89.7]}} &  6.4 &  5 &  0 \\
 6 &  \model{senet154} &  95.5 {\footnotesize \textcolor{gray}{[95.3, 95.7]}} &  89.0 {\footnotesize \textcolor{gray}{[88.4, 89.6]}} &  6.5 &  6 &  0 \\
 7 &  \model{inception\_resnet\_v2\_tf} &  95.2 {\footnotesize \textcolor{gray}{[95.1, 95.4]}} &  88.4 {\footnotesize \textcolor{gray}{[87.7, 89.0]}} &  6.9 &  9 &  -2 \\
 8 &  \model{inception\_v4\_tf} &  95.2 {\footnotesize \textcolor{gray}{[95.0, 95.4]}} &  88.3 {\footnotesize \textcolor{gray}{[87.6, 88.9]}} &  6.9 &  10 &  -2 \\
 9 &  \model{inceptionresnetv2} &  95.1 {\footnotesize \textcolor{gray}{[94.9, 95.3]}} &  88.5 {\footnotesize \textcolor{gray}{[87.8, 89.1]}} &  6.7 &  8 &  1 \\
 10 &  \model{se\_resnext101\_32x4d} &  95.0 {\footnotesize \textcolor{gray}{[94.8, 95.2]}} &  88.0 {\footnotesize \textcolor{gray}{[87.4, 88.7]}} &  7.0 &  11 &  -1 \\
 11 &  \model{inceptionv4} &  94.9 {\footnotesize \textcolor{gray}{[94.7, 95.1]}} &  88.7 {\footnotesize \textcolor{gray}{[88.1, 89.3]}} &  6.2 &  7 &  4 \\
 12 &  \model{dpn107} &  94.7 {\footnotesize \textcolor{gray}{[94.5, 94.9]}} &  87.6 {\footnotesize \textcolor{gray}{[86.9, 88.2]}} &  7.1 &  13 &  -1 \\
 13 &  \model{dpn92} &  94.6 {\footnotesize \textcolor{gray}{[94.4, 94.8]}} &  87.2 {\footnotesize \textcolor{gray}{[86.5, 87.8]}} &  7.5 &  17 &  -4 \\
 14 &  \model{dpn131} &  94.6 {\footnotesize \textcolor{gray}{[94.4, 94.8]}} &  87.0 {\footnotesize \textcolor{gray}{[86.3, 87.7]}} &  7.6 &  19 &  -5 \\
 15 &  \model{dpn98} &  94.5 {\footnotesize \textcolor{gray}{[94.3, 94.7]}} &  87.2 {\footnotesize \textcolor{gray}{[86.5, 87.8]}} &  7.3 &  16 &  -1 \\
 16 &  \model{se\_resnext50\_32x4d} &  94.4 {\footnotesize \textcolor{gray}{[94.2, 94.6]}} &  87.6 {\footnotesize \textcolor{gray}{[87.0, 88.3]}} &  6.8 &  12 &  4 \\
 17 &  \model{se\_resnet152} &  94.4 {\footnotesize \textcolor{gray}{[94.2, 94.6]}} &  87.4 {\footnotesize \textcolor{gray}{[86.7, 88.0]}} &  7.0 &  15 &  2 \\
 18 &  \model{xception} &  94.3 {\footnotesize \textcolor{gray}{[94.1, 94.5]}} &  87.0 {\footnotesize \textcolor{gray}{[86.3, 87.7]}} &  7.3 &  20 &  -2 \\
 19 &  \model{se\_resnet101} &  94.3 {\footnotesize \textcolor{gray}{[94.1, 94.5]}} &  87.1 {\footnotesize \textcolor{gray}{[86.4, 87.7]}} &  7.2 &  18 &  1 \\
 20 &  \model{resnext101\_64x4d} &  94.3 {\footnotesize \textcolor{gray}{[94.0, 94.5]}} &  86.9 {\footnotesize \textcolor{gray}{[86.2, 87.5]}} &  7.4 &  22 &  -2 \\
 21 &  \model{resnet\_v2\_152\_tf} &  94.1 {\footnotesize \textcolor{gray}{[93.9, 94.3]}} &  86.9 {\footnotesize \textcolor{gray}{[86.2, 87.5]}} &  7.2 &  21 &  0 \\
 22 &  \model{resnet152} &  94.0 {\footnotesize \textcolor{gray}{[93.8, 94.3]}} &  87.6 {\footnotesize \textcolor{gray}{[86.9, 88.2]}} &  6.5 &  14 &  8 \\
 23 &  \model{inception\_v3\_tf} &  93.9 {\footnotesize \textcolor{gray}{[93.7, 94.1]}} &  86.4 {\footnotesize \textcolor{gray}{[85.7, 87.0]}} &  7.6 &  23 &  0 \\
 24 &  \model{resnext101\_32x4d} &  93.9 {\footnotesize \textcolor{gray}{[93.7, 94.1]}} &  86.2 {\footnotesize \textcolor{gray}{[85.5, 86.8]}} &  7.7 &  25 &  -1 \\
 25 &  \model{se\_resnet50} &  93.8 {\footnotesize \textcolor{gray}{[93.5, 94.0]}} &  86.3 {\footnotesize \textcolor{gray}{[85.6, 87.0]}} &  7.4 &  24 &  1 \\
 26 &  \model{resnet\_v2\_101\_tf} &  93.7 {\footnotesize \textcolor{gray}{[93.5, 93.9]}} &  86.1 {\footnotesize \textcolor{gray}{[85.4, 86.8]}} &  7.6 &  27 &  -1 \\
 27 &  \model{fbresnet152} &  93.6 {\footnotesize \textcolor{gray}{[93.4, 93.8]}} &  86.1 {\footnotesize \textcolor{gray}{[85.4, 86.7]}} &  7.5 &  28 &  -1 \\
 28 &  \model{dpn68b} &  93.6 {\footnotesize \textcolor{gray}{[93.4, 93.8]}} &  85.3 {\footnotesize \textcolor{gray}{[84.6, 86.0]}} &  8.3 &  33 &  -5 \\
 29 &  \model{densenet161} &  93.6 {\footnotesize \textcolor{gray}{[93.3, 93.8]}} &  86.1 {\footnotesize \textcolor{gray}{[85.4, 86.8]}} &  7.4 &  26 &  3 \\
 30 &  \model{resnet101} &  93.5 {\footnotesize \textcolor{gray}{[93.3, 93.8]}} &  86.0 {\footnotesize \textcolor{gray}{[85.3, 86.7]}} &  7.6 &  30 &  0 \\
 31 &  \model{inception\_v3} &  93.5 {\footnotesize \textcolor{gray}{[93.3, 93.7]}} &  85.9 {\footnotesize \textcolor{gray}{[85.2, 86.6]}} &  7.6 &  31 &  0 \\
 32 &  \model{inceptionv3} &  93.4 {\footnotesize \textcolor{gray}{[93.2, 93.6]}} &  86.1 {\footnotesize \textcolor{gray}{[85.4, 86.7]}} &  7.4 &  29 &  3 \\
 33 &  \model{densenet201} &  93.4 {\footnotesize \textcolor{gray}{[93.1, 93.6]}} &  85.3 {\footnotesize \textcolor{gray}{[84.6, 86.0]}} &  8.1 &  34 &  -1 \\
\bottomrule
\end{tabular}

  \label{tab:imagenetv2-b-33_top5_full_results}
  \centering
\end{table*}
\begin{table*}[ht!]
  \rowcolors{3}{white}{gray!15}
  \begin{tabular}{rp{4.75cm}rrrrr}
\toprule 
\multicolumn{7}{c}{\textbf{ImageNet Top-5 \datasetb}} \\ 
\midrule
\multicolumn{1}{l}{Orig.} &                               &                                        &                                          &    & \multicolumn{1}{l}{New} &  \\ 
 \multicolumn{1}{l}{Rank} & Model & Orig. Accuracy & New Accuracy & Gap & \multicolumn{1}{l}{Rank} & $\Delta$ Rank \\
\midrule
 34 &  \model{resnet\_v1\_152\_tf} &  93.2 {\footnotesize \textcolor{gray}{[92.9, 93.4]}} &  85.4 {\footnotesize \textcolor{gray}{[84.6, 86.0]}} &  7.8 &  32 &  2 \\
 35 &  \model{resnet\_v1\_101\_tf} &  92.9 {\footnotesize \textcolor{gray}{[92.7, 93.1]}} &  85.2 {\footnotesize \textcolor{gray}{[84.5, 85.9]}} &  7.7 &  35 &  0 \\
 36 &  \model{resnet50} &  92.9 {\footnotesize \textcolor{gray}{[92.6, 93.1]}} &  84.7 {\footnotesize \textcolor{gray}{[83.9, 85.4]}} &  8.2 &  38 &  -2 \\
 37 &  \model{resnet\_v2\_50\_tf} &  92.8 {\footnotesize \textcolor{gray}{[92.6, 93.1]}} &  84.4 {\footnotesize \textcolor{gray}{[83.6, 85.1]}} &  8.5 &  40 &  -3 \\
 38 &  \model{densenet169} &  92.8 {\footnotesize \textcolor{gray}{[92.6, 93.0]}} &  84.7 {\footnotesize \textcolor{gray}{[84.0, 85.4]}} &  8.1 &  37 &  1 \\
 39 &  \model{dpn68} &  92.8 {\footnotesize \textcolor{gray}{[92.5, 93.0]}} &  84.6 {\footnotesize \textcolor{gray}{[83.9, 85.3]}} &  8.2 &  39 &  0 \\
 40 &  \model{cafferesnet101} &  92.8 {\footnotesize \textcolor{gray}{[92.5, 93.0]}} &  84.9 {\footnotesize \textcolor{gray}{[84.1, 85.6]}} &  7.9 &  36 &  4 \\
 41 &  \model{resnet\_v1\_50\_tf} &  92.2 {\footnotesize \textcolor{gray}{[92.0, 92.4]}} &  84.1 {\footnotesize \textcolor{gray}{[83.4, 84.8]}} &  8.1 &  41 &  0 \\
 42 &  \model{densenet121} &  92.0 {\footnotesize \textcolor{gray}{[91.7, 92.2]}} &  83.8 {\footnotesize \textcolor{gray}{[83.1, 84.5]}} &  8.2 &  42 &  0 \\
 43 &  \model{pnasnet\_mobile\_tf} &  91.9 {\footnotesize \textcolor{gray}{[91.6, 92.1]}} &  83.1 {\footnotesize \textcolor{gray}{[82.4, 83.8]}} &  8.8 &  46 &  -3 \\
 44 &  \model{vgg19\_bn} &  91.8 {\footnotesize \textcolor{gray}{[91.6, 92.1]}} &  83.5 {\footnotesize \textcolor{gray}{[82.7, 84.2]}} &  8.4 &  43 &  1 \\
 45 &  \model{inception\_v2\_tf} &  91.8 {\footnotesize \textcolor{gray}{[91.5, 92.0]}} &  83.1 {\footnotesize \textcolor{gray}{[82.3, 83.8]}} &  8.7 &  47 &  -2 \\
 46 &  \model{nasnetamobile} &  91.7 {\footnotesize \textcolor{gray}{[91.5, 92.0]}} &  83.4 {\footnotesize \textcolor{gray}{[82.6, 84.1]}} &  8.4 &  45 &  1 \\
 47 &  \model{nasnet\_mobile\_tf} &  91.6 {\footnotesize \textcolor{gray}{[91.3, 91.8]}} &  82.2 {\footnotesize \textcolor{gray}{[81.4, 82.9]}} &  9.4 &  50 &  -3 \\
 48 &  \model{bninception} &  91.6 {\footnotesize \textcolor{gray}{[91.3, 91.8]}} &  83.4 {\footnotesize \textcolor{gray}{[82.7, 84.2]}} &  8.1 &  44 &  4 \\
 49 &  \model{vgg16\_bn} &  91.5 {\footnotesize \textcolor{gray}{[91.3, 91.8]}} &  83.0 {\footnotesize \textcolor{gray}{[82.2, 83.7]}} &  8.6 &  48 &  1 \\
 50 &  \model{resnet34} &  91.4 {\footnotesize \textcolor{gray}{[91.2, 91.7]}} &  82.7 {\footnotesize \textcolor{gray}{[82.0, 83.5]}} &  8.7 &  49 &  1 \\
 51 &  \model{vgg19} &  90.9 {\footnotesize \textcolor{gray}{[90.6, 91.1]}} &  81.5 {\footnotesize \textcolor{gray}{[80.7, 82.2]}} &  9.4 &  52 &  -1 \\
 52 &  \model{vgg16} &  90.4 {\footnotesize \textcolor{gray}{[90.1, 90.6]}} &  81.7 {\footnotesize \textcolor{gray}{[80.9, 82.4]}} &  8.7 &  51 &  1 \\
 53 &  \model{vgg13\_bn} &  90.4 {\footnotesize \textcolor{gray}{[90.1, 90.6]}} &  81.1 {\footnotesize \textcolor{gray}{[80.3, 81.9]}} &  9.3 &  53 &  0 \\
 54 &  \model{mobilenet\_v1\_tf} &  90.0 {\footnotesize \textcolor{gray}{[89.7, 90.2]}} &  79.4 {\footnotesize \textcolor{gray}{[78.6, 80.1]}} &  10.6 &  60 &  -6 \\
 56 &  \model{vgg\_19\_tf} &  89.8 {\footnotesize \textcolor{gray}{[89.6, 90.1]}} &  80.7 {\footnotesize \textcolor{gray}{[79.9, 81.4]}} &  9.2 &  54 &  2 \\
 55 &  \model{vgg\_16\_tf} &  89.8 {\footnotesize \textcolor{gray}{[89.6, 90.1]}} &  80.5 {\footnotesize \textcolor{gray}{[79.7, 81.3]}} &  9.3 &  55 &  0 \\
 57 &  \model{vgg11\_bn} &  89.8 {\footnotesize \textcolor{gray}{[89.5, 90.1]}} &  80.0 {\footnotesize \textcolor{gray}{[79.2, 80.8]}} &  9.8 &  58 &  -1 \\
 58 &  \model{inception\_v1\_tf} &  89.6 {\footnotesize \textcolor{gray}{[89.4, 89.9]}} &  80.1 {\footnotesize \textcolor{gray}{[79.3, 80.9]}} &  9.5 &  57 &  1 \\
 59 &  \model{vgg13} &  89.2 {\footnotesize \textcolor{gray}{[89.0, 89.5]}} &  79.5 {\footnotesize \textcolor{gray}{[78.7, 80.3]}} &  9.7 &  59 &  0 \\
 60 &  \model{resnet18} &  89.1 {\footnotesize \textcolor{gray}{[88.8, 89.3]}} &  80.2 {\footnotesize \textcolor{gray}{[79.4, 81.0]}} &  8.9 &  56 &  4 \\
 61 &  \model{vgg11} &  88.6 {\footnotesize \textcolor{gray}{[88.3, 88.9]}} &  78.8 {\footnotesize \textcolor{gray}{[78.0, 79.6]}} &  9.8 &  61 &  0 \\
 62 &  \model{squeezenet1\_1} &  80.6 {\footnotesize \textcolor{gray}{[80.3, 81.0]}} &  69.0 {\footnotesize \textcolor{gray}{[68.1, 69.9]}} &  11.6 &  62 &  0 \\
 63 &  \model{squeezenet1\_0} &  80.4 {\footnotesize \textcolor{gray}{[80.1, 80.8]}} &  68.5 {\footnotesize \textcolor{gray}{[67.6, 69.4]}} &  11.9 &  63 &  0 \\
 64 &  \model{alexnet} &  79.1 {\footnotesize \textcolor{gray}{[78.7, 79.4]}} &  67.4 {\footnotesize \textcolor{gray}{[66.5, 68.3]}} &  11.7 &  64 &  0 \\
 65 &  \model{fv\_64k} &  55.7 {\footnotesize \textcolor{gray}{[55.3, 56.2]}} &  42.6 {\footnotesize \textcolor{gray}{[41.6, 43.6]}} &  13.2 &  65 &  0 \\
 66 &  \model{fv\_16k} &  49.9 {\footnotesize \textcolor{gray}{[49.5, 50.4]}} &  37.5 {\footnotesize \textcolor{gray}{[36.6, 38.5]}} &  12.4 &  66 &  0 \\
 67 &  \model{fv\_4k} &  41.3 {\footnotesize \textcolor{gray}{[40.8, 41.7]}} &  31.0 {\footnotesize \textcolor{gray}{[30.1, 31.9]}} &  10.3 &  67 &  0 \\
\bottomrule
\end{tabular}

  \centering
\end{table*}

\begin{table*}[ht!]
  \rowcolors{3}{white}{gray!15}
  \caption{Top-1 model accuracy on the original ImageNet validation set and our new test set \dataseta{}.
  $\Delta$ Rank is the relative difference in the ranking from the original test set to the new test set.
  For example, $\Delta \text{Rank} = -2$ means that a model dropped by two places on the new test set compared to the original test set.
  The confidence intervals are 95\% Clopper-Pearson intervals.
  Due to space constraints, references for the models can be found in Appendix \ref{apx:imagenet_model_descriptions}.
  The second part of the table can be found on the following page.
  }
  \label{tab:imagenetv2-a-44_top1_full_results}
  \begin{tabular}{rp{4.75cm}rrrrr}
\toprule 
\multicolumn{7}{c}{\textbf{ImageNet Top-1 \dataseta}} \\ 
\midrule
\multicolumn{1}{l}{Orig.} &                               &                                        &                                          &   & \multicolumn{1}{l}{New} &  \\ 
 \multicolumn{1}{l}{Rank} & Model & Orig. Accuracy & New Accuracy & Gap & \multicolumn{1}{l}{Rank} & $\Delta$ Rank \\
\midrule
 1 &  \model{pnasnet\_large\_tf} &  82.9 {\footnotesize \textcolor{gray}{[82.5, 83.2]}} &  80.2 {\footnotesize \textcolor{gray}{[79.4, 80.9]}} &  2.7 &  2 &  -1 \\
 2 &  \model{pnasnet5large} &  82.7 {\footnotesize \textcolor{gray}{[82.4, 83.1]}} &  80.3 {\footnotesize \textcolor{gray}{[79.5, 81.1]}} &  2.4 &  1 &  1 \\
 3 &  \model{nasnet\_large\_tf} &  82.7 {\footnotesize \textcolor{gray}{[82.4, 83.0]}} &  80.1 {\footnotesize \textcolor{gray}{[79.3, 80.9]}} &  2.6 &  3 &  0 \\
 4 &  \model{nasnetalarge} &  82.5 {\footnotesize \textcolor{gray}{[82.2, 82.8]}} &  80.0 {\footnotesize \textcolor{gray}{[79.2, 80.8]}} &  2.5 &  4 &  0 \\
 5 &  \model{senet154} &  81.3 {\footnotesize \textcolor{gray}{[81.0, 81.6]}} &  78.7 {\footnotesize \textcolor{gray}{[77.8, 79.5]}} &  2.6 &  5 &  0 \\
 6 &  \model{polynet} &  80.9 {\footnotesize \textcolor{gray}{[80.5, 81.2]}} &  78.5 {\footnotesize \textcolor{gray}{[77.7, 79.3]}} &  2.3 &  6 &  0 \\
 7 &  \model{inception\_resnet\_v2\_tf} &  80.4 {\footnotesize \textcolor{gray}{[80.0, 80.7]}} &  77.9 {\footnotesize \textcolor{gray}{[77.1, 78.7]}} &  2.5 &  8 &  -1 \\
 8 &  \model{inceptionresnetv2} &  80.3 {\footnotesize \textcolor{gray}{[79.9, 80.6]}} &  78.0 {\footnotesize \textcolor{gray}{[77.2, 78.8]}} &  2.3 &  7 &  1 \\
 9 &  \model{se\_resnext101\_32x4d} &  80.2 {\footnotesize \textcolor{gray}{[79.9, 80.6]}} &  77.6 {\footnotesize \textcolor{gray}{[76.8, 78.5]}} &  2.6 &  11 &  -2 \\
 10 &  \model{inception\_v4\_tf} &  80.2 {\footnotesize \textcolor{gray}{[79.8, 80.5]}} &  77.8 {\footnotesize \textcolor{gray}{[77.0, 78.6]}} &  2.4 &  10 &  0 \\
 11 &  \model{inceptionv4} &  80.1 {\footnotesize \textcolor{gray}{[79.7, 80.4]}} &  77.9 {\footnotesize \textcolor{gray}{[77.0, 78.7]}} &  2.2 &  9 &  2 \\
 12 &  \model{dpn107} &  79.7 {\footnotesize \textcolor{gray}{[79.4, 80.1]}} &  76.6 {\footnotesize \textcolor{gray}{[75.8, 77.5]}} &  3.1 &  12 &  0 \\
 13 &  \model{dpn131} &  79.4 {\footnotesize \textcolor{gray}{[79.1, 79.8]}} &  76.6 {\footnotesize \textcolor{gray}{[75.7, 77.4]}} &  2.9 &  13 &  0 \\
 14 &  \model{dpn92} &  79.4 {\footnotesize \textcolor{gray}{[79.0, 79.8]}} &  76.3 {\footnotesize \textcolor{gray}{[75.5, 77.1]}} &  3.1 &  17 &  -3 \\
 15 &  \model{dpn98} &  79.2 {\footnotesize \textcolor{gray}{[78.9, 79.6]}} &  76.3 {\footnotesize \textcolor{gray}{[75.5, 77.2]}} &  2.9 &  16 &  -1 \\
 16 &  \model{se\_resnext50\_32x4d} &  79.1 {\footnotesize \textcolor{gray}{[78.7, 79.4]}} &  76.5 {\footnotesize \textcolor{gray}{[75.7, 77.3]}} &  2.6 &  14 &  2 \\
 17 &  \model{resnext101\_64x4d} &  79.0 {\footnotesize \textcolor{gray}{[78.6, 79.3]}} &  75.6 {\footnotesize \textcolor{gray}{[74.7, 76.4]}} &  3.4 &  20 &  -3 \\
 18 &  \model{xception} &  78.8 {\footnotesize \textcolor{gray}{[78.5, 79.2]}} &  76.4 {\footnotesize \textcolor{gray}{[75.5, 77.2]}} &  2.5 &  15 &  3 \\
 19 &  \model{se\_resnet152} &  78.7 {\footnotesize \textcolor{gray}{[78.3, 79.0]}} &  76.1 {\footnotesize \textcolor{gray}{[75.3, 76.9]}} &  2.5 &  18 &  1 \\
 20 &  \model{se\_resnet101} &  78.4 {\footnotesize \textcolor{gray}{[78.0, 78.8]}} &  75.8 {\footnotesize \textcolor{gray}{[75.0, 76.7]}} &  2.6 &  19 &  1 \\
 21 &  \model{resnet152} &  78.3 {\footnotesize \textcolor{gray}{[77.9, 78.7]}} &  75.3 {\footnotesize \textcolor{gray}{[74.5, 76.2]}} &  3.0 &  22 &  -1 \\
 22 &  \model{resnext101\_32x4d} &  78.2 {\footnotesize \textcolor{gray}{[77.8, 78.5]}} &  75.4 {\footnotesize \textcolor{gray}{[74.5, 76.2]}} &  2.8 &  21 &  1 \\
 23 &  \model{inception\_v3\_tf} &  78.0 {\footnotesize \textcolor{gray}{[77.6, 78.3]}} &  75.0 {\footnotesize \textcolor{gray}{[74.2, 75.9]}} &  2.9 &  24 &  -1 \\
 24 &  \model{resnet\_v2\_152\_tf} &  77.8 {\footnotesize \textcolor{gray}{[77.4, 78.1]}} &  75.2 {\footnotesize \textcolor{gray}{[74.4, 76.1]}} &  2.6 &  23 &  1 \\
 25 &  \model{se\_resnet50} &  77.6 {\footnotesize \textcolor{gray}{[77.3, 78.0]}} &  74.2 {\footnotesize \textcolor{gray}{[73.3, 75.1]}} &  3.4 &  30 &  -5 \\
 26 &  \model{fbresnet152} &  77.4 {\footnotesize \textcolor{gray}{[77.0, 77.8]}} &  74.8 {\footnotesize \textcolor{gray}{[74.0, 75.7]}} &  2.6 &  25 &  1 \\
 27 &  \model{resnet101} &  77.4 {\footnotesize \textcolor{gray}{[77.0, 77.7]}} &  74.5 {\footnotesize \textcolor{gray}{[73.6, 75.3]}} &  2.9 &  29 &  -2 \\
 28 &  \model{inceptionv3} &  77.3 {\footnotesize \textcolor{gray}{[77.0, 77.7]}} &  74.5 {\footnotesize \textcolor{gray}{[73.6, 75.4]}} &  2.8 &  28 &  0 \\
 29 &  \model{inception\_v3} &  77.2 {\footnotesize \textcolor{gray}{[76.8, 77.6]}} &  74.7 {\footnotesize \textcolor{gray}{[73.8, 75.6]}} &  2.5 &  26 &  3 \\
 30 &  \model{densenet161} &  77.1 {\footnotesize \textcolor{gray}{[76.8, 77.5]}} &  74.6 {\footnotesize \textcolor{gray}{[73.7, 75.4]}} &  2.6 &  27 &  3 \\
 31 &  \model{dpn68b} &  77.0 {\footnotesize \textcolor{gray}{[76.7, 77.4]}} &  73.8 {\footnotesize \textcolor{gray}{[72.9, 74.7]}} &  3.2 &  33 &  -2 \\
 32 &  \model{resnet\_v2\_101\_tf} &  77.0 {\footnotesize \textcolor{gray}{[76.6, 77.3]}} &  74.0 {\footnotesize \textcolor{gray}{[73.1, 74.8]}} &  3.0 &  31 &  1 \\
 33 &  \model{densenet201} &  76.9 {\footnotesize \textcolor{gray}{[76.5, 77.3]}} &  73.9 {\footnotesize \textcolor{gray}{[73.1, 74.8]}} &  3.0 &  32 &  1 \\
\bottomrule
\end{tabular}

  \centering
\end{table*}
\begin{table*}[ht!]
  \rowcolors{3}{white}{gray!15}
  \begin{tabular}{rp{4.75cm}rrrrr}
\toprule 
\multicolumn{7}{c}{\textbf{ImageNet Top-1 \dataseta}} \\ 
\midrule
\multicolumn{1}{l}{Orig.} &                               &                                        &                                          &   & \multicolumn{1}{l}{New} &  \\ 
 \multicolumn{1}{l}{Rank} & Model & Orig. Accuracy & New Accuracy & Gap & \multicolumn{1}{l}{Rank} & $\Delta$ Rank \\
\midrule
 34 &  \model{resnet\_v1\_152\_tf} &  76.8 {\footnotesize \textcolor{gray}{[76.4, 77.2]}} &  73.7 {\footnotesize \textcolor{gray}{[72.9, 74.6]}} &  3.1 &  34 &  0 \\
 35 &  \model{resnet\_v1\_101\_tf} &  76.4 {\footnotesize \textcolor{gray}{[76.0, 76.8]}} &  73.4 {\footnotesize \textcolor{gray}{[72.5, 74.2]}} &  3.0 &  35 &  0 \\
 36 &  \model{cafferesnet101} &  76.2 {\footnotesize \textcolor{gray}{[75.8, 76.6]}} &  72.9 {\footnotesize \textcolor{gray}{[72.0, 73.7]}} &  3.3 &  37 &  -1 \\
 37 &  \model{resnet50} &  76.1 {\footnotesize \textcolor{gray}{[75.8, 76.5]}} &  72.7 {\footnotesize \textcolor{gray}{[71.8, 73.6]}} &  3.4 &  38 &  -1 \\
 38 &  \model{dpn68} &  75.9 {\footnotesize \textcolor{gray}{[75.5, 76.2]}} &  73.0 {\footnotesize \textcolor{gray}{[72.1, 73.8]}} &  2.9 &  36 &  2 \\
 39 &  \model{densenet169} &  75.6 {\footnotesize \textcolor{gray}{[75.2, 76.0]}} &  72.3 {\footnotesize \textcolor{gray}{[71.4, 73.1]}} &  3.3 &  40 &  -1 \\
 40 &  \model{resnet\_v2\_50\_tf} &  75.6 {\footnotesize \textcolor{gray}{[75.2, 76.0]}} &  72.3 {\footnotesize \textcolor{gray}{[71.4, 73.2]}} &  3.3 &  39 &  1 \\
 41 &  \model{resnet\_v1\_50\_tf} &  75.2 {\footnotesize \textcolor{gray}{[74.8, 75.6]}} &  71.9 {\footnotesize \textcolor{gray}{[71.0, 72.8]}} &  3.3 &  41 &  0 \\
 42 &  \model{densenet121} &  74.4 {\footnotesize \textcolor{gray}{[74.0, 74.8]}} &  70.5 {\footnotesize \textcolor{gray}{[69.6, 71.4]}} &  3.9 &  47 &  -5 \\
 43 &  \model{vgg19\_bn} &  74.2 {\footnotesize \textcolor{gray}{[73.8, 74.6]}} &  71.4 {\footnotesize \textcolor{gray}{[70.5, 72.3]}} &  2.8 &  42 &  1 \\
 44 &  \model{pnasnet\_mobile\_tf} &  74.1 {\footnotesize \textcolor{gray}{[73.8, 74.5]}} &  70.6 {\footnotesize \textcolor{gray}{[69.7, 71.5]}} &  3.6 &  46 &  -2 \\
 45 &  \model{nasnetamobile} &  74.1 {\footnotesize \textcolor{gray}{[73.7, 74.5]}} &  70.9 {\footnotesize \textcolor{gray}{[70.0, 71.8]}} &  3.2 &  45 &  0 \\
 46 &  \model{inception\_v2\_tf} &  74.0 {\footnotesize \textcolor{gray}{[73.6, 74.4]}} &  71.1 {\footnotesize \textcolor{gray}{[70.2, 72.0]}} &  2.9 &  44 &  2 \\
 47 &  \model{nasnet\_mobile\_tf} &  74.0 {\footnotesize \textcolor{gray}{[73.6, 74.4]}} &  70.0 {\footnotesize \textcolor{gray}{[69.0, 70.8]}} &  4.0 &  50 &  -3 \\
 48 &  \model{bninception} &  73.5 {\footnotesize \textcolor{gray}{[73.1, 73.9]}} &  71.3 {\footnotesize \textcolor{gray}{[70.4, 72.2]}} &  2.2 &  43 &  5 \\
 49 &  \model{vgg16\_bn} &  73.4 {\footnotesize \textcolor{gray}{[73.0, 73.7]}} &  70.2 {\footnotesize \textcolor{gray}{[69.3, 71.1]}} &  3.1 &  48 &  1 \\
 50 &  \model{resnet34} &  73.3 {\footnotesize \textcolor{gray}{[72.9, 73.7]}} &  70.2 {\footnotesize \textcolor{gray}{[69.2, 71.0]}} &  3.2 &  49 &  1 \\
 51 &  \model{vgg19} &  72.4 {\footnotesize \textcolor{gray}{[72.0, 72.8]}} &  68.7 {\footnotesize \textcolor{gray}{[67.8, 69.6]}} &  3.7 &  51 &  0 \\
 52 &  \model{vgg16} &  71.6 {\footnotesize \textcolor{gray}{[71.2, 72.0]}} &  68.0 {\footnotesize \textcolor{gray}{[67.0, 68.9]}} &  3.6 &  52 &  0 \\
 53 &  \model{vgg13\_bn} &  71.6 {\footnotesize \textcolor{gray}{[71.2, 72.0]}} &  67.3 {\footnotesize \textcolor{gray}{[66.4, 68.2]}} &  4.3 &  55 &  -2 \\
 54 &  \model{mobilenet\_v1\_tf} &  71.0 {\footnotesize \textcolor{gray}{[70.6, 71.4]}} &  66.1 {\footnotesize \textcolor{gray}{[65.2, 67.0]}} &  4.9 &  59 &  -5 \\
 55 &  \model{vgg\_19\_tf} &  71.0 {\footnotesize \textcolor{gray}{[70.6, 71.4]}} &  67.4 {\footnotesize \textcolor{gray}{[66.5, 68.3]}} &  3.6 &  54 &  1 \\
 56 &  \model{vgg\_16\_tf} &  70.9 {\footnotesize \textcolor{gray}{[70.5, 71.3]}} &  67.6 {\footnotesize \textcolor{gray}{[66.7, 68.5]}} &  3.3 &  53 &  3 \\
 57 &  \model{vgg11\_bn} &  70.4 {\footnotesize \textcolor{gray}{[70.0, 70.8]}} &  66.4 {\footnotesize \textcolor{gray}{[65.5, 67.3]}} &  4.0 &  58 &  -1 \\
 58 &  \model{vgg13} &  69.9 {\footnotesize \textcolor{gray}{[69.5, 70.3]}} &  66.0 {\footnotesize \textcolor{gray}{[65.0, 66.9]}} &  4.0 &  60 &  -2 \\
 59 &  \model{inception\_v1\_tf} &  69.8 {\footnotesize \textcolor{gray}{[69.4, 70.2]}} &  66.4 {\footnotesize \textcolor{gray}{[65.5, 67.4]}} &  3.3 &  57 &  2 \\
 60 &  \model{resnet18} &  69.8 {\footnotesize \textcolor{gray}{[69.4, 70.2]}} &  66.6 {\footnotesize \textcolor{gray}{[65.7, 67.5]}} &  3.2 &  56 &  4 \\
 61 &  \model{vgg11} &  69.0 {\footnotesize \textcolor{gray}{[68.6, 69.4]}} &  64.6 {\footnotesize \textcolor{gray}{[63.7, 65.6]}} &  4.4 &  61 &  0 \\
 62 &  \model{squeezenet1\_1} &  58.2 {\footnotesize \textcolor{gray}{[57.7, 58.6]}} &  54.4 {\footnotesize \textcolor{gray}{[53.4, 55.4]}} &  3.8 &  62 &  0 \\
 63 &  \model{squeezenet1\_0} &  58.1 {\footnotesize \textcolor{gray}{[57.7, 58.5]}} &  53.4 {\footnotesize \textcolor{gray}{[52.4, 54.4]}} &  4.7 &  63 &  0 \\
 64 &  \model{alexnet} &  56.5 {\footnotesize \textcolor{gray}{[56.1, 57.0]}} &  51.3 {\footnotesize \textcolor{gray}{[50.3, 52.3]}} &  5.2 &  64 &  0 \\
 65 &  \model{fv\_64k} &  35.1 {\footnotesize \textcolor{gray}{[34.7, 35.5]}} &  29.1 {\footnotesize \textcolor{gray}{[28.2, 30.0]}} &  6.0 &  65 &  0 \\
 66 &  \model{fv\_16k} &  28.3 {\footnotesize \textcolor{gray}{[27.9, 28.7]}} &  23.4 {\footnotesize \textcolor{gray}{[22.5, 24.2]}} &  5.0 &  66 &  0 \\
 67 &  \model{fv\_4k} &  21.2 {\footnotesize \textcolor{gray}{[20.8, 21.5]}} &  17.8 {\footnotesize \textcolor{gray}{[17.0, 18.5]}} &  3.4 &  67 &  0 \\
\bottomrule
\end{tabular}

  \centering
\end{table*}

\begin{table*}[ht!]
  \rowcolors{3}{white}{gray!15}
  \caption{Top-5 model accuracy on the original ImageNet validation set and our new test set \dataseta{}.
  $\Delta$ Rank is the relative difference in the ranking from the original test set to the new test set.
  For example, $\Delta \text{Rank} = -2$ means that a model dropped by two places on the new test set compared to the original test set.
  The confidence intervals are 95\% Clopper-Pearson intervals.
  Due to space constraints, references for the models can be found in Appendix \ref{apx:imagenet_model_descriptions}.
  The second part of the table can be found on the following page.
  }
  \begin{tabular}{rp{4.75cm}rrrrr}
\toprule 
\multicolumn{7}{c}{\textbf{ImageNet Top-5 \dataseta}} \\ 
\midrule
\multicolumn{1}{l}{Orig.} &                               &                                        &                                          &   & \multicolumn{1}{l}{New} &  \\ 
 \multicolumn{1}{l}{Rank} & Model & Orig. Accuracy & New Accuracy & Gap & \multicolumn{1}{l}{Rank} & $\Delta$ Rank \\
\midrule
 1 &  \model{pnasnet\_large\_tf} &  96.2 {\footnotesize \textcolor{gray}{[96.0, 96.3]}} &  95.6 {\footnotesize \textcolor{gray}{[95.2, 96.0]}} &  0.6 &  2 &  -1 \\
 2 &  \model{nasnet\_large\_tf} &  96.2 {\footnotesize \textcolor{gray}{[96.0, 96.3]}} &  95.7 {\footnotesize \textcolor{gray}{[95.2, 96.0]}} &  0.5 &  1 &  1 \\
 3 &  \model{nasnetalarge} &  96.0 {\footnotesize \textcolor{gray}{[95.8, 96.2]}} &  95.3 {\footnotesize \textcolor{gray}{[94.9, 95.8]}} &  0.7 &  4 &  -1 \\
 4 &  \model{pnasnet5large} &  96.0 {\footnotesize \textcolor{gray}{[95.8, 96.2]}} &  95.5 {\footnotesize \textcolor{gray}{[95.0, 95.9]}} &  0.5 &  3 &  1 \\
 5 &  \model{polynet} &  95.6 {\footnotesize \textcolor{gray}{[95.4, 95.7]}} &  94.9 {\footnotesize \textcolor{gray}{[94.4, 95.3]}} &  0.7 &  5 &  0 \\
 6 &  \model{senet154} &  95.5 {\footnotesize \textcolor{gray}{[95.3, 95.7]}} &  94.8 {\footnotesize \textcolor{gray}{[94.3, 95.2]}} &  0.7 &  6 &  0 \\
 7 &  \model{inception\_resnet\_v2\_tf} &  95.2 {\footnotesize \textcolor{gray}{[95.1, 95.4]}} &  94.7 {\footnotesize \textcolor{gray}{[94.2, 95.1]}} &  0.6 &  7 &  0 \\
 8 &  \model{inception\_v4\_tf} &  95.2 {\footnotesize \textcolor{gray}{[95.0, 95.4]}} &  94.4 {\footnotesize \textcolor{gray}{[94.0, 94.9]}} &  0.8 &  9 &  -1 \\
 9 &  \model{inceptionresnetv2} &  95.1 {\footnotesize \textcolor{gray}{[94.9, 95.3]}} &  94.5 {\footnotesize \textcolor{gray}{[94.1, 95.0]}} &  0.6 &  8 &  1 \\
 10 &  \model{se\_resnext101\_32x4d} &  95.0 {\footnotesize \textcolor{gray}{[94.8, 95.2]}} &  94.3 {\footnotesize \textcolor{gray}{[93.8, 94.7]}} &  0.7 &  11 &  -1 \\
 11 &  \model{inceptionv4} &  94.9 {\footnotesize \textcolor{gray}{[94.7, 95.1]}} &  94.3 {\footnotesize \textcolor{gray}{[93.8, 94.7]}} &  0.6 &  10 &  1 \\
 12 &  \model{dpn107} &  94.7 {\footnotesize \textcolor{gray}{[94.5, 94.9]}} &  93.7 {\footnotesize \textcolor{gray}{[93.2, 94.2]}} &  1.0 &  12 &  0 \\
 13 &  \model{dpn92} &  94.6 {\footnotesize \textcolor{gray}{[94.4, 94.8]}} &  93.7 {\footnotesize \textcolor{gray}{[93.2, 94.2]}} &  0.9 &  14 &  -1 \\
 14 &  \model{dpn131} &  94.6 {\footnotesize \textcolor{gray}{[94.4, 94.8]}} &  93.5 {\footnotesize \textcolor{gray}{[92.9, 93.9]}} &  1.1 &  20 &  -6 \\
 15 &  \model{dpn98} &  94.5 {\footnotesize \textcolor{gray}{[94.3, 94.7]}} &  93.6 {\footnotesize \textcolor{gray}{[93.1, 94.1]}} &  0.9 &  17 &  -2 \\
 16 &  \model{se\_resnext50\_32x4d} &  94.4 {\footnotesize \textcolor{gray}{[94.2, 94.6]}} &  93.6 {\footnotesize \textcolor{gray}{[93.1, 94.1]}} &  0.8 &  16 &  0 \\
 17 &  \model{se\_resnet152} &  94.4 {\footnotesize \textcolor{gray}{[94.2, 94.6]}} &  93.7 {\footnotesize \textcolor{gray}{[93.2, 94.2]}} &  0.7 &  13 &  4 \\
 18 &  \model{xception} &  94.3 {\footnotesize \textcolor{gray}{[94.1, 94.5]}} &  93.6 {\footnotesize \textcolor{gray}{[93.1, 94.1]}} &  0.7 &  18 &  0 \\
 19 &  \model{se\_resnet101} &  94.3 {\footnotesize \textcolor{gray}{[94.1, 94.5]}} &  93.6 {\footnotesize \textcolor{gray}{[93.1, 94.0]}} &  0.7 &  19 &  0 \\
 20 &  \model{resnext101\_64x4d} &  94.3 {\footnotesize \textcolor{gray}{[94.0, 94.5]}} &  93.3 {\footnotesize \textcolor{gray}{[92.8, 93.8]}} &  0.9 &  22 &  -2 \\
 21 &  \model{resnet\_v2\_152\_tf} &  94.1 {\footnotesize \textcolor{gray}{[93.9, 94.3]}} &  93.4 {\footnotesize \textcolor{gray}{[92.9, 93.9]}} &  0.7 &  21 &  0 \\
 22 &  \model{resnet152} &  94.0 {\footnotesize \textcolor{gray}{[93.8, 94.3]}} &  93.7 {\footnotesize \textcolor{gray}{[93.2, 94.2]}} &  0.4 &  15 &  7 \\
 23 &  \model{inception\_v3\_tf} &  93.9 {\footnotesize \textcolor{gray}{[93.7, 94.1]}} &  92.8 {\footnotesize \textcolor{gray}{[92.3, 93.3]}} &  1.1 &  25 &  -2 \\
 24 &  \model{resnext101\_32x4d} &  93.9 {\footnotesize \textcolor{gray}{[93.7, 94.1]}} &  92.7 {\footnotesize \textcolor{gray}{[92.2, 93.2]}} &  1.2 &  28 &  -4 \\
 25 &  \model{se\_resnet50} &  93.8 {\footnotesize \textcolor{gray}{[93.5, 94.0]}} &  93.0 {\footnotesize \textcolor{gray}{[92.4, 93.5]}} &  0.8 &  24 &  1 \\
 26 &  \model{resnet\_v2\_101\_tf} &  93.7 {\footnotesize \textcolor{gray}{[93.5, 93.9]}} &  93.2 {\footnotesize \textcolor{gray}{[92.7, 93.7]}} &  0.5 &  23 &  3 \\
 27 &  \model{fbresnet152} &  93.6 {\footnotesize \textcolor{gray}{[93.4, 93.8]}} &  92.7 {\footnotesize \textcolor{gray}{[92.1, 93.2]}} &  0.9 &  29 &  -2 \\
 28 &  \model{dpn68b} &  93.6 {\footnotesize \textcolor{gray}{[93.4, 93.8]}} &  92.7 {\footnotesize \textcolor{gray}{[92.1, 93.2]}} &  0.9 &  31 &  -3 \\
 29 &  \model{densenet161} &  93.6 {\footnotesize \textcolor{gray}{[93.3, 93.8]}} &  92.8 {\footnotesize \textcolor{gray}{[92.3, 93.3]}} &  0.8 &  26 &  3 \\
 30 &  \model{resnet101} &  93.5 {\footnotesize \textcolor{gray}{[93.3, 93.8]}} &  92.8 {\footnotesize \textcolor{gray}{[92.3, 93.3]}} &  0.8 &  27 &  3 \\
 31 &  \model{inception\_v3} &  93.5 {\footnotesize \textcolor{gray}{[93.3, 93.7]}} &  92.7 {\footnotesize \textcolor{gray}{[92.1, 93.2]}} &  0.9 &  30 &  1 \\
 32 &  \model{inceptionv3} &  93.4 {\footnotesize \textcolor{gray}{[93.2, 93.6]}} &  92.6 {\footnotesize \textcolor{gray}{[92.1, 93.1]}} &  0.8 &  32 &  0 \\
 33 &  \model{densenet201} &  93.4 {\footnotesize \textcolor{gray}{[93.1, 93.6]}} &  92.4 {\footnotesize \textcolor{gray}{[91.9, 92.9]}} &  1.0 &  33 &  0 \\
\bottomrule
\end{tabular}

  \label{tab:imagenetv2-a-44_top5_full_results}
  \centering
\end{table*}
\begin{table*}[ht!]
  \rowcolors{3}{white}{gray!15}
  \begin{tabular}{rp{4.75cm}rrrrr}
\toprule 
\multicolumn{7}{c}{\textbf{ImageNet Top-5 \dataseta}} \\ 
\midrule
\multicolumn{1}{l}{Orig.} &                               &                                        &                                          &   & \multicolumn{1}{l}{New} &  \\ 
 \multicolumn{1}{l}{Rank} & Model & Orig. Accuracy & New Accuracy & Gap & \multicolumn{1}{l}{Rank} & $\Delta$ Rank \\
\midrule
 34 &  \model{resnet\_v1\_152\_tf} &  93.2 {\footnotesize \textcolor{gray}{[92.9, 93.4]}} &  92.2 {\footnotesize \textcolor{gray}{[91.7, 92.7]}} &  1.0 &  34 &  0 \\
 35 &  \model{resnet\_v1\_101\_tf} &  92.9 {\footnotesize \textcolor{gray}{[92.7, 93.1]}} &  92.0 {\footnotesize \textcolor{gray}{[91.5, 92.5]}} &  0.9 &  36 &  -1 \\
 36 &  \model{resnet50} &  92.9 {\footnotesize \textcolor{gray}{[92.6, 93.1]}} &  92.0 {\footnotesize \textcolor{gray}{[91.5, 92.5]}} &  0.9 &  37 &  -1 \\
 37 &  \model{resnet\_v2\_50\_tf} &  92.8 {\footnotesize \textcolor{gray}{[92.6, 93.1]}} &  91.9 {\footnotesize \textcolor{gray}{[91.4, 92.5]}} &  0.9 &  38 &  -1 \\
 38 &  \model{densenet169} &  92.8 {\footnotesize \textcolor{gray}{[92.6, 93.0]}} &  91.9 {\footnotesize \textcolor{gray}{[91.4, 92.4]}} &  0.9 &  39 &  -1 \\
 39 &  \model{dpn68} &  92.8 {\footnotesize \textcolor{gray}{[92.5, 93.0]}} &  92.1 {\footnotesize \textcolor{gray}{[91.5, 92.6]}} &  0.7 &  35 &  4 \\
 40 &  \model{cafferesnet101} &  92.8 {\footnotesize \textcolor{gray}{[92.5, 93.0]}} &  91.6 {\footnotesize \textcolor{gray}{[91.1, 92.2]}} &  1.1 &  40 &  0 \\
 41 &  \model{resnet\_v1\_50\_tf} &  92.2 {\footnotesize \textcolor{gray}{[92.0, 92.4]}} &  91.1 {\footnotesize \textcolor{gray}{[90.6, 91.7]}} &  1.0 &  41 &  0 \\
 42 &  \model{densenet121} &  92.0 {\footnotesize \textcolor{gray}{[91.7, 92.2]}} &  91.1 {\footnotesize \textcolor{gray}{[90.5, 91.6]}} &  0.9 &  42 &  0 \\
 43 &  \model{pnasnet\_mobile\_tf} &  91.9 {\footnotesize \textcolor{gray}{[91.6, 92.1]}} &  90.7 {\footnotesize \textcolor{gray}{[90.1, 91.3]}} &  1.1 &  47 &  -4 \\
 44 &  \model{vgg19\_bn} &  91.8 {\footnotesize \textcolor{gray}{[91.6, 92.1]}} &  91.0 {\footnotesize \textcolor{gray}{[90.4, 91.5]}} &  0.9 &  44 &  0 \\
 45 &  \model{inception\_v2\_tf} &  91.8 {\footnotesize \textcolor{gray}{[91.5, 92.0]}} &  91.0 {\footnotesize \textcolor{gray}{[90.5, 91.6]}} &  0.7 &  43 &  2 \\
 46 &  \model{nasnetamobile} &  91.7 {\footnotesize \textcolor{gray}{[91.5, 92.0]}} &  90.9 {\footnotesize \textcolor{gray}{[90.3, 91.4]}} &  0.9 &  46 &  0 \\
 47 &  \model{nasnet\_mobile\_tf} &  91.6 {\footnotesize \textcolor{gray}{[91.3, 91.8]}} &  90.1 {\footnotesize \textcolor{gray}{[89.5, 90.7]}} &  1.4 &  50 &  -3 \\
 48 &  \model{bninception} &  91.6 {\footnotesize \textcolor{gray}{[91.3, 91.8]}} &  90.9 {\footnotesize \textcolor{gray}{[90.3, 91.5]}} &  0.7 &  45 &  3 \\
 49 &  \model{vgg16\_bn} &  91.5 {\footnotesize \textcolor{gray}{[91.3, 91.8]}} &  90.4 {\footnotesize \textcolor{gray}{[89.8, 90.9]}} &  1.1 &  49 &  0 \\
 50 &  \model{resnet34} &  91.4 {\footnotesize \textcolor{gray}{[91.2, 91.7]}} &  90.5 {\footnotesize \textcolor{gray}{[89.9, 91.0]}} &  1.0 &  48 &  2 \\
 51 &  \model{vgg19} &  90.9 {\footnotesize \textcolor{gray}{[90.6, 91.1]}} &  89.7 {\footnotesize \textcolor{gray}{[89.1, 90.3]}} &  1.2 &  51 &  0 \\
 52 &  \model{vgg16} &  90.4 {\footnotesize \textcolor{gray}{[90.1, 90.6]}} &  88.8 {\footnotesize \textcolor{gray}{[88.1, 89.4]}} &  1.6 &  53 &  -1 \\
 53 &  \model{vgg13\_bn} &  90.4 {\footnotesize \textcolor{gray}{[90.1, 90.6]}} &  89.0 {\footnotesize \textcolor{gray}{[88.3, 89.6]}} &  1.4 &  52 &  1 \\
 54 &  \model{mobilenet\_v1\_tf} &  90.0 {\footnotesize \textcolor{gray}{[89.7, 90.2]}} &  87.6 {\footnotesize \textcolor{gray}{[86.9, 88.2]}} &  2.4 &  60 &  -6 \\
 56 &  \model{vgg\_19\_tf} &  89.8 {\footnotesize \textcolor{gray}{[89.6, 90.1]}} &  88.5 {\footnotesize \textcolor{gray}{[87.8, 89.1]}} &  1.4 &  55 &  1 \\
 55 &  \model{vgg\_16\_tf} &  89.8 {\footnotesize \textcolor{gray}{[89.6, 90.1]}} &  88.6 {\footnotesize \textcolor{gray}{[87.9, 89.2]}} &  1.3 &  54 &  1 \\
 57 &  \model{vgg11\_bn} &  89.8 {\footnotesize \textcolor{gray}{[89.5, 90.1]}} &  88.3 {\footnotesize \textcolor{gray}{[87.6, 88.9]}} &  1.5 &  56 &  1 \\
 58 &  \model{inception\_v1\_tf} &  89.6 {\footnotesize \textcolor{gray}{[89.4, 89.9]}} &  88.1 {\footnotesize \textcolor{gray}{[87.4, 88.7]}} &  1.5 &  57 &  1 \\
 59 &  \model{vgg13} &  89.2 {\footnotesize \textcolor{gray}{[89.0, 89.5]}} &  87.6 {\footnotesize \textcolor{gray}{[86.9, 88.2]}} &  1.6 &  59 &  0 \\
 60 &  \model{resnet18} &  89.1 {\footnotesize \textcolor{gray}{[88.8, 89.3]}} &  88.1 {\footnotesize \textcolor{gray}{[87.4, 88.7]}} &  1.0 &  58 &  2 \\
 61 &  \model{vgg11} &  88.6 {\footnotesize \textcolor{gray}{[88.3, 88.9]}} &  86.9 {\footnotesize \textcolor{gray}{[86.2, 87.5]}} &  1.7 &  61 &  0 \\
 62 &  \model{squeezenet1\_1} &  80.6 {\footnotesize \textcolor{gray}{[80.3, 81.0]}} &  78.0 {\footnotesize \textcolor{gray}{[77.2, 78.8]}} &  2.6 &  62 &  0 \\
 63 &  \model{squeezenet1\_0} &  80.4 {\footnotesize \textcolor{gray}{[80.1, 80.8]}} &  77.7 {\footnotesize \textcolor{gray}{[76.9, 78.5]}} &  2.7 &  63 &  0 \\
 64 &  \model{alexnet} &  79.1 {\footnotesize \textcolor{gray}{[78.7, 79.4]}} &  75.9 {\footnotesize \textcolor{gray}{[75.0, 76.7]}} &  3.2 &  64 &  0 \\
 65 &  \model{fv\_64k} &  55.7 {\footnotesize \textcolor{gray}{[55.3, 56.2]}} &  49.8 {\footnotesize \textcolor{gray}{[48.8, 50.7]}} &  6.0 &  65 &  0 \\
 66 &  \model{fv\_16k} &  49.9 {\footnotesize \textcolor{gray}{[49.5, 50.4]}} &  44.2 {\footnotesize \textcolor{gray}{[43.2, 45.2]}} &  5.7 &  66 &  0 \\
 67 &  \model{fv\_4k} &  41.3 {\footnotesize \textcolor{gray}{[40.8, 41.7]}} &  36.5 {\footnotesize \textcolor{gray}{[35.6, 37.5]}} &  4.8 &  67 &  0 \\
\bottomrule
\end{tabular}

  \centering
\end{table*}

\begin{table*}[ht!]
  \rowcolors{3}{white}{gray!15}
  \caption{Top-1 model accuracy on the original ImageNet validation set and our new test set \datasetc{}.
  $\Delta$ Rank is the relative difference in the ranking from the original test set to the new test set.
  For example, $\Delta \text{Rank} = -2$ means that a model dropped by two places on the new test set compared to the original test set.
  The confidence intervals are 95\% Clopper-Pearson intervals.
  Due to space constraints, references for the models can be found in Appendix \ref{apx:imagenet_model_descriptions}.
  The second part of the table can be found on the following page.
  }
  \label{tab:imagenetv2-c-12_top1_full_results}
  \begin{tabular}{rp{4.75cm}rrrrr}
\toprule 
\multicolumn{7}{c}{\textbf{ImageNet Top-1 \datasetc}} \\ 
\midrule
\multicolumn{1}{l}{Orig.} &                               &                                        &                                          &    & \multicolumn{1}{l}{New} &  \\ 
 \multicolumn{1}{l}{Rank} & Model & Orig. Accuracy & New Accuracy & Gap & \multicolumn{1}{l}{Rank} & $\Delta$ Rank \\
\midrule
 1 &  \model{pnasnet\_large\_tf} &  82.9 {\footnotesize \textcolor{gray}{[82.5, 83.2]}} &  83.9 {\footnotesize \textcolor{gray}{[83.2, 84.6]}} &  -1.0 &  3 &  -2 \\
 2 &  \model{pnasnet5large} &  82.7 {\footnotesize \textcolor{gray}{[82.4, 83.1]}} &  83.9 {\footnotesize \textcolor{gray}{[83.1, 84.6]}} &  -1.1 &  4 &  -2 \\
 3 &  \model{nasnet\_large\_tf} &  82.7 {\footnotesize \textcolor{gray}{[82.4, 83.0]}} &  84.0 {\footnotesize \textcolor{gray}{[83.3, 84.7]}} &  -1.3 &  2 &  1 \\
 4 &  \model{nasnetalarge} &  82.5 {\footnotesize \textcolor{gray}{[82.2, 82.8]}} &  84.2 {\footnotesize \textcolor{gray}{[83.4, 84.9]}} &  -1.7 &  1 &  3 \\
 5 &  \model{senet154} &  81.3 {\footnotesize \textcolor{gray}{[81.0, 81.6]}} &  82.8 {\footnotesize \textcolor{gray}{[82.1, 83.6]}} &  -1.5 &  6 &  -1 \\
 6 &  \model{polynet} &  80.9 {\footnotesize \textcolor{gray}{[80.5, 81.2]}} &  83.0 {\footnotesize \textcolor{gray}{[82.2, 83.7]}} &  -2.1 &  5 &  1 \\
 7 &  \model{inception\_resnet\_v2\_tf} &  80.4 {\footnotesize \textcolor{gray}{[80.0, 80.7]}} &  82.5 {\footnotesize \textcolor{gray}{[81.7, 83.2]}} &  -2.1 &  8 &  -1 \\
 8 &  \model{inceptionresnetv2} &  80.3 {\footnotesize \textcolor{gray}{[79.9, 80.6]}} &  82.8 {\footnotesize \textcolor{gray}{[82.0, 83.5]}} &  -2.5 &  7 &  1 \\
 9 &  \model{se\_resnext101\_32x4d} &  80.2 {\footnotesize \textcolor{gray}{[79.9, 80.6]}} &  82.2 {\footnotesize \textcolor{gray}{[81.5, 83.0]}} &  -2.0 &  11 &  -2 \\
 10 &  \model{inception\_v4\_tf} &  80.2 {\footnotesize \textcolor{gray}{[79.8, 80.5]}} &  82.3 {\footnotesize \textcolor{gray}{[81.5, 83.0]}} &  -2.1 &  9 &  1 \\
 11 &  \model{inceptionv4} &  80.1 {\footnotesize \textcolor{gray}{[79.7, 80.4]}} &  82.3 {\footnotesize \textcolor{gray}{[81.5, 83.0]}} &  -2.2 &  10 &  1 \\
 12 &  \model{dpn107} &  79.7 {\footnotesize \textcolor{gray}{[79.4, 80.1]}} &  81.4 {\footnotesize \textcolor{gray}{[80.6, 82.1]}} &  -1.6 &  13 &  -1 \\
 13 &  \model{dpn131} &  79.4 {\footnotesize \textcolor{gray}{[79.1, 79.8]}} &  81.3 {\footnotesize \textcolor{gray}{[80.5, 82.1]}} &  -1.9 &  15 &  -2 \\
 14 &  \model{dpn92} &  79.4 {\footnotesize \textcolor{gray}{[79.0, 79.8]}} &  81.2 {\footnotesize \textcolor{gray}{[80.5, 82.0]}} &  -1.8 &  16 &  -2 \\
 15 &  \model{dpn98} &  79.2 {\footnotesize \textcolor{gray}{[78.9, 79.6]}} &  81.5 {\footnotesize \textcolor{gray}{[80.7, 82.3]}} &  -2.3 &  12 &  3 \\
 16 &  \model{se\_resnext50\_32x4d} &  79.1 {\footnotesize \textcolor{gray}{[78.7, 79.4]}} &  81.4 {\footnotesize \textcolor{gray}{[80.6, 82.1]}} &  -2.3 &  14 &  2 \\
 17 &  \model{resnext101\_64x4d} &  79.0 {\footnotesize \textcolor{gray}{[78.6, 79.3]}} &  80.3 {\footnotesize \textcolor{gray}{[79.5, 81.0]}} &  -1.3 &  22 &  -5 \\
 18 &  \model{xception} &  78.8 {\footnotesize \textcolor{gray}{[78.5, 79.2]}} &  81.0 {\footnotesize \textcolor{gray}{[80.2, 81.8]}} &  -2.2 &  18 &  0 \\
 19 &  \model{se\_resnet152} &  78.7 {\footnotesize \textcolor{gray}{[78.3, 79.0]}} &  81.0 {\footnotesize \textcolor{gray}{[80.3, 81.8]}} &  -2.4 &  17 &  2 \\
 20 &  \model{se\_resnet101} &  78.4 {\footnotesize \textcolor{gray}{[78.0, 78.8]}} &  80.5 {\footnotesize \textcolor{gray}{[79.7, 81.3]}} &  -2.1 &  19 &  1 \\
 21 &  \model{resnet152} &  78.3 {\footnotesize \textcolor{gray}{[77.9, 78.7]}} &  80.3 {\footnotesize \textcolor{gray}{[79.5, 81.1]}} &  -2.0 &  21 &  0 \\
 22 &  \model{resnext101\_32x4d} &  78.2 {\footnotesize \textcolor{gray}{[77.8, 78.5]}} &  79.9 {\footnotesize \textcolor{gray}{[79.1, 80.6]}} &  -1.7 &  26 &  -4 \\
 23 &  \model{inception\_v3\_tf} &  78.0 {\footnotesize \textcolor{gray}{[77.6, 78.3]}} &  80.1 {\footnotesize \textcolor{gray}{[79.3, 80.9]}} &  -2.1 &  23 &  0 \\
 24 &  \model{resnet\_v2\_152\_tf} &  77.8 {\footnotesize \textcolor{gray}{[77.4, 78.1]}} &  80.3 {\footnotesize \textcolor{gray}{[79.5, 81.1]}} &  -2.6 &  20 &  4 \\
 25 &  \model{se\_resnet50} &  77.6 {\footnotesize \textcolor{gray}{[77.3, 78.0]}} &  79.4 {\footnotesize \textcolor{gray}{[78.6, 80.2]}} &  -1.8 &  31 &  -6 \\
 26 &  \model{fbresnet152} &  77.4 {\footnotesize \textcolor{gray}{[77.0, 77.8]}} &  80.1 {\footnotesize \textcolor{gray}{[79.3, 80.9]}} &  -2.7 &  24 &  2 \\
 27 &  \model{resnet101} &  77.4 {\footnotesize \textcolor{gray}{[77.0, 77.7]}} &  79.0 {\footnotesize \textcolor{gray}{[78.2, 79.8]}} &  -1.7 &  32 &  -5 \\
 28 &  \model{inceptionv3} &  77.3 {\footnotesize \textcolor{gray}{[77.0, 77.7]}} &  79.6 {\footnotesize \textcolor{gray}{[78.8, 80.4]}} &  -2.3 &  27 &  1 \\
 29 &  \model{inception\_v3} &  77.2 {\footnotesize \textcolor{gray}{[76.8, 77.6]}} &  79.6 {\footnotesize \textcolor{gray}{[78.8, 80.4]}} &  -2.4 &  28 &  1 \\
 30 &  \model{densenet161} &  77.1 {\footnotesize \textcolor{gray}{[76.8, 77.5]}} &  79.5 {\footnotesize \textcolor{gray}{[78.7, 80.3]}} &  -2.4 &  29 &  1 \\
 31 &  \model{dpn68b} &  77.0 {\footnotesize \textcolor{gray}{[76.7, 77.4]}} &  79.4 {\footnotesize \textcolor{gray}{[78.6, 80.2]}} &  -2.4 &  30 &  1 \\
 32 &  \model{resnet\_v2\_101\_tf} &  77.0 {\footnotesize \textcolor{gray}{[76.6, 77.3]}} &  80.1 {\footnotesize \textcolor{gray}{[79.3, 80.8]}} &  -3.1 &  25 &  7 \\
 33 &  \model{densenet201} &  76.9 {\footnotesize \textcolor{gray}{[76.5, 77.3]}} &  79.0 {\footnotesize \textcolor{gray}{[78.1, 79.7]}} &  -2.1 &  34 &  -1 \\
\bottomrule
\end{tabular}

  \centering
\end{table*}
\begin{table*}[ht!]
  \rowcolors{3}{white}{gray!15}
  \begin{tabular}{rp{4.75cm}rrrrr}
\toprule 
\multicolumn{7}{c}{\textbf{ImageNet Top-1 \datasetc}} \\ 
\midrule
\multicolumn{1}{l}{Orig.} &                               &                                        &                                          &    & \multicolumn{1}{l}{New} &  \\ 
 \multicolumn{1}{l}{Rank} & Model & Orig. Accuracy & New Accuracy & Gap & \multicolumn{1}{l}{Rank} & $\Delta$ Rank \\
\midrule
 34 &  \model{resnet\_v1\_152\_tf} &  76.8 {\footnotesize \textcolor{gray}{[76.4, 77.2]}} &  79.0 {\footnotesize \textcolor{gray}{[78.2, 79.8]}} &  -2.2 &  33 &  1 \\
 35 &  \model{resnet\_v1\_101\_tf} &  76.4 {\footnotesize \textcolor{gray}{[76.0, 76.8]}} &  78.6 {\footnotesize \textcolor{gray}{[77.8, 79.4]}} &  -2.2 &  35 &  0 \\
 36 &  \model{cafferesnet101} &  76.2 {\footnotesize \textcolor{gray}{[75.8, 76.6]}} &  78.3 {\footnotesize \textcolor{gray}{[77.4, 79.1]}} &  -2.1 &  37 &  -1 \\
 37 &  \model{resnet50} &  76.1 {\footnotesize \textcolor{gray}{[75.8, 76.5]}} &  78.1 {\footnotesize \textcolor{gray}{[77.3, 78.9]}} &  -2.0 &  38 &  -1 \\
 38 &  \model{dpn68} &  75.9 {\footnotesize \textcolor{gray}{[75.5, 76.2]}} &  78.4 {\footnotesize \textcolor{gray}{[77.6, 79.2]}} &  -2.6 &  36 &  2 \\
 39 &  \model{densenet169} &  75.6 {\footnotesize \textcolor{gray}{[75.2, 76.0]}} &  78.0 {\footnotesize \textcolor{gray}{[77.2, 78.8]}} &  -2.4 &  39 &  0 \\
 40 &  \model{resnet\_v2\_50\_tf} &  75.6 {\footnotesize \textcolor{gray}{[75.2, 76.0]}} &  78.0 {\footnotesize \textcolor{gray}{[77.2, 78.8]}} &  -2.4 &  40 &  0 \\
 41 &  \model{resnet\_v1\_50\_tf} &  75.2 {\footnotesize \textcolor{gray}{[74.8, 75.6]}} &  77.0 {\footnotesize \textcolor{gray}{[76.2, 77.9]}} &  -1.8 &  41 &  0 \\
 42 &  \model{densenet121} &  74.4 {\footnotesize \textcolor{gray}{[74.0, 74.8]}} &  76.8 {\footnotesize \textcolor{gray}{[75.9, 77.6]}} &  -2.3 &  45 &  -3 \\
 43 &  \model{vgg19\_bn} &  74.2 {\footnotesize \textcolor{gray}{[73.8, 74.6]}} &  76.6 {\footnotesize \textcolor{gray}{[75.7, 77.4]}} &  -2.3 &  46 &  -3 \\
 44 &  \model{pnasnet\_mobile\_tf} &  74.1 {\footnotesize \textcolor{gray}{[73.8, 74.5]}} &  76.8 {\footnotesize \textcolor{gray}{[76.0, 77.6]}} &  -2.7 &  44 &  0 \\
 45 &  \model{nasnetamobile} &  74.1 {\footnotesize \textcolor{gray}{[73.7, 74.5]}} &  76.4 {\footnotesize \textcolor{gray}{[75.5, 77.2]}} &  -2.3 &  47 &  -2 \\
 46 &  \model{inception\_v2\_tf} &  74.0 {\footnotesize \textcolor{gray}{[73.6, 74.4]}} &  77.0 {\footnotesize \textcolor{gray}{[76.1, 77.8]}} &  -3.0 &  43 &  3 \\
 47 &  \model{nasnet\_mobile\_tf} &  74.0 {\footnotesize \textcolor{gray}{[73.6, 74.4]}} &  76.0 {\footnotesize \textcolor{gray}{[75.1, 76.8]}} &  -2.0 &  49 &  -2 \\
 48 &  \model{bninception} &  73.5 {\footnotesize \textcolor{gray}{[73.1, 73.9]}} &  77.0 {\footnotesize \textcolor{gray}{[76.1, 77.8]}} &  -3.4 &  42 &  6 \\
 49 &  \model{vgg16\_bn} &  73.4 {\footnotesize \textcolor{gray}{[73.0, 73.7]}} &  75.9 {\footnotesize \textcolor{gray}{[75.1, 76.8]}} &  -2.6 &  50 &  -1 \\
 50 &  \model{resnet34} &  73.3 {\footnotesize \textcolor{gray}{[72.9, 73.7]}} &  76.3 {\footnotesize \textcolor{gray}{[75.4, 77.1]}} &  -3.0 &  48 &  2 \\
 51 &  \model{vgg19} &  72.4 {\footnotesize \textcolor{gray}{[72.0, 72.8]}} &  74.2 {\footnotesize \textcolor{gray}{[73.3, 75.0]}} &  -1.8 &  51 &  0 \\
 52 &  \model{vgg16} &  71.6 {\footnotesize \textcolor{gray}{[71.2, 72.0]}} &  73.9 {\footnotesize \textcolor{gray}{[73.0, 74.7]}} &  -2.3 &  52 &  0 \\
 53 &  \model{vgg13\_bn} &  71.6 {\footnotesize \textcolor{gray}{[71.2, 72.0]}} &  73.5 {\footnotesize \textcolor{gray}{[72.7, 74.4]}} &  -1.9 &  55 &  -2 \\
 54 &  \model{mobilenet\_v1\_tf} &  71.0 {\footnotesize \textcolor{gray}{[70.6, 71.4]}} &  72.4 {\footnotesize \textcolor{gray}{[71.5, 73.3]}} &  -1.4 &  59 &  -5 \\
 55 &  \model{vgg\_19\_tf} &  71.0 {\footnotesize \textcolor{gray}{[70.6, 71.4]}} &  73.6 {\footnotesize \textcolor{gray}{[72.7, 74.5]}} &  -2.6 &  53 &  2 \\
 56 &  \model{vgg\_16\_tf} &  70.9 {\footnotesize \textcolor{gray}{[70.5, 71.3]}} &  73.5 {\footnotesize \textcolor{gray}{[72.7, 74.4]}} &  -2.6 &  54 &  2 \\
 57 &  \model{vgg11\_bn} &  70.4 {\footnotesize \textcolor{gray}{[70.0, 70.8]}} &  73.0 {\footnotesize \textcolor{gray}{[72.1, 73.8]}} &  -2.6 &  58 &  -1 \\
 58 &  \model{vgg13} &  69.9 {\footnotesize \textcolor{gray}{[69.5, 70.3]}} &  72.0 {\footnotesize \textcolor{gray}{[71.1, 72.9]}} &  -2.1 &  60 &  -2 \\
 59 &  \model{inception\_v1\_tf} &  69.8 {\footnotesize \textcolor{gray}{[69.4, 70.2]}} &  73.1 {\footnotesize \textcolor{gray}{[72.2, 73.9]}} &  -3.3 &  56 &  3 \\
 60 &  \model{resnet18} &  69.8 {\footnotesize \textcolor{gray}{[69.4, 70.2]}} &  73.0 {\footnotesize \textcolor{gray}{[72.2, 73.9]}} &  -3.3 &  57 &  3 \\
 61 &  \model{vgg11} &  69.0 {\footnotesize \textcolor{gray}{[68.6, 69.4]}} &  70.8 {\footnotesize \textcolor{gray}{[69.9, 71.7]}} &  -1.8 &  61 &  0 \\
 62 &  \model{squeezenet1\_1} &  58.2 {\footnotesize \textcolor{gray}{[57.7, 58.6]}} &  61.7 {\footnotesize \textcolor{gray}{[60.7, 62.6]}} &  -3.5 &  62 &  0 \\
 63 &  \model{squeezenet1\_0} &  58.1 {\footnotesize \textcolor{gray}{[57.7, 58.5]}} &  60.7 {\footnotesize \textcolor{gray}{[59.7, 61.7]}} &  -2.6 &  63 &  0 \\
 64 &  \model{alexnet} &  56.5 {\footnotesize \textcolor{gray}{[56.1, 57.0]}} &  58.2 {\footnotesize \textcolor{gray}{[57.2, 59.1]}} &  -1.7 &  64 &  0 \\
 65 &  \model{fv\_64k} &  35.1 {\footnotesize \textcolor{gray}{[34.7, 35.5]}} &  34.2 {\footnotesize \textcolor{gray}{[33.3, 35.2]}} &  0.8 &  65 &  0 \\
 66 &  \model{fv\_16k} &  28.3 {\footnotesize \textcolor{gray}{[27.9, 28.7]}} &  27.4 {\footnotesize \textcolor{gray}{[26.6, 28.3]}} &  0.9 &  66 &  0 \\
 67 &  \model{fv\_4k} &  21.2 {\footnotesize \textcolor{gray}{[20.8, 21.5]}} &  21.1 {\footnotesize \textcolor{gray}{[20.3, 21.9]}} &  0.1 &  67 &  0 \\
\bottomrule
\end{tabular}

  \centering
\end{table*}

\begin{table*}[ht!]
  \rowcolors{3}{white}{gray!15}
  \caption{Top-5 model accuracy on the original ImageNet validation set and our new test set \datasetc{}.
  $\Delta$ Rank is the relative difference in the ranking from the original test set to the new test set.
  For example, $\Delta \text{Rank} = -2$ means that a model dropped by two places on the new test set compared to the original test set.
  The confidence intervals are 95\% Clopper-Pearson intervals.
  Due to space constraints, references for the models can be found in Appendix \ref{apx:imagenet_model_descriptions}.
  The second part of the table can be found on the following page.
  }
  \begin{tabular}{rp{4.75cm}rrrrr}
\toprule 
\multicolumn{7}{c}{\textbf{ImageNet Top-5 \datasetc}} \\ 
\midrule
\multicolumn{1}{l}{Orig.} &                               &                                        &                                          &    & \multicolumn{1}{l}{New} &  \\ 
 \multicolumn{1}{l}{Rank} & Model & Orig. Accuracy & New Accuracy & Gap & \multicolumn{1}{l}{Rank} & $\Delta$ Rank \\
\midrule
 1 &  \model{pnasnet\_large\_tf} &  96.2 {\footnotesize \textcolor{gray}{[96.0, 96.3]}} &  97.2 {\footnotesize \textcolor{gray}{[96.9, 97.5]}} &  -1.0 &  2 &  -1 \\
 2 &  \model{nasnet\_large\_tf} &  96.2 {\footnotesize \textcolor{gray}{[96.0, 96.3]}} &  97.2 {\footnotesize \textcolor{gray}{[96.9, 97.5]}} &  -1.0 &  1 &  1 \\
 3 &  \model{nasnetalarge} &  96.0 {\footnotesize \textcolor{gray}{[95.8, 96.2]}} &  97.1 {\footnotesize \textcolor{gray}{[96.7, 97.4]}} &  -1.1 &  3 &  0 \\
 4 &  \model{pnasnet5large} &  96.0 {\footnotesize \textcolor{gray}{[95.8, 96.2]}} &  96.9 {\footnotesize \textcolor{gray}{[96.6, 97.2]}} &  -0.9 &  4 &  0 \\
 5 &  \model{polynet} &  95.6 {\footnotesize \textcolor{gray}{[95.4, 95.7]}} &  96.8 {\footnotesize \textcolor{gray}{[96.4, 97.1]}} &  -1.2 &  5 &  0 \\
 6 &  \model{senet154} &  95.5 {\footnotesize \textcolor{gray}{[95.3, 95.7]}} &  96.6 {\footnotesize \textcolor{gray}{[96.2, 97.0]}} &  -1.1 &  8 &  -2 \\
 7 &  \model{inception\_resnet\_v2\_tf} &  95.2 {\footnotesize \textcolor{gray}{[95.1, 95.4]}} &  96.8 {\footnotesize \textcolor{gray}{[96.4, 97.1]}} &  -1.5 &  6 &  1 \\
 8 &  \model{inception\_v4\_tf} &  95.2 {\footnotesize \textcolor{gray}{[95.0, 95.4]}} &  96.5 {\footnotesize \textcolor{gray}{[96.1, 96.9]}} &  -1.3 &  9 &  -1 \\
 9 &  \model{inceptionresnetv2} &  95.1 {\footnotesize \textcolor{gray}{[94.9, 95.3]}} &  96.7 {\footnotesize \textcolor{gray}{[96.3, 97.0]}} &  -1.5 &  7 &  2 \\
 10 &  \model{se\_resnext101\_32x4d} &  95.0 {\footnotesize \textcolor{gray}{[94.8, 95.2]}} &  96.2 {\footnotesize \textcolor{gray}{[95.8, 96.6]}} &  -1.2 &  11 &  -1 \\
 11 &  \model{inceptionv4} &  94.9 {\footnotesize \textcolor{gray}{[94.7, 95.1]}} &  96.4 {\footnotesize \textcolor{gray}{[96.0, 96.7]}} &  -1.5 &  10 &  1 \\
 12 &  \model{dpn107} &  94.7 {\footnotesize \textcolor{gray}{[94.5, 94.9]}} &  96.0 {\footnotesize \textcolor{gray}{[95.6, 96.4]}} &  -1.4 &  13 &  -1 \\
 13 &  \model{dpn92} &  94.6 {\footnotesize \textcolor{gray}{[94.4, 94.8]}} &  95.9 {\footnotesize \textcolor{gray}{[95.5, 96.3]}} &  -1.3 &  17 &  -4 \\
 14 &  \model{dpn131} &  94.6 {\footnotesize \textcolor{gray}{[94.4, 94.8]}} &  96.0 {\footnotesize \textcolor{gray}{[95.6, 96.4]}} &  -1.5 &  14 &  0 \\
 15 &  \model{dpn98} &  94.5 {\footnotesize \textcolor{gray}{[94.3, 94.7]}} &  96.0 {\footnotesize \textcolor{gray}{[95.6, 96.4]}} &  -1.5 &  15 &  0 \\
 16 &  \model{se\_resnext50\_32x4d} &  94.4 {\footnotesize \textcolor{gray}{[94.2, 94.6]}} &  95.9 {\footnotesize \textcolor{gray}{[95.5, 96.3]}} &  -1.5 &  18 &  -2 \\
 17 &  \model{se\_resnet152} &  94.4 {\footnotesize \textcolor{gray}{[94.2, 94.6]}} &  95.9 {\footnotesize \textcolor{gray}{[95.5, 96.3]}} &  -1.5 &  19 &  -2 \\
 18 &  \model{xception} &  94.3 {\footnotesize \textcolor{gray}{[94.1, 94.5]}} &  95.9 {\footnotesize \textcolor{gray}{[95.5, 96.3]}} &  -1.6 &  20 &  -2 \\
 19 &  \model{se\_resnet101} &  94.3 {\footnotesize \textcolor{gray}{[94.1, 94.5]}} &  95.9 {\footnotesize \textcolor{gray}{[95.5, 96.3]}} &  -1.6 &  21 &  -2 \\
 20 &  \model{resnext101\_64x4d} &  94.3 {\footnotesize \textcolor{gray}{[94.0, 94.5]}} &  95.7 {\footnotesize \textcolor{gray}{[95.3, 96.1]}} &  -1.5 &  23 &  -3 \\
 21 &  \model{resnet\_v2\_152\_tf} &  94.1 {\footnotesize \textcolor{gray}{[93.9, 94.3]}} &  96.0 {\footnotesize \textcolor{gray}{[95.6, 96.3]}} &  -1.9 &  16 &  5 \\
 22 &  \model{resnet152} &  94.0 {\footnotesize \textcolor{gray}{[93.8, 94.3]}} &  96.2 {\footnotesize \textcolor{gray}{[95.8, 96.5]}} &  -2.1 &  12 &  10 \\
 23 &  \model{inception\_v3\_tf} &  93.9 {\footnotesize \textcolor{gray}{[93.7, 94.1]}} &  95.5 {\footnotesize \textcolor{gray}{[95.1, 95.9]}} &  -1.5 &  25 &  -2 \\
 24 &  \model{resnext101\_32x4d} &  93.9 {\footnotesize \textcolor{gray}{[93.7, 94.1]}} &  95.2 {\footnotesize \textcolor{gray}{[94.8, 95.6]}} &  -1.3 &  31 &  -7 \\
 25 &  \model{se\_resnet50} &  93.8 {\footnotesize \textcolor{gray}{[93.5, 94.0]}} &  95.5 {\footnotesize \textcolor{gray}{[95.1, 95.9]}} &  -1.8 &  24 &  1 \\
 26 &  \model{resnet\_v2\_101\_tf} &  93.7 {\footnotesize \textcolor{gray}{[93.5, 93.9]}} &  95.8 {\footnotesize \textcolor{gray}{[95.4, 96.2]}} &  -2.1 &  22 &  4 \\
 27 &  \model{fbresnet152} &  93.6 {\footnotesize \textcolor{gray}{[93.4, 93.8]}} &  95.2 {\footnotesize \textcolor{gray}{[94.8, 95.7]}} &  -1.7 &  28 &  -1 \\
 28 &  \model{dpn68b} &  93.6 {\footnotesize \textcolor{gray}{[93.4, 93.8]}} &  95.2 {\footnotesize \textcolor{gray}{[94.8, 95.6]}} &  -1.6 &  32 &  -4 \\
 29 &  \model{densenet161} &  93.6 {\footnotesize \textcolor{gray}{[93.3, 93.8]}} &  95.2 {\footnotesize \textcolor{gray}{[94.8, 95.6]}} &  -1.7 &  29 &  0 \\
 30 &  \model{resnet101} &  93.5 {\footnotesize \textcolor{gray}{[93.3, 93.8]}} &  95.4 {\footnotesize \textcolor{gray}{[95.0, 95.8]}} &  -1.9 &  26 &  4 \\
 31 &  \model{inception\_v3} &  93.5 {\footnotesize \textcolor{gray}{[93.3, 93.7]}} &  95.1 {\footnotesize \textcolor{gray}{[94.7, 95.5]}} &  -1.6 &  34 &  -3 \\
 32 &  \model{inceptionv3} &  93.4 {\footnotesize \textcolor{gray}{[93.2, 93.6]}} &  95.2 {\footnotesize \textcolor{gray}{[94.8, 95.6]}} &  -1.8 &  30 &  2 \\
 33 &  \model{densenet201} &  93.4 {\footnotesize \textcolor{gray}{[93.1, 93.6]}} &  95.2 {\footnotesize \textcolor{gray}{[94.8, 95.7]}} &  -1.9 &  27 &  6 \\
\bottomrule
\end{tabular}

  \label{tab:imagenetv2-c-12_top5_full_results}
  \centering
\end{table*}
\begin{table*}[ht!]
  \rowcolors{3}{white}{gray!15}
  \begin{tabular}{rp{4.75cm}rrrrr}
\toprule 
\multicolumn{7}{c}{\textbf{ImageNet Top-5 \datasetc}} \\ 
\midrule
\multicolumn{1}{l}{Orig.} &                               &                                        &                                          &    & \multicolumn{1}{l}{New} &  \\ 
 \multicolumn{1}{l}{Rank} & Model & Orig. Accuracy & New Accuracy & Gap & \multicolumn{1}{l}{Rank} & $\Delta$ Rank \\
\midrule
 34 &  \model{resnet\_v1\_152\_tf} &  93.2 {\footnotesize \textcolor{gray}{[92.9, 93.4]}} &  95.2 {\footnotesize \textcolor{gray}{[94.7, 95.6]}} &  -2.0 &  33 &  1 \\
 35 &  \model{resnet\_v1\_101\_tf} &  92.9 {\footnotesize \textcolor{gray}{[92.7, 93.1]}} &  94.9 {\footnotesize \textcolor{gray}{[94.4, 95.3]}} &  -2.0 &  35 &  0 \\
 36 &  \model{resnet50} &  92.9 {\footnotesize \textcolor{gray}{[92.6, 93.1]}} &  94.7 {\footnotesize \textcolor{gray}{[94.2, 95.1]}} &  -1.8 &  39 &  -3 \\
 37 &  \model{resnet\_v2\_50\_tf} &  92.8 {\footnotesize \textcolor{gray}{[92.6, 93.1]}} &  94.8 {\footnotesize \textcolor{gray}{[94.3, 95.2]}} &  -1.9 &  37 &  0 \\
 38 &  \model{densenet169} &  92.8 {\footnotesize \textcolor{gray}{[92.6, 93.0]}} &  94.7 {\footnotesize \textcolor{gray}{[94.2, 95.1]}} &  -1.9 &  38 &  0 \\
 39 &  \model{dpn68} &  92.8 {\footnotesize \textcolor{gray}{[92.5, 93.0]}} &  94.8 {\footnotesize \textcolor{gray}{[94.3, 95.2]}} &  -2.0 &  36 &  3 \\
 40 &  \model{cafferesnet101} &  92.8 {\footnotesize \textcolor{gray}{[92.5, 93.0]}} &  94.6 {\footnotesize \textcolor{gray}{[94.1, 95.0]}} &  -1.8 &  40 &  0 \\
 41 &  \model{resnet\_v1\_50\_tf} &  92.2 {\footnotesize \textcolor{gray}{[92.0, 92.4]}} &  94.2 {\footnotesize \textcolor{gray}{[93.8, 94.7]}} &  -2.1 &  41 &  0 \\
 42 &  \model{densenet121} &  92.0 {\footnotesize \textcolor{gray}{[91.7, 92.2]}} &  94.0 {\footnotesize \textcolor{gray}{[93.5, 94.5]}} &  -2.0 &  46 &  -4 \\
 43 &  \model{pnasnet\_mobile\_tf} &  91.9 {\footnotesize \textcolor{gray}{[91.6, 92.1]}} &  94.1 {\footnotesize \textcolor{gray}{[93.6, 94.5]}} &  -2.2 &  44 &  -1 \\
 44 &  \model{vgg19\_bn} &  91.8 {\footnotesize \textcolor{gray}{[91.6, 92.1]}} &  94.0 {\footnotesize \textcolor{gray}{[93.5, 94.4]}} &  -2.1 &  47 &  -3 \\
 45 &  \model{inception\_v2\_tf} &  91.8 {\footnotesize \textcolor{gray}{[91.5, 92.0]}} &  94.2 {\footnotesize \textcolor{gray}{[93.7, 94.7]}} &  -2.5 &  42 &  3 \\
 46 &  \model{nasnetamobile} &  91.7 {\footnotesize \textcolor{gray}{[91.5, 92.0]}} &  94.1 {\footnotesize \textcolor{gray}{[93.6, 94.5]}} &  -2.3 &  43 &  3 \\
 47 &  \model{nasnet\_mobile\_tf} &  91.6 {\footnotesize \textcolor{gray}{[91.3, 91.8]}} &  93.8 {\footnotesize \textcolor{gray}{[93.4, 94.3]}} &  -2.3 &  49 &  -2 \\
 48 &  \model{bninception} &  91.6 {\footnotesize \textcolor{gray}{[91.3, 91.8]}} &  94.0 {\footnotesize \textcolor{gray}{[93.6, 94.5]}} &  -2.5 &  45 &  3 \\
 49 &  \model{vgg16\_bn} &  91.5 {\footnotesize \textcolor{gray}{[91.3, 91.8]}} &  93.7 {\footnotesize \textcolor{gray}{[93.2, 94.1]}} &  -2.1 &  50 &  -1 \\
 50 &  \model{resnet34} &  91.4 {\footnotesize \textcolor{gray}{[91.2, 91.7]}} &  93.9 {\footnotesize \textcolor{gray}{[93.4, 94.3]}} &  -2.5 &  48 &  2 \\
 51 &  \model{vgg19} &  90.9 {\footnotesize \textcolor{gray}{[90.6, 91.1]}} &  92.8 {\footnotesize \textcolor{gray}{[92.2, 93.3]}} &  -1.9 &  51 &  0 \\
 52 &  \model{vgg16} &  90.4 {\footnotesize \textcolor{gray}{[90.1, 90.6]}} &  92.5 {\footnotesize \textcolor{gray}{[92.0, 93.0]}} &  -2.1 &  53 &  -1 \\
 53 &  \model{vgg13\_bn} &  90.4 {\footnotesize \textcolor{gray}{[90.1, 90.6]}} &  92.6 {\footnotesize \textcolor{gray}{[92.1, 93.1]}} &  -2.2 &  52 &  1 \\
 54 &  \model{mobilenet\_v1\_tf} &  90.0 {\footnotesize \textcolor{gray}{[89.7, 90.2]}} &  91.4 {\footnotesize \textcolor{gray}{[90.8, 91.9]}} &  -1.4 &  59 &  -5 \\
 56 &  \model{vgg\_19\_tf} &  89.8 {\footnotesize \textcolor{gray}{[89.6, 90.1]}} &  92.1 {\footnotesize \textcolor{gray}{[91.5, 92.6]}} &  -2.2 &  56 &  0 \\
 55 &  \model{vgg\_16\_tf} &  89.8 {\footnotesize \textcolor{gray}{[89.6, 90.1]}} &  92.2 {\footnotesize \textcolor{gray}{[91.6, 92.7]}} &  -2.3 &  54 &  1 \\
 57 &  \model{vgg11\_bn} &  89.8 {\footnotesize \textcolor{gray}{[89.5, 90.1]}} &  91.9 {\footnotesize \textcolor{gray}{[91.4, 92.5]}} &  -2.1 &  58 &  -1 \\
 58 &  \model{inception\_v1\_tf} &  89.6 {\footnotesize \textcolor{gray}{[89.4, 89.9]}} &  92.1 {\footnotesize \textcolor{gray}{[91.6, 92.6]}} &  -2.5 &  55 &  3 \\
 59 &  \model{vgg13} &  89.2 {\footnotesize \textcolor{gray}{[89.0, 89.5]}} &  91.4 {\footnotesize \textcolor{gray}{[90.8, 91.9]}} &  -2.2 &  60 &  -1 \\
 60 &  \model{resnet18} &  89.1 {\footnotesize \textcolor{gray}{[88.8, 89.3]}} &  92.0 {\footnotesize \textcolor{gray}{[91.4, 92.5]}} &  -2.9 &  57 &  3 \\
 61 &  \model{vgg11} &  88.6 {\footnotesize \textcolor{gray}{[88.3, 88.9]}} &  91.0 {\footnotesize \textcolor{gray}{[90.4, 91.5]}} &  -2.4 &  61 &  0 \\
 62 &  \model{squeezenet1\_1} &  80.6 {\footnotesize \textcolor{gray}{[80.3, 81.0]}} &  83.9 {\footnotesize \textcolor{gray}{[83.1, 84.6]}} &  -3.2 &  62 &  0 \\
 63 &  \model{squeezenet1\_0} &  80.4 {\footnotesize \textcolor{gray}{[80.1, 80.8]}} &  83.5 {\footnotesize \textcolor{gray}{[82.8, 84.3]}} &  -3.1 &  63 &  0 \\
 64 &  \model{alexnet} &  79.1 {\footnotesize \textcolor{gray}{[78.7, 79.4]}} &  81.8 {\footnotesize \textcolor{gray}{[81.0, 82.6]}} &  -2.7 &  64 &  0 \\
 65 &  \model{fv\_64k} &  55.7 {\footnotesize \textcolor{gray}{[55.3, 56.2]}} &  55.9 {\footnotesize \textcolor{gray}{[54.9, 56.8]}} &  -0.1 &  65 &  0 \\
 66 &  \model{fv\_16k} &  49.9 {\footnotesize \textcolor{gray}{[49.5, 50.4]}} &  49.8 {\footnotesize \textcolor{gray}{[48.8, 50.8]}} &  0.1 &  66 &  0 \\
 67 &  \model{fv\_4k} &  41.3 {\footnotesize \textcolor{gray}{[40.8, 41.7]}} &  41.9 {\footnotesize \textcolor{gray}{[40.9, 42.8]}} &  -0.6 &  67 &  0 \\
\bottomrule
\end{tabular}

  \centering
\end{table*}

\newpage
\clearpage

\subsubsection{Accuracy Plots for All ImageNet Test Sets}
Figure \ref{fig:imagenet_plotpage} shows the top-1 and top-5 accuracies for our three test sets and all convolutional networks in our model testbed.
Figure \ref{fig:imagenet_probit_plotpage} shows the accuracies for all models (including Fisher Vector models) with a probit scale on the axes.

\begin{figure*}[ht!]
  \newcommand{\ppwm}{0.43}
  \newcommand{\pphp}{.5cm}
  \centering
  \hspace{\pphp}
  \begin{subfigure}{\ppwm\textwidth}
    \includegraphics[width=\linewidth]{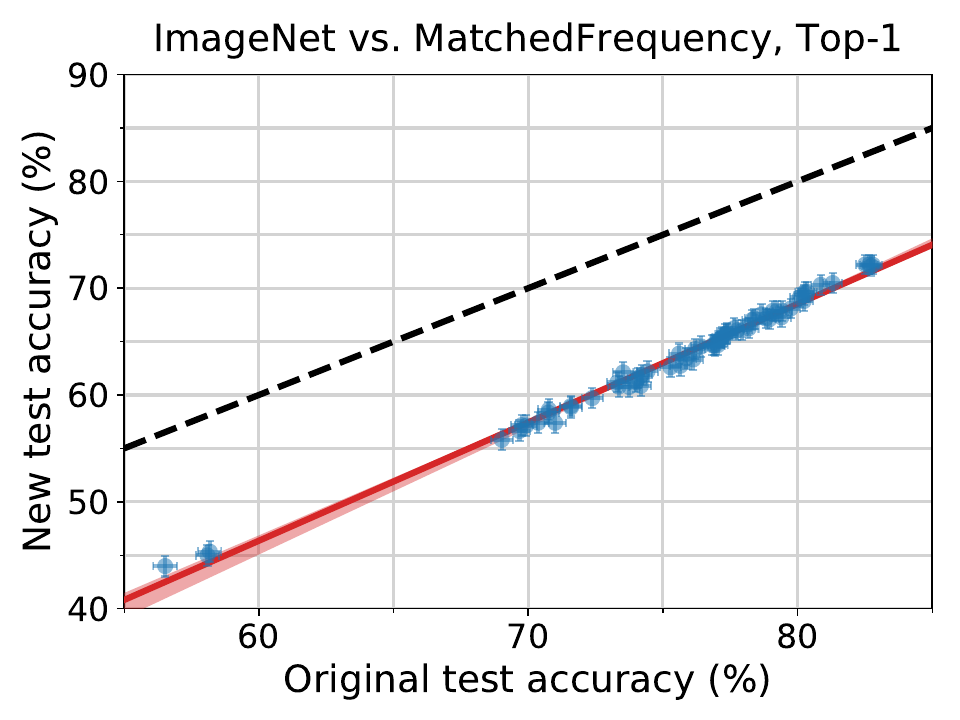}
  \end{subfigure}
  \hfill
  \begin{subfigure}{\ppwm\textwidth}
    \includegraphics[width=\linewidth]{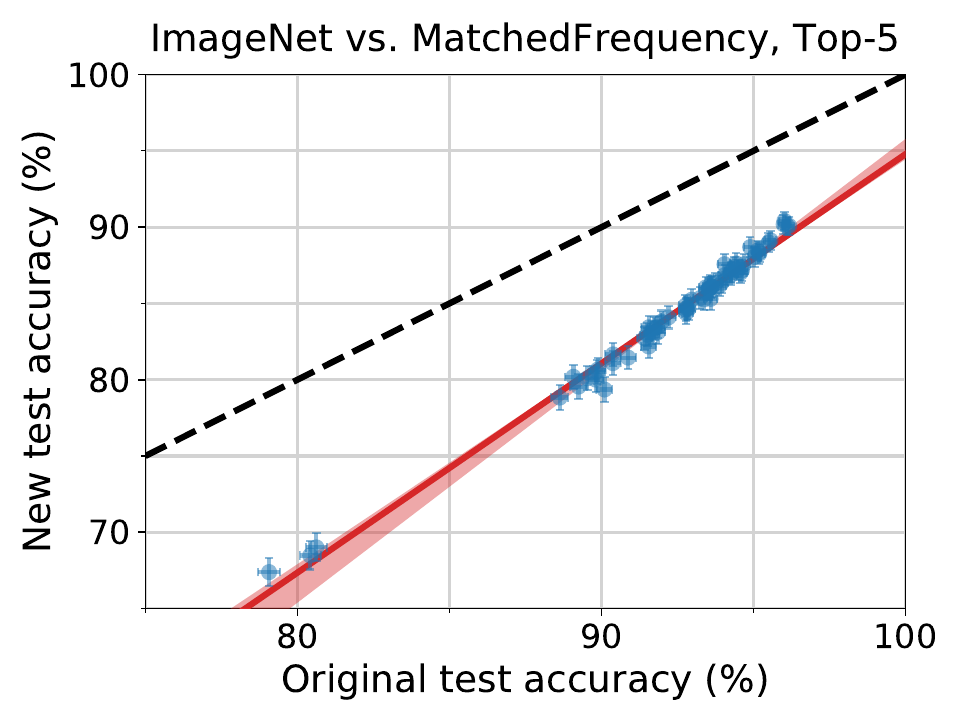}
  \end{subfigure}
  \hspace{\pphp} $ $ \\[.2cm]
  \hspace{\pphp}
  \begin{subfigure}{\ppwm\textwidth}
    \includegraphics[width=\linewidth]{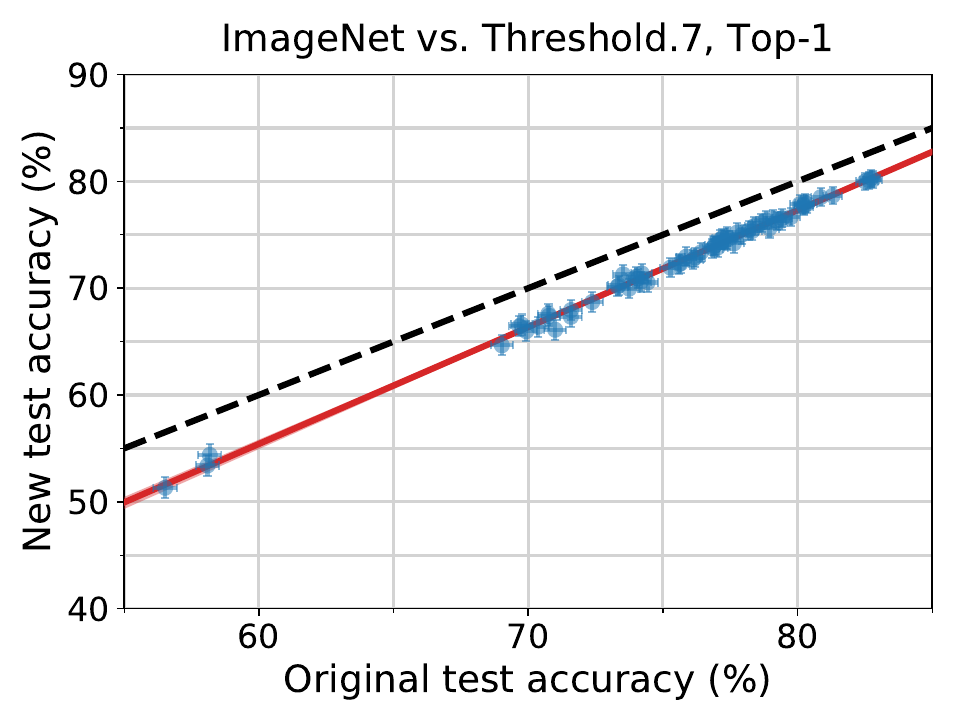}
  \end{subfigure}
  \hfill
  \begin{subfigure}{\ppwm\textwidth}
    \includegraphics[width=\linewidth]{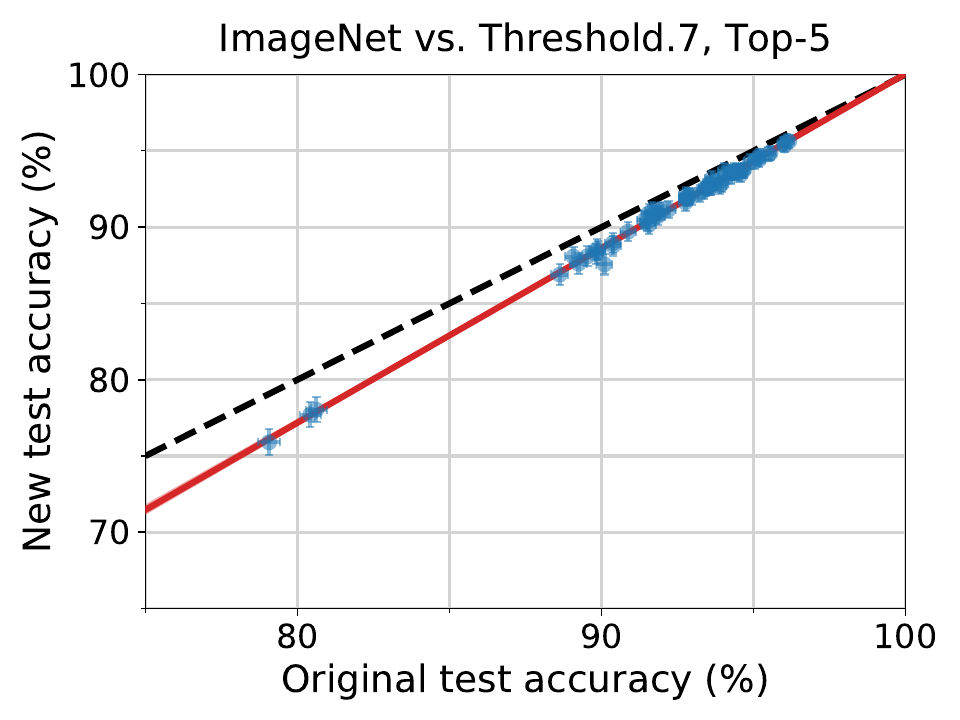}
  \end{subfigure}
  \hspace{\pphp} $ $ \\[.2cm]
  \hspace{\pphp}
  \begin{subfigure}{\ppwm\textwidth}
    \includegraphics[width=\linewidth]{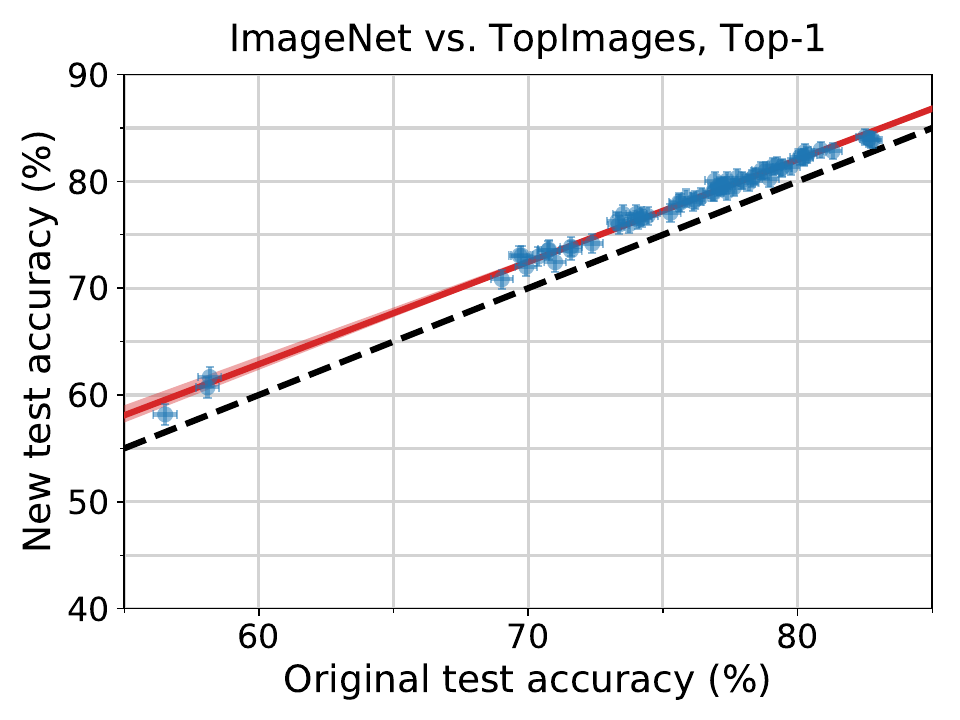}
  \end{subfigure}
  \hfill
  \begin{subfigure}{\ppwm\textwidth}
    \includegraphics[width=\linewidth]{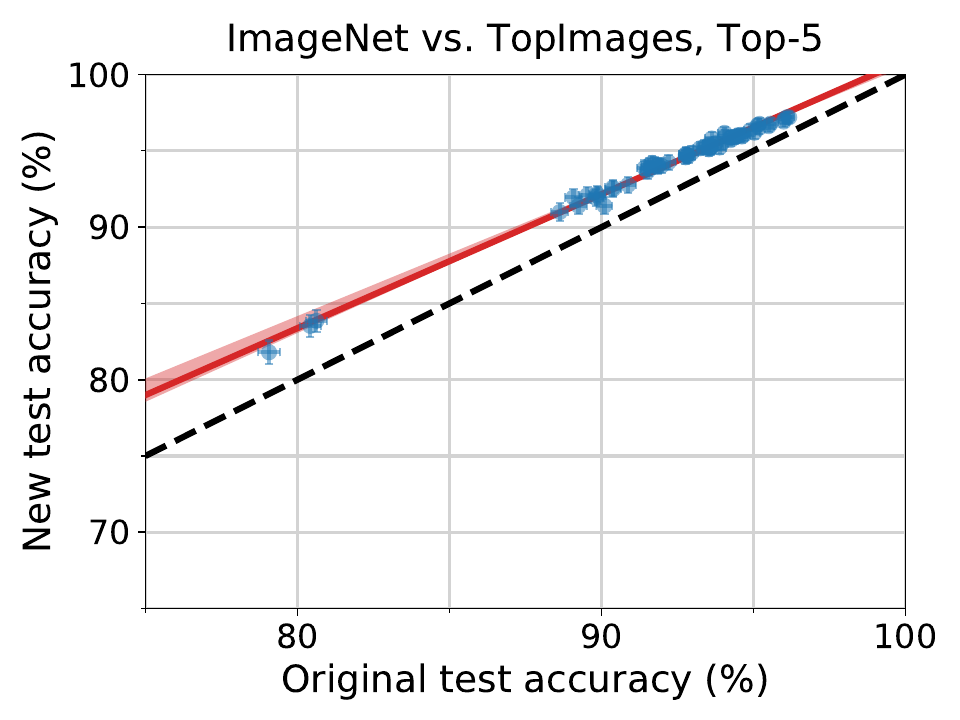}
  \end{subfigure} \hspace{\pphp} $ $  \\[-.1cm]
  \begin{subfigure}{\textwidth}
    \centering
    \includegraphics[width=.75\linewidth]{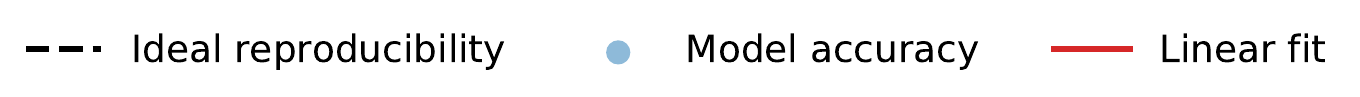}
  \end{subfigure}
  \vspace{-.6cm}
  \caption{
    \small
    Model accuracy on the original ImageNet validation set vs.\ our new test sets.
    See Section \ref{sec:imagenet_details} for a description of these test sets.
    Each data point corresponds to one model in our testbed (shown with 95\% Clopper-Pearson confidence intervals).
    The red shaded region is a 95\% confidence region for the linear fit from 100,000 bootstrap samples.
    For \datasetb{}, the accuracies on the new test set are significantly below the original accuracies.
    The accuracies for \dataseta{} are still below the original counterpart, but for \datasetc{} they improve over the original test accuracies.
    This shows that small variations in the data generation process can have significant impact on the accuracy scores.
    As for CIFAR-10, all plots reveal an approximatly linear relationship between original and new test accuracy.
    Only the slope for the top-5 accuracies on \datasetc{} is significantly smaller than 1 (0.88, 95\% confidence interval from 100,000 bootstrap samples: $[0.81, 0.91]$).
    It is unclear if this is a sign of adaptive overfitting or due to the models approaching the 100\% accuracy regime.
    Investigating this further is an interesting question for future work.
    \vspace{-.3cm}
  }
  \label{fig:imagenet_plotpage}
\end{figure*}

\begin{figure*}[ht!]
  \newcommand{\ppwm}{0.43}
  \newcommand{\pphp}{.5cm}
  \centering
  \hspace{\pphp}
  \begin{subfigure}{\ppwm\textwidth}
    \includegraphics[width=\linewidth]{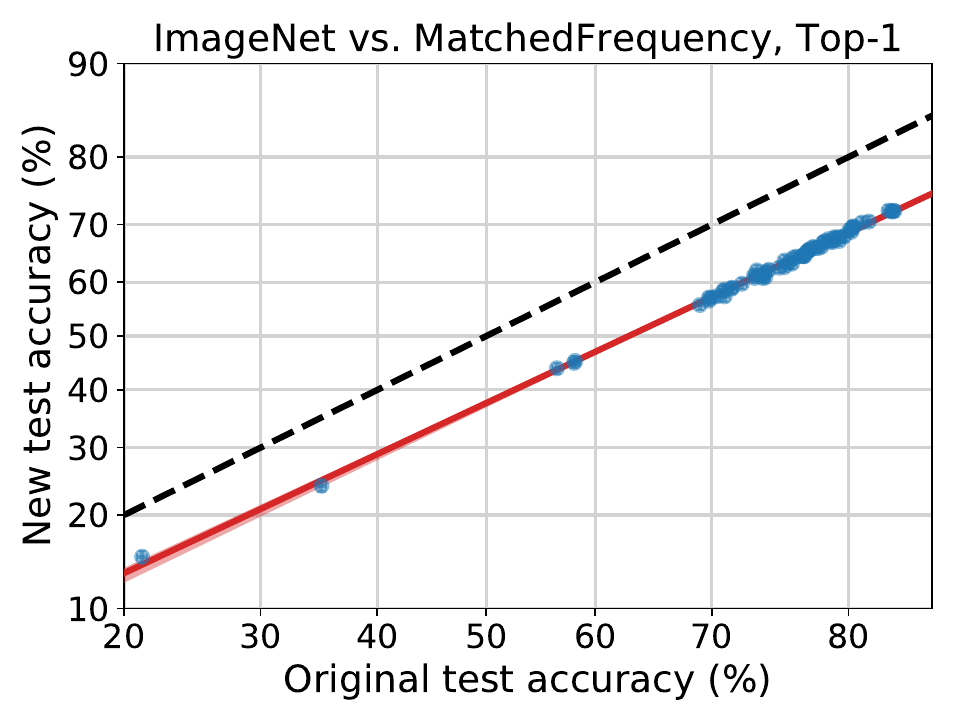}
  \end{subfigure}
  \hfill
  \begin{subfigure}{\ppwm\textwidth}
    \includegraphics[width=\linewidth]{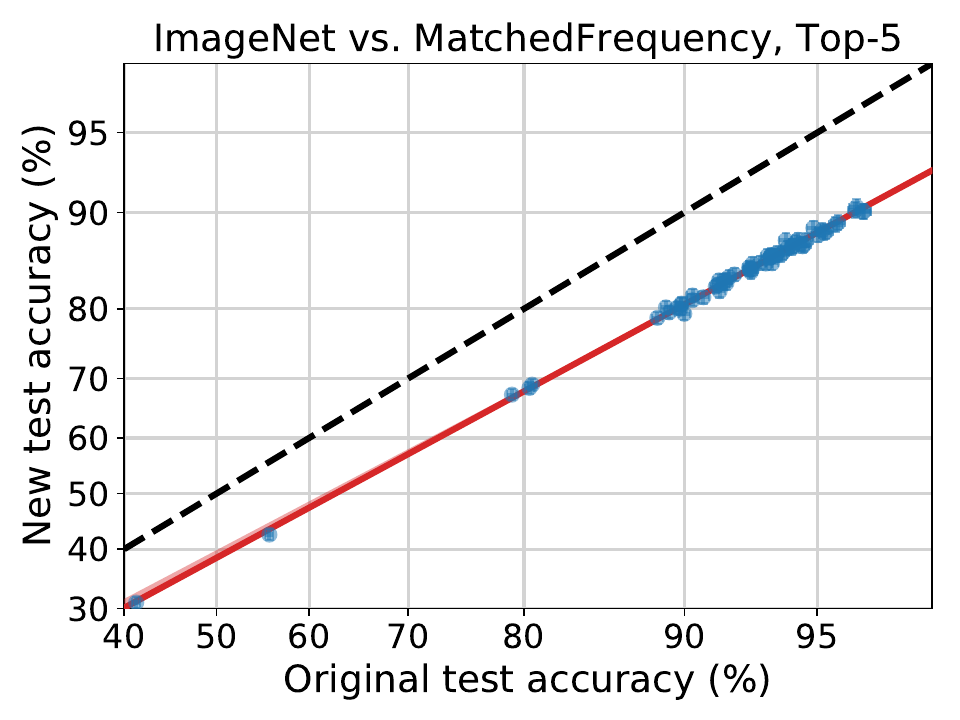}
  \end{subfigure}
  \hspace{\pphp} $ $ \\[.2cm]
  \hspace{\pphp}
  \begin{subfigure}{\ppwm\textwidth}
    \includegraphics[width=\linewidth]{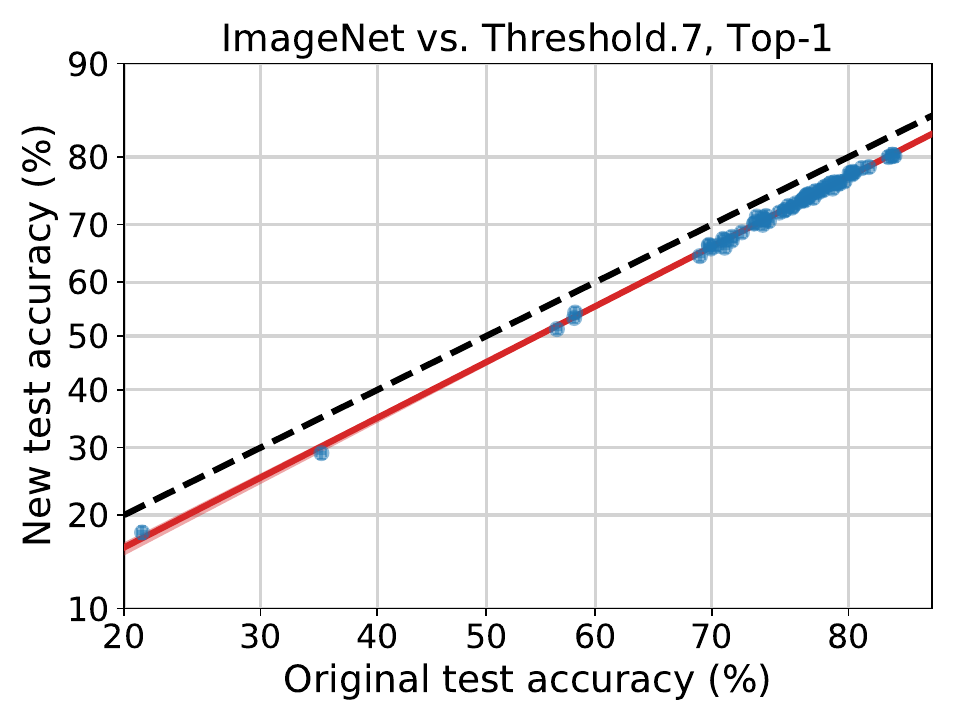}
  \end{subfigure}
  \hfill
  \begin{subfigure}{\ppwm\textwidth}
    \includegraphics[width=\linewidth]{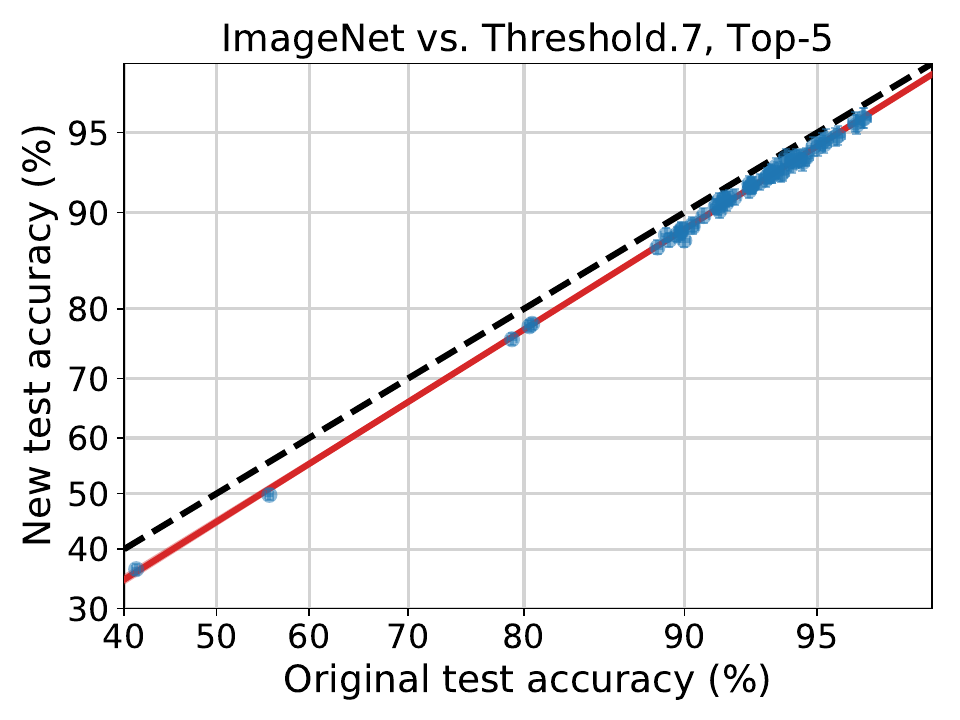}
  \end{subfigure}
  \hspace{\pphp} $ $ \\[.2cm]
  \hspace{\pphp}
  \begin{subfigure}{\ppwm\textwidth}
    \includegraphics[width=\linewidth]{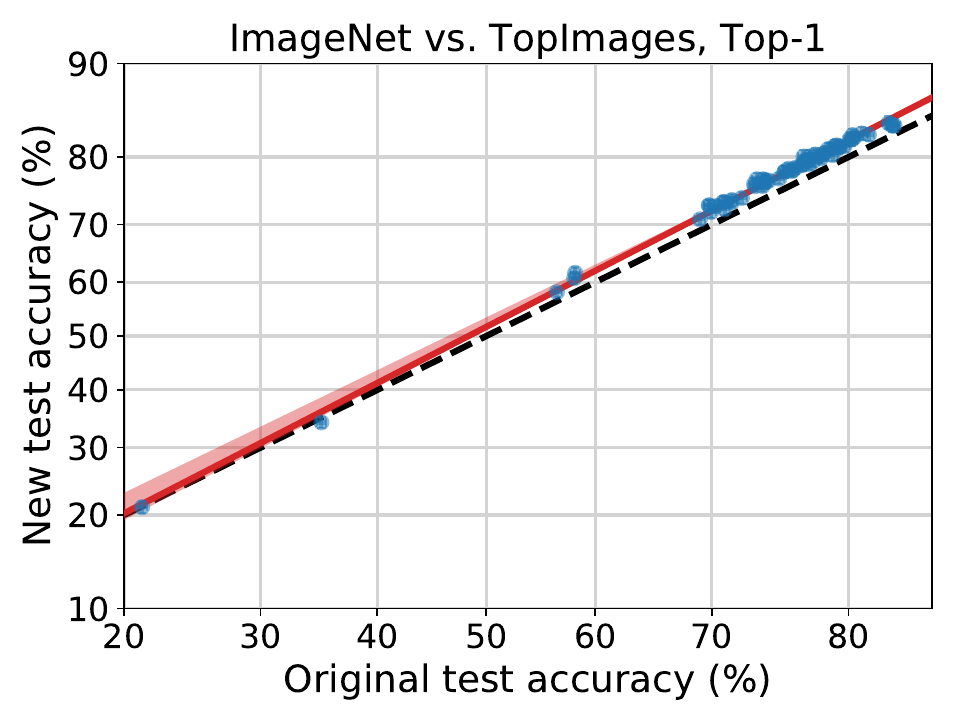}
  \end{subfigure}
  \hfill
  \begin{subfigure}{\ppwm\textwidth}
    \includegraphics[width=\linewidth]{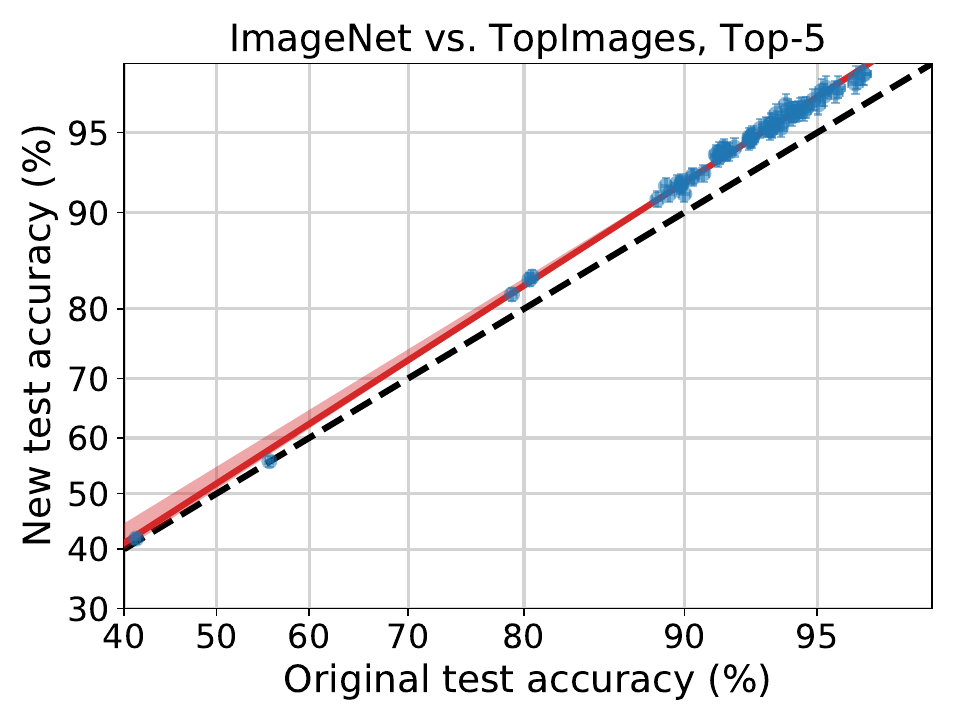}
  \end{subfigure} \hspace{\pphp} $ $  \\[-.1cm]
  \begin{subfigure}{\textwidth}
    \centering
    \includegraphics[width=.75\linewidth]{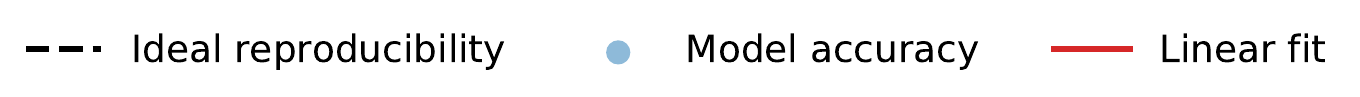}
  \end{subfigure}
  \vspace{-.6cm}
  \caption{
    \small
    Model accuracy on the original ImageNet validation set vs.\ our new test sets.
    The structure of the plots is similar to Figure \ref{fig:imagenet_plotpage} and we refer the reader to the description there.
    In contrast to Figure \ref{fig:imagenet_plotpage}, the plots here contain also the Fisher Vector models.
    Moreover, the axes are scaled according to the probit transformation, i.e., accuracy $\alpha$ appears at $\Phi^{-1}(\alpha)$, where $\Phi$ is the Gaussian CDF.
    For all three datasets and both top-1 and top-5 accuracy, the plots reveal a good linear fit in the probit domain spanning around 60 percentage points of accuracy.
    All plots include a 95\% confidence region for the linear fit as in Figure \ref{fig:imagenet_plotpage}, but the red shaded region is hard to see in some of the plots due to its small size.
    \vspace{-.3cm}
  }
  \label{fig:imagenet_probit_plotpage}
\end{figure*}

\subsubsection{Example Images}
Figure \ref{fig:testexamples_imagenet} shows randomly selected images for three randomly selected classes for both the original ImageNet validation set and our three new test sets.

\begin{figure*}
   \centering
   \setlength{\imagedim}{2cm}
   \setlength{\imagexspacing}{0.3cm}
   \setlength{\imageyspacing}{0.3cm}
   \begin{subfigure}[t]{\textwidth}
   \begin{tikzpicture}
   \tikzstyle{img}=[inner sep=0pt,outer sep=0pt];
   \node [img] (image0) {\includegraphics[width=\imagedim, height=\imagedim]{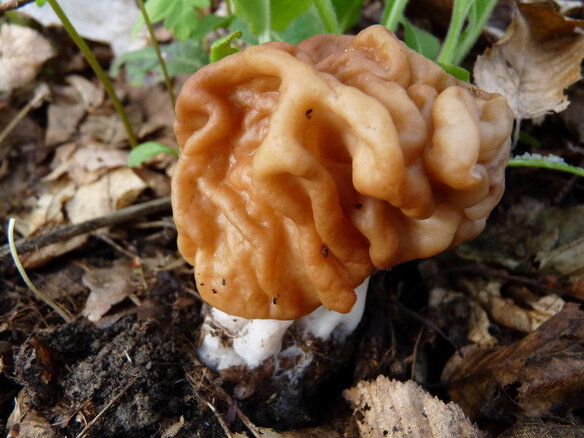}};
   \node [img,anchor=west,at=(image0.east),xshift=\imagexspacing] (image1)
   {\includegraphics[width=\imagedim, height=\imagedim]{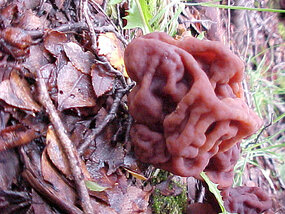}};
   \node [img,anchor=north,at=(image0.south), yshift=-\imageyspacing] (image2)
   {\includegraphics[width=\imagedim, height=\imagedim]{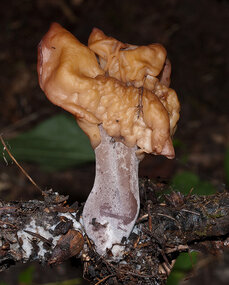}};
   \node [img,anchor=west,at=(image2.east),xshift=\imagexspacing] (image3)
   {\includegraphics[width=\imagedim, height=\imagedim]{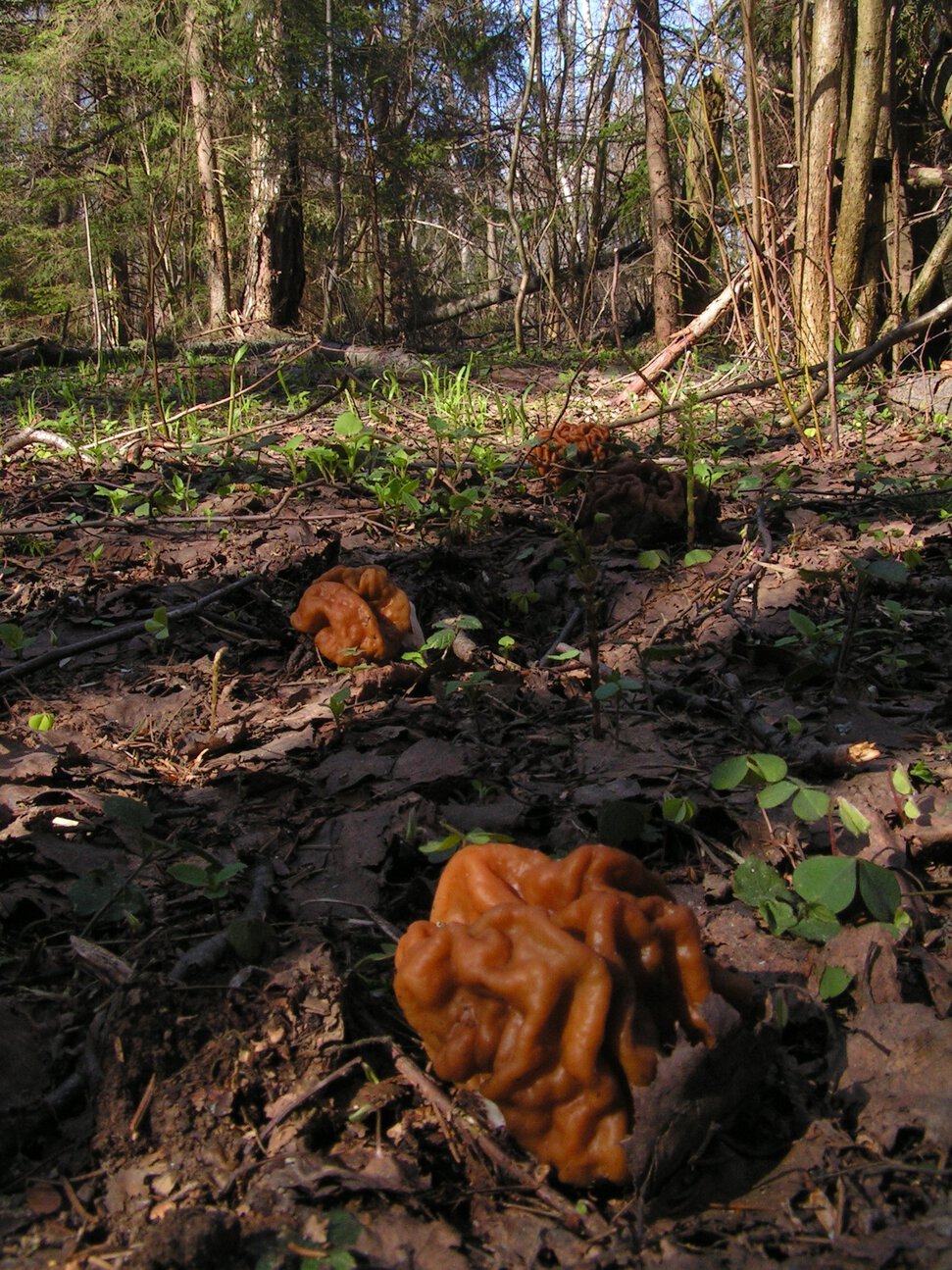}};

   \node [img,anchor=east,at=(image0.west),xshift=-4.0\imagexspacing] (image4)
   {\includegraphics[width=\imagedim, height=\imagedim]{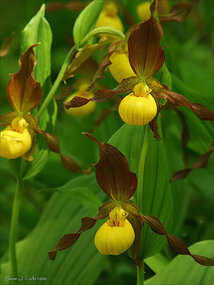}};
   \node [img,anchor=east,at=(image4.west),xshift=-\imagexspacing] (image5)
   {\includegraphics[width=\imagedim, height=\imagedim]{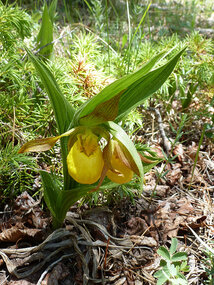}};
   \node [img,anchor=north,at=(image4.south), yshift=-\imageyspacing] (image6)
   {\includegraphics[width=\imagedim, height=\imagedim]{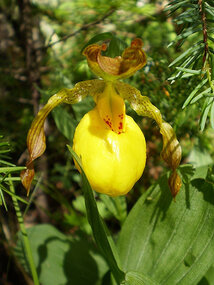}};
   \node [img,anchor=east,at=(image6.west),xshift=-\imagexspacing] (image7)
   {\includegraphics[width=\imagedim, height=\imagedim]{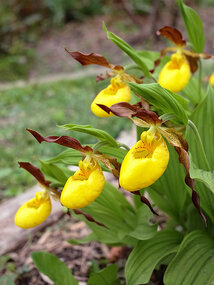}};

   \node [img,anchor=west,at=(image1.east),xshift=4.0\imagexspacing] (image8)
   {\includegraphics[width=\imagedim, height=\imagedim]{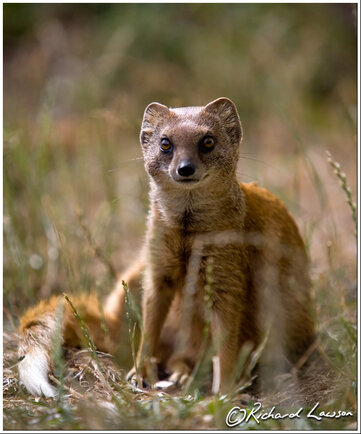}};
   \node [img,anchor=west,at=(image8.east),xshift=\imagexspacing] (image9)
   {\includegraphics[width=\imagedim, height=\imagedim]{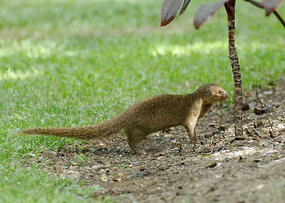}};
   \node [img,anchor=north,at=(image8.south), yshift=-\imageyspacing] (image10)
   {\includegraphics[width=\imagedim, height=\imagedim]{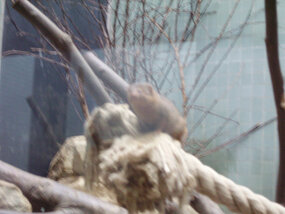}};
   \node [img,anchor=west,at=(image10.east),xshift=\imagexspacing] (image11)
   {\includegraphics[width=\imagedim, height=\imagedim]{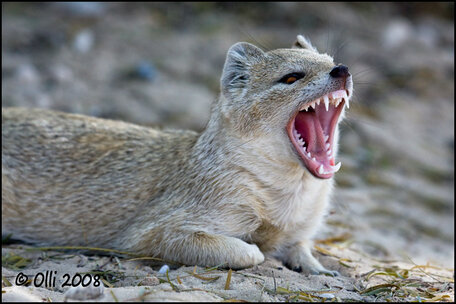}};
   \end{tikzpicture}
    \centering
    \vspace{-.1cm}
     \caption{Test Set A}
   \end{subfigure}

   \vspace{.2cm}

   \begin{subfigure}[t]{\textwidth}
   \begin{tikzpicture}
   \tikzstyle{img}=[inner sep=0pt,outer sep=0pt];
   \node [img] (image0) {\includegraphics[width=\imagedim, height=\imagedim]{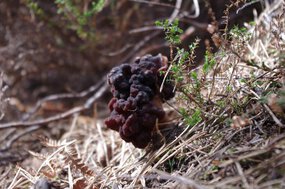}};
   \node [img,anchor=west,at=(image0.east),xshift=\imagexspacing] (image1)
   {\includegraphics[width=\imagedim, height=\imagedim]{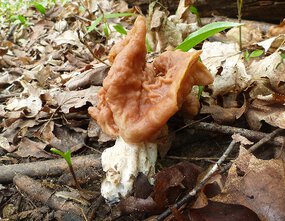}};
   \node [img,anchor=north,at=(image0.south), yshift=-\imageyspacing] (image2)
   {\includegraphics[width=\imagedim, height=\imagedim]{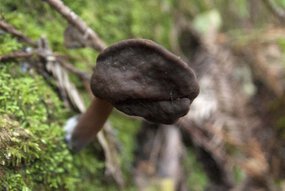}};
   \node [img,anchor=west,at=(image2.east),xshift=\imagexspacing] (image3)
   {\includegraphics[width=\imagedim, height=\imagedim]{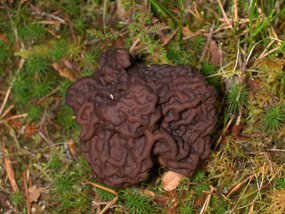}};

   \node [img,anchor=east,at=(image0.west),xshift=-4.0\imagexspacing] (image4)
   {\includegraphics[width=\imagedim, height=\imagedim]{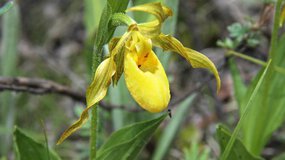}};
   \node [img,anchor=east,at=(image4.west),xshift=-\imagexspacing] (image5)
   {\includegraphics[width=\imagedim, height=\imagedim]{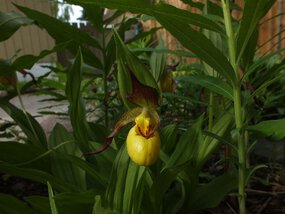}};
   \node [img,anchor=north,at=(image4.south), yshift=-\imageyspacing] (image6)
   {\includegraphics[width=\imagedim, height=\imagedim]{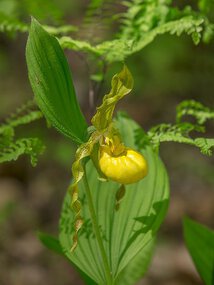}};
   \node [img,anchor=east,at=(image6.west),xshift=-\imagexspacing] (image7)
   {\includegraphics[width=\imagedim, height=\imagedim]{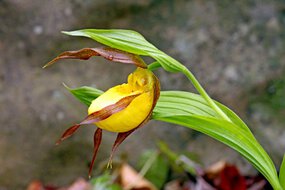}};

   \node [img,anchor=west,at=(image1.east),xshift=4.0\imagexspacing] (image8)
   {\includegraphics[width=\imagedim, height=\imagedim]{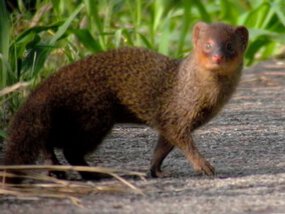}};
   \node [img,anchor=west,at=(image8.east),xshift=\imagexspacing] (image9)
   {\includegraphics[width=\imagedim, height=\imagedim]{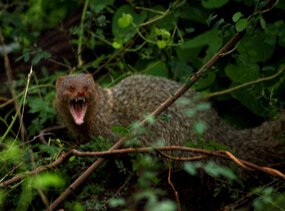}};
   \node [img,anchor=north,at=(image8.south), yshift=-\imageyspacing] (image10)
   {\includegraphics[width=\imagedim, height=\imagedim]{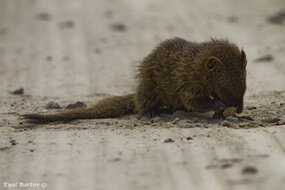}};
   \node [img,anchor=west,at=(image10.east),xshift=\imagexspacing] (image11)
   {\includegraphics[width=\imagedim, height=\imagedim]{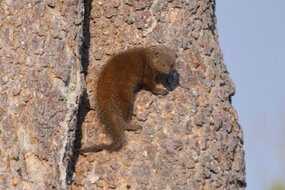}};
   \end{tikzpicture}
    \centering
    \vspace{-.1cm}
    \caption{Test Set B}
   \end{subfigure}
   
   \vspace{.2cm}

   \begin{subfigure}[t]{\textwidth}
   \begin{tikzpicture}
   \tikzstyle{img}=[inner sep=0pt,outer sep=0pt];
   \node [img] (image0) {\includegraphics[width=\imagedim, height=\imagedim]{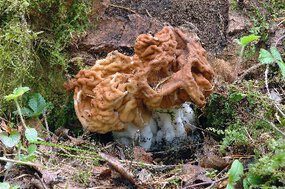}};
   \node [img,anchor=west,at=(image0.east),xshift=\imagexspacing] (image1)
   {\includegraphics[width=\imagedim, height=\imagedim]{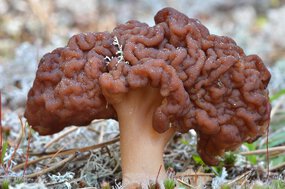}};
   \node [img,anchor=north,at=(image0.south), yshift=-\imageyspacing] (image2)
   {\includegraphics[width=\imagedim, height=\imagedim]{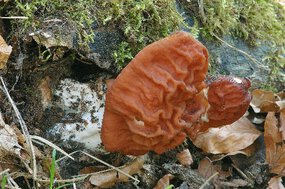}};
   \node [img,anchor=west,at=(image2.east),xshift=\imagexspacing] (image3)
   {\includegraphics[width=\imagedim, height=\imagedim]{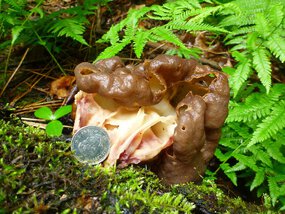}};

   \node [img,anchor=east,at=(image0.west),xshift=-4.0\imagexspacing] (image4)
   {\includegraphics[width=\imagedim, height=\imagedim]{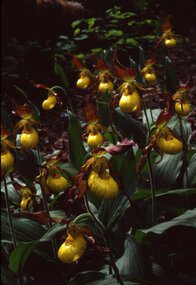}};
   \node [img,anchor=east,at=(image4.west),xshift=-\imagexspacing] (image5)
   {\includegraphics[width=\imagedim, height=\imagedim]{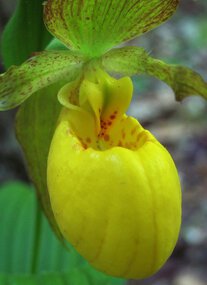}};
   \node [img,anchor=north,at=(image4.south), yshift=-\imageyspacing] (image6)
   {\includegraphics[width=\imagedim, height=\imagedim]{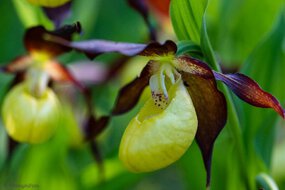}};
   \node [img,anchor=east,at=(image6.west),xshift=-\imagexspacing] (image7)
   {\includegraphics[width=\imagedim, height=\imagedim]{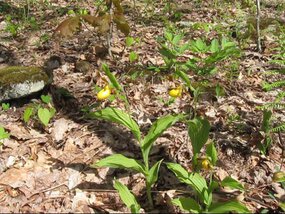}};

   \node [img,anchor=west,at=(image1.east),xshift=4.0\imagexspacing] (image8)
   {\includegraphics[width=\imagedim, height=\imagedim]{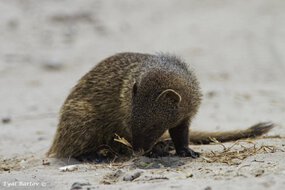}};
   \node [img,anchor=west,at=(image8.east),xshift=\imagexspacing] (image9)
   {\includegraphics[width=\imagedim, height=\imagedim]{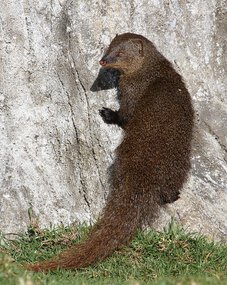}};
   \node [img,anchor=north,at=(image8.south), yshift=-\imageyspacing] (image10)
   {\includegraphics[width=\imagedim, height=\imagedim]{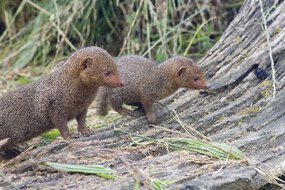}};
   \node [img,anchor=west,at=(image10.east),xshift=\imagexspacing] (image11)
   {\includegraphics[width=\imagedim, height=\imagedim]{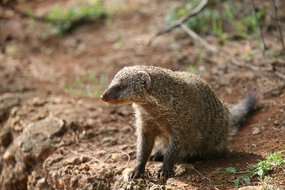}};
   \end{tikzpicture}
    \centering
    \vspace{-.1cm}
    \caption{Test Set C}
   \end{subfigure}
   
   \vspace{.2cm}

   \begin{subfigure}[t]{\textwidth}
   \begin{tikzpicture}
   \tikzstyle{img}=[inner sep=0pt,outer sep=0pt];
   \node [img] (image0) {\includegraphics[width=\imagedim, height=\imagedim]{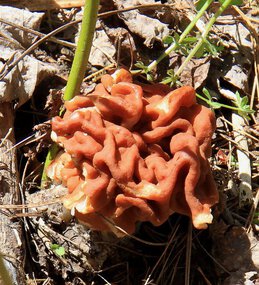}};
   \node [img,anchor=west,at=(image0.east),xshift=\imagexspacing] (image1)
   {\includegraphics[width=\imagedim, height=\imagedim]{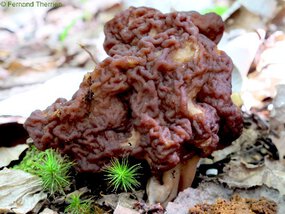}};
   \node [img,anchor=north,at=(image0.south), yshift=-\imageyspacing] (image2)
   {\includegraphics[width=\imagedim, height=\imagedim]{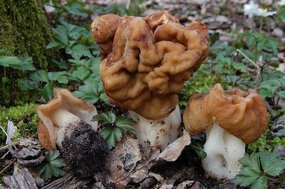}};
   \node [img,anchor=west,at=(image2.east),xshift=\imagexspacing] (image3)
   {\includegraphics[width=\imagedim, height=\imagedim]{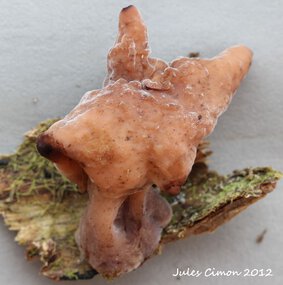}};

   \node [img,anchor=east,at=(image0.west),xshift=-4.0\imagexspacing] (image4)
   {\includegraphics[width=\imagedim, height=\imagedim]{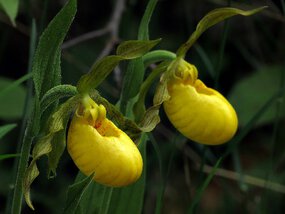}};
   \node [img,anchor=east,at=(image4.west),xshift=-\imagexspacing] (image5)
   {\includegraphics[width=\imagedim, height=\imagedim]{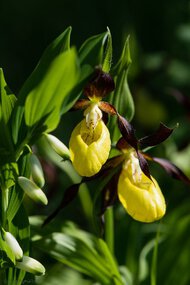}};
   \node [img,anchor=north,at=(image4.south), yshift=-\imageyspacing] (image6)
   {\includegraphics[width=\imagedim, height=\imagedim]{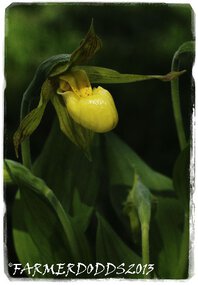}};
   \node [img,anchor=east,at=(image6.west),xshift=-\imagexspacing] (image7)
   {\includegraphics[width=\imagedim, height=\imagedim]{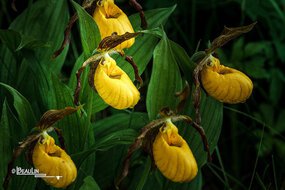}};

   \node [img,anchor=west,at=(image1.east),xshift=4.0\imagexspacing] (image8)
   {\includegraphics[width=\imagedim, height=\imagedim]{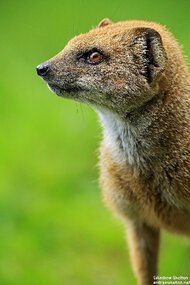}};
   \node [img,anchor=west,at=(image8.east),xshift=\imagexspacing] (image9)
   {\includegraphics[width=\imagedim, height=\imagedim]{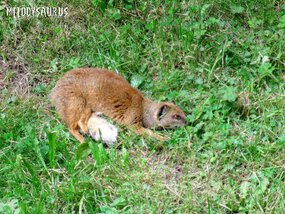}};
   \node [img,anchor=north,at=(image8.south), yshift=-\imageyspacing] (image10)
   {\includegraphics[width=\imagedim, height=\imagedim]{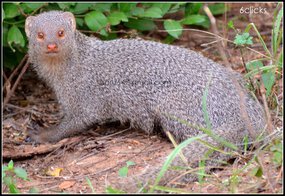}};
   \node [img,anchor=west,at=(image10.east),xshift=\imagexspacing] (image11)
   {\includegraphics[width=\imagedim, height=\imagedim]{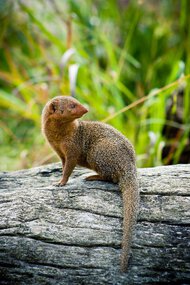}};
   \end{tikzpicture}
    \centering
    \vspace{-.1cm}
     \caption{Test Set D}
   \end{subfigure}

   \caption{\small Randomly selected images from the original ImageNet validation set and our new ImageNet test sets.
   We display four images from three randomly selected classes for each of the four datasets (the original validation set and our three test sets described in Section \ref{sec:imagenet_details}).
   The displayed classes are ``Cypripedium calceolus'', ``gyromitra'', and ``mongoose''. 
   The following footnote reveals which datasets correspond to original and new ImageNet test sets. \protect \footnotemark}
   \label{fig:testexamples_imagenet}
   \end{figure*}
   \footnotetext{Test Set A is the original validation set, Test Set B is the \datasetb dataset, Test Set C is the \dataseta, Test set D is \datasetc.}

\subsubsection{Effect of Selection Frequency on Model Accuracy}
\label{sec:rainbow}
To better understand how the selection frequency of an image impacts the model accuracies, Figures \ref{fig:rainbow_plot}, \ref{fig:rainbow_plot_original}, and \ref{fig:rainbow_plot_both} show model accuracies stratified into five selection frequency bins.

\begin{figure*}[ht!]
  \centering
  \begin{subfigure}{.74\textwidth}
    \includegraphics[width=\linewidth]{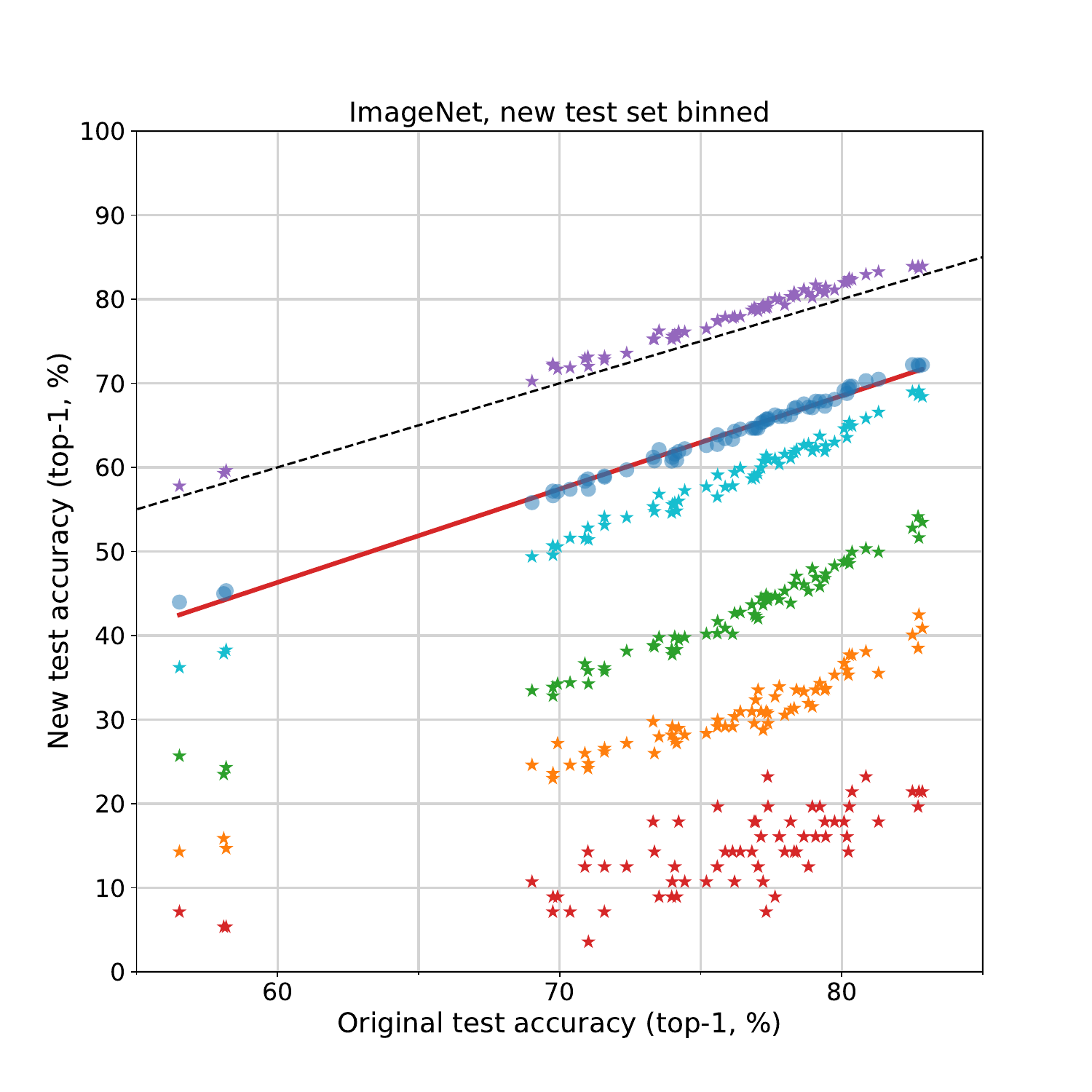}
  \end{subfigure}
  \begin{subfigure}{0.24\textwidth}
    \includegraphics[width=\linewidth]{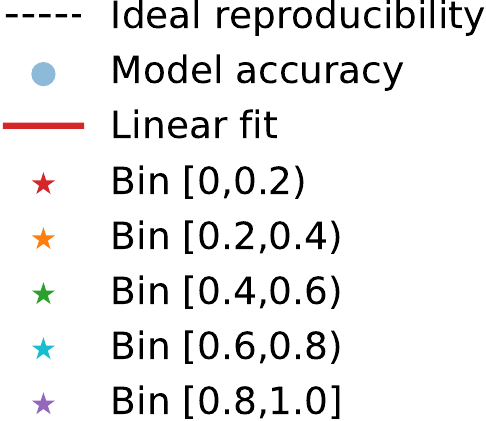}
  \end{subfigure} \\
  \caption{Model accuracy on the original ImageNet validation set vs.\ accuracy on our new test set \datasetb{}, stratified into five selection frequency bins.
  Every bin contains the images with MTurk selection frequency falling into the corresponding range.
    Each data point corresponds to one model and one of the five frequency bins (indicated by the different colors).
    The x-value of each data point is given by the model's accuracy on the entire original validation set.
    The y-value is given by the model's accuracy on our new test images falling into the respective selection frequency bin.
  The plot shows that the selection frequency has strong influence on the model accuracy.
  For instance, images with selection frequencies in the $[0.4, 0.6)$ bin lead to an average model accuracy about 20\% lower than for the entire test set \datasetb{}, and 30\% lower than the original validation set.
  We remark that we manually reviewed all images in \datasetb{} to ensure that (almost) all images have the correct class label, regardless of selection frequency bin.}
  \label{fig:rainbow_plot}
\end{figure*}

\begin{figure*}[ht!]
  \centering
  \begin{subfigure}{.74\textwidth}
    \includegraphics[width=\linewidth]{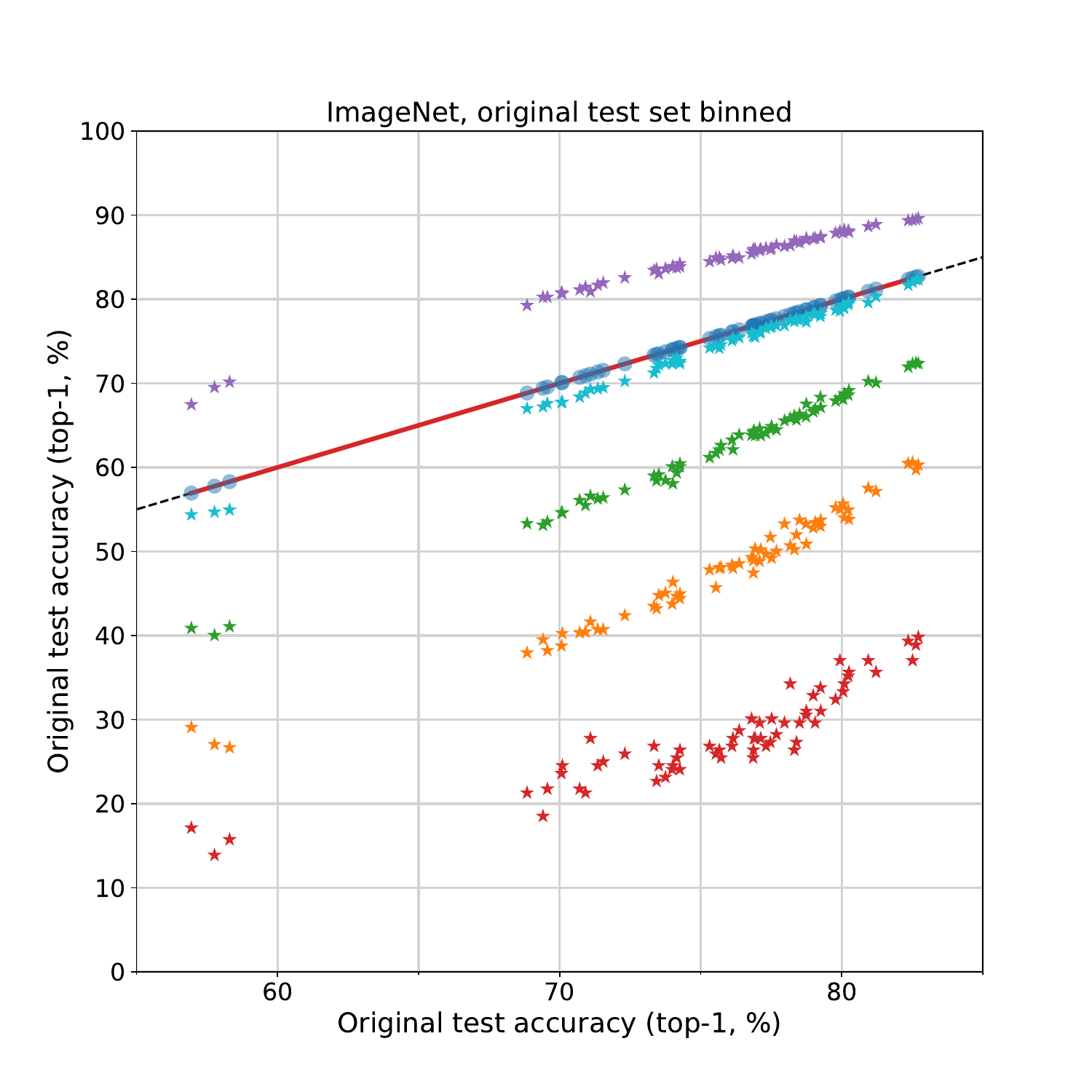}
  \end{subfigure}
  \begin{subfigure}{0.24\textwidth}
    \includegraphics[width=\linewidth]{figures/rainbow_plot/rainbow_imagenetv2-b-33_bin_new_legend_cropped.pdf}
  \end{subfigure} \\
  \caption{Model accuracy on the original ImageNet validation set stratified into five selection frequency bins.
    This plot has a similar structure as Figure \ref{fig:rainbow_plot} above, but contains the original validation set accuracy on both axes (as before, the images are binned on the y-axis and not binned on the x-axis, i.e., the x-value is the accuracy on the entire validation set).
  The plot shows that the selection frequency has strong influence on the model accuracy on the original ImageNet validation set as well.
  For instance, images with selection frequencies in the $[0.4, 0.6)$ bin lead to an average model accuracy about 10 -- 15\% lower than for the entire validation set.
}
  \label{fig:rainbow_plot_original}
\end{figure*}

\begin{figure*}[ht!]
  \centering
  \begin{subfigure}{.74\textwidth}
    \includegraphics[width=\linewidth]{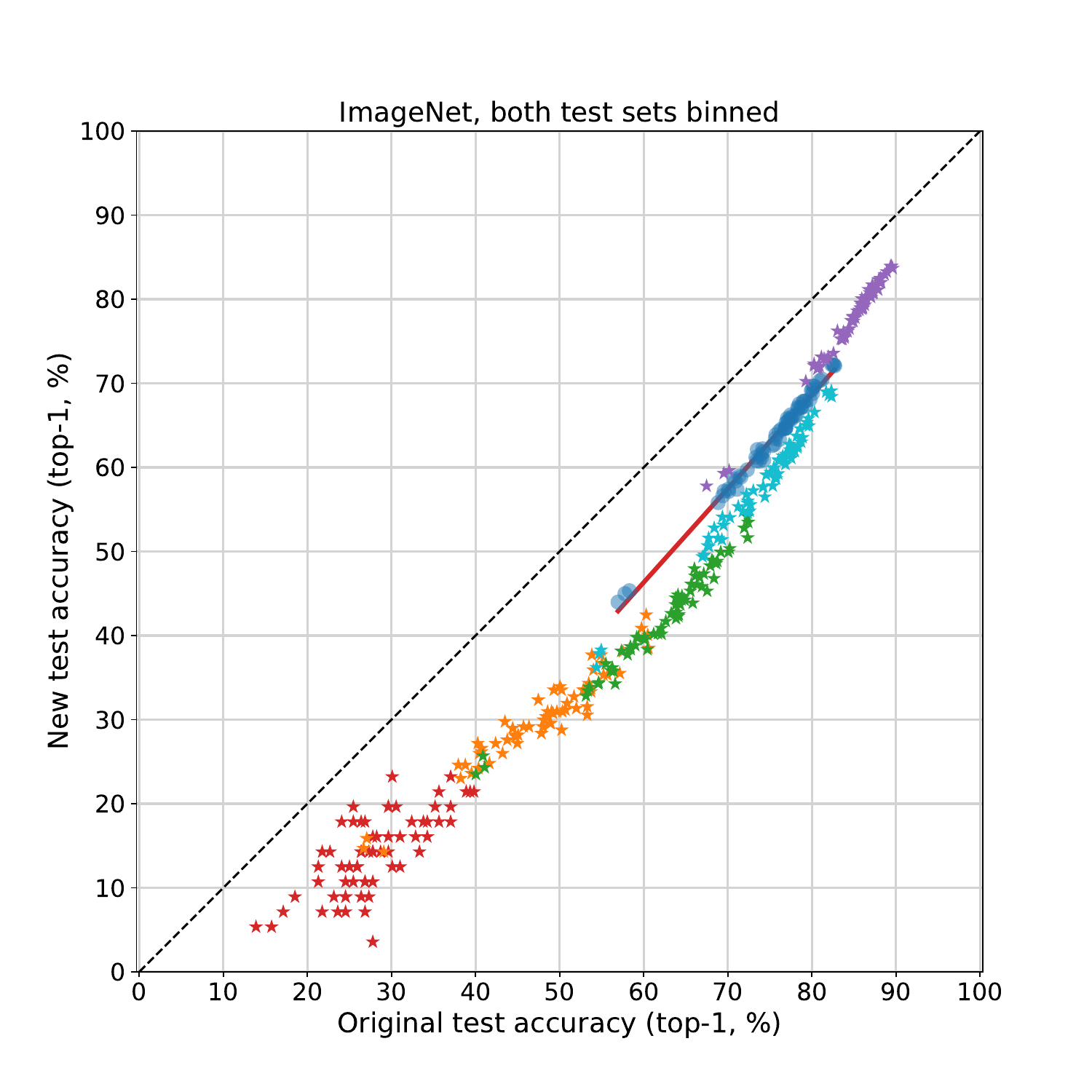}
  \end{subfigure}
  \begin{subfigure}{0.24\textwidth}
    \includegraphics[width=\linewidth]{figures/rainbow_plot/rainbow_imagenetv2-b-33_bin_new_legend_cropped.pdf}
  \end{subfigure} \\
  \caption{Model accuracy on the original ImageNet validation set vs.\ accuracy on our new test set \datasetb{}.
    In contrast to the preceding Figures \ref{fig:rainbow_plot} and \ref{fig:rainbow_plot_original}, both original and new test accuracy is now stratified into five selection frequency bins.
    Each data point corresponds to the accuracy achieved by one model on the images from one of the five frequency bins (indicated by the different colors).
    The plot shows that the model accuracies in the various bins are strongly correlated, but the accuracy on images in our new test is consistently lower.
    The gap is largest for images in the middle frequency bins (about 20\% accuracy difference) and smallest for images in the lowest and highest frequency bins (5 -- 10 \% difference).
  }
  \label{fig:rainbow_plot_both}
\end{figure*}

\subsubsection{Ambiguous Class Examples}
\label{app:ambiguous_imagenet}

Figure \ref{fig:ambiguous_examples_imagenet} shows randomly selected images from the original ImageNet validation set for three pairs of classes with ambiguous class boundaries.
We remark that several more classes in ImageNet have ill-defined boundaries.
The three pairs of classes here were chosen only as illustrative examples.

The following list shows names and definitions for the three class pairs:

\begin{itemize}
  \item Pair 1
  \begin{enumerate}
  \item[a.] \class{projectile, missile}: ``a weapon that is forcibly thrown or projected at a targets but is not self-propelled''
  \item[b.] \class{missile}: ``a rocket carrying a warhead of conventional or nuclear explosives; may be ballistic or directed by remote control''
  \end{enumerate} 
  \item Pair 2
  \begin{enumerate}
  \item[c.] \class{tusker}: ``any mammal with prominent tusks (especially an elephant or wild boar)''
  \item[d.] \class{Indian elephant, Elephas maximus}: ``Asian elephant having smaller ears and tusks primarily in the male''
  \end{enumerate} 
  \item Pair 3
  \begin{enumerate}
  \item[e.] \class{screen, CRT screen}: ``the display that is electronically created on the surface of the large end of a cathode-ray tube''
  \item[f.] \class{monitor}: ``electronic equipment that is used to check the quality or content of electronic transmissions''
  \end{enumerate}
\end{itemize}

% "n04008634,n03773504,n01871265,n02504013,n04152593,n03782006"
\begin{figure*}
  \centering
  \setlength{\imagedim}{2cm}
  \setlength{\imagexspacing}{0.3cm}
  \setlength{\imageyspacing}{0.3cm}
  \vspace{-.5cm}
  \begin{subfigure}[t]{0.49\textwidth}
  \begin{tikzpicture}
  \tikzstyle{img}=[inner sep=0pt,outer sep=0pt];
  \node [img] (image0) {\includegraphics[width=\imagedim, height=\imagedim]{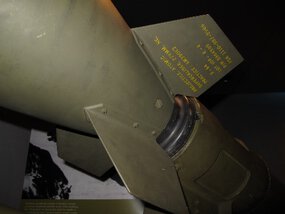}};
  \node [img,anchor=west,at=(image0.east),xshift=\imagexspacing] (image1)
  {\includegraphics[width=\imagedim, height=\imagedim]{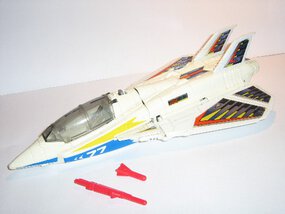}};
  \node [img,anchor=west,at=(image1.east), xshift=\imagexspacing] (image2)
  {\includegraphics[width=\imagedim, height=\imagedim]{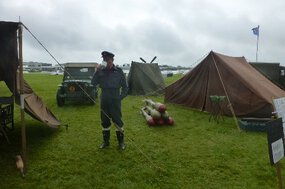}};
  \node [img,anchor=north,at=(image0.south),yshift=-\imagexspacing] (image3)
  {\includegraphics[width=\imagedim, height=\imagedim]{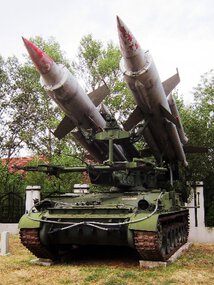}};
  \node [img,anchor=west,at=(image3.east),xshift=\imagexspacing] (image4)
  {\includegraphics[width=\imagedim, height=\imagedim]{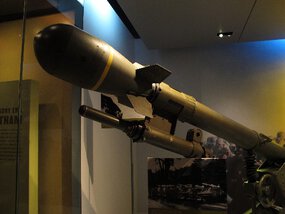}};
  \node [img,anchor=west,at=(image4.east),xshift=\imagexspacing] (image5)
  {\includegraphics[width=\imagedim, height=\imagedim]{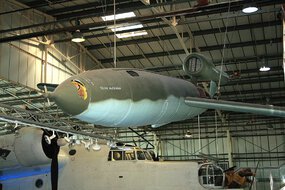}};
  \node [img,anchor=north,at=(image3.south),yshift=-\imagexspacing] (image6)
  {\includegraphics[width=\imagedim, height=\imagedim]{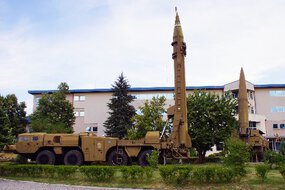}};
  \node [img,anchor=west,at=(image6.east),xshift=\imagexspacing] (image7)
  {\includegraphics[width=\imagedim, height=\imagedim]{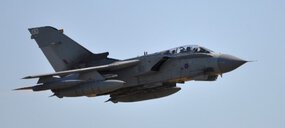}};
  \node [img,anchor=west,at=(image7.east),xshift=\imagexspacing] (image8)
  {\includegraphics[width=\imagedim, height=\imagedim]{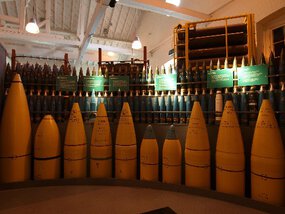}};
  \end{tikzpicture}
   \centering
   \vspace{-.1cm}
   \caption{\class{projectile, missile}} %, ""a weapon that is forcibly thrown or projected at a targets but is not self-propelled"}
  \end{subfigure}
  \begin{subfigure}[t]{0.49\textwidth}
    \begin{tikzpicture}
    \tikzstyle{img}=[inner sep=0pt,outer sep=0pt];
    \node [img] (image0) {\includegraphics[width=\imagedim, height=\imagedim]{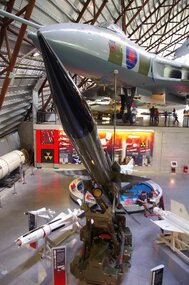}};
    \node [img,anchor=west,at=(image0.east),xshift=\imagexspacing] (image1)
    {\includegraphics[width=\imagedim, height=\imagedim]{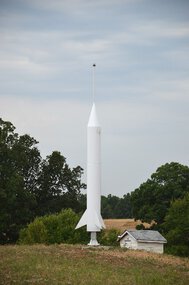}};
    \node [img,anchor=west,at=(image1.east), xshift=\imagexspacing] (image2)
    {\includegraphics[width=\imagedim, height=\imagedim]{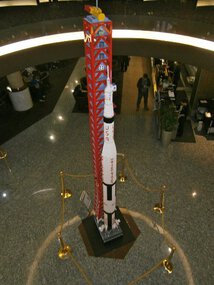}};
    \node [img,anchor=north,at=(image0.south),yshift=-\imagexspacing] (image3)
    {\includegraphics[width=\imagedim, height=\imagedim]{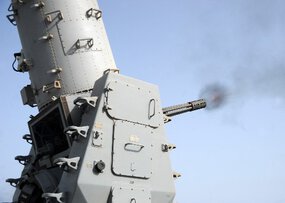}};
    \node [img,anchor=west,at=(image3.east),xshift=\imagexspacing] (image4)
    {\includegraphics[width=\imagedim, height=\imagedim]{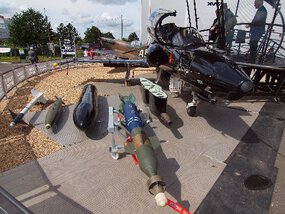}};
    \node [img,anchor=west,at=(image4.east),xshift=\imagexspacing] (image5)
    {\includegraphics[width=\imagedim, height=\imagedim]{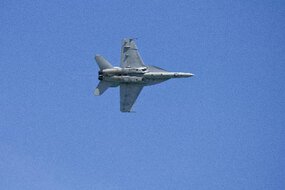}};
    \node [img,anchor=north,at=(image3.south),yshift=-\imagexspacing] (image6)
    {\includegraphics[width=\imagedim, height=\imagedim]{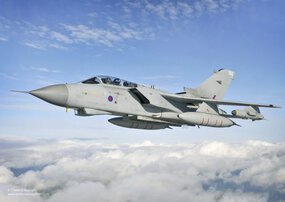}};
    \node [img,anchor=west,at=(image6.east),xshift=\imagexspacing] (image7)
    {\includegraphics[width=\imagedim, height=\imagedim]{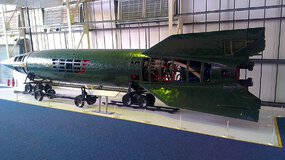}};
    \node [img,anchor=west,at=(image7.east),xshift=\imagexspacing] (image8)
    {\includegraphics[width=\imagedim, height=\imagedim]{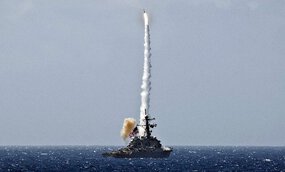}};
    \end{tikzpicture}
     \centering
     \vspace{-.1cm}
     \caption{\class{missile}}%, "a rocket carrying a warhead of conventional or nuclear explosives; may be ballistic or directed by remote control"}
    \end{subfigure}
  \vspace{.3cm}

  \begin{subfigure}[t]{0.49\textwidth}
    \begin{tikzpicture}
    \tikzstyle{img}=[inner sep=0pt,outer sep=0pt];
    \node [img] (image0) {\includegraphics[width=\imagedim, height=\imagedim]{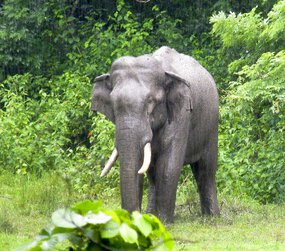}};
    \node [img,anchor=west,at=(image0.east),xshift=\imagexspacing] (image1)
    {\includegraphics[width=\imagedim, height=\imagedim]{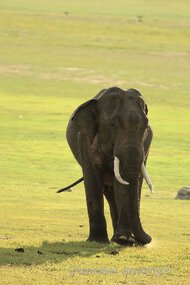}};
    \node [img,anchor=west,at=(image1.east), xshift=\imagexspacing] (image2)
    {\includegraphics[width=\imagedim, height=\imagedim]{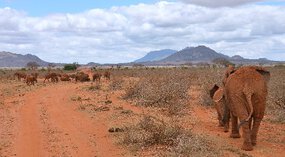}};
    \node [img,anchor=north,at=(image0.south),yshift=-\imagexspacing] (image3)
    {\includegraphics[width=\imagedim, height=\imagedim]{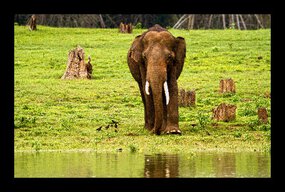}};
    \node [img,anchor=west,at=(image3.east),xshift=\imagexspacing] (image4)
    {\includegraphics[width=\imagedim, height=\imagedim]{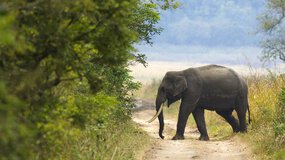}};
    \node [img,anchor=west,at=(image4.east),xshift=\imagexspacing] (image5)
    {\includegraphics[width=\imagedim, height=\imagedim]{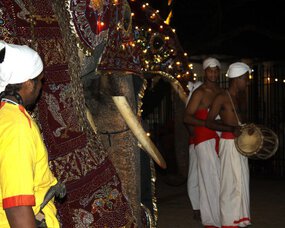}};
    \node [img,anchor=north,at=(image3.south),yshift=-\imagexspacing] (image6)
    {\includegraphics[width=\imagedim, height=\imagedim]{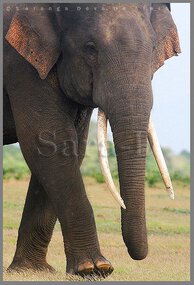}};
    \node [img,anchor=west,at=(image6.east),xshift=\imagexspacing] (image7)
    {\includegraphics[width=\imagedim, height=\imagedim]{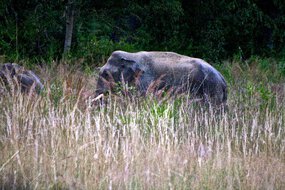}};
    \node [img,anchor=west,at=(image7.east),xshift=\imagexspacing] (image8)
    {\includegraphics[width=\imagedim, height=\imagedim]{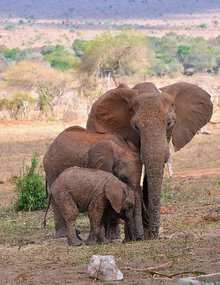}};
    \end{tikzpicture}
     \centering
     \vspace{-.1cm}
     \caption{\class{tusker}}%, "any mammal with prominent tusks (especially an elephant or wild boar)"}
    \end{subfigure}
    \begin{subfigure}[t]{0.49\textwidth}
      \begin{tikzpicture}
      \tikzstyle{img}=[inner sep=0pt,outer sep=0pt];
      \node [img] (image0) {\includegraphics[width=\imagedim, height=\imagedim]{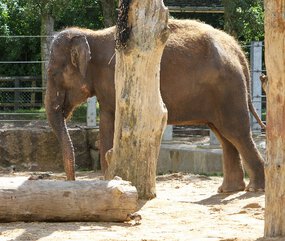}};
      \node [img,anchor=west,at=(image0.east),xshift=\imagexspacing] (image1)
      {\includegraphics[width=\imagedim, height=\imagedim]{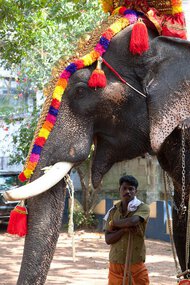}};
      \node [img,anchor=west,at=(image1.east), xshift=\imagexspacing] (image2)
      {\includegraphics[width=\imagedim, height=\imagedim]{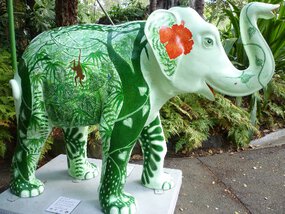}};
      \node [img,anchor=north,at=(image0.south),yshift=-\imagexspacing] (image3)
      {\includegraphics[width=\imagedim, height=\imagedim]{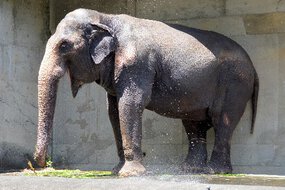}};
      \node [img,anchor=west,at=(image3.east),xshift=\imagexspacing] (image4)
      {\includegraphics[width=\imagedim, height=\imagedim]{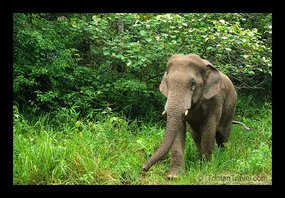}};
      \node [img,anchor=west,at=(image4.east),xshift=\imagexspacing] (image5)
      {\includegraphics[width=\imagedim, height=\imagedim]{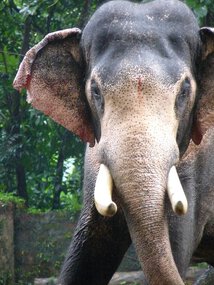}};
      \node [img,anchor=north,at=(image3.south),yshift=-\imagexspacing] (image6)
      {\includegraphics[width=\imagedim, height=\imagedim]{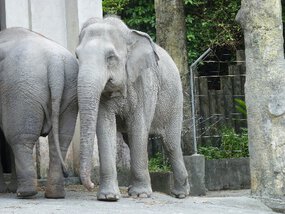}};
      \node [img,anchor=west,at=(image6.east),xshift=\imagexspacing] (image7)
      {\includegraphics[width=\imagedim, height=\imagedim]{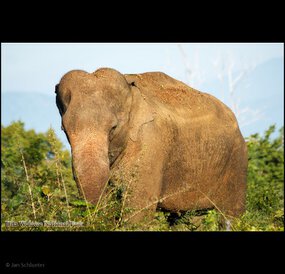}};
      \node [img,anchor=west,at=(image7.east),xshift=\imagexspacing] (image8)
      {\includegraphics[width=\imagedim, height=\imagedim]{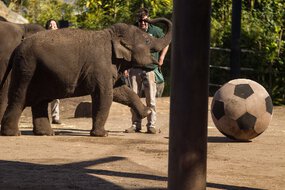}};
      \end{tikzpicture}
       \centering
       \vspace{-.1cm}
       \caption{\class{Indian elephant, Elephas maximus}}%, "Asian elephant having smaller ears and tusks primarily in the male"}
      \end{subfigure}
      \vspace{.3cm}

    \begin{subfigure}[t]{0.49\textwidth}
      \begin{tikzpicture}
      \tikzstyle{img}=[inner sep=0pt,outer sep=0pt];
      \node [img] (image0) {\includegraphics[width=\imagedim, height=\imagedim]{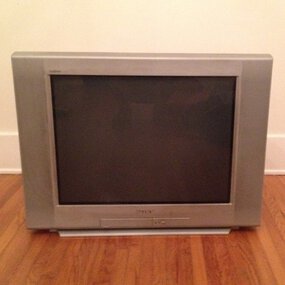}};
      \node [img,anchor=west,at=(image0.east),xshift=\imagexspacing] (image1)
      {\includegraphics[width=\imagedim, height=\imagedim]{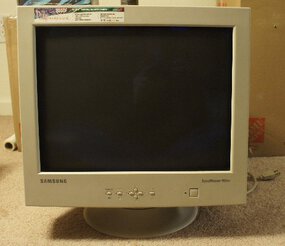}};
      \node [img,anchor=west,at=(image1.east), xshift=\imagexspacing] (image2)
      {\includegraphics[width=\imagedim, height=\imagedim]{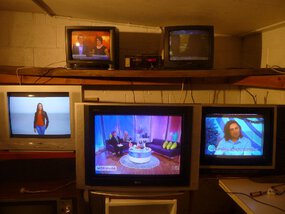}};
      \node [img,anchor=north,at=(image0.south),yshift=-\imagexspacing] (image3)
      {\includegraphics[width=\imagedim, height=\imagedim]{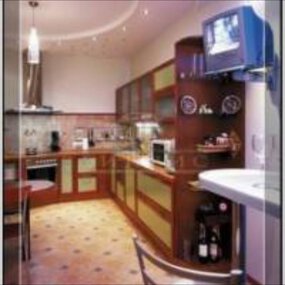}};
      \node [img,anchor=west,at=(image3.east),xshift=\imagexspacing] (image4)
      {\includegraphics[width=\imagedim, height=\imagedim]{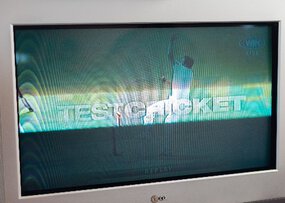}};
      \node [img,anchor=west,at=(image4.east),xshift=\imagexspacing] (image5)
      {\includegraphics[width=\imagedim, height=\imagedim]{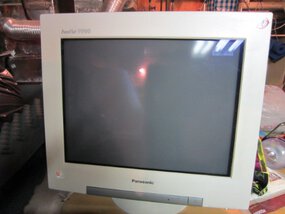}};
      \node [img,anchor=north,at=(image3.south),yshift=-\imagexspacing] (image6)
      {\includegraphics[width=\imagedim, height=\imagedim]{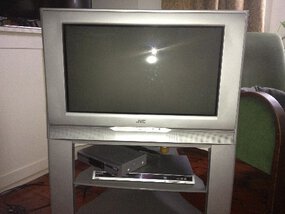}};
      \node [img,anchor=west,at=(image6.east),xshift=\imagexspacing] (image7)
      {\includegraphics[width=\imagedim, height=\imagedim]{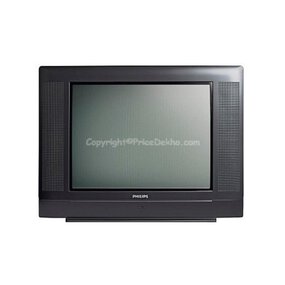}};
      \node [img,anchor=west,at=(image7.east),xshift=\imagexspacing] (image8)
      {\includegraphics[width=\imagedim, height=\imagedim]{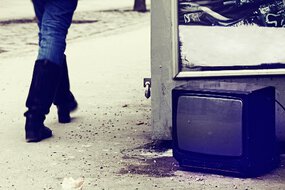}};
      \end{tikzpicture}
       \centering
       \vspace{-.1cm}
       \caption{\class{screen, CRT screen}} %, "the display that is electronically created on the surface of the large end of a cathode-ray tube"}
      \end{subfigure}
      \begin{subfigure}[t]{0.49\textwidth}
        \begin{tikzpicture}
        \tikzstyle{img}=[inner sep=0pt,outer sep=0pt];
        \node [img] (image0) {\includegraphics[width=\imagedim, height=\imagedim]{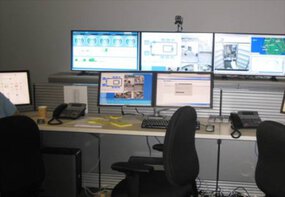}};
        \node [img,anchor=west,at=(image0.east),xshift=\imagexspacing] (image1)
        {\includegraphics[width=\imagedim, height=\imagedim]{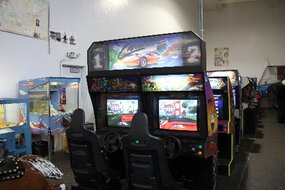}};
        \node [img,anchor=west,at=(image1.east), xshift=\imagexspacing] (image2)
        {\includegraphics[width=\imagedim, height=\imagedim]{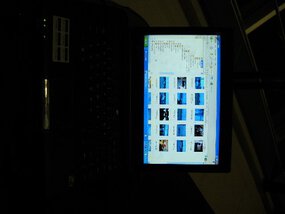}};
        \node [img,anchor=north,at=(image0.south),yshift=-\imagexspacing] (image3)
        {\includegraphics[width=\imagedim, height=\imagedim]{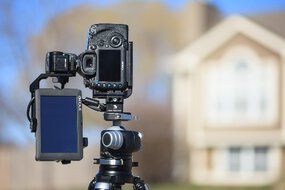}};
        \node [img,anchor=west,at=(image3.east),xshift=\imagexspacing] (image4)
        {\includegraphics[width=\imagedim, height=\imagedim]{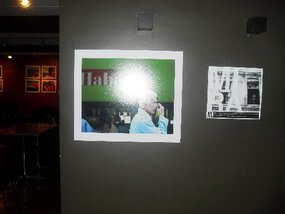}};
        \node [img,anchor=west,at=(image4.east),xshift=\imagexspacing] (image5)
        {\includegraphics[width=\imagedim, height=\imagedim]{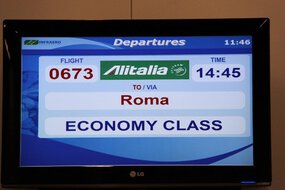}};
        \node [img,anchor=north,at=(image3.south),yshift=-\imagexspacing] (image6)
        {\includegraphics[width=\imagedim, height=\imagedim]{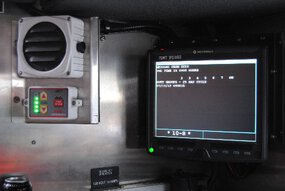}};
        \node [img,anchor=west,at=(image6.east),xshift=\imagexspacing] (image7)
        {\includegraphics[width=\imagedim, height=\imagedim]{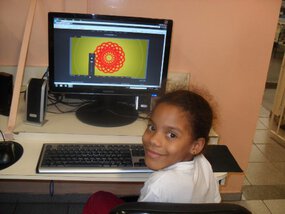}};
        \node [img,anchor=west,at=(image7.east),xshift=\imagexspacing] (image8)
        {\includegraphics[width=\imagedim, height=\imagedim]{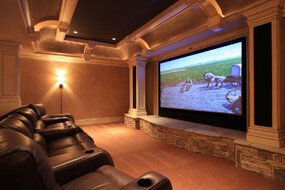}};
        \end{tikzpicture}
         \centering
         \vspace{-.1cm}
         \caption{\class{monitor}} %, "electronic equipment that is used to check the quality or content of electronic transmissions"}
        \end{subfigure}
    \vspace{-.2cm}
  \caption{\small Random images from the original ImageNet validation set for three pairs of classes with ambiguous class boundaries.}
  \label{fig:ambiguous_examples_imagenet}
  \end{figure*}

}{}

\iftoggle{showtodos}{
  \section{List of ToDos}
  \listoftodos
}{}

\end{document}